\documentclass[10pt,twocolumn,letterpaper]{article}
\usepackage[T1]{fontenc}

\usepackage[pagenumbers]{wacv} 

\definecolor{wacvblue}{rgb}{0.21,0.49,0.74}
\usepackage[ruled,vlined,linesnumbered,algo2e]{algorithm2e}
\usepackage[pagebackref,breaklinks,colorlinks,allcolors=wacvblue]{hyperref}
\usepackage{graphicx}    
\usepackage{subcaption} 
\usepackage{booktabs}   
\usepackage{rotating}   
\usepackage{amsmath}    
\usepackage{stfloats}
\usepackage{tcolorbox}
\tcbuselibrary{skins, breakable}
\usepackage{fancyvrb}
\usepackage{fvextra}
\usepackage{xcolor}
\usepackage{xcolor}
\usepackage{fvextra}    
\usepackage{subcaption}
\usepackage{tcolorbox}
\tcbuselibrary{skins, breakable}

\usepackage{xcolor}
\usepackage{listings}
\usepackage{caption}

\definecolor{backgray}{rgb}{0.95,0.95,0.95}
\definecolor{backblue}{rgb}{0.92,0.95,0.99}
\definecolor{codepurple}{rgb}{0.5,0,0.35}
\definecolor{codegreen}{rgb}{0,0.5,0}

\def\wacvPaperID{284} 
\def\confName{WACV}
\def\confYear{2027}

\title{BayesCrop: Real-Time Probabilistic Multi-Crop Video Reframing}

\author{
Cheng-Han Lee\textsuperscript{1}\,\,\,\,\, 
Rajesh Sureddi\textsuperscript{1}\,\,\,\,\, 
Neil Birkbeck\textsuperscript{2}\,\,\,\,\,
Yilin Wang\textsuperscript{2}\,\,\,\,\,
Balu Adsumilli\textsuperscript{2}\,\,\,\,\,
Alan Bovik\textsuperscript{3}\\\\
\centerline{
\textsuperscript{1}The University of Texas at Austin\,\,\,\,\,
\textsuperscript{2}YouTube, Google\,\,\,\,\,
\textsuperscript{3}University of Colorado Boulder
}
}

\begin{document}
\maketitle

\begin{abstract}
Automatic video reframing is increasingly important for adapting landscape content to heterogeneous display formats. Existing methods typically predict a single deterministic crop, despite the subjective nature of reframing and the presence of multiple plausible regions of interest. We present \textbf{BayesCrop}, a real-time probabilistic framework for multi-crop video reframing. BayesCrop represents crop preferences using a two-component spatially isotropic Gaussian mixture over crop position and scale. Each component defines a representative reframing mode together with its local uncertainty, enabling both dual-view prediction and sampling of additional plausible crop candidates. For efficient video processing, we introduce an anchor-and-propagate strategy that performs neural prediction only on sparse anchor frames and propagates crop trajectories using motion vectors available from the compressed video bitstream. The resulting trajectories can be combined into a single output using a lightweight periodic view-switching strategy. Experiments on LIVE-YT-MC, LIVE-YT-VC++, and a multi-target long-video benchmark demonstrate competitive single-output localization, high multi-crop reference coverage, diverse yet coherent crop hypotheses, and stable perceptual quality. BayesCrop achieves real-time throughput of 112 FPS at the default operating point, while sampling from the learned distributions further improves reference coverage beyond the two representative crop means. The source code is publicly available at  \href{https://github.com/steven413d/BayesCrop}{\textcolor{blue}{https://github.com/steven413d/BayesCrop}}.
\end{abstract}
\section{Introduction}

\begin{figure}[t]
\centering
\includegraphics[width=1.0\linewidth]{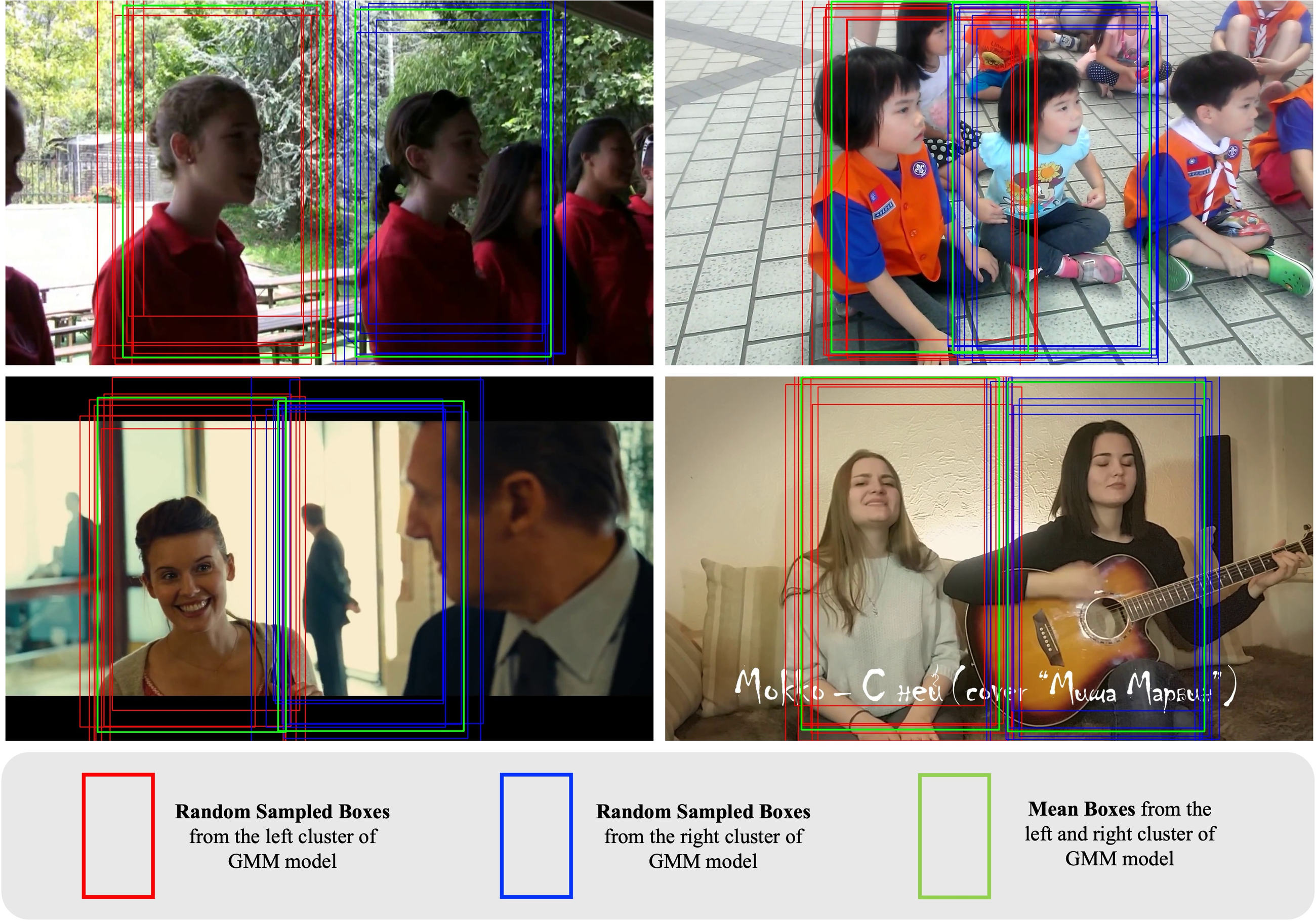}
\caption{
\textbf{Probabilistic ROI prediction on LIVE-YT-VC++ \cite{lee2026subjective}.} Green boxes denote predicted
GMM means; red and blue boxes are samples from the two crop modes.
}
\label{fig:random}
\end{figure}

\begin{figure*}[t]
    \centering
    \includegraphics[width=0.9\linewidth]{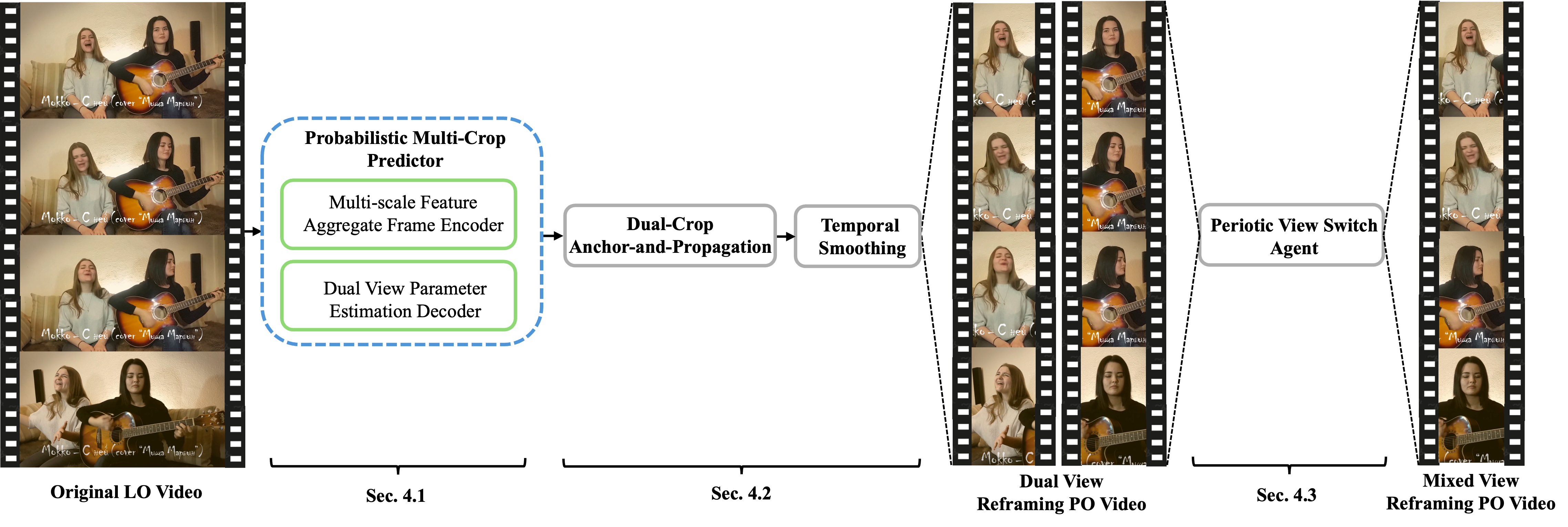}
    \caption{
    \textbf{Overview of BayesCrop.}
    Given a landscape-orientation (LO) video, BayesCrop predicts two
probabilistic crop modes on anchor frames and propagates them using
compressed-domain motion vectors. Gaussian temporal smoothing produces two stable crop trajectories, which are alternated by a lightweight Periodic View Switch Agent to generate the final mixed-view portrait-orientation (PO) output.}
    \label{fig1}
\end{figure*}

The rapid growth of short-form video and heterogeneous display formats
has increased the demand for automatic video reframing. Landscape
content must often be converted into portrait formats while preserving
important subjects, visual composition, and temporal coherence.
However, reframing is inherently subjective: a frame may contain
multiple meaningful regions of interest, and different viewers may
prefer different crops. Representing such ambiguity with a single
``optimal'' bounding box therefore oversimplifies the set of plausible
reframing choices.

We instead formulate reframing as a \emph{probabilistic multi-crop}
problem. As illustrated in Fig.~\ref{fig:random}, multiple crop
hypotheses can concentrate around meaningful subjects while retaining
local variation in position and scale. This motivates
\textbf{BayesCrop}, which represents crop parameters $(x,y,r)$ using a
two-component spatially isotropic Gaussian mixture. Each component
defines a representative crop through its mean and a local candidate
distribution through its variance. The model therefore supports both
two compact reframing modes and additional plausible crops sampled from
the predicted distributions.

To enable efficient video processing, BayesCrop adopts an
\emph{anchor-and-propagate} strategy. Full crop prediction is performed
only on sparse anchor frames, while intermediate crop trajectories are
propagated using motion vectors available directly from the compressed
video bitstream. This avoids additional optical-flow or tracking
networks and substantially reduces repeated neural inference while
maintaining temporally coherent trajectories. The resulting two crop
streams can be converted into a single reframed video using a lightweight
\emph{Periodic View Switch Agent}, which alternates between the two
trajectories over temporal segments without additional semantic models.

To supervise probabilistic prediction, we construct
\textbf{LIVE-YT-MC}, a pseudo-labeled dataset derived from subjective
human crop annotations. Rather than collapsing multiple opinions into
one target, LIVE-YT-MC preserves multiple crop modes and their local
variation in $(x,y,r)$ space.

We evaluate BayesCrop from both single-output and multi-crop
perspectives. On LIVE-YT-VC++, BayesCrop achieves competitive
single-output localization while substantially improving set-level
reference coverage and operating in real time. Sampling additional
candidates from the predicted GMM further improves Best-of-$N$ coverage,
demonstrating the utility of modeling continuous crop distributions
rather than only two deterministic boxes. Experiments on longer
multi-target videos further show that BayesCrop maintains visual quality
and temporal stability while providing diverse, content-relevant
reframing.

Our contributions are:
\begin{itemize}

    \item \textbf{Probabilistic Multi-Crop Reframing.}
    We formulate subjective video reframing as distributional prediction
    and introduce BayesCrop, which models two complementary crop modes
    and supports sampling additional plausible candidates.

    \item \textbf{Human-Preference Distribution Modeling.}
    We construct LIVE-YT-MC from human-subjective crop annotations,
    preserving multimodal crop preferences and local uncertainty.

    \item \textbf{Efficient Anchor-and-Propagate Inference.}
    We combine sparse neural prediction with compressed-domain motion
    propagation to produce temporally coherent crop trajectories in
    real time.

    \item \textbf{Multi-Crop Coverage and Diversity.}
    We evaluate both representative crops and sampled candidates,
    showing improved reference coverage and diverse yet coherent
    reframing hypotheses.
\end{itemize}

\section{Related Works}

\subsection{Image Cropping}
Image cropping methods can be broadly grouped into attention-guided,
aesthetic-aware, and regression-based approaches. Attention-guided
methods preserve salient regions using visual cues or eye-tracking
information \cite{chen2003visual,santella2006gaze}, while later work
incorporates higher-level semantic or emotional salience
\cite{zhang2020emotion}. Aesthetic-aware methods rank candidate crops
according to composition quality
\cite{chen2017learning,wang2017deep,zeng2020grid,wang2023image}, whereas
regression-based methods directly predict crop coordinates
\cite{guo2018automatic,hong2021composing}. Recent approaches further
address varying aspect ratios \cite{li2020learning} and transformer-based
spatial modeling \cite{pan2021transview}. Most existing methods,
however, still produce a single deterministic crop rather than modeling
multiple plausible composition preferences.

\subsection{Video Reframing Benchmarks}
Existing video reframing datasets differ in scale and annotation strategy.
RetargetVid \cite{apostolidis2021fast} provides manually annotated videos
but is limited in scale and resolution, while H2V-142K
\cite{zhu2021horizontal} contains more frames but relatively limited
video diversity. LIVE-YT-VC \cite{lee2024subjective} introduced a
large-scale subjective study of portrait-region preferences, and
LIVE-YT-VC++ \cite{lee2026subjective} further aggregates multiple human opinions into temporally
stable references. More recently, LOPOV \cite{gu2025dynamic} generates
annotations using VLMs and image-composition models. In contrast,
LIVE-YT-MC is derived from human-subjective LIVE-YT-VC annotations and
explicitly preserves multiple crop modes and their local variation.

\subsection{Video Reframing Methods}
Video reframing methods commonly combine saliency, subject detection,
and temporal tracking. Representative approaches include H2V
\cite{zhu2021horizontal}, SmartVidCrop \cite{apostolidis2021fast}, and
FEVR \cite{gu2025dynamic}, which primarily formulate reframing as
deterministic single-ROI prediction. Spatio-temporal grounding models
such as STCAT \cite{jin2022embracing} and CG-STVG
\cite{gu2024context} can also be adapted for reframing, but rely on
heavier architectures and remain single-view. Agent-based methods such
as RAVA \cite{cao2024reframe} provide flexible editing at higher
computational cost. BayesCrop instead models multiple plausible crop
modes and combines sparse semantic prediction with compressed-domain
motion propagation for efficient real-time reframing.

\section{LIVE-YT-MC: LIVE YouTube Multi-Crop Dataset}

\subsection{Probabilistic Multi-Crop Modeling}

\begin{figure*}[t]
\centering
\includegraphics[width=0.9\linewidth]{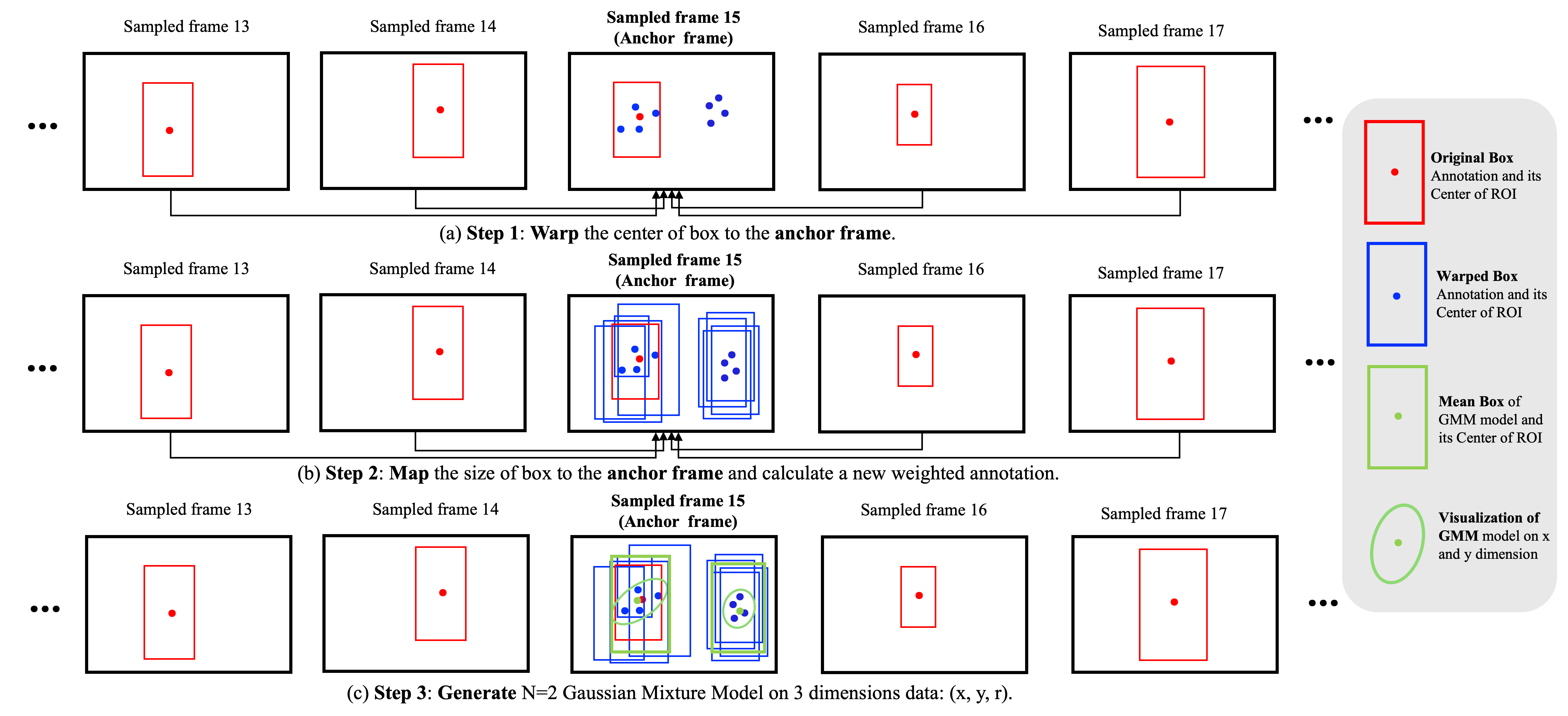}
\caption{
\textbf{Pseudo-label generation pipeline for LIVE-YT-MC within a sliding temporal window.} The three steps are: (1) temporal warping of human-opinion crops (red), (2) aggregation of warped candidates within the window (blue), and (3) two-component GMM fitting, where green boxes denote representative crop means. For visualization, the figure uses a window size of 5, while the actual data-generation pipeline uses a window size of 15.
}
\label{fig2}
\end{figure*}

Conventional reframing benchmarks typically assign a single crop to each
frame, despite the subjective nature of visual composition. Different
viewers may prefer different subjects, crop locations, or scales while
producing equally plausible reframing results. We therefore extend
LIVE-YT-VC into \textbf{LIVE-YT-MC}, where crop preferences are modeled
in the parameter space $\mathbf{z}=(x,y,r)$, with $(x,y)$ denoting the
crop center and $r$ the crop scale.

Rather than collapsing multiple opinions into one deterministic target,
LIVE-YT-MC represents them using a two-component Gaussian Mixture Model
(GMM). Each component describes a dominant reframing mode together with
its local variation, providing supervision for both representative crop
prediction and probabilistic sampling.

\subsection{Video Sources and Human Annotations}

LIVE-YT-MC is derived from the human-subjective annotations of
LIVE-YT-VC. The source videos originate from YouTube-UGC
\cite{wang2019youtube} and LSVQ \cite{ying2021patch}. We retain 224
YouTube-UGC videos and 1,576 LSVQ videos at 1080p resolution, yielding
1,800 six-second clips with 180 frames each.

For LIVE-YT-VC, 30 frames are uniformly sampled from each clip for
subjective annotation. Human participants identify visually important
regions according to their perceptual and aesthetic preferences,
naturally producing variation across subjects, locations, and crop
scales. LIVE-YT-MC preserves this variability as a distribution of
plausible reframing choices.

\subsection{Data Curation}

\subsubsection{\textbf{Subjective Pseudo-Label Generation}}

As illustrated in Fig.~\ref{fig2}, we transform sparse human-opinion
annotations into temporally dense probabilistic crop targets through
three steps: temporal warping, sliding-window aggregation, and GMM
fitting.

\textbf{Iterative Warping and Anchoring.}
Given an annotated crop center $c_t=(x_t,y_t)$, we propagate it across
neighboring frames using iterative pixel-level tracking and map the
resulting trajectory into a reference anchor frame:
\begin{equation}
c_t^{\,\mathrm{anchor}}
=
\mathcal{W}_{t\rightarrow\mathrm{anchor}}(c_t),
\end{equation}
where $\mathcal{W}_{t\rightarrow\mathrm{anchor}}$ denotes the temporal
warping operator. This alignment converts sparse frame-level annotations
into a denser set of warped candidates associated with the anchor frame.

\textbf{Sliding-Window Aggregation.}
Around each anchor frame, we collect warped annotations from a local
temporal window. This aggregation incorporates short-term motion context
and reduces sensitivity to sparse annotation, producing a dense point
cloud of plausible crop candidates in the neighborhood of each anchor.

\textbf{Two-Mode Clustering and GMM Fitting.}
The aggregated candidates may contain multiple reframing modes together
with occasional tracking outliers. We first apply K-Means with $K=2$ to
identify two dominant crop groups and suppress unreliable warped samples.
We then fit a two-component GMM over
$\mathbf{z}=[x,y,r]^T$:
\begin{equation}
p(\mathbf{z})
=
\sum_{k=1}^{2}
\pi_k
\mathcal{N}
\left(
\mathbf{z}\mid
\boldsymbol{\mu}_k,
\boldsymbol{\Sigma}_k
\right).
\end{equation}

We choose $K=2$ as a compact representation of subjective reframing
ambiguity. A single component collapses competing visual focuses, while
larger $K$ increases the number of parameters to predict and can
fragment limited annotations into weakly supported, overlapping, or
redundant modes. Two components therefore provide a practical balance
between preserving alternative reframing choices and maintaining compact,
stable supervision.

Here, $\boldsymbol{\mu}_k$ and $\boldsymbol{\Sigma}_k$ denote the mean
and covariance of the $k$-th reframing mode, while $\pi_k$ represents
its relative mass in the aggregated human-opinion distribution. We treat
$\pi_k$ as a soft preference cue rather than a hard indicator of a
uniquely correct crop. The component means provide representative crop
hypotheses, while their variances define local neighborhoods from which
additional plausible candidates can be sampled.

\subsubsection{\textbf{Spatially Isotropic GMM Simplification}}

Directly predicting a full covariance matrix introduces additional
cross-dimensional parameters and requires the network to estimate
correlations between crop center and scale. To obtain a more compact and
stable supervision target, we simplify each fitted covariance while retaining a conservative estimate of local spatial uncertainty.

For a fitted component,
\begin{equation}
\boldsymbol{\Sigma}
=
\begin{bmatrix}
\sigma_x^2 & \sigma_{xy} & \sigma_{xr} \\
\sigma_{yx} & \sigma_y^2 & \sigma_{yr} \\
\sigma_{rx} & \sigma_{ry} & \sigma_r^2
\end{bmatrix},
\end{equation}
we extract the spatial sub-matrix $\boldsymbol{\Sigma}_{xy}$ and compute
its minimum eigenvalue $\lambda_{\min}$. The covariance used for
supervision is then
\begin{equation}
\boldsymbol{\Sigma}_{\mathrm{iso}}
=
\begin{bmatrix}
\lambda_{\min} & 0 & 0 \\
0 & \lambda_{\min} & 0 \\
0 & 0 & \sigma_r^2
\end{bmatrix}.
\end{equation}

Thus, the $(x,y)$ dimensions share a common spatial variance, while the
crop-scale variance $\sigma_r^2$ is retained independently. Using
$\lambda_{\min}$ yields a conservative circular approximation governed by the narrower principal direction of the fitted spatial distribution. This removes cross-covariance terms and reduces prediction complexity while preserving the local uncertainty of each reframing mode.

\begin{figure}[t]
\centering
\includegraphics[width=1.0\linewidth]{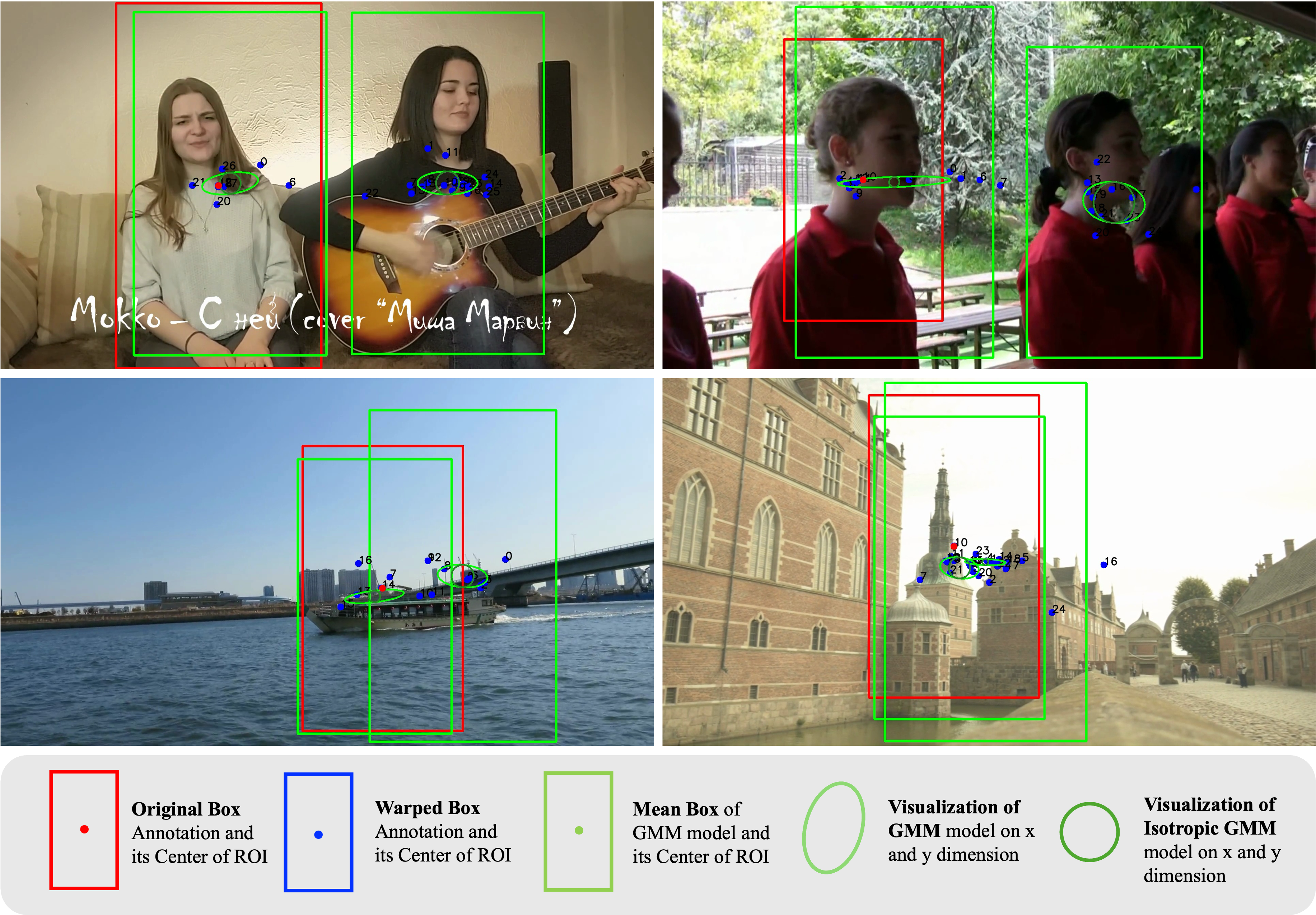}
\caption{
\textbf{GMM-based ROI modeling in LIVE-YT-MC.} Elliptical contours show the
original fitted spatial covariance, while circular contours show the
spatially isotropic representation used for supervision. The examples
illustrate both separated crop modes and local scale variation.
}
\label{fig3}
\end{figure}

As shown in Fig.~\ref{fig3}, the representation captures both spatially separated crop modes and local variation in crop scale around similar regions of interest.

\section{Proposed Method}

BayesCrop is an efficient probabilistic multi-crop reframing framework.
Given sparsely sampled anchor frames, it predicts two complementary crop
distributions and propagates their representative crops using
compressed-domain motion information to obtain temporally coherent
trajectories. For single-output reframing, a lightweight Periodic View
Switch Agent alternates between the two trajectories over temporal
segments.

\subsection{Probabilistic Multi-Crop Predictor}

\begin{figure*}[t]
\centering
\includegraphics[width=1.0\linewidth]{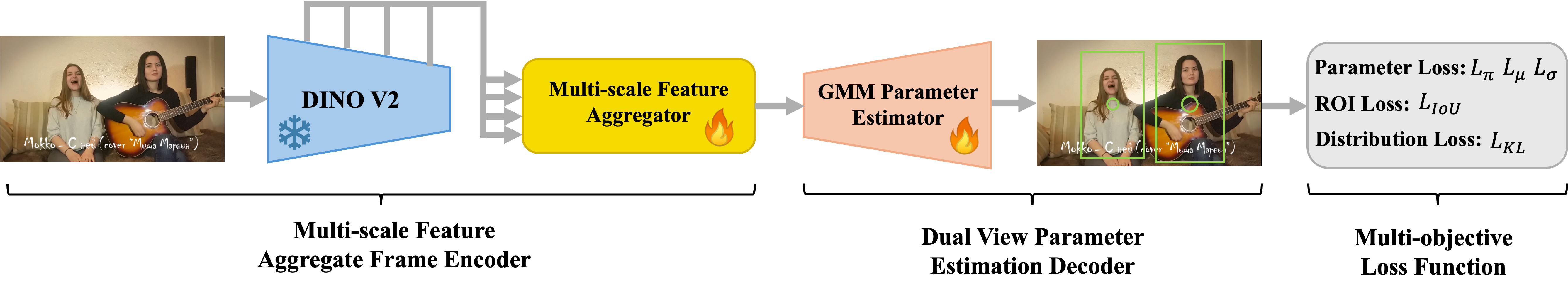}
\caption{
\textbf{Architecture of  Probabilistic Multi-Crop Predictor.} A frozen DINOv2 backbone and multi-scale
feature aggregator feed two learnable crop queries that predict the
parameters $(\pi,\mu,\sigma)$ of a two-component spatially isotropic GMM.
}
\label{fig4}
\end{figure*}

\subsubsection{\textbf{Model Design}}

As shown in Fig.~\ref{fig4}, BayesCrop consists of a frozen
DINOv2~\cite{oquab2023dinov2} visual encoder, a multi-scale feature
aggregator, and a lightweight GMM parameter estimator. Features from
four intermediate transformer layers are concatenated and projected by
a $1\times1$ convolution to form a compact fused representation.

Following DETR~\cite{carion2020end}, two learnable crop queries interact
with the fused features, with each query representing one reframing mode.
Three lightweight MLP heads predict
\begin{equation}
\left\{
\pi_k,\,
\mu_k,\,
\sigma_k
\right\}_{k=1}^{2},
\end{equation}
where $\pi_k$ denotes the mixture weight, $\mu_k=(x,y,r)$ defines the
representative crop, and $\sigma_k$ describes its local uncertainty.
The two means provide representative crop hypotheses, while the
predicted variances define continuous neighborhoods from which
additional plausible candidates can be sampled.

\subsubsection{\textbf{Multi-Objective Training}}

The predictor is trained using complementary parameter, geometric, and
distributional objectives:
\begin{equation}
\mathcal{L}_{\mathrm{total}}
=
\mathcal{L}_{\mathrm{param}}
+
\mathcal{L}_{\mathrm{roi}}
+
\mathcal{L}_{\mathrm{dist}}.
\end{equation}

The parameter loss directly supervises the GMM parameters:
\begin{equation}
\mathcal{L}_{\mathrm{param}}
=
\lambda_{\pi}\mathcal{L}_{\pi}
+
\lambda_{\mu}\mathcal{L}_{\mu}
+
\lambda_{\sigma}\mathcal{L}_{\sigma},
\end{equation}
where $\mathcal{L}_{\pi}$ uses focal BCE for the mixture weights and
$\mathcal{L}_{\mu}$ and $\mathcal{L}_{\sigma}$ use focal-weighted
regression losses. The mixture weight $\pi$ is interpreted as a soft
preference mass rather than a hard class label.

To improve representative crop localization, we additionally apply
\begin{equation}
\mathcal{L}_{\mathrm{roi}}
=
\lambda_{\mathrm{roi}}
\left[
1-\mathrm{GIoU}(B_{\mu},B_{\mathrm{gt}})
\right],
\end{equation}
where $B_{\mu}$ is reconstructed from the predicted mean crop. Finally,
distributional consistency is encouraged through
\begin{equation}
\mathcal{L}_{\mathrm{dist}}
=
\lambda_{\mathrm{KL}}
D_{\mathrm{KL}}
\left[
\mathcal{N}(\mu_{\mathrm{gt}},\sigma_{\mathrm{gt}}^2)
\parallel
\mathcal{N}(\mu,\sigma^2)
\right].
\end{equation}
Together, these objectives supervise both representative crop geometry
and local distributional variation.

For detailed formulation of all loss functions, please refer to the Supp.

\subsection{Anchor-and-Propagate Video Reframing}

\SetAlCapHSkip{0pt}
\SetAlgoCaptionSeparator{:}

\begin{algorithm2e}[t]
\caption{\textbf{Dual-Crop Anchor-and-Propagate Reframing}}
\label{alg}
\SetKwInOut{Input}{Input}
\SetKwInOut{Output}{Output}

\Input{
Video sequence $\mathcal{V}=\{f_1,\ldots,f_N\}$,
motion-vector fields $\{\mathbf{MV}_t\}_{t=2}^{N}$,
refresh interval $R$
}
\Output{
Dual crop trajectories $\mathcal{B}=\{B_t^1,B_t^2\}_{t=1}^{N}$
}

\For{$t=1$ \KwTo $N$}{

    $\mathrm{refresh}
    \leftarrow
    (t \bmod R = 0)
    \lor
    \mathrm{IsIFrame}(f_t)
    \lor
    \mathrm{BoundaryCheck}(\mathcal{B}_{t-1})$\;

    \eIf{$\mathrm{refresh}$}{
        \tcp{Semantic anchor prediction}

        $\{\pi_k,\mu_k,\sigma_k\}_{k=1}^{2}
        \leftarrow \mathrm{BayesCrop}(f_t)$\;

        $\{B^{tmp,1}_t,B^{tmp,2}_t\}
        \leftarrow
        \mathrm{BoxFromMean}(\{\mu_1,\mu_2\})$\;
    }{
        \tcp{Compressed-domain propagation}

        \For{$k\in\{1,2\}$}{

            $\Omega_t^k
            \leftarrow
            \{
            (\mathbf{mv}_i,p_i)
            \mid
            \mathbf{mv}_i \in \mathbf{MV}_t,\;
            p_i \in B^k_{t-1}
            \}$\;

            $g_i^k
            \leftarrow
            \exp
            \left(
            -\frac{\|p_i-c^k_{t-1}\|_2^2}{2\sigma_m^2}
            \right)$\;

            $\widetilde{\Omega}_t^k
            \leftarrow
            \{
            (\mathbf{mv}_i,p_i)\in\Omega_t^k
            \mid
            g_i^k > \tau_m
            \}$\;

            $\Delta c_t^k
            \leftarrow
            \mathrm{MedianMotion}(\widetilde{\Omega}_t^k)$\;

            $s_t^k
            \leftarrow
            \mathrm{MedianScale}
            (\widetilde{\Omega}_t^k,\Delta c_t^k,c^k_{t-1})$\;

            $B^{tmp,k}_t
            \leftarrow
            \mathrm{UpdateBox}
            (B^k_{t-1},\Delta c_t^k,s_t^k)$\;
        }
    }

    \For{$k\in\{1,2\}$}{
        $B^{res,k}_t
        \leftarrow
        \mathrm{BoundaryRescue}(B^{tmp,k}_t)$\;

        $B^k_t
        \leftarrow
        \mathrm{TemporalSmooth}(B^{res,k}_t)$\;
    }
}

\Return $\mathcal{B}$\;
\end{algorithm2e}

Running the visual predictor on every frame is redundant and costly. BayesCrop therefore performs full prediction only on anchor or refresh frames and propagates the two crop trajectories between anchors using compressed-domain motion vectors.

Algorithm~\ref{alg} summarizes the propagation procedure. On intermediate frames, in-box motion vectors are filtered by a center-biased Gaussian threshold. Their median motion estimates translation, while the same set estimates scale change. The updated crops are then boundary-corrected and temporally smoothed.

Anchor prediction is triggered periodically, at I-frames, or when propagation becomes invalid, limiting drift while reducing repeated backbone evaluations.

\subsection{Periodic View Switch Agent}
For single-output reframing, we evenly divide the video into $S$ temporal segments and alternate between the two trajectories: ROI$_1 \rightarrow$ ROI$_2 \rightarrow$ ROI$_1 \rightarrow \cdots$. Each ROI is retained throughout its segment, producing stable view changes with negligible selection overhead. The switching policy is independent of the predicted mixture weights $\pi$.
\section{Experiments}
Detailed definitions of all metrics are provided in the Supp.
\subsection{Datasets}

\subsubsection{\textbf{LIVE-YT-MC}}
We construct LIVE-YT-MC from 200 videos randomly sampled from the 1,800 source videos of LIVE-YT-VC~\cite{lee2024subjective}. Using the pseudo-label generation pipeline in Sec.~3, we obtain 32,000 training and 4,000 testing instances, with the split performed at the video level to avoid content leakage. Each instance contains a two-component probabilistic crop annotation consisting of mixture weights, crop means, and spatial/scale variances, and is used to train and analyze the BayesCrop predictor.

\subsubsection{\textbf{LIVE-YT-VC++}}
LIVE-YT-VC++~\cite{lee2026subjective} extends LIVE-YT-VC by temporally aggregating neighboring human-opinion annotations to obtain denser and more stable reference crop trajectories. We use 200 videos from this benchmark as the primary evaluation set for video reframing, reporting single-output localization accuracy, multi-crop reference coverage, and system throughput to assess selected-crop quality, subjective-reference coverage, and the efficiency of the anchor-and-propagate pipeline. For LIVE-YT-VC++, we use causal Gaussian smoothing to preserve the online nature of the reframing pipeline. 

\subsubsection{\textbf{Multi-Target Long Video Benchmark}}
To evaluate longer and more semantically complex videos, we construct a Multi-Target Long Video Benchmark from MiraData~\cite{ju2024miradata}, selecting 114 videos with an average duration of approximately 24 seconds whose captions describe multiple spatially distinct subjects or regions of interest. These videos are disjoint from the training data and are evaluated using VBench~\cite{huang2024vbench} metrics, including Aesthetic Quality, Temporal Smoothness, and Text-Video Alignment, to assess the perceptual quality and semantic consistency of the final reframed videos produced by the Periodic View Switch Agent. For the long-video benchmark, we apply centered non-causal Gaussian smoothing to the complete trajectory, since this evaluation focuses on final rendered video quality rather than streaming latency.

\subsection{Baselines}

We compare BayesCrop with representative geometric, video-reframing,
and spatio-temporal grounding baselines. \textbf{CenterCrop} serves as a
parameter-free geometric baseline, while \textbf{SmartVidCrop}
\cite{apostolidis2021fast} provides an efficient saliency-based
reframing baseline with approximately 8.33M parameters. We also compare
against the stronger grounding-based models \textbf{STCAT}
\cite{jin2022embracing} and \textbf{CG-STVG}
\cite{gu2024context}, containing approximately 207.5M and 238.9M
parameters, respectively, and adapt them for ROI prediction under our evaluation setting. For the long-video benchmark, we additionally use \textbf{Gemini 3 Flash} \cite{google_gemini_3_flash_2026} as a direct MLLM cropping baseline that predicts crop coordinates from sampled video frames.

\subsection{Ablation Study of the Probabilistic Predictor on LIVE-YT-MC}

\begin{table}[t]
\centering
\caption{
\textbf{Ablation study of the BayesCrop predictor on LIVE-YT-MC.}
We evaluate spatial localization, mixture-weight prediction, and
distribution matching.
Lower Wasserstein distance indicates better distributional agreement.
}
\label{tab:main_results}
\resizebox{\columnwidth}{!}{
\begin{tabular}{lcccc}
\toprule
\textbf{Method}
& $\pi$-mIoU $\uparrow$
& $\pi$ Acc. $\uparrow$
& $W_2(x,y)$ $\downarrow$
& $W_2(x,y,r)$ $\downarrow$ \\
\midrule
CenterCrop
& 46.6 & 27.1 & 0.13 & 0.20 \\
\midrule
BayesCrop w/o Aggregator
& 54.8 & \underline{62.4} & \underline{0.10} & 0.14 \\

BayesCrop w/o IoU Loss
& 49.8 & 57.6 & 0.11 & 0.16 \\

BayesCrop w/o KL Loss
& \underline{56.6} & 62.2 & \underline{0.10} & \textbf{0.13} \\

\textbf{BayesCrop (Full)}
& \textbf{58.3}
& \textbf{63.4}
& \textbf{0.09}
& \textbf{0.13} \\
\bottomrule
\end{tabular}
}
\end{table}

Table~\ref{tab:main_results} analyzes the major components of the probabilistic predictor.
The complete BayesCrop model achieves the highest $\pi$-mIoU of 58.3 and mixture-selection accuracy of 63.4, while obtaining the lowest spatial Wasserstein distance of 0.09.

Removing the multi-scale feature aggregator reduces $\pi$-mIoU from 58.3 to 54.8, demonstrating the benefit of combining intermediate DINOv2 representations for crop localization. Removing the ROI loss produces the largest localization degradation, reducing $\pi$-mIoU to 49.8.
In contrast, removing the KL objective primarily affects distribution matching, confirming that direct parameter and geometric supervision alone do not fully capture the spread of the pseudo-label distribution. Together, these results support the use of complementary localization
and distributional objectives.

\subsection{Effect of the Periodic Refresh Interval}

\begin{table}[t]
\centering
\caption{
\textbf{Ablation study of the periodic refresh interval $R$ on LIVE-YT-VC++.}
We evaluate the trade-off among localization accuracy, perceptual
quality, temporal smoothness, and system throughput.
}
\label{tab:ablation_refresh_rate}
\resizebox{\columnwidth}{!}{
\begin{tabular}{lcccc}
\toprule
\textbf{Refresh Interval}
& \textbf{mIoU} $\uparrow$
& \textbf{Aesthetic} $\uparrow$
& \textbf{Temp. Smooth.} $\uparrow$
& \textbf{FPS} $\uparrow$ \\
\midrule
$R=1$  & 52.2 & 45.4 & \textbf{96.0} & 73 \\
$R=5$  & \textbf{52.7} & \textbf{45.7} & \underline{95.9} & 112 \\
$R=10$ & \underline{52.4} & 45.4 & \textbf{96.0} & \underline{116} \\
$R=15$ & 52.0 & \underline{45.5} & \textbf{96.0} & \textbf{120} \\
\bottomrule
\end{tabular}
}
\end{table}

Table~\ref{tab:ablation_refresh_rate} studies the effect of the periodic refresh interval $R$. Increasing $R$ substantially improves throughput by reducing the frequency of full BayesCrop inference, while localization accuracy remains relatively stable across a broad range of settings.

The best localization accuracy is obtained at $R=5$, reaching
52.7 mIoU together with the highest Aesthetic score of 45.7 and 95.9 Temporal Smoothness.
Compared with frame-by-frame inference ($R=1$), this setting increases throughput from 73 FPS to 112 FPS while slightly improving localization accuracy.
Further increasing the interval to $R=10$ or $R=15$ raises throughput to 116 and 120 FPS, respectively, with only modest changes in perceptual quality and temporal smoothness, but with a small decline in mIoU. We therefore adopt $R=5$ as the default operating point for subsequent experiments.

\subsection{Comparison with State-of-the-Art Methods on LIVE-YT-VC++}

\begin{table}[t]
\centering
\caption{
\textbf{Comparison with state-of-the-art methods on LIVE-YT-VC++.}
Single-output mIoU evaluates one selected crop per frame.
Because BayesCrop explicitly predicts two plausible crop modes,
we additionally report Best-of-2 IoU as a set-level reference coverage
metric.
Best-of-2 and conventional single-output mIoU measure different
properties and should not be interpreted as identical evaluation
protocols.
}
\label{tab:sota_comparison}
\resizebox{\columnwidth}{!}{
\begin{tabular}{lcccc}
\toprule
\textbf{Method}
& \textbf{\# Params} $\downarrow$
& \textbf{Single mIoU} $\uparrow$
& \textbf{Best-of-2} $\uparrow$
& \textbf{FPS} $\uparrow$ \\
\midrule
CenterCrop
& -- & 47.7 & -- & -- \\

SmartVidCrop~\cite{apostolidis2021fast}
& \textbf{8.33M} & 48.6 & -- & 16.9 \\

STCAT~\cite{jin2022embracing}
& 207.5M & \underline{58.2} & -- & \underline{44.1} \\

CG-STVG~\cite{gu2024context}
& 238.9M & \textbf{59.1} & -- & 31.9 \\

\midrule

\textbf{BayesCrop} ($\hat{\pi}$-selected, $R=5$)
& \underline{26.0M} & 55.2 & -- & \textbf{112} \\

\textbf{BayesCrop} (Two-View Set, $R=5$)
& \underline{26.0M} & -- & \textbf{68.4} & \textbf{112} \\

\bottomrule
\end{tabular}
}
\end{table}

Table~\ref{tab:sota_comparison} compares BayesCrop with existing
reframing and spatio-temporal grounding approaches.
Among deterministic single-output methods, CG-STVG obtains the highest
localization accuracy at 59.1 mIoU, followed by STCAT at 58.2.
BayesCrop reaches 55.2 mIoU when a single crop is selected using the
predicted mixture weights $\hat{\pi}$.

BayesCrop, however, targets a different accuracy--diversity--efficiency
operating point.
Its 26.0M-parameter predictor is approximately $8.0\times$ smaller than
STCAT and $9.2\times$ smaller than CG-STVG, while achieving 112 FPS,
compared with 44.1 FPS and 31.9 FPS, respectively.
SmartVidCrop uses fewer parameters, but achieves substantially lower
single-output localization accuracy and throughput.

More importantly, BayesCrop explicitly preserves two plausible
reframing hypotheses.
When the complete two-view candidate set is evaluated using Best-of-2
IoU, BayesCrop achieves 68.4 reference coverage.
This result indicates that retaining multiple hypotheses substantially
increases the likelihood that the predicted candidate set contains a
crop consistent with the subjective reference.
We emphasize that the 68.4 Best-of-2 result characterizes multi-crop
coverage rather than conventional single-output localization accuracy.

\subsection{Analysis of View Selection and Multi-Crop Coverage}

\begin{table}[t]
\centering
\caption{
\textbf{Analysis of BayesCrop view-selection strategies on LIVE-YT-VC++.}
The first four rows evaluate single-output policies, while Best-of-2 evaluates the complete two-view candidate set.
}
\label{tab:selection_analysis}
\begin{tabular}{lcc}
\toprule
\textbf{Output / Selection Strategy}
& \textbf{Evaluation}
& \textbf{IoU} $\uparrow$ \\
\midrule

Always Left View
& Single-output
& 52.7 \\

Always Right View
& Single-output
& 53.3 \\

Random Selection
& Single-output
& 53.1 \\

$\hat{\pi}$-based Selection
& Single-output
& \textbf{55.2} \\

\midrule

Best-of-2
& Set Coverage
& \textbf{68.4} \\

\bottomrule
\end{tabular}
\end{table}

Table~\ref{tab:selection_analysis} analyzes single-view selection and multi-crop coverage. Fixed or random selection achieves 52.7-53.3 mIoU, while selecting the mode with the larger predicted mixture weight improves single-output accuracy to 55.2, indicating that $\hat{\pi}$ provides a useful soft preference cue.

Best-of-2 coverage further increases to 68.4, demonstrating the benefit of retaining multiple plausible crop hypotheses. The gap between $\hat{\pi}$-based selection and Best-of-2 also reflects the ambiguity of
ranking subjective reframing choices against a single reference crop.

\begin{figure*}[t]
    \centering
    \def\fw{0.245\textwidth} 
    \setlength{\tabcolsep}{1pt} 
    
    \begin{subfigure}{0.32\textwidth}
        \centering
        \begin{tabular}{cccc}
            \textbf{3s} & \textbf{10s} & \textbf{17s} & \textbf{20s} \\
            \includegraphics[width=\fw]{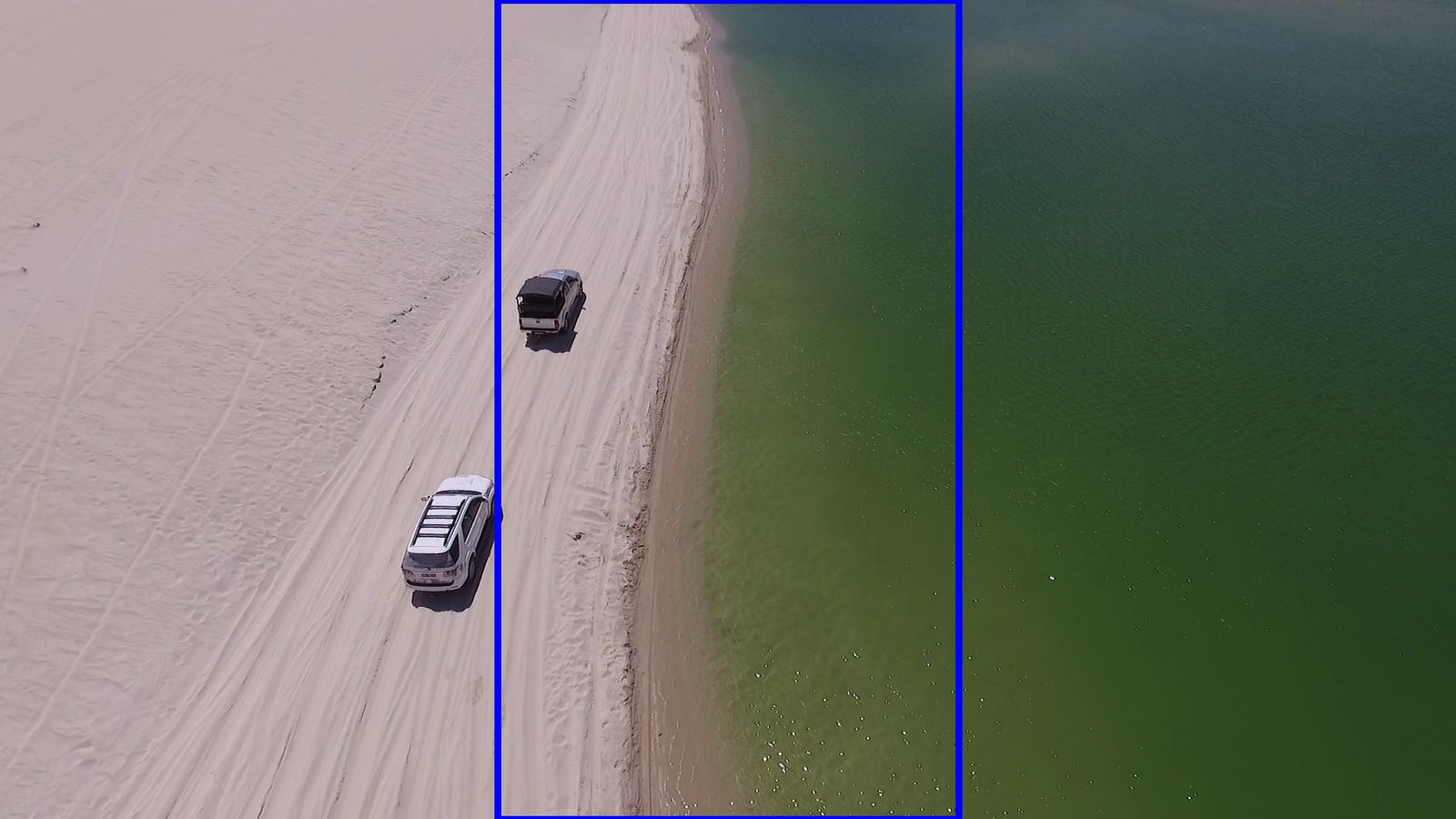} &
            \includegraphics[width=\fw]{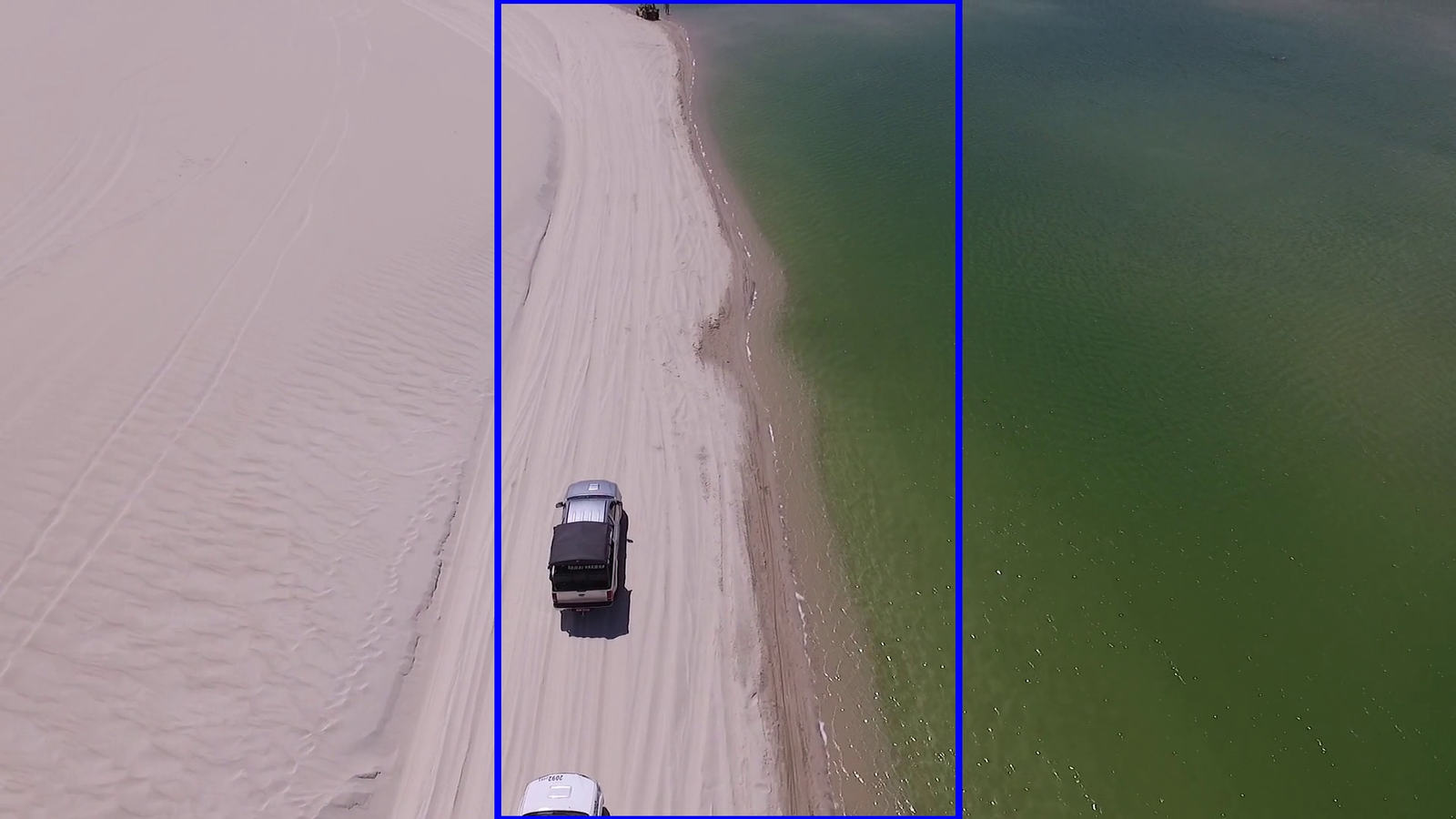} &
            \includegraphics[width=\fw]{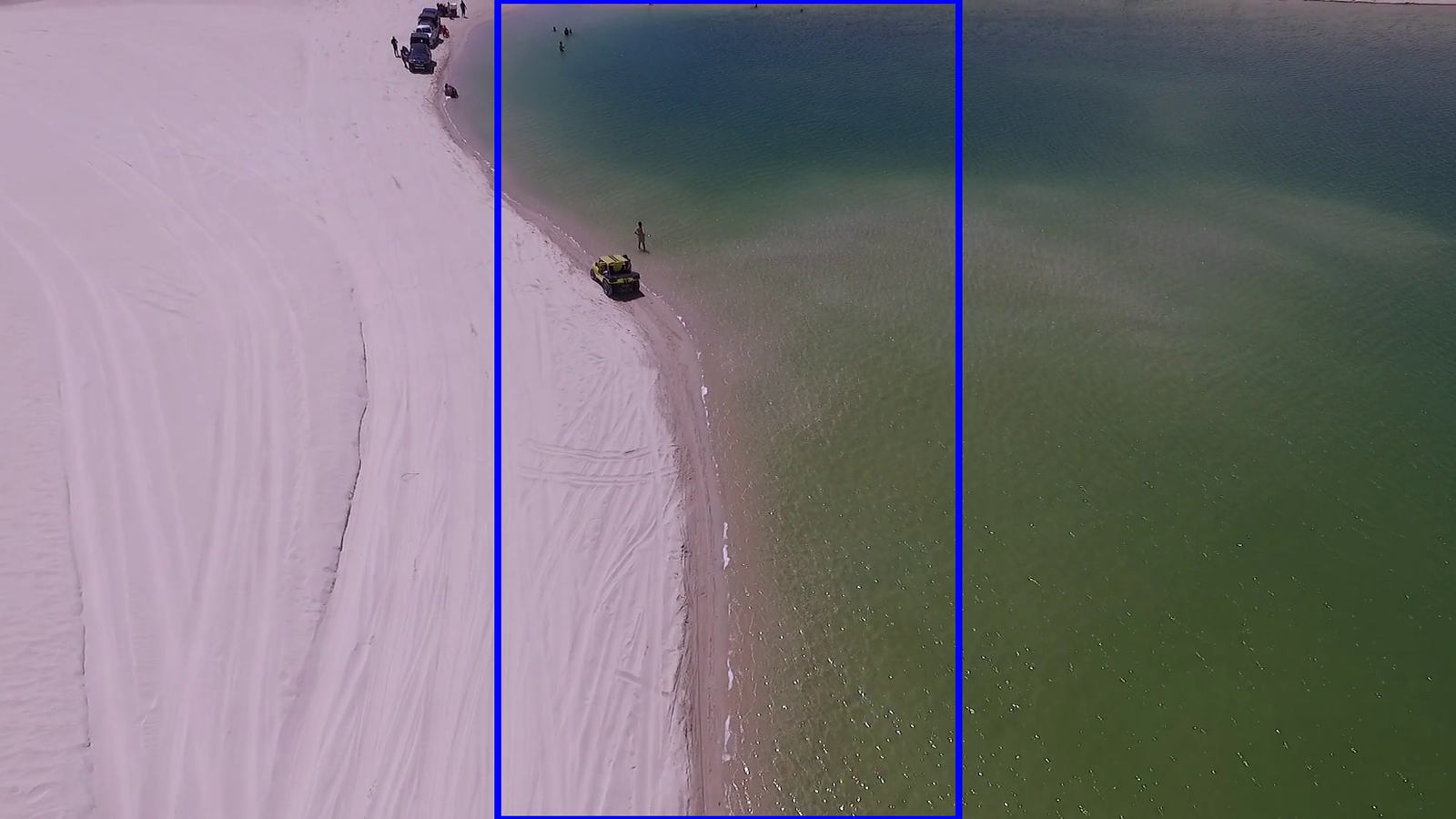} &
            \includegraphics[width=\fw]{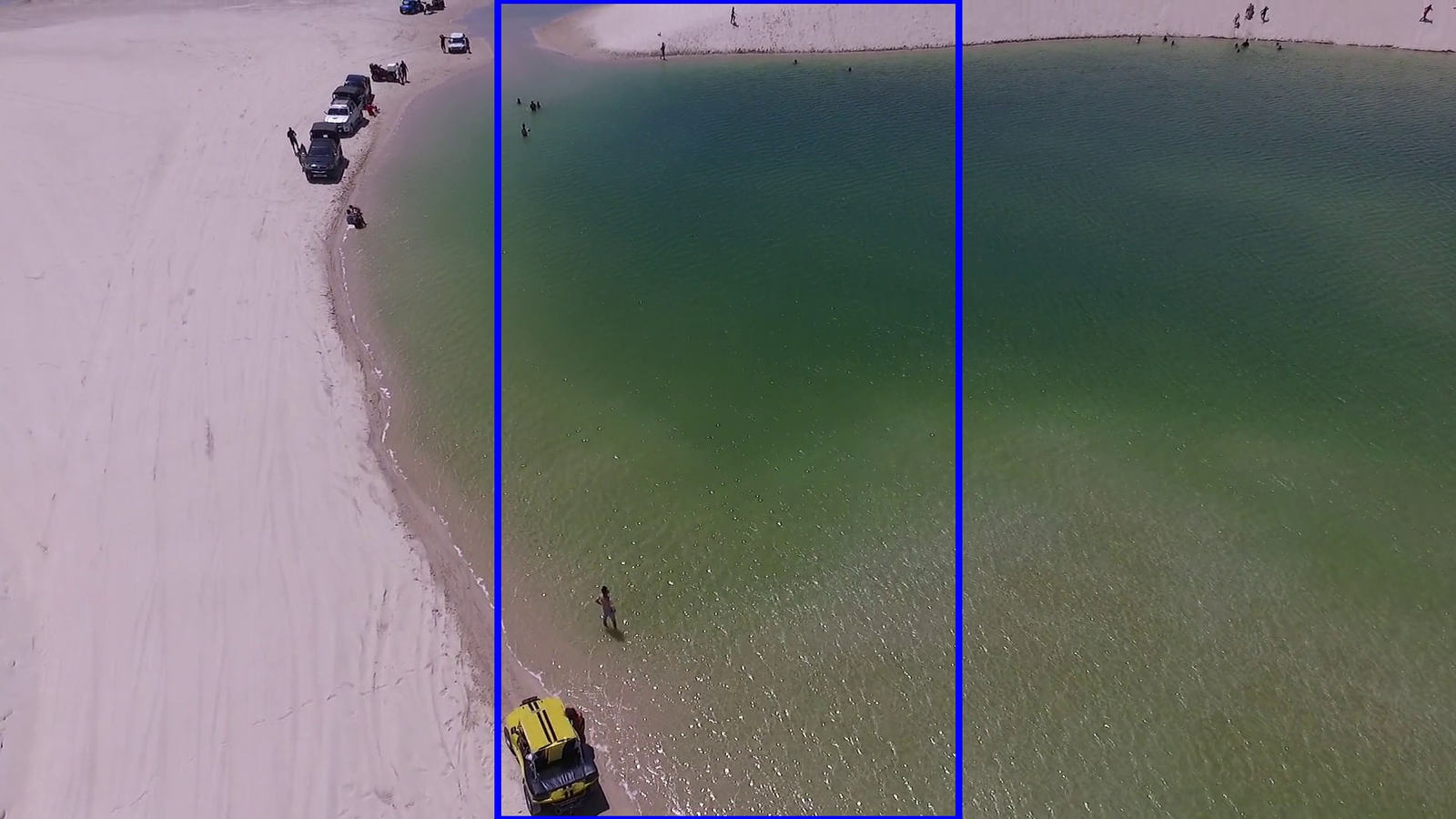} \\
            \includegraphics[width=\fw]{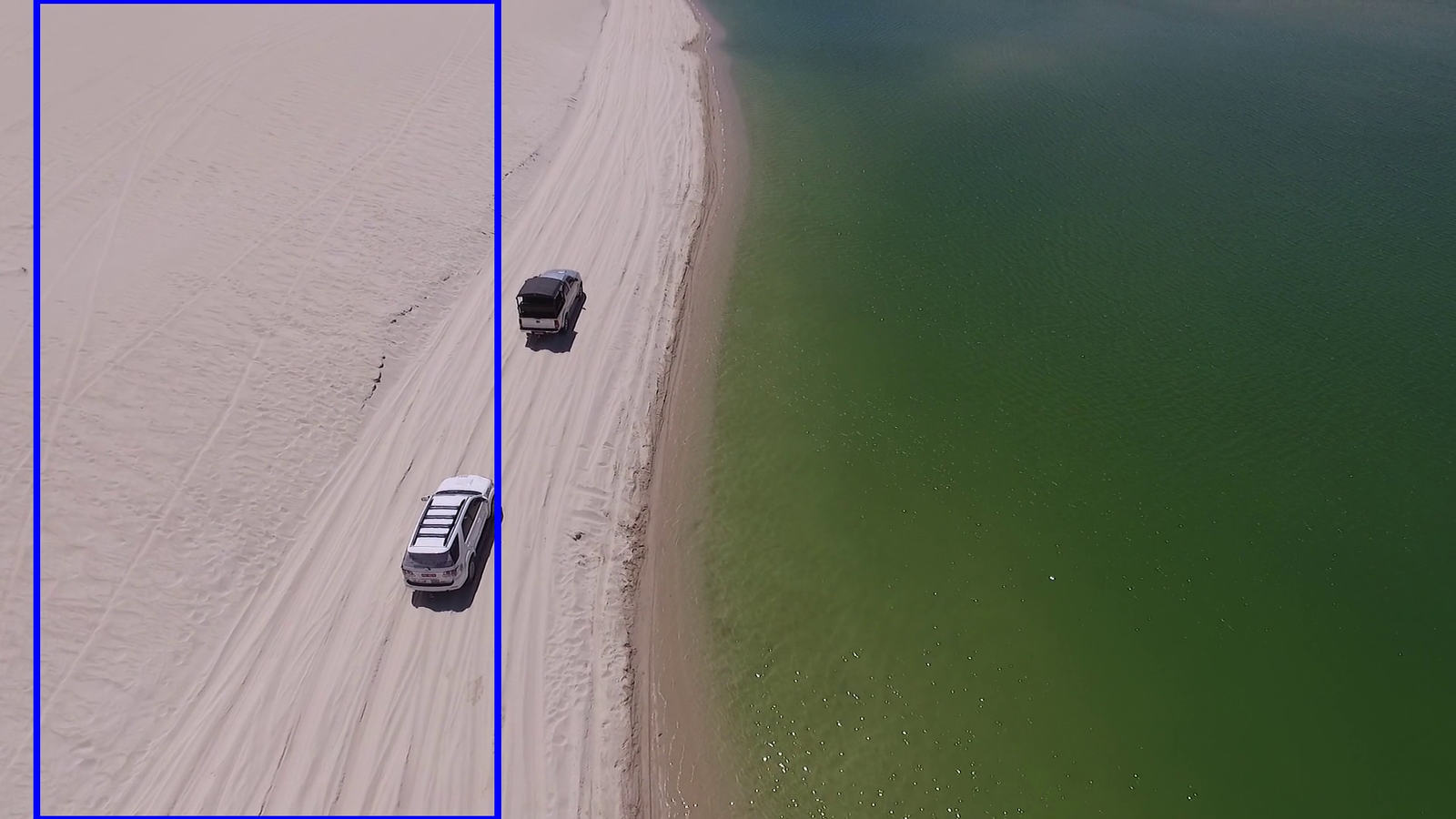} &
            \includegraphics[width=\fw]{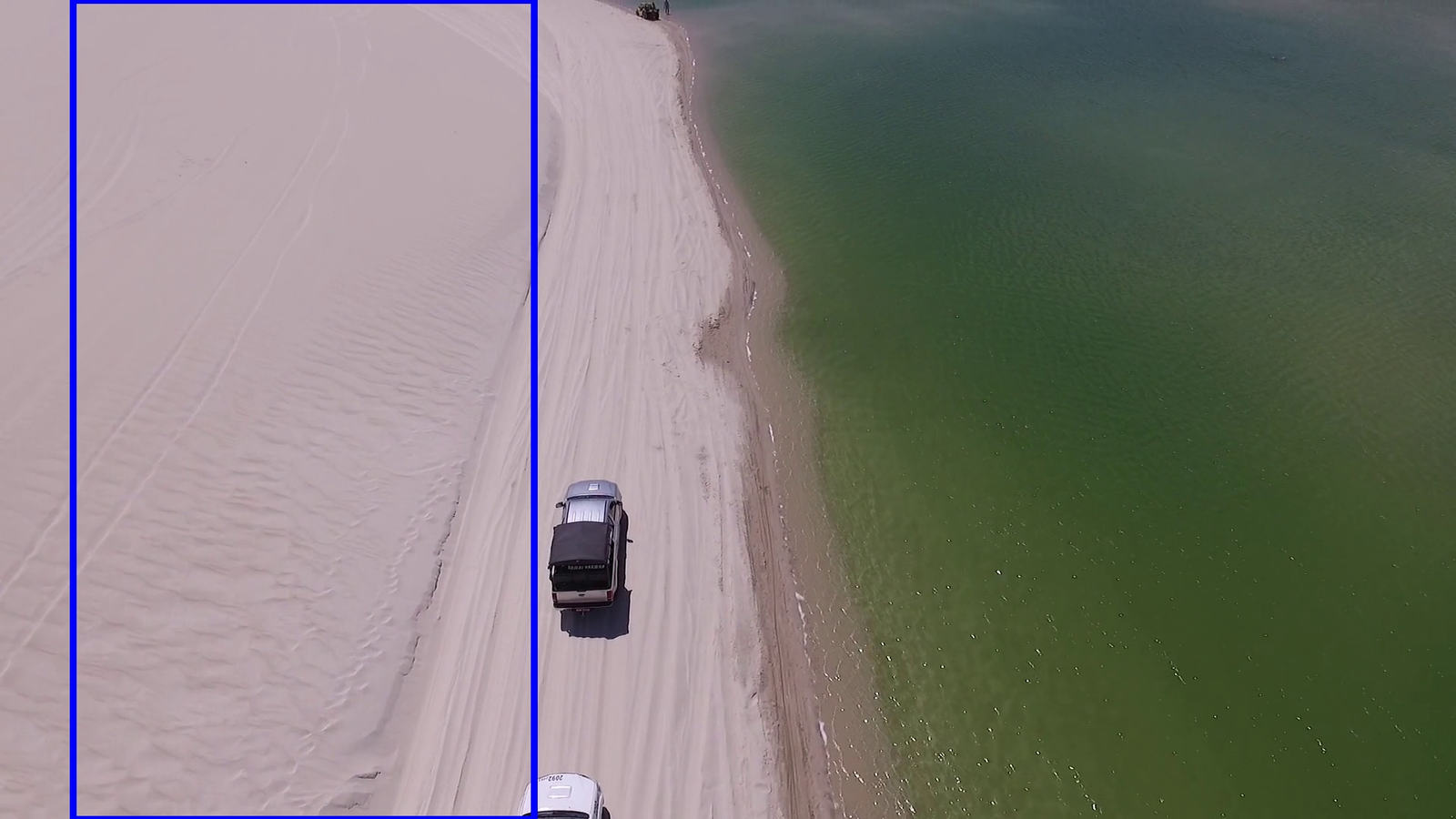} &
            \includegraphics[width=\fw]{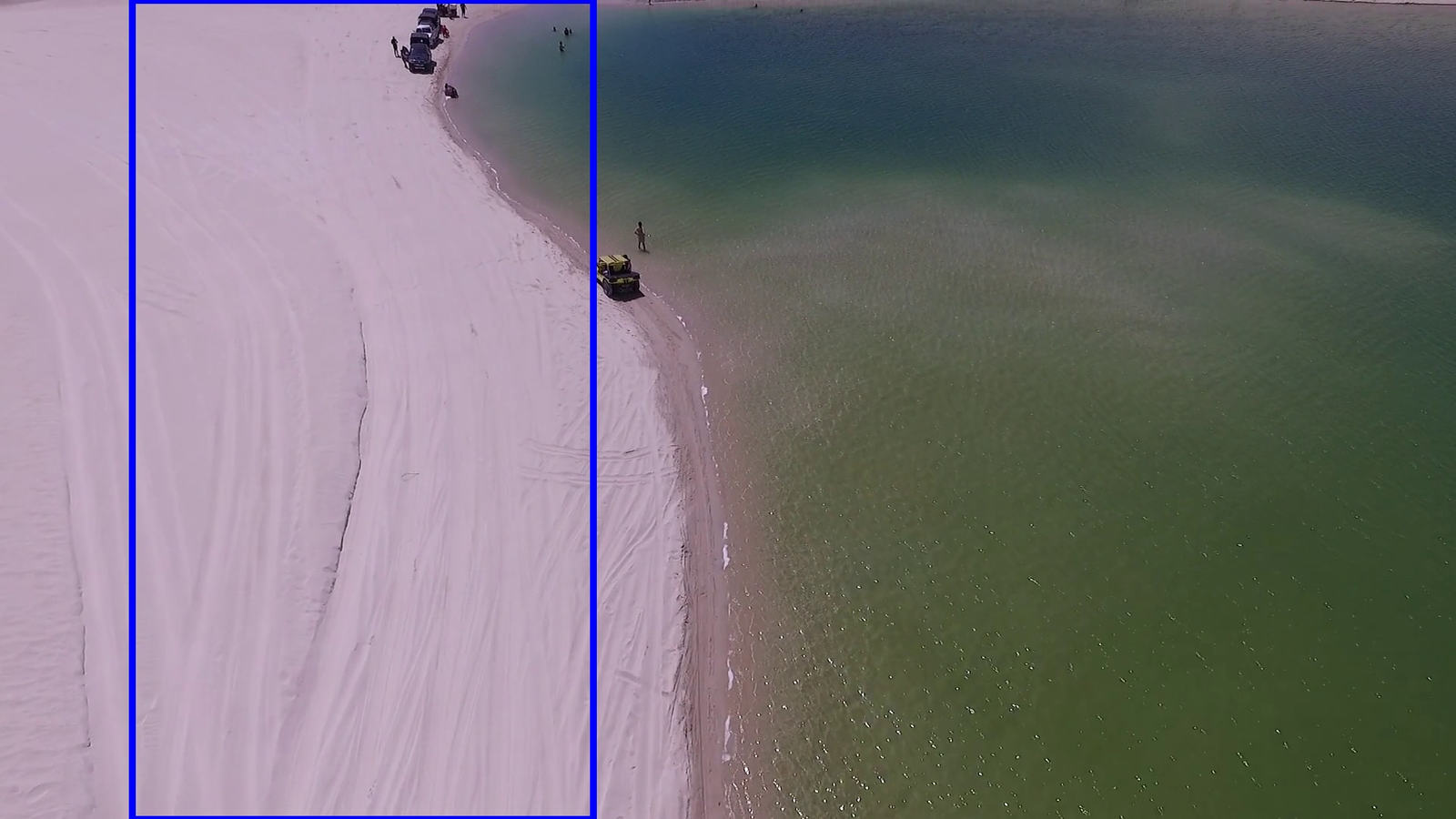} &
            \includegraphics[width=\fw]{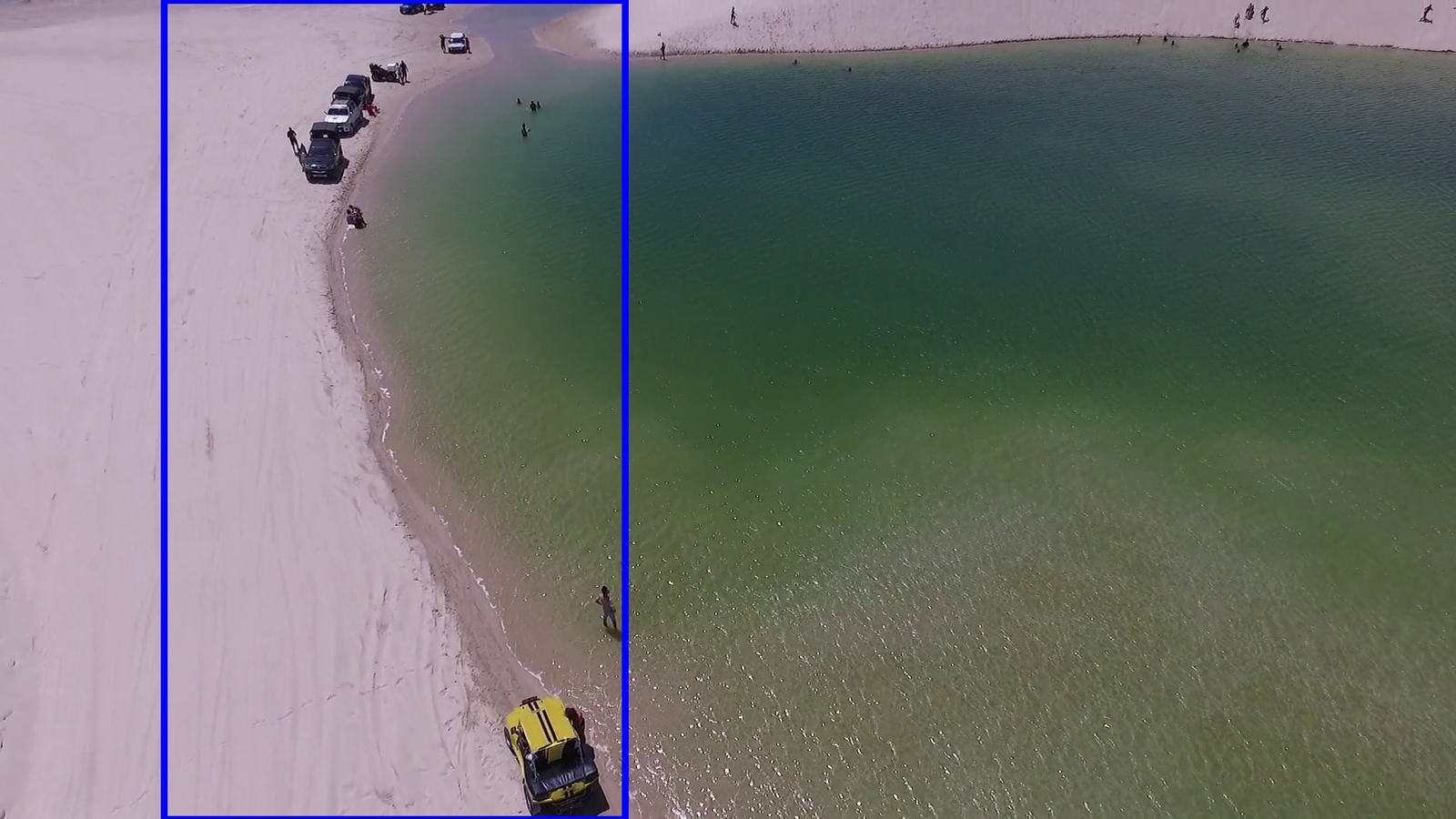} \\
            \includegraphics[width=\fw]{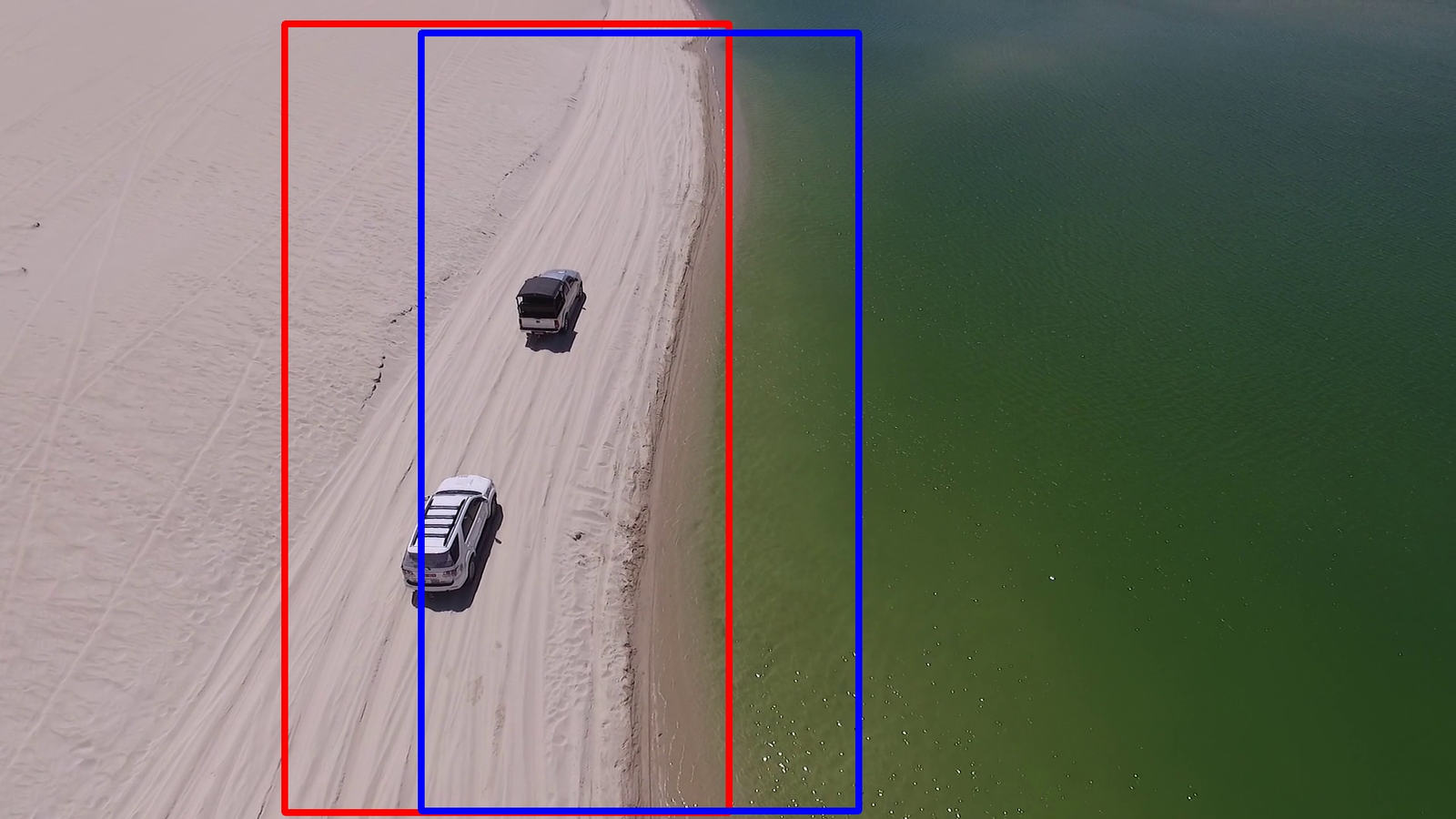} &
            \includegraphics[width=\fw]{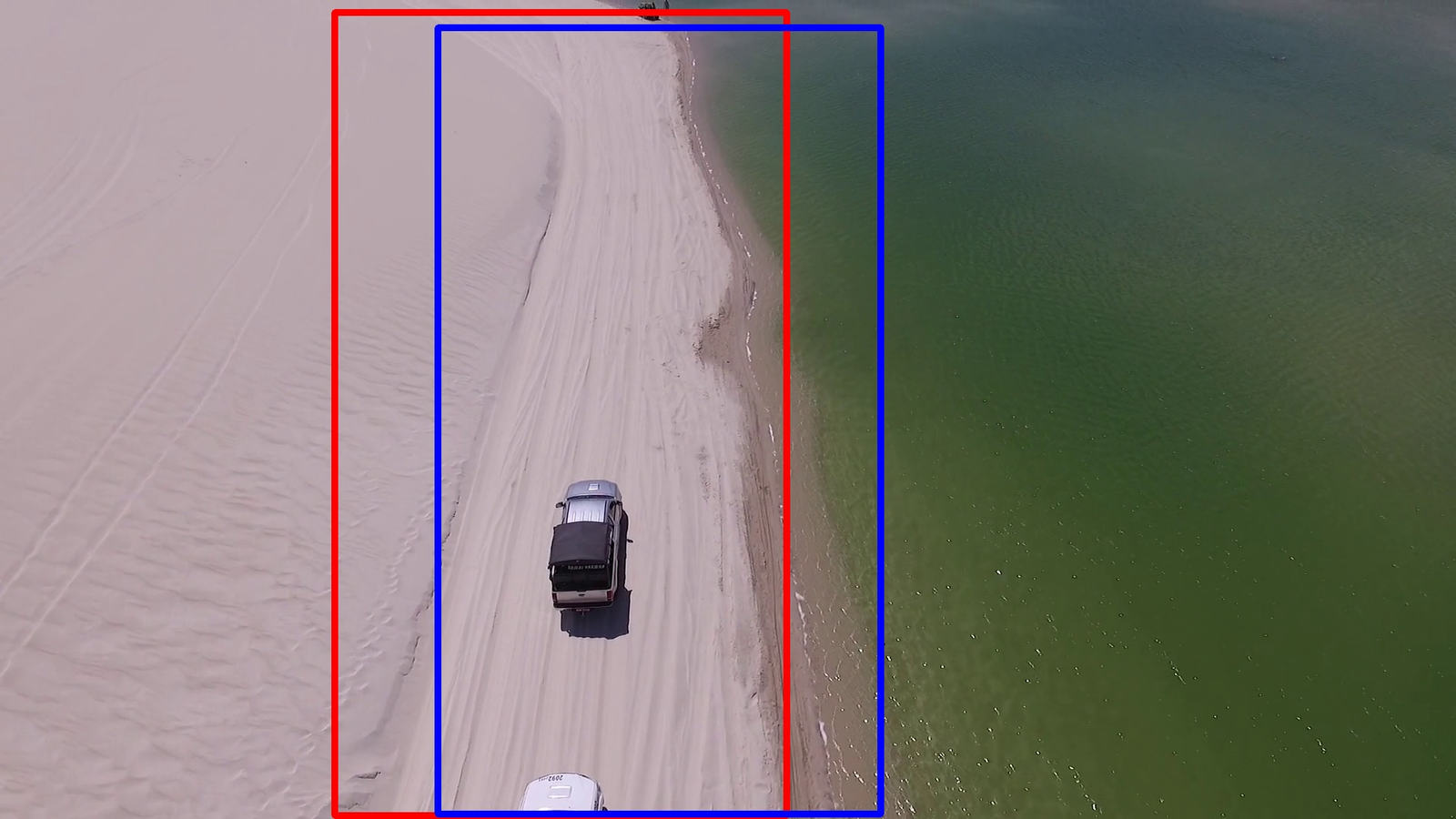} &
            \includegraphics[width=\fw]{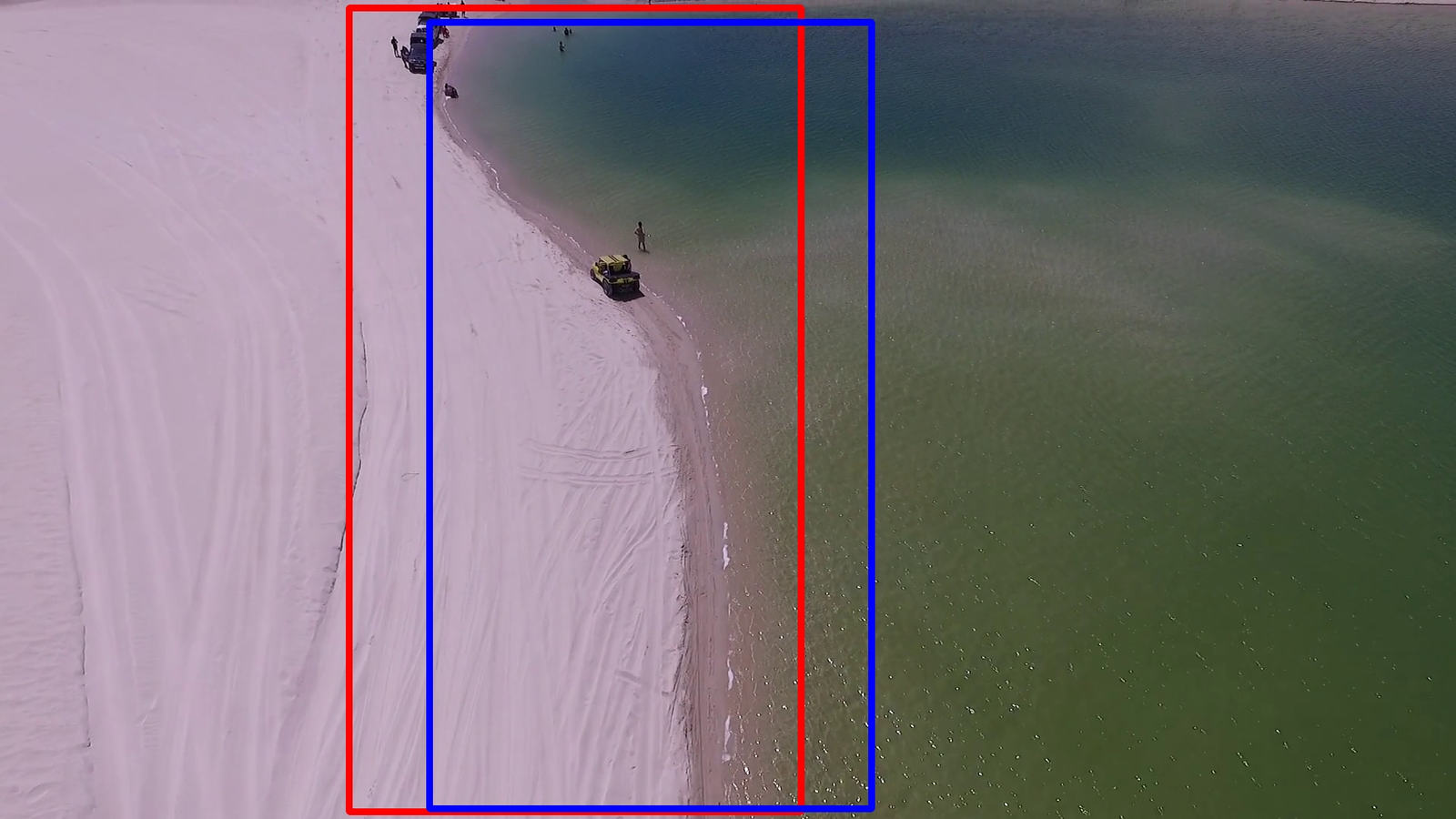} &
            \includegraphics[width=\fw]{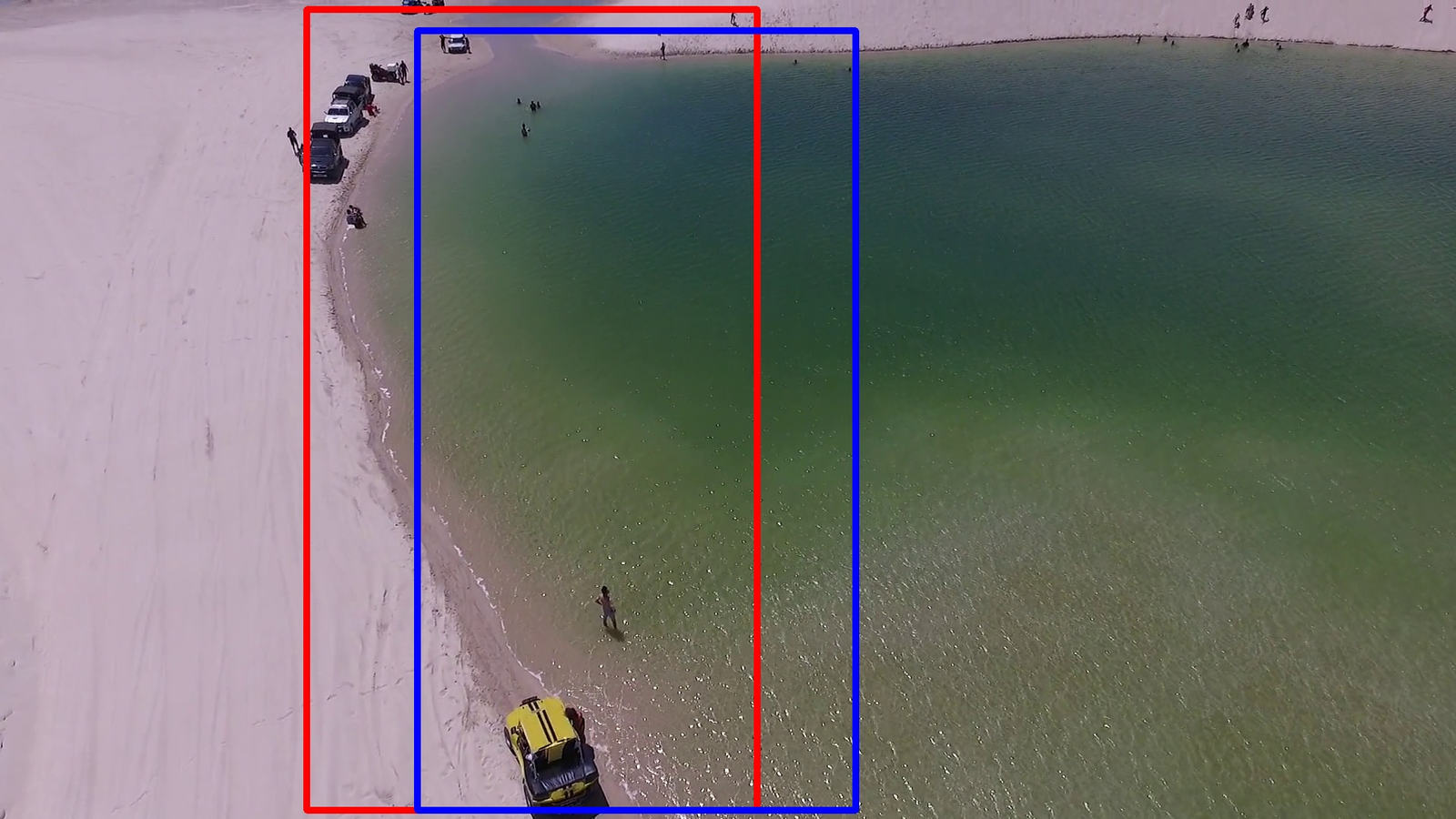} \\
        \end{tabular}
        \caption{Video ID: 4005402}
    \end{subfigure}
    \hfill
    \begin{subfigure}{0.32\textwidth}
        \centering
        \begin{tabular}{cccc}
            \textbf{3s} & \textbf{9s} & \textbf{15s} & \textbf{18s} \\
            \includegraphics[width=\fw]{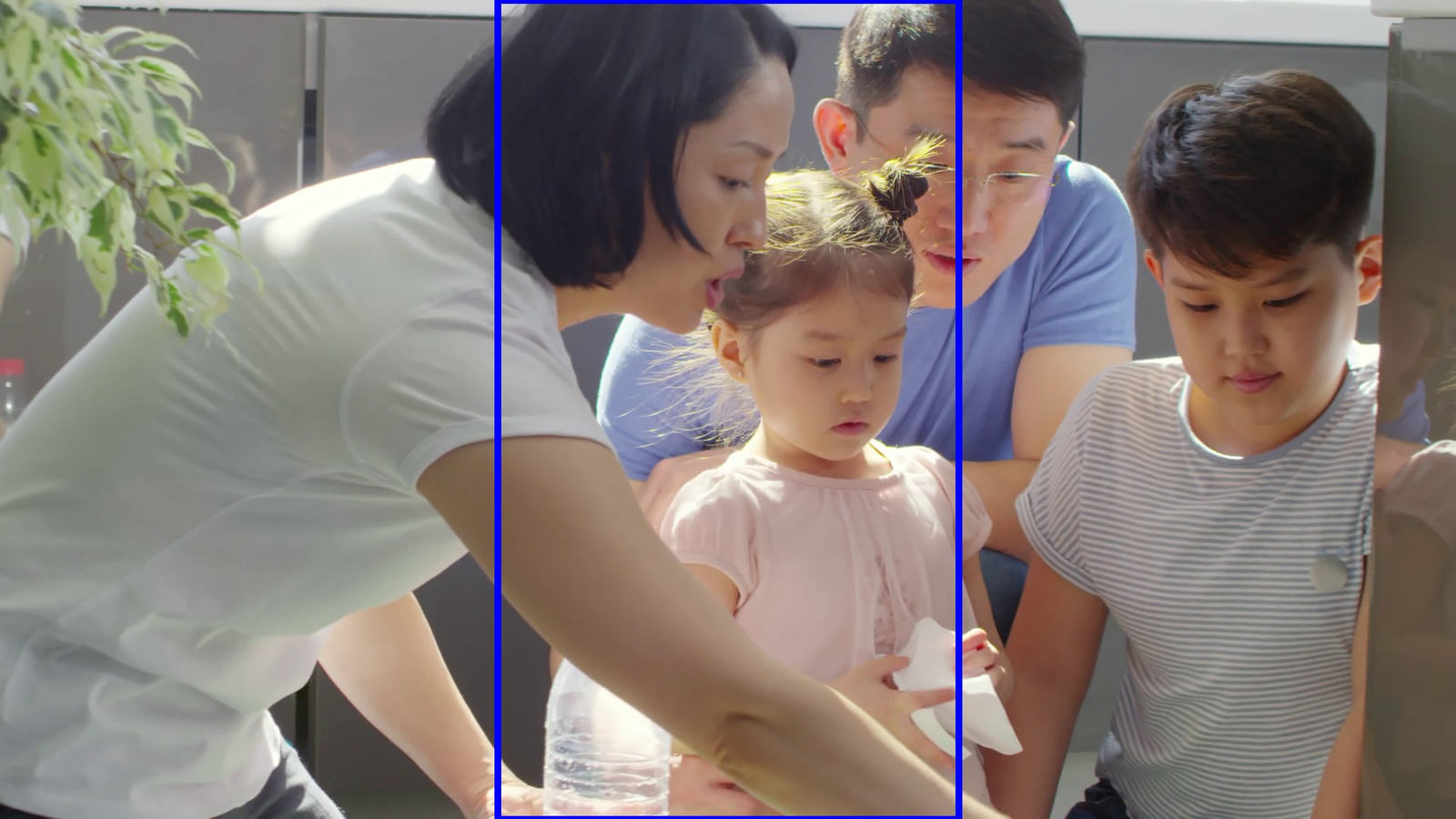} &
            \includegraphics[width=\fw]{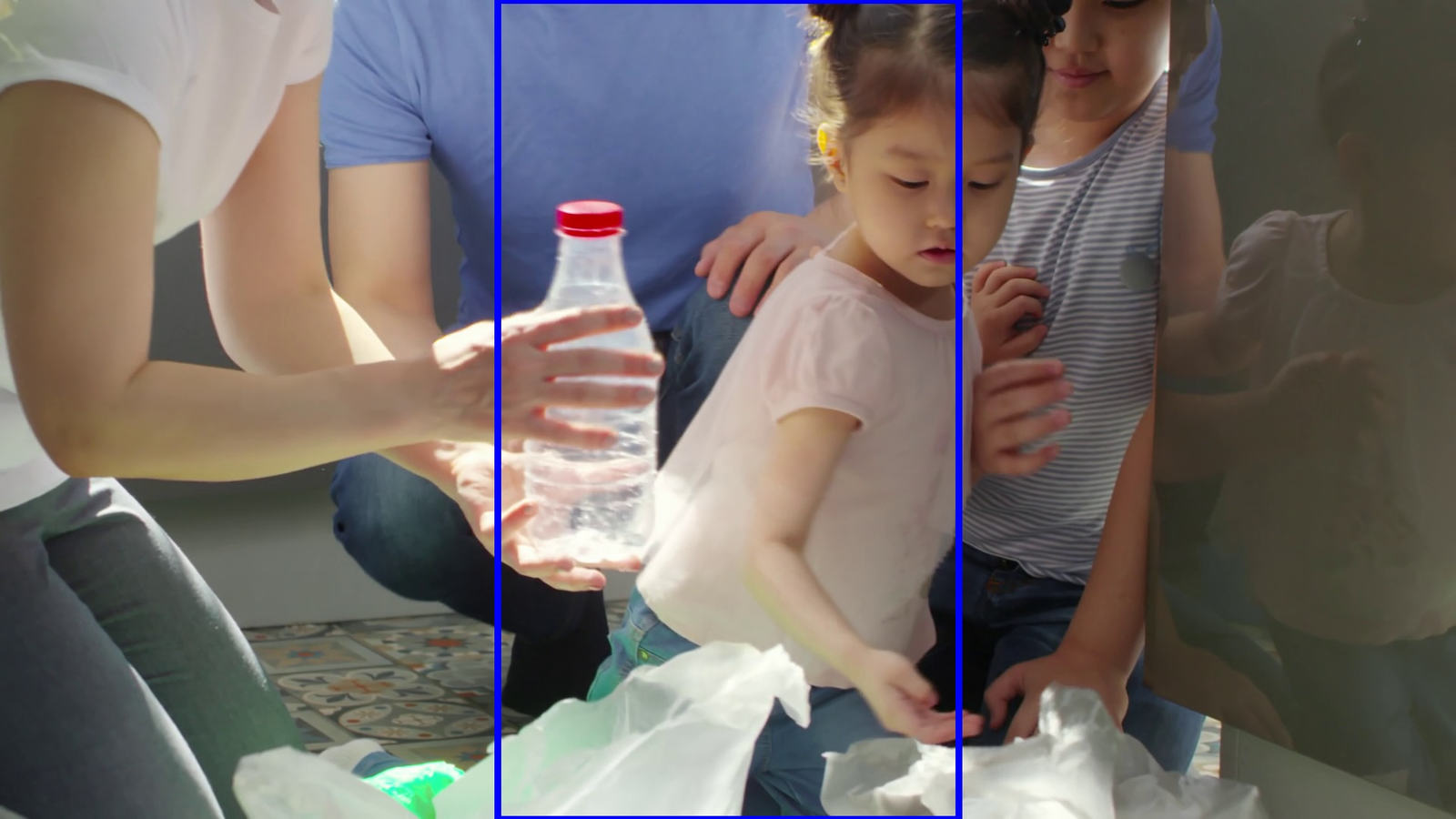} &
            \includegraphics[width=\fw]{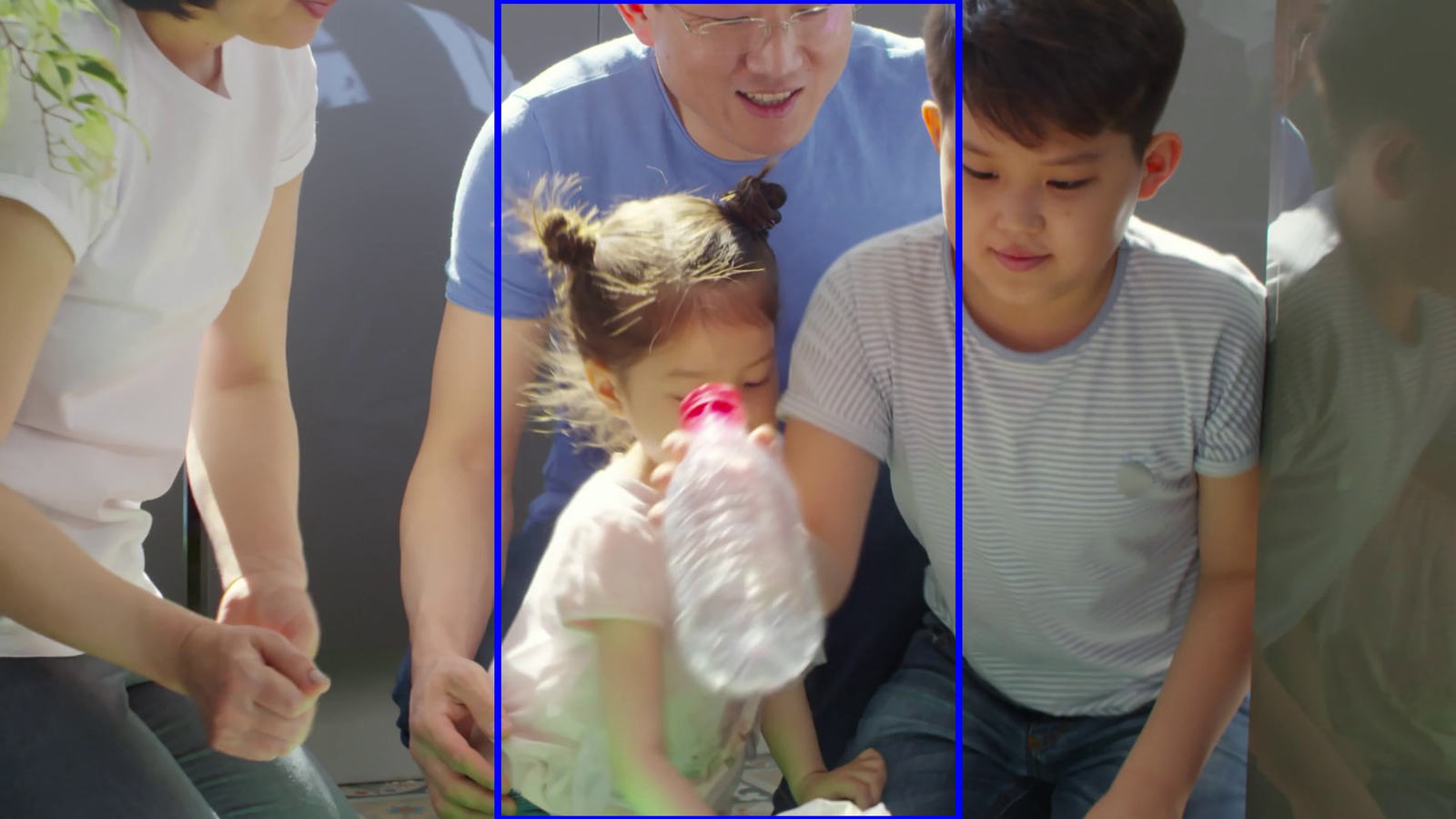} &
            \includegraphics[width=\fw]{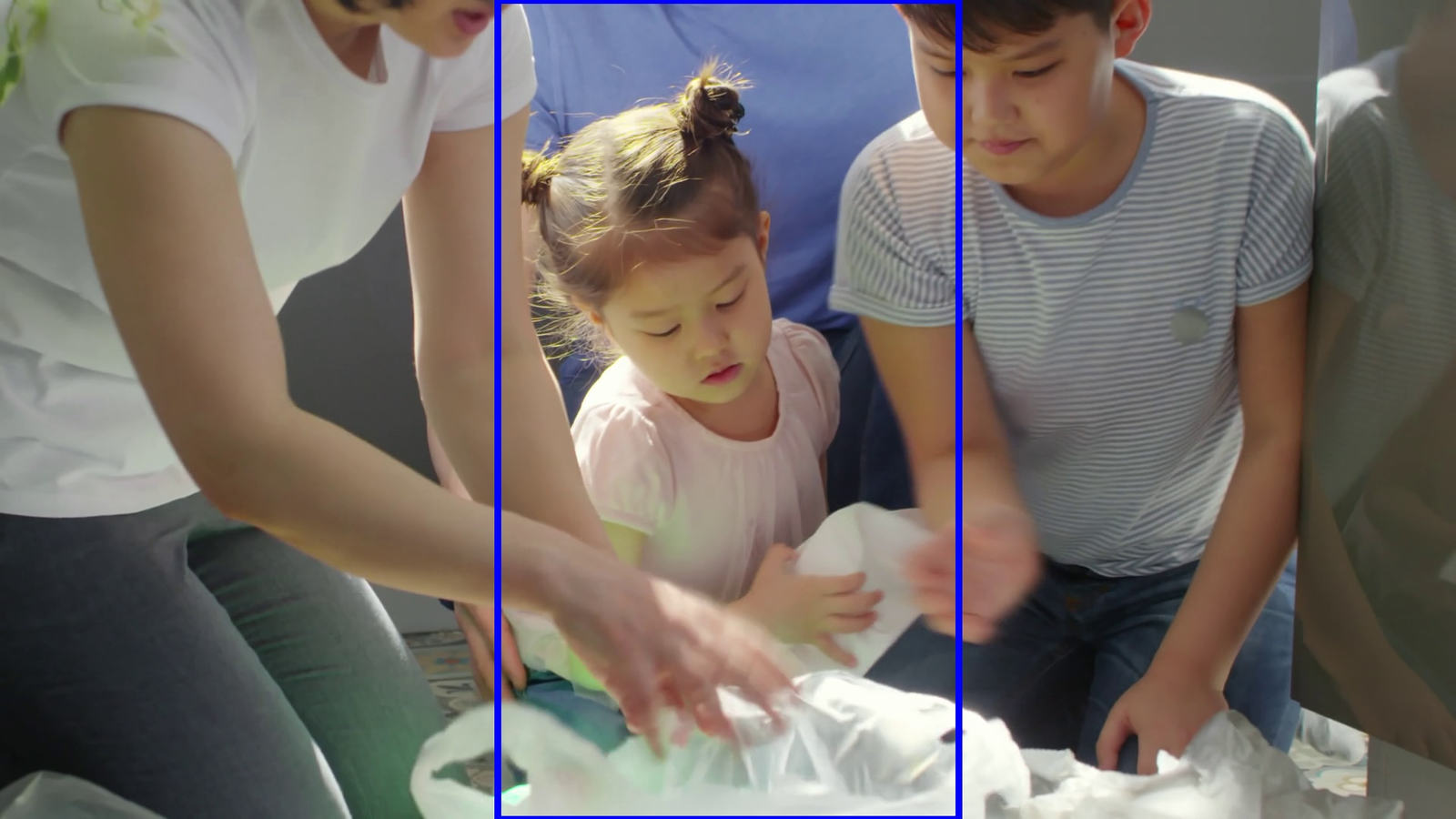} \\
            \includegraphics[width=\fw]{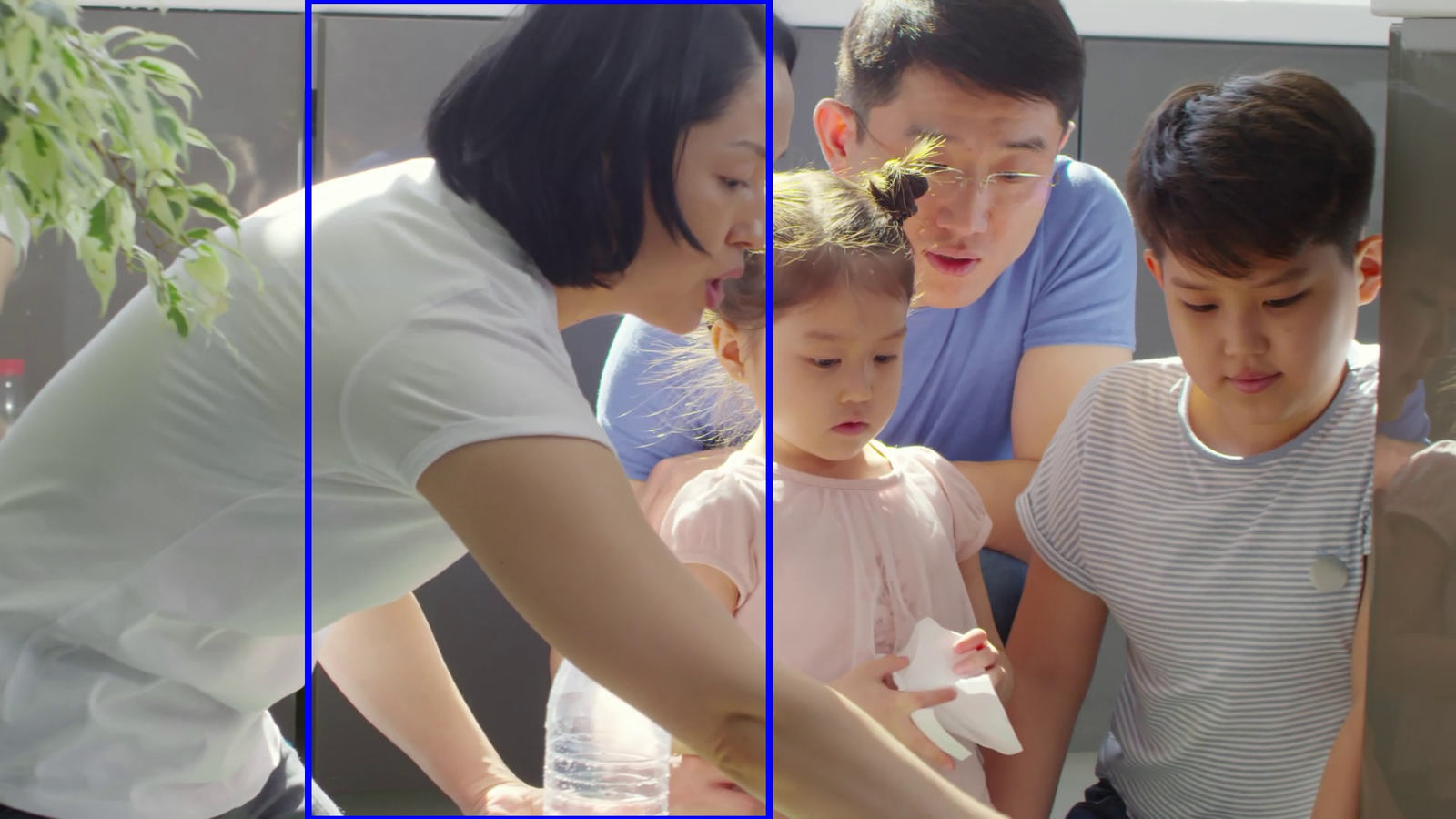} &
            \includegraphics[width=\fw]{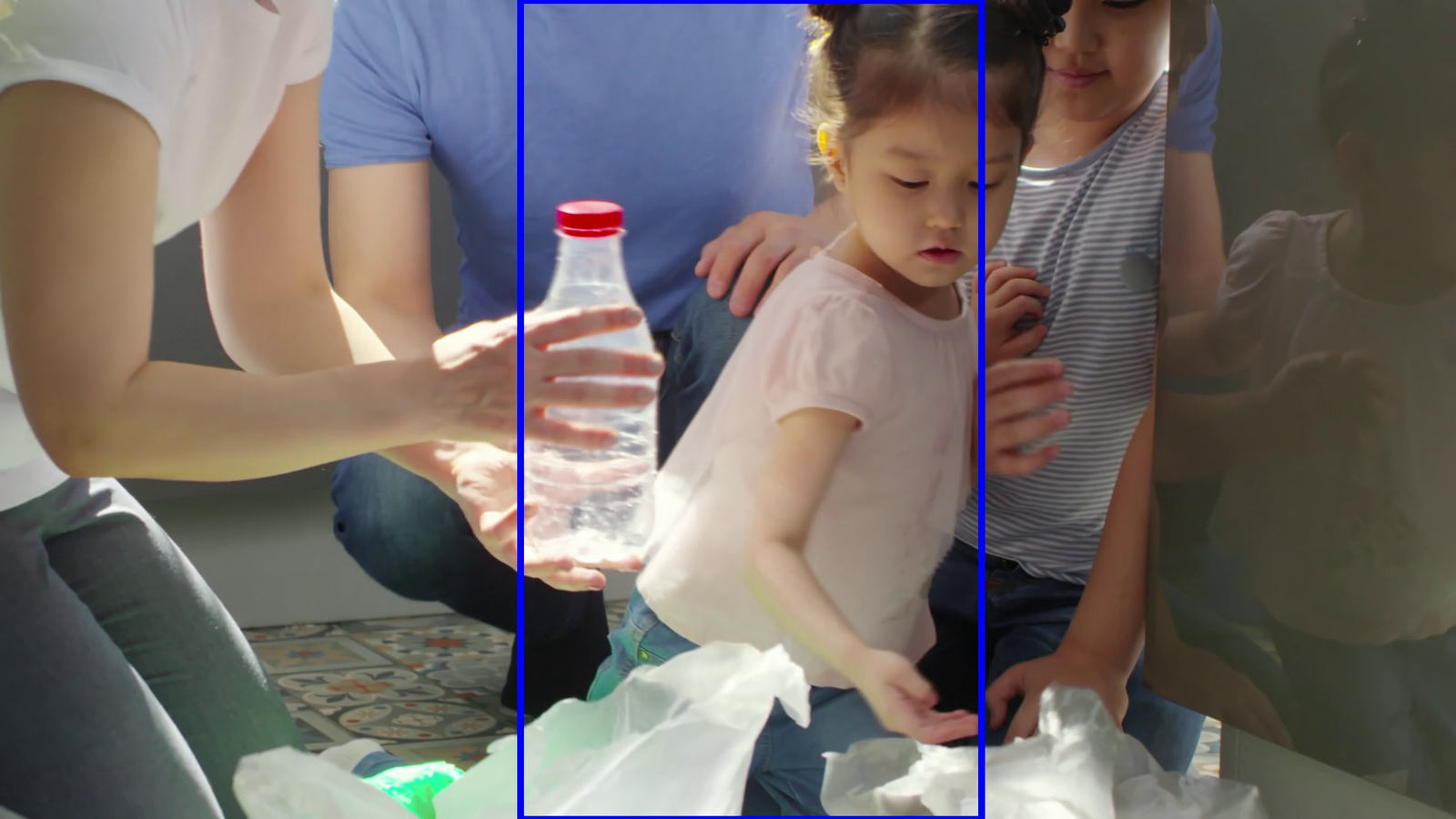} &
            \includegraphics[width=\fw]{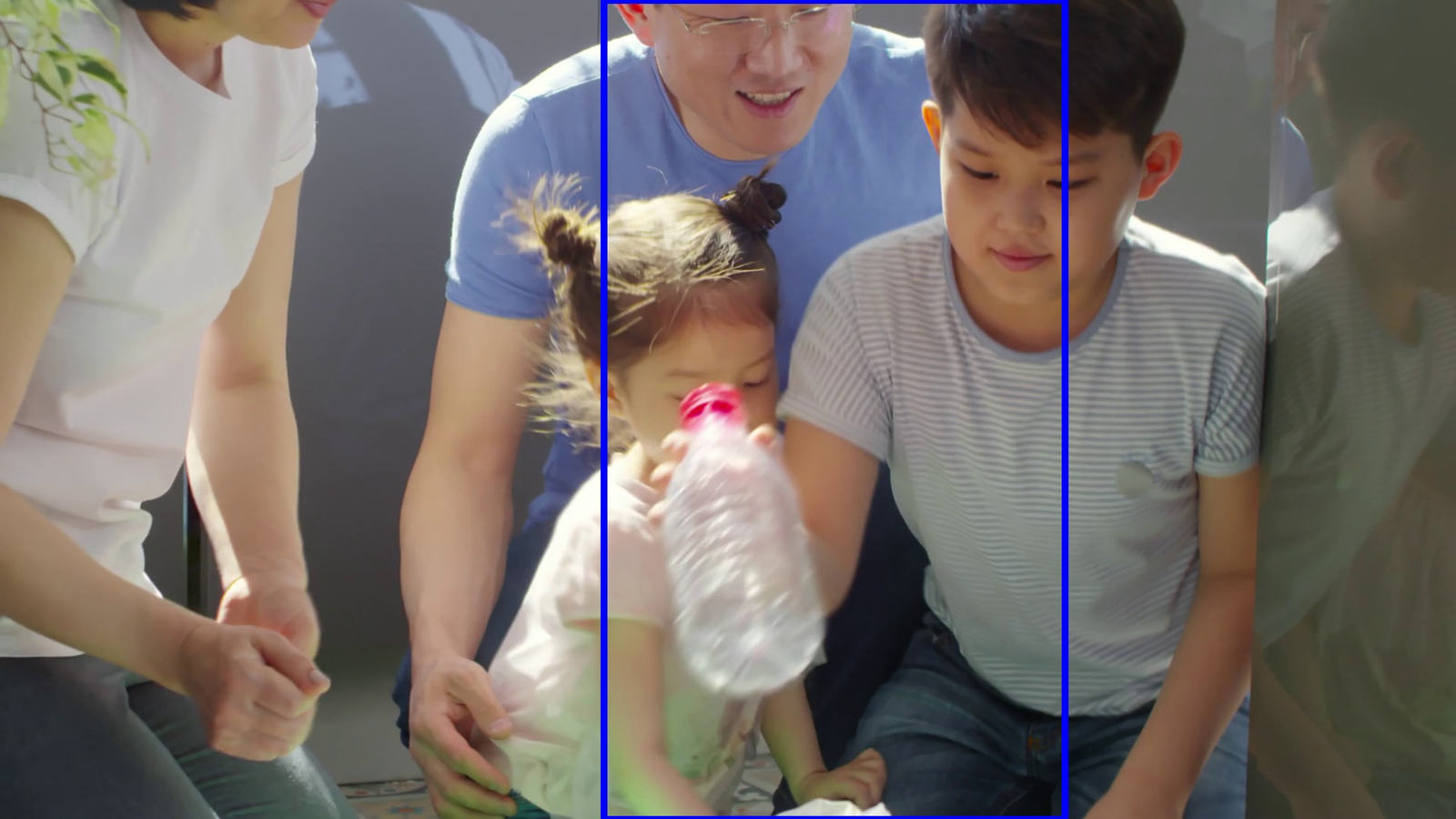} &
            \includegraphics[width=\fw]{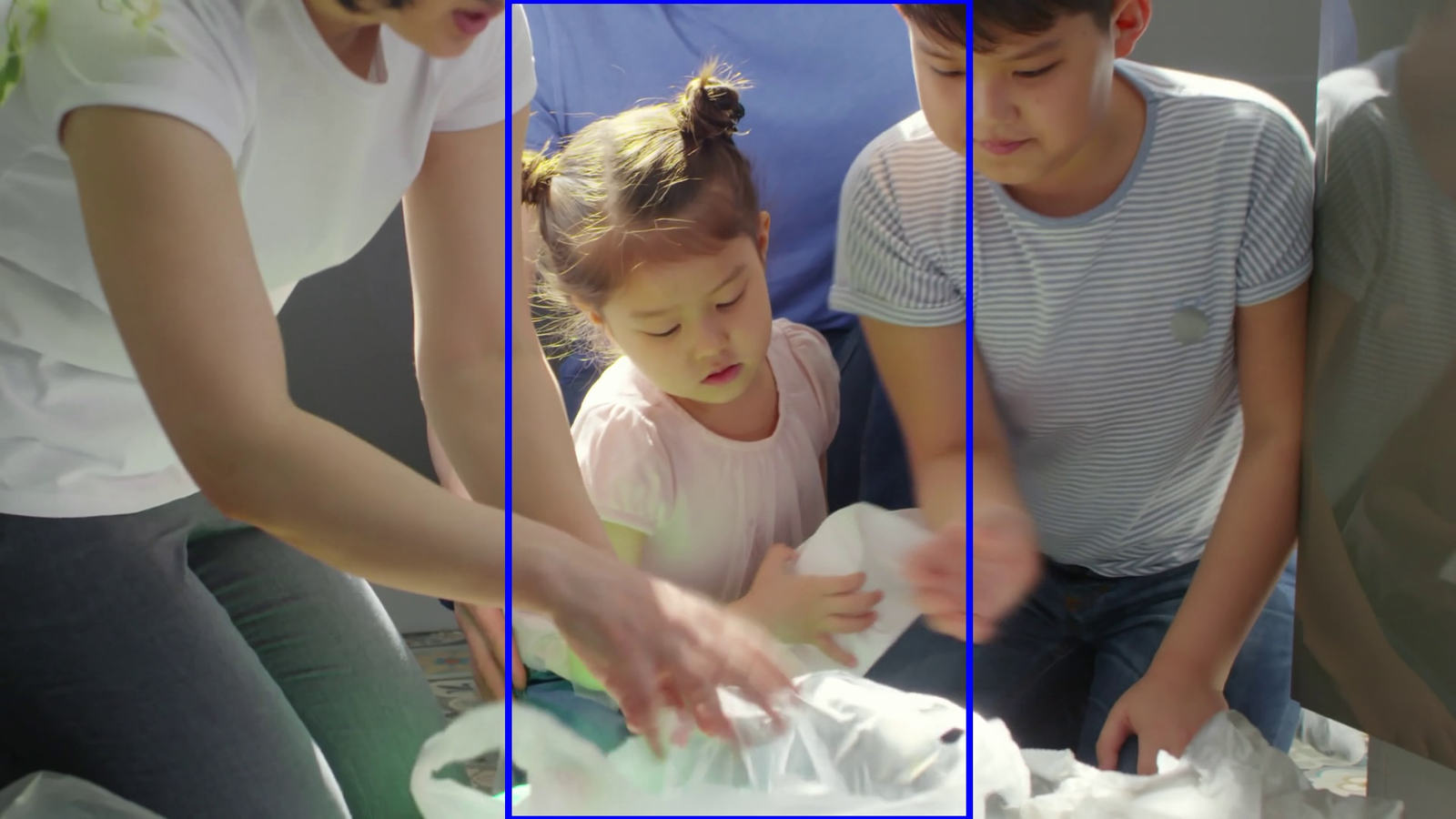} \\
            \includegraphics[width=\fw]{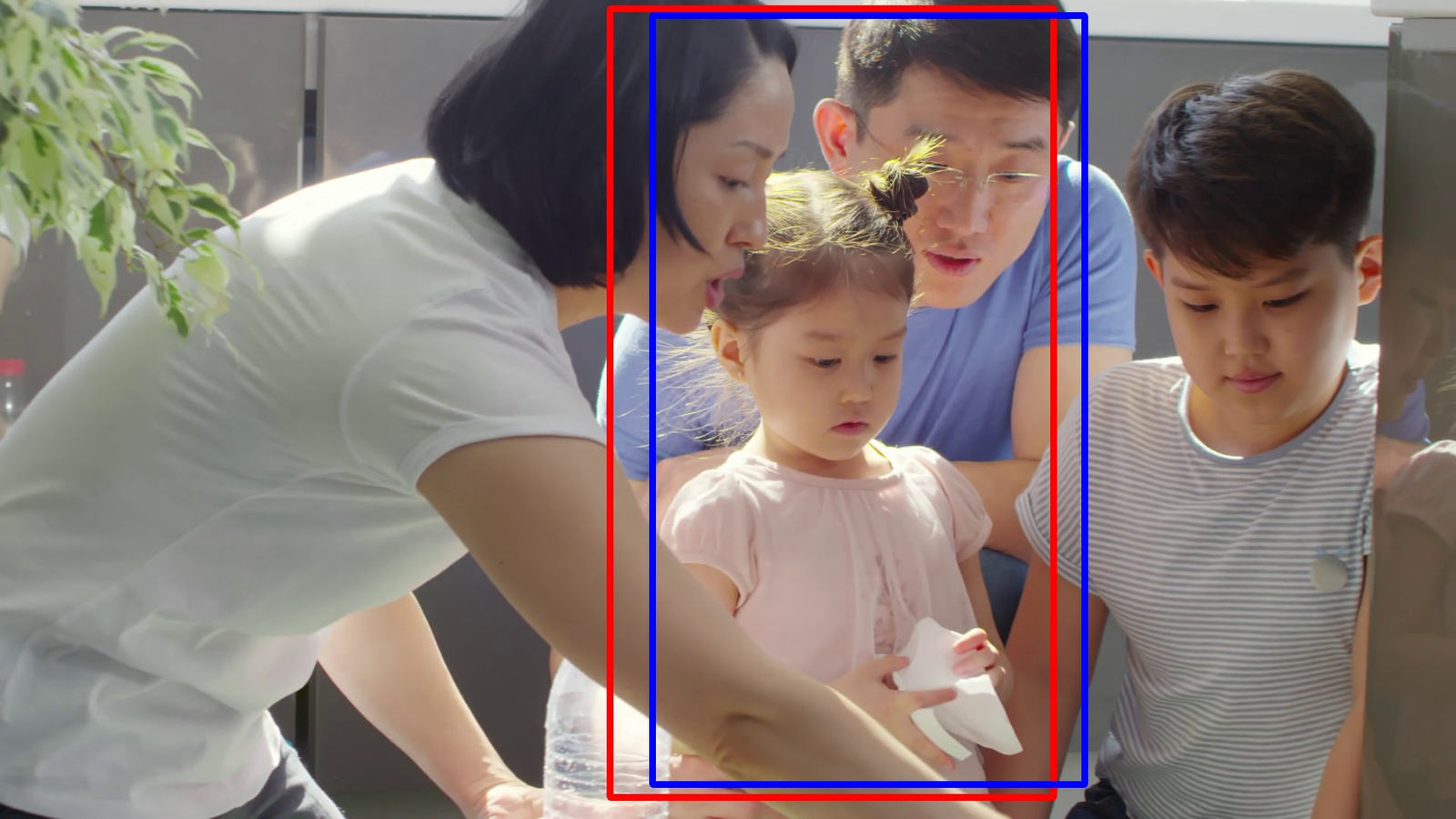} &
            \includegraphics[width=\fw]{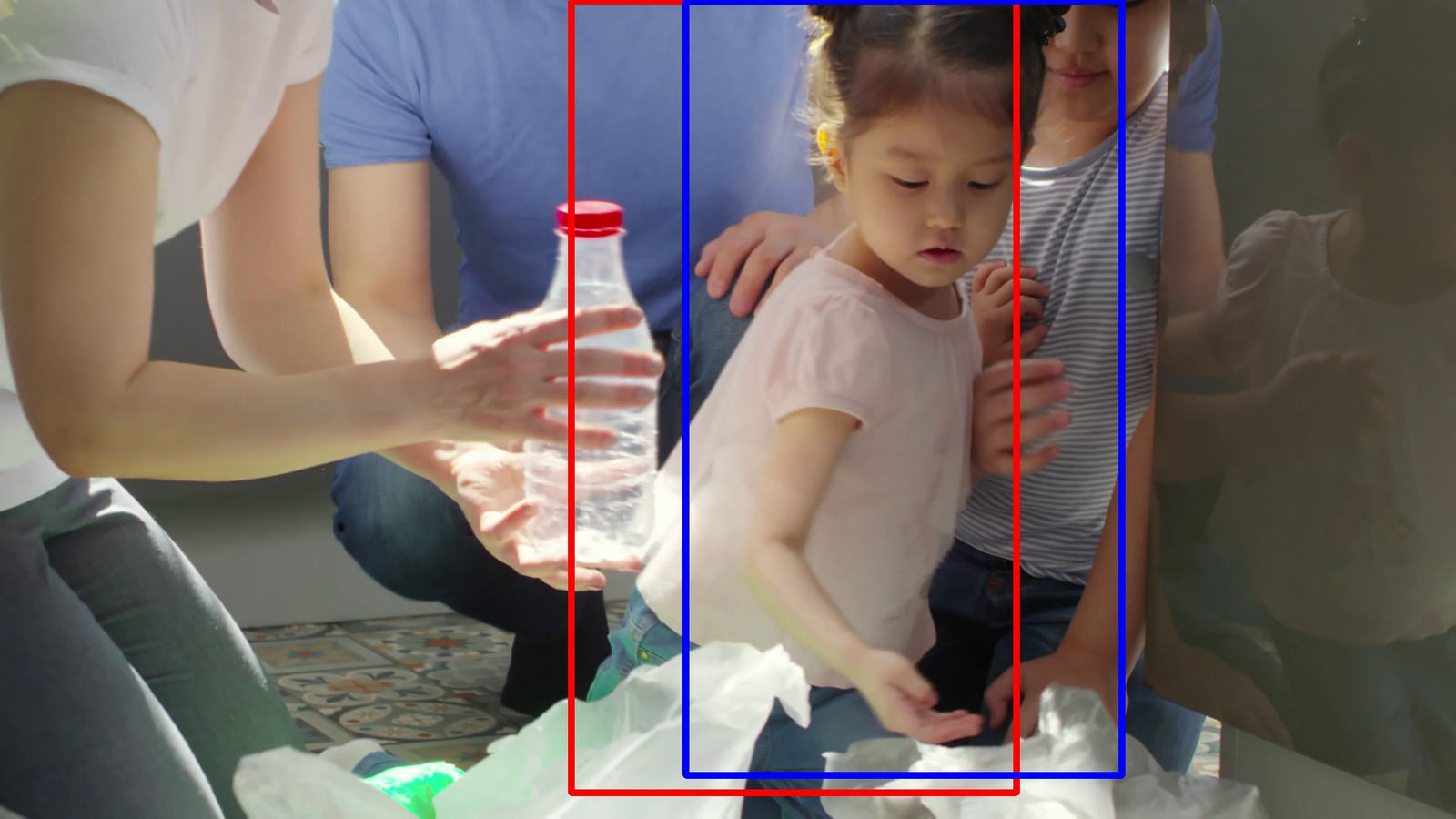} &
            \includegraphics[width=\fw]{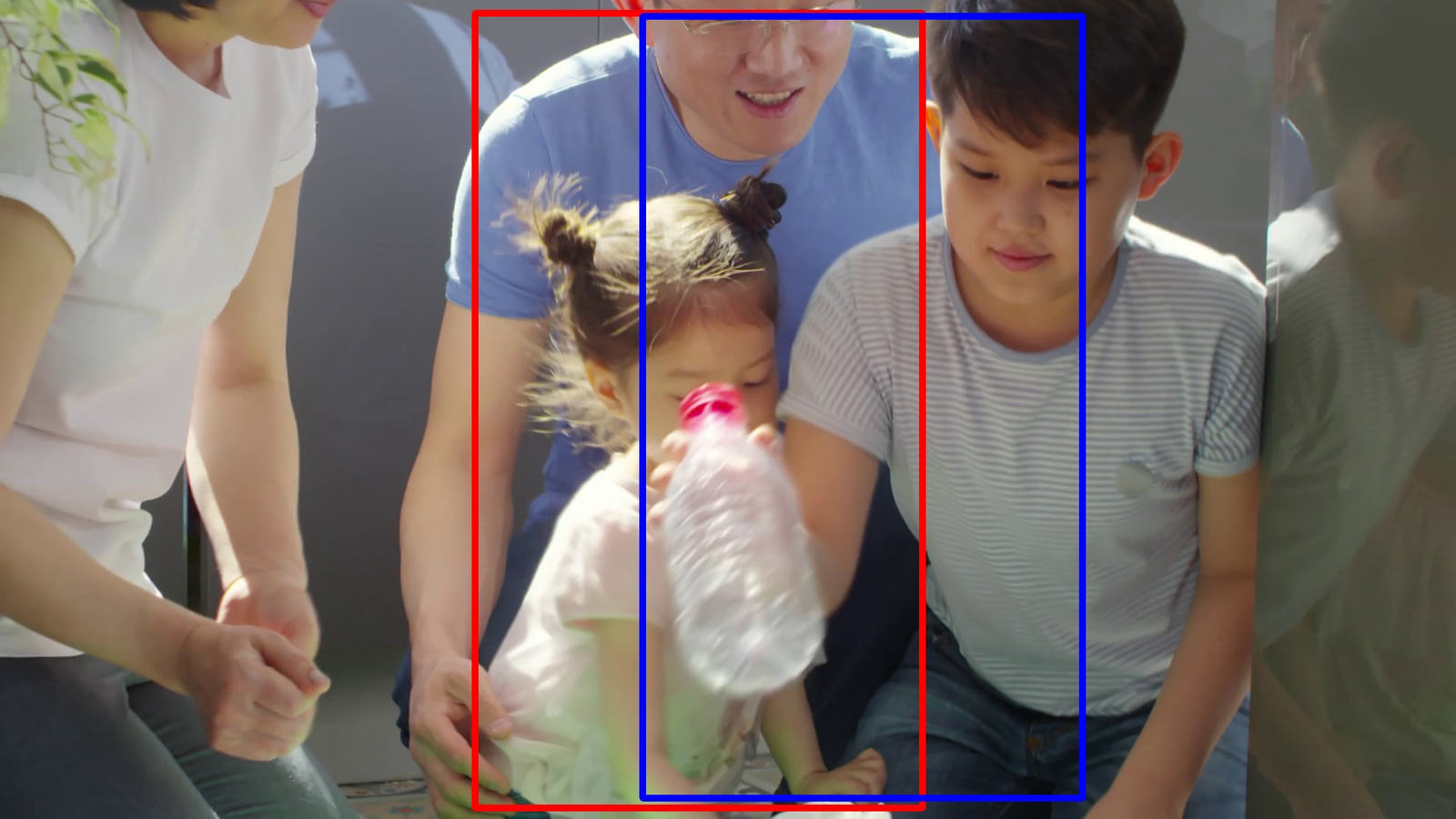} &
            \includegraphics[width=\fw]{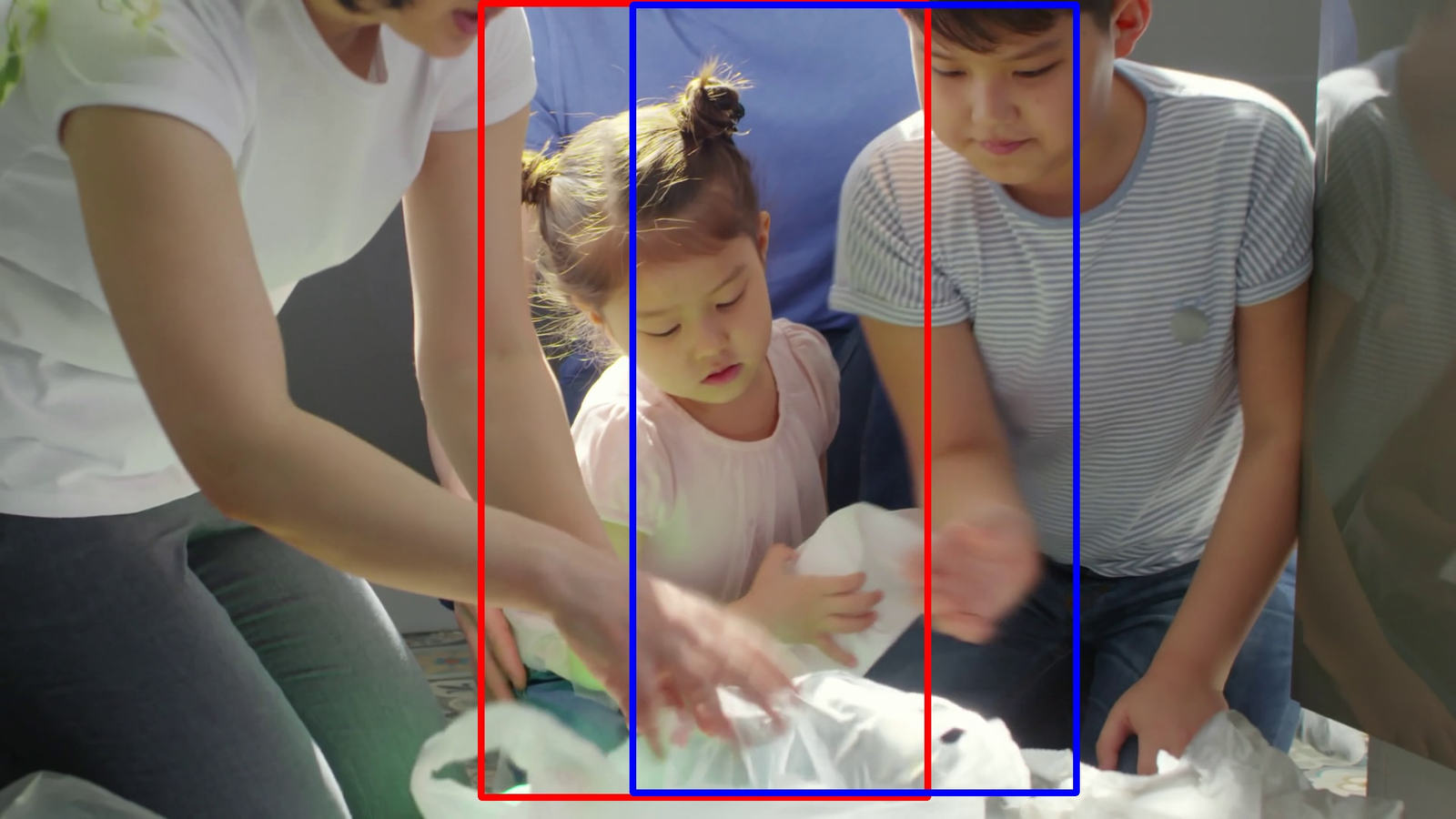} \\
        \end{tabular}
        \caption{Video ID: 4008756}
    \end{subfigure}
    \hfill
    \begin{subfigure}{0.32\textwidth}
        \centering
        \begin{tabular}{cccc}
            \textbf{8s} & \textbf{12s} & \textbf{16s} & \textbf{20s} \\
            \includegraphics[width=\fw]{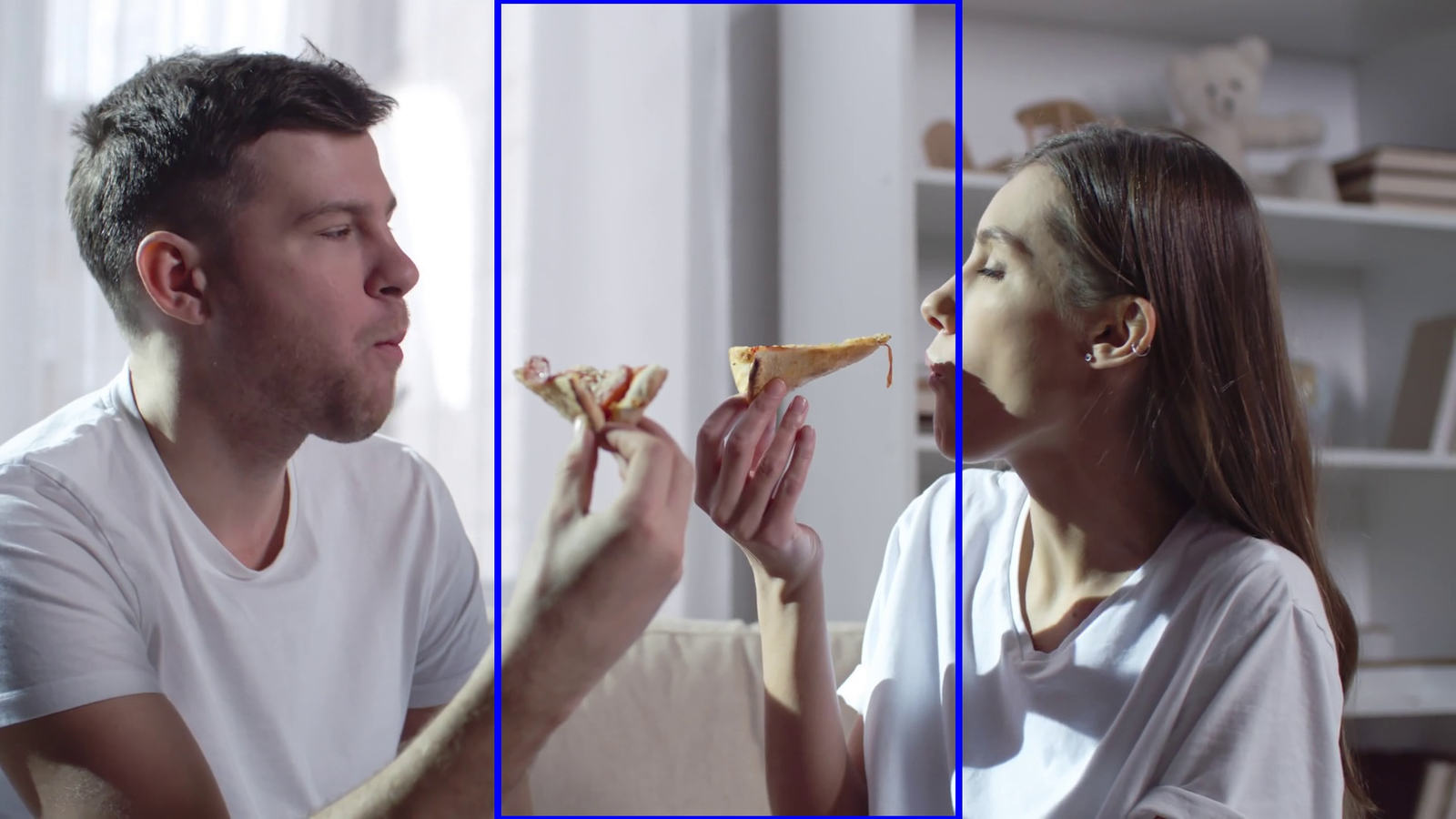} &
            \includegraphics[width=\fw]{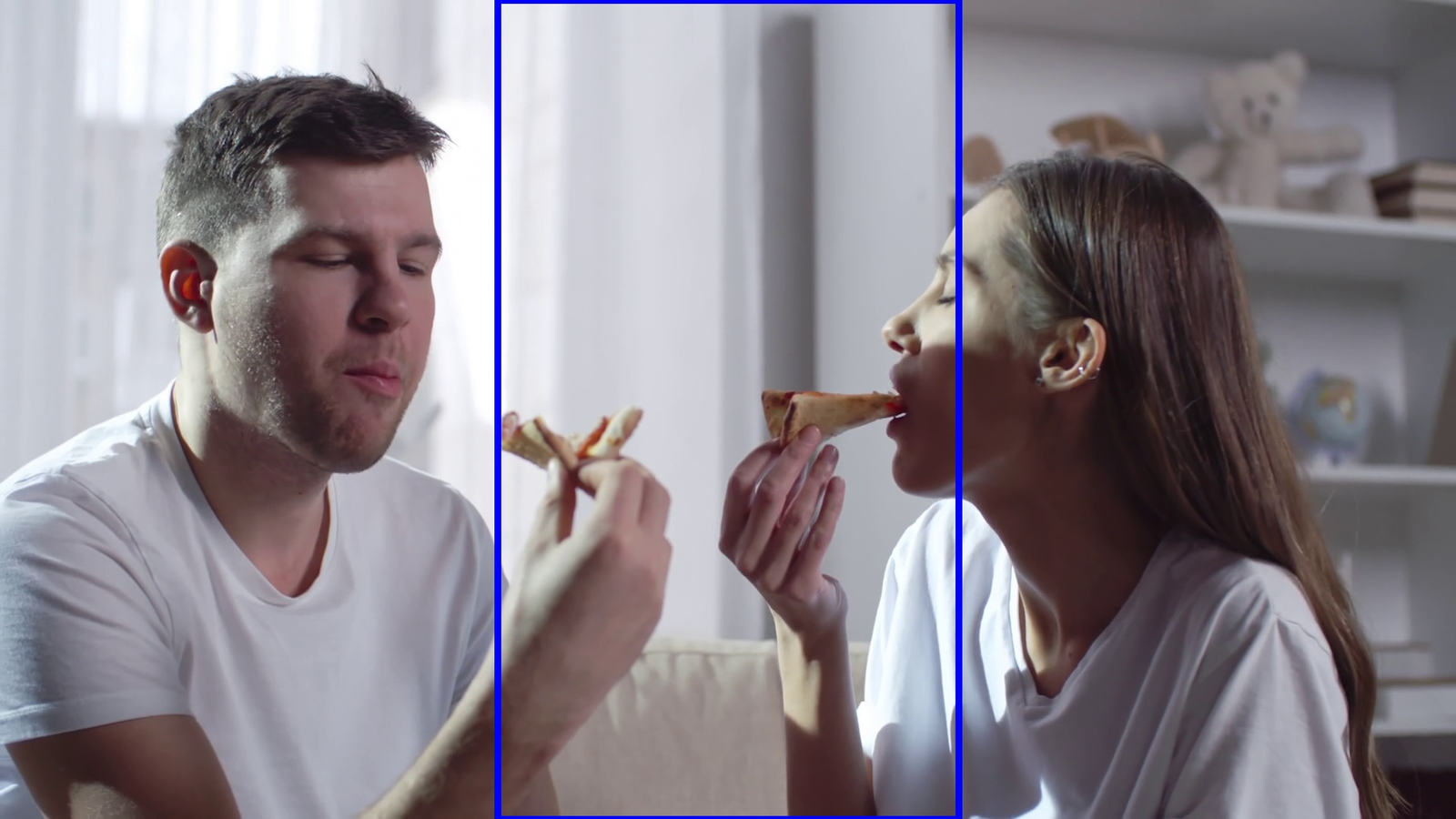} &
            \includegraphics[width=\fw]{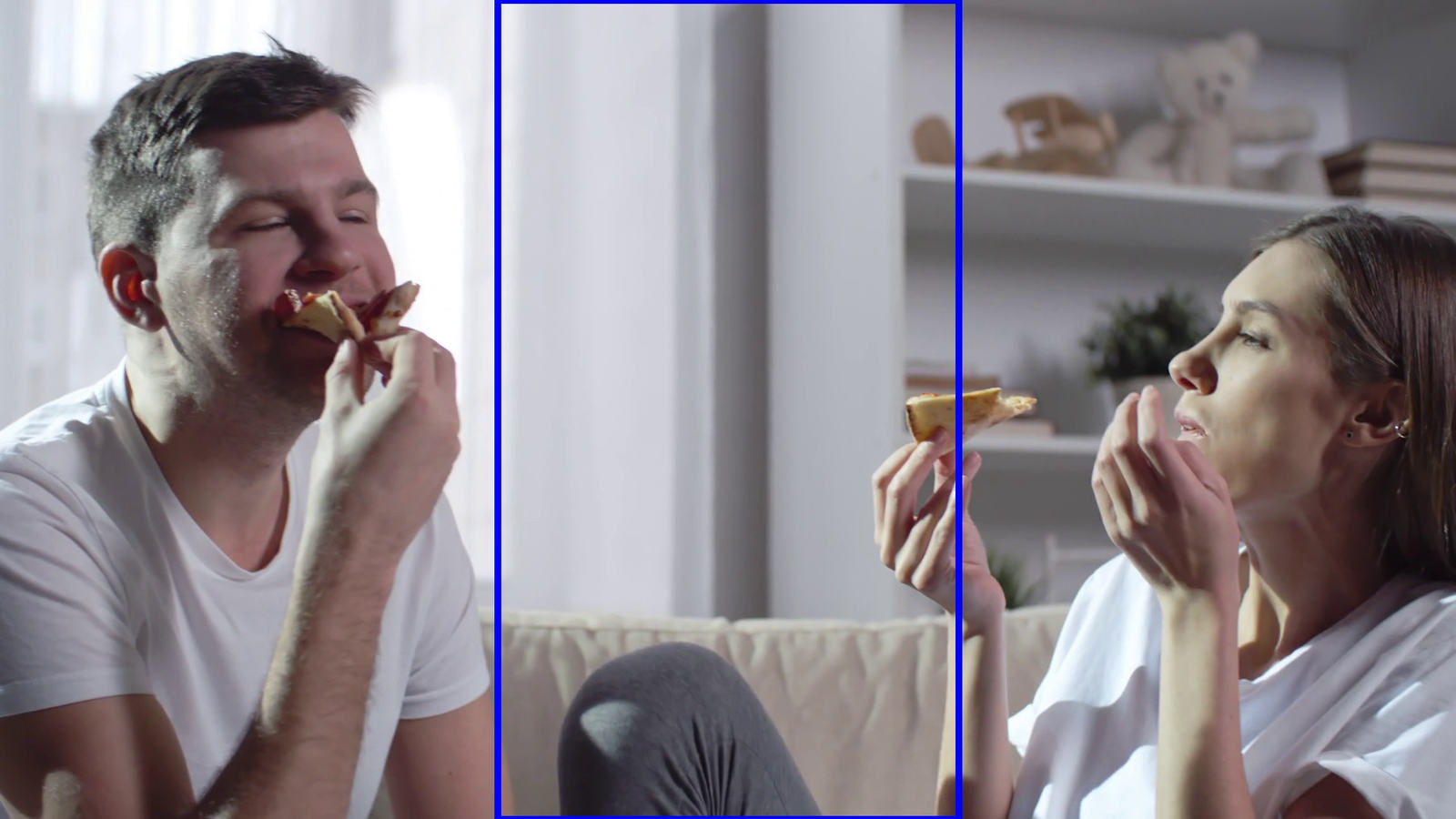} &
            \includegraphics[width=\fw]{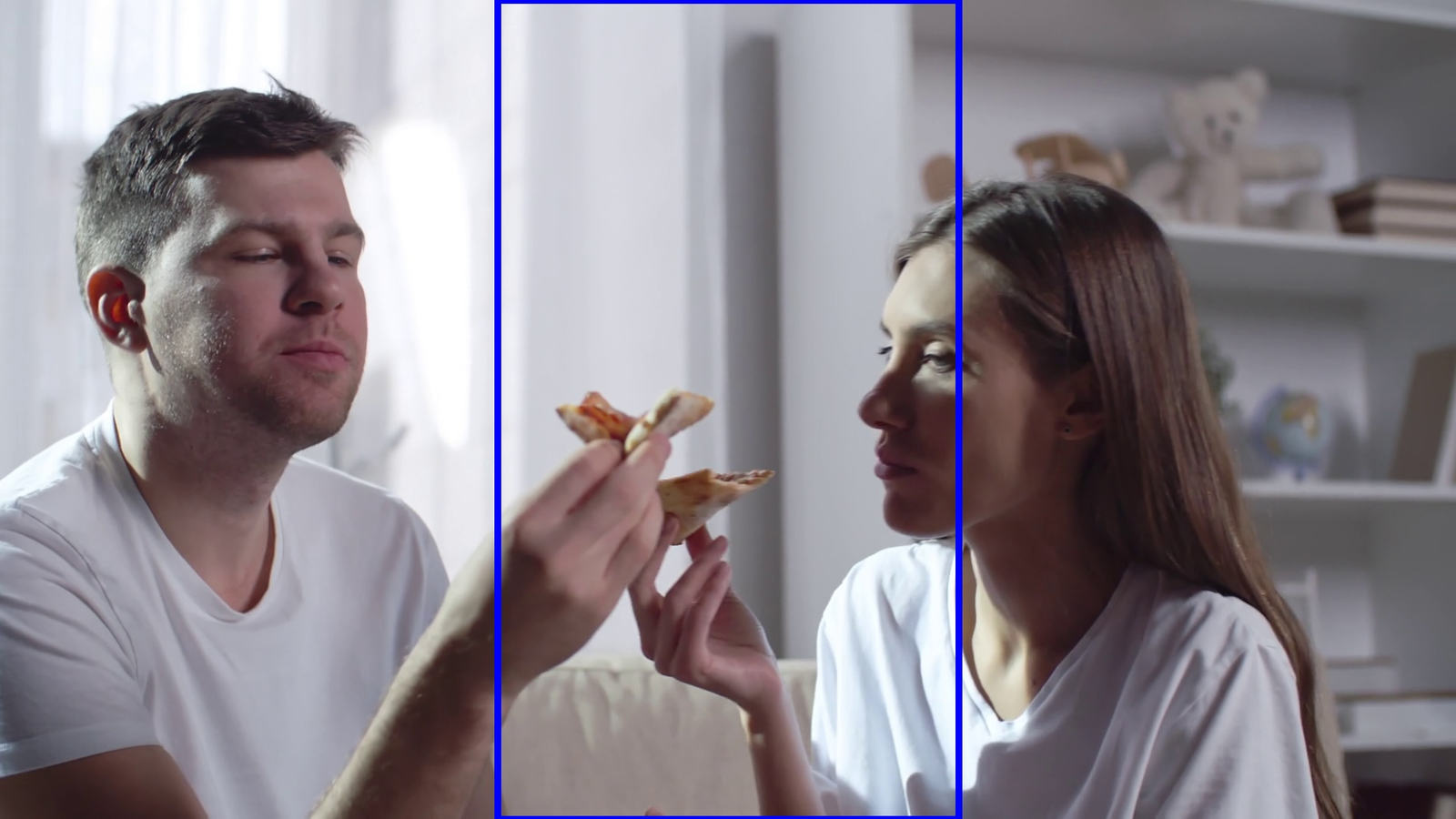} \\
            \includegraphics[width=\fw]{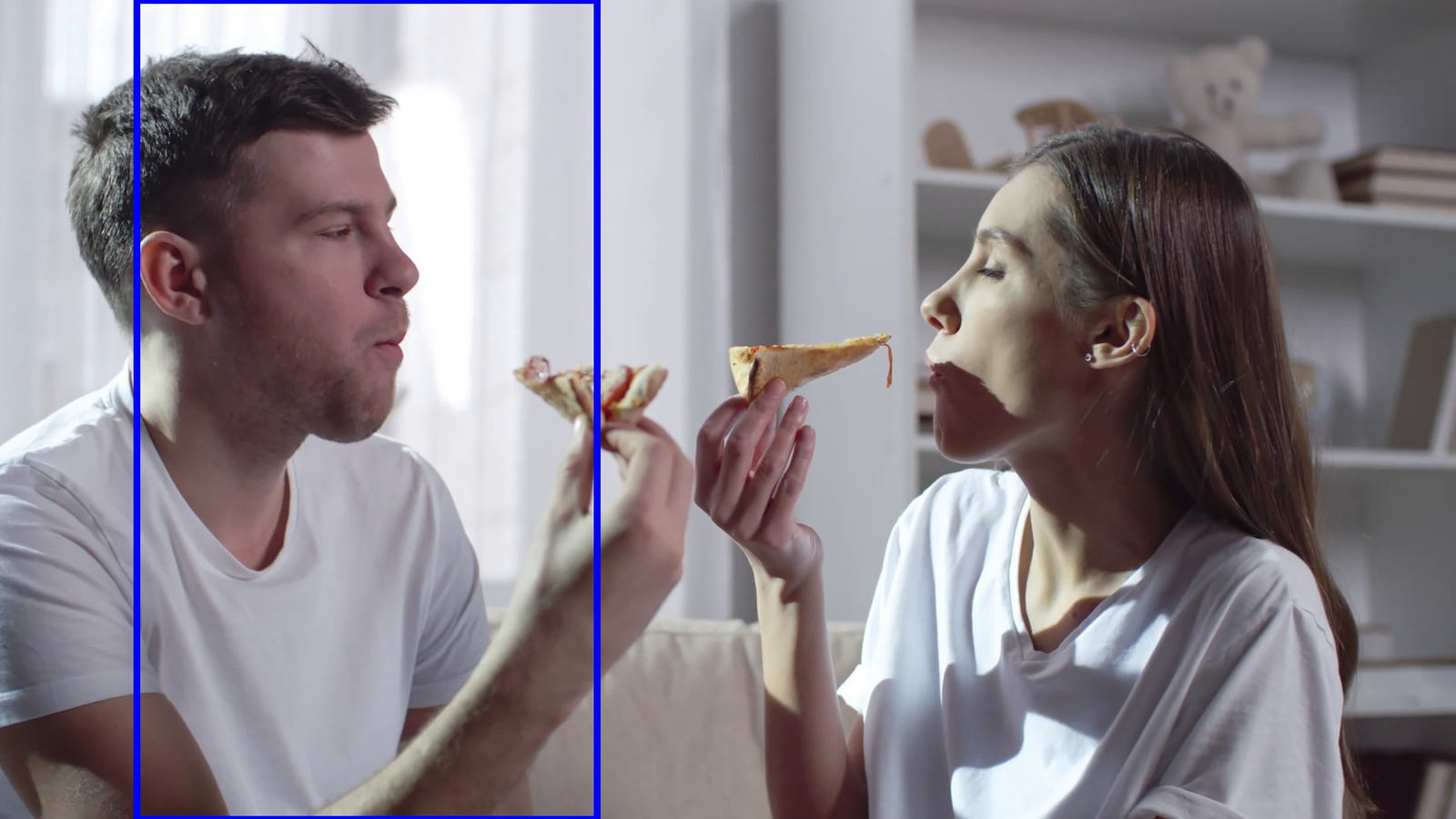} &
            \includegraphics[width=\fw]{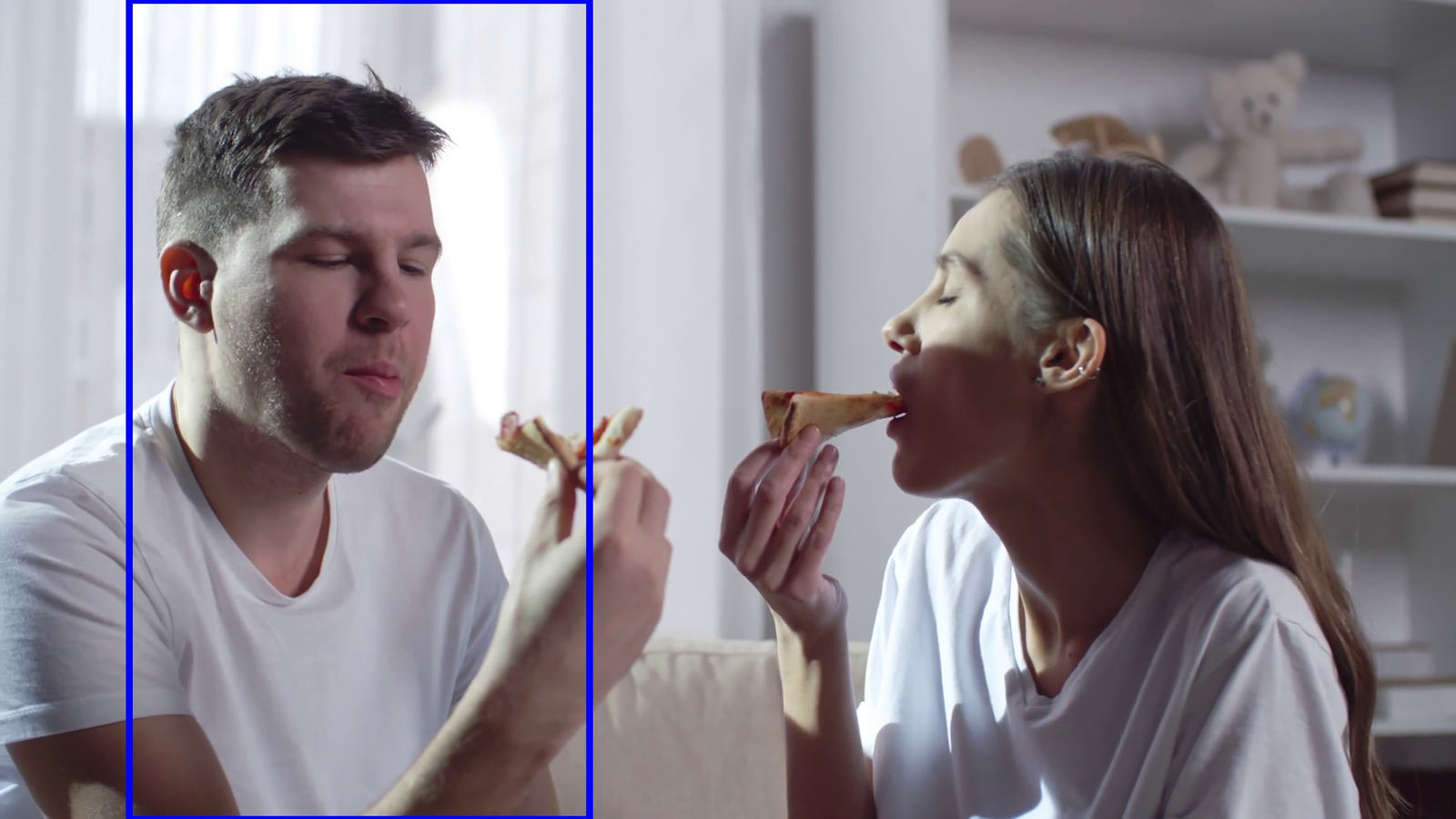} &
            \includegraphics[width=\fw]{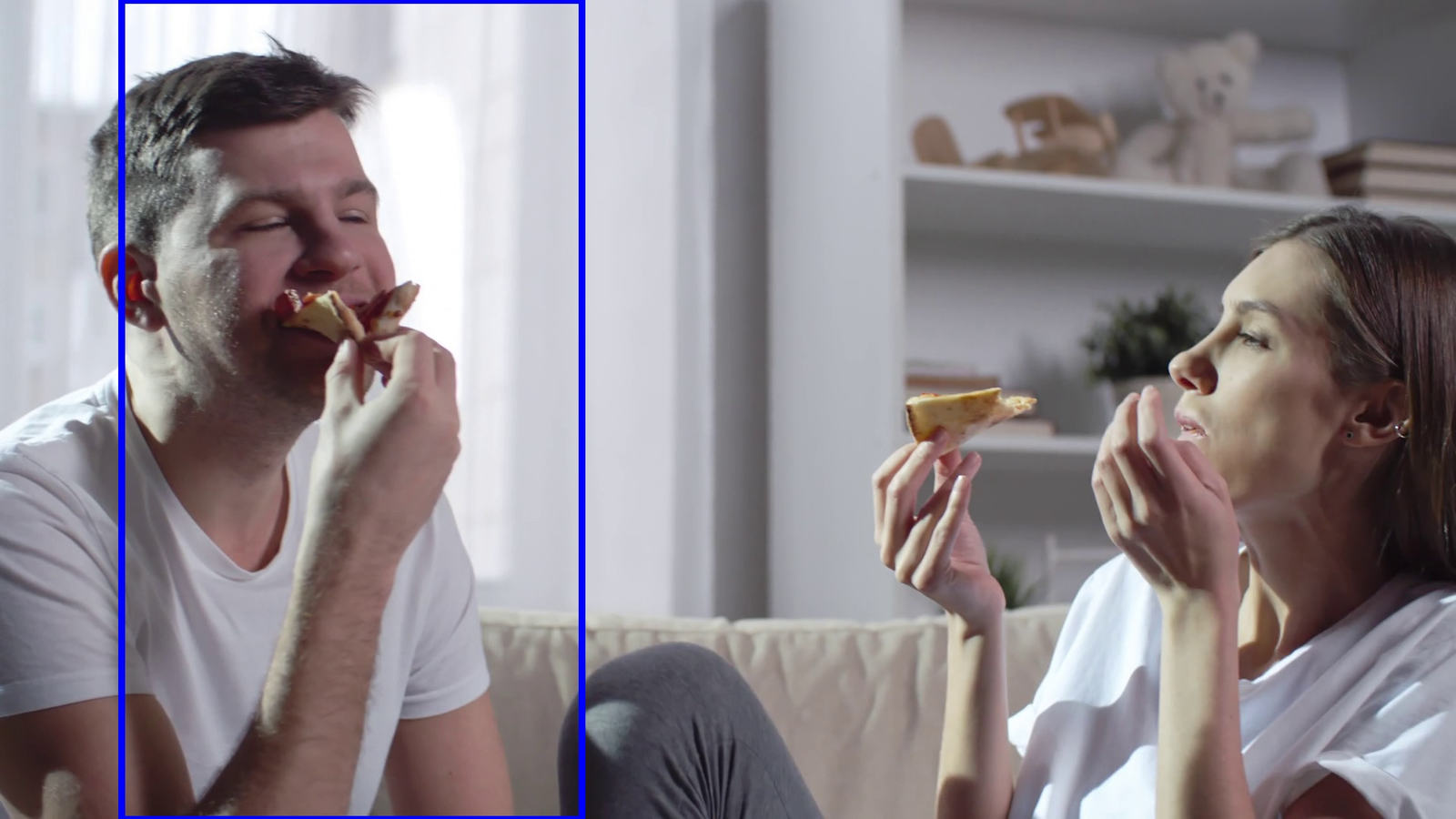} &
            \includegraphics[width=\fw]{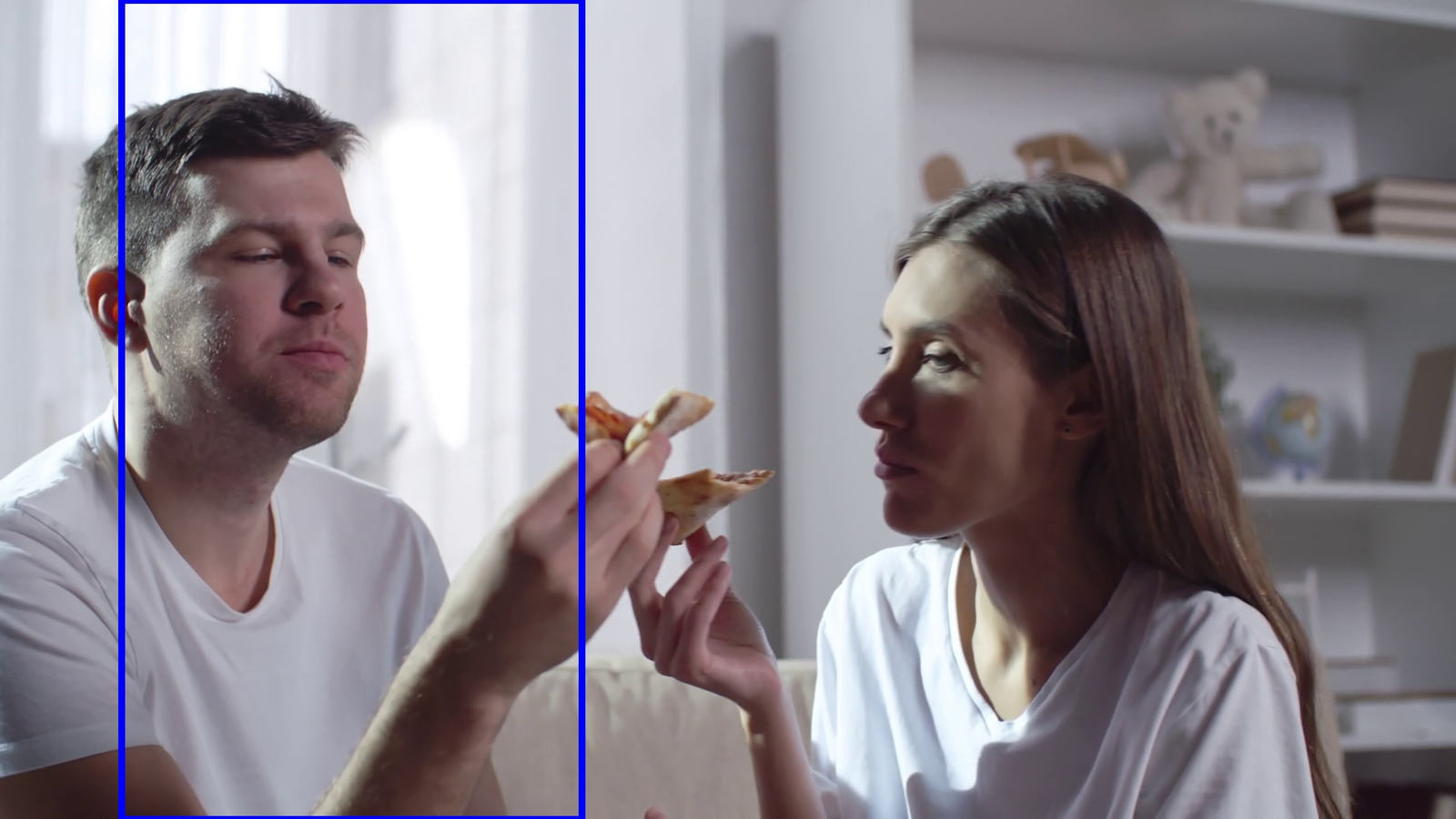} \\
            \includegraphics[width=\fw]{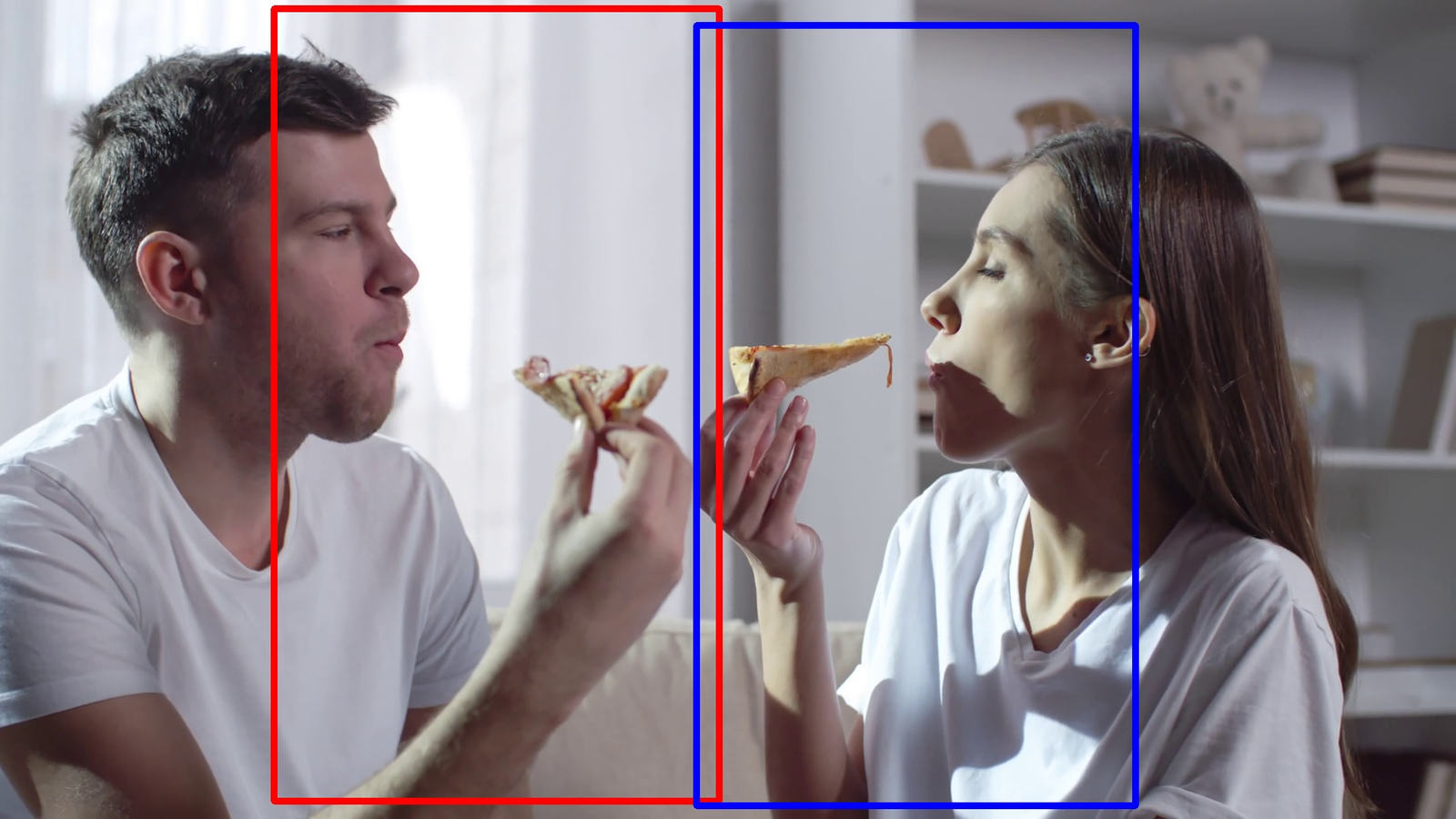} &
            \includegraphics[width=\fw]{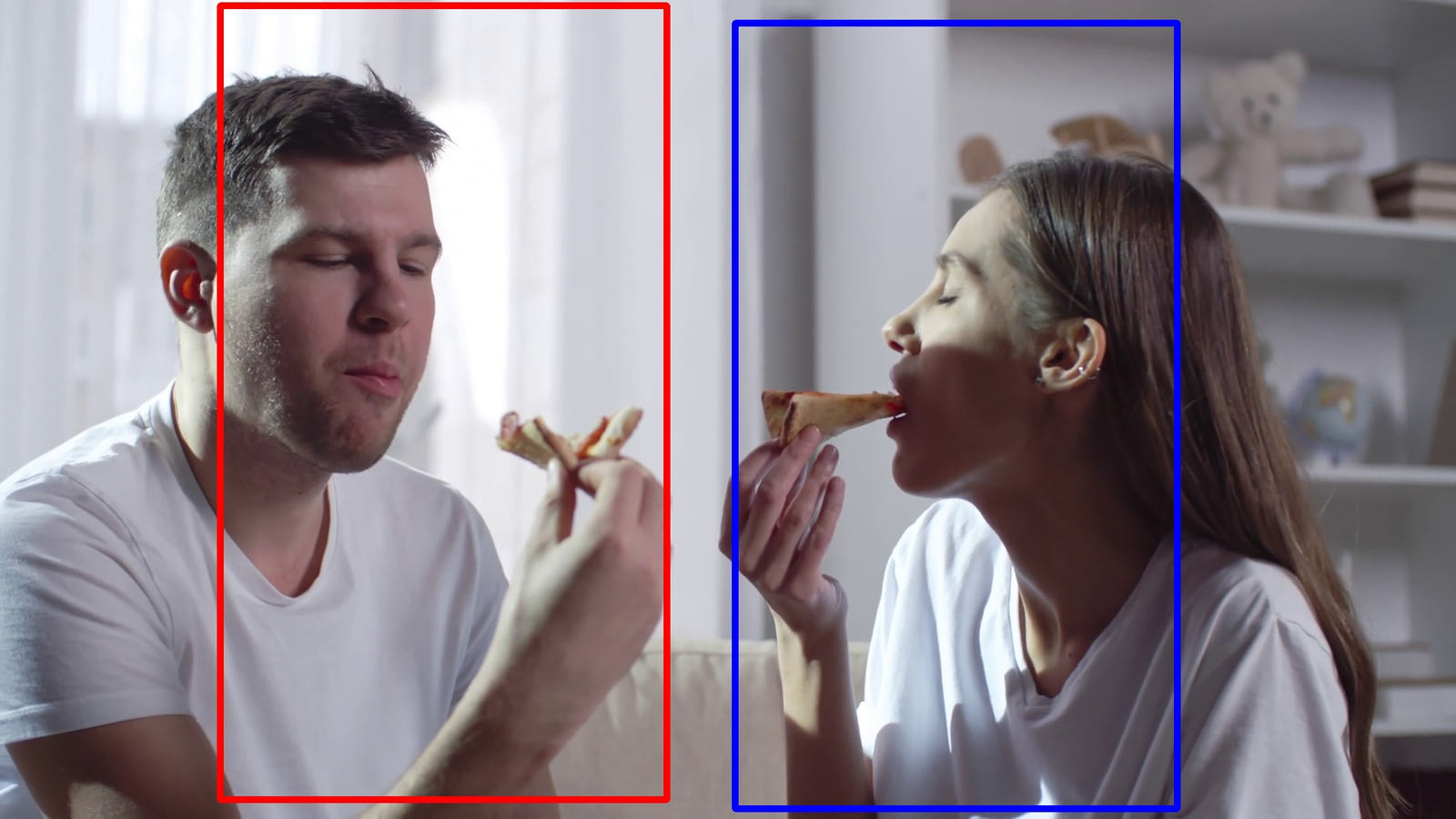} &
            \includegraphics[width=\fw]{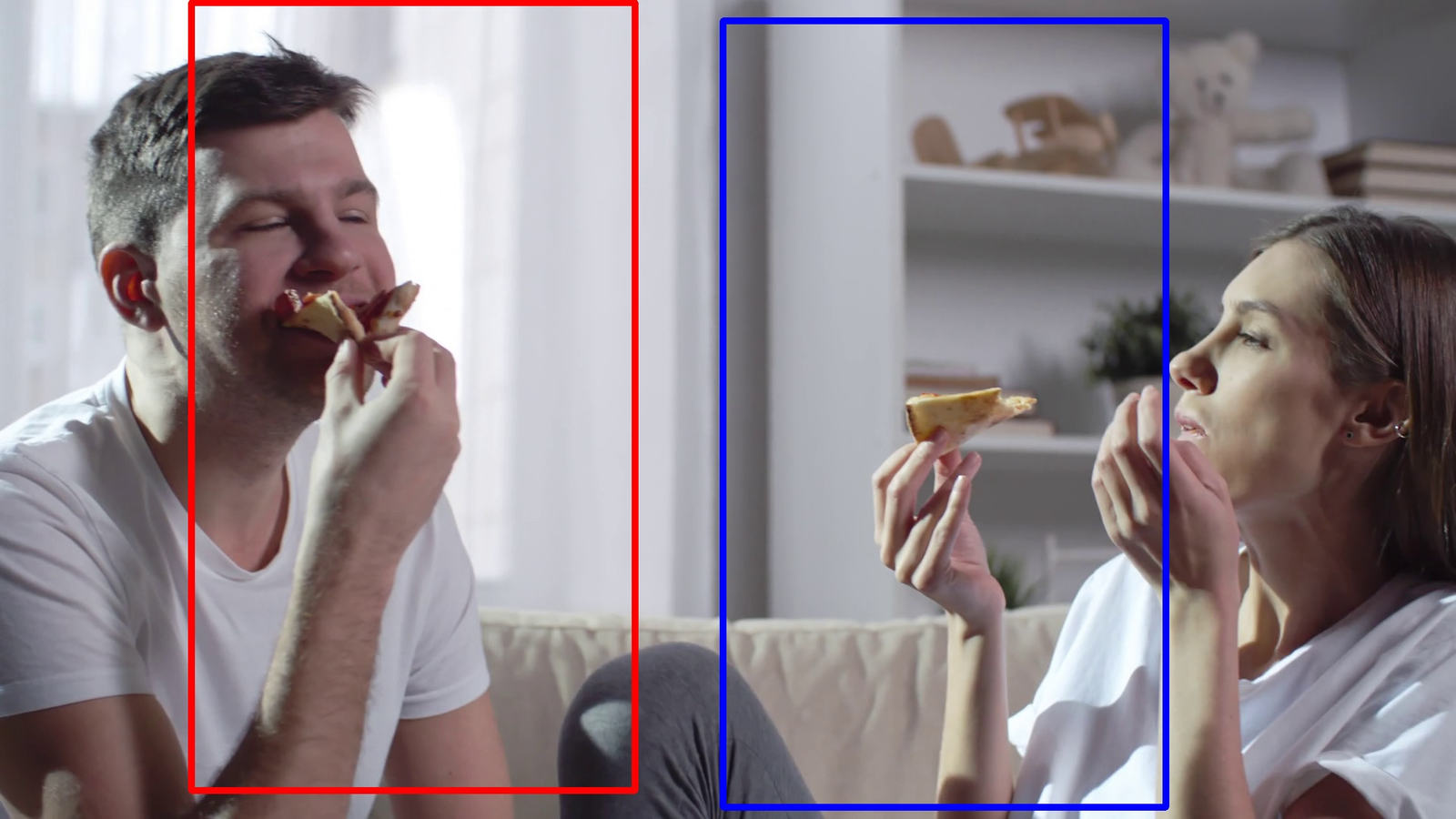} &
            \includegraphics[width=\fw]{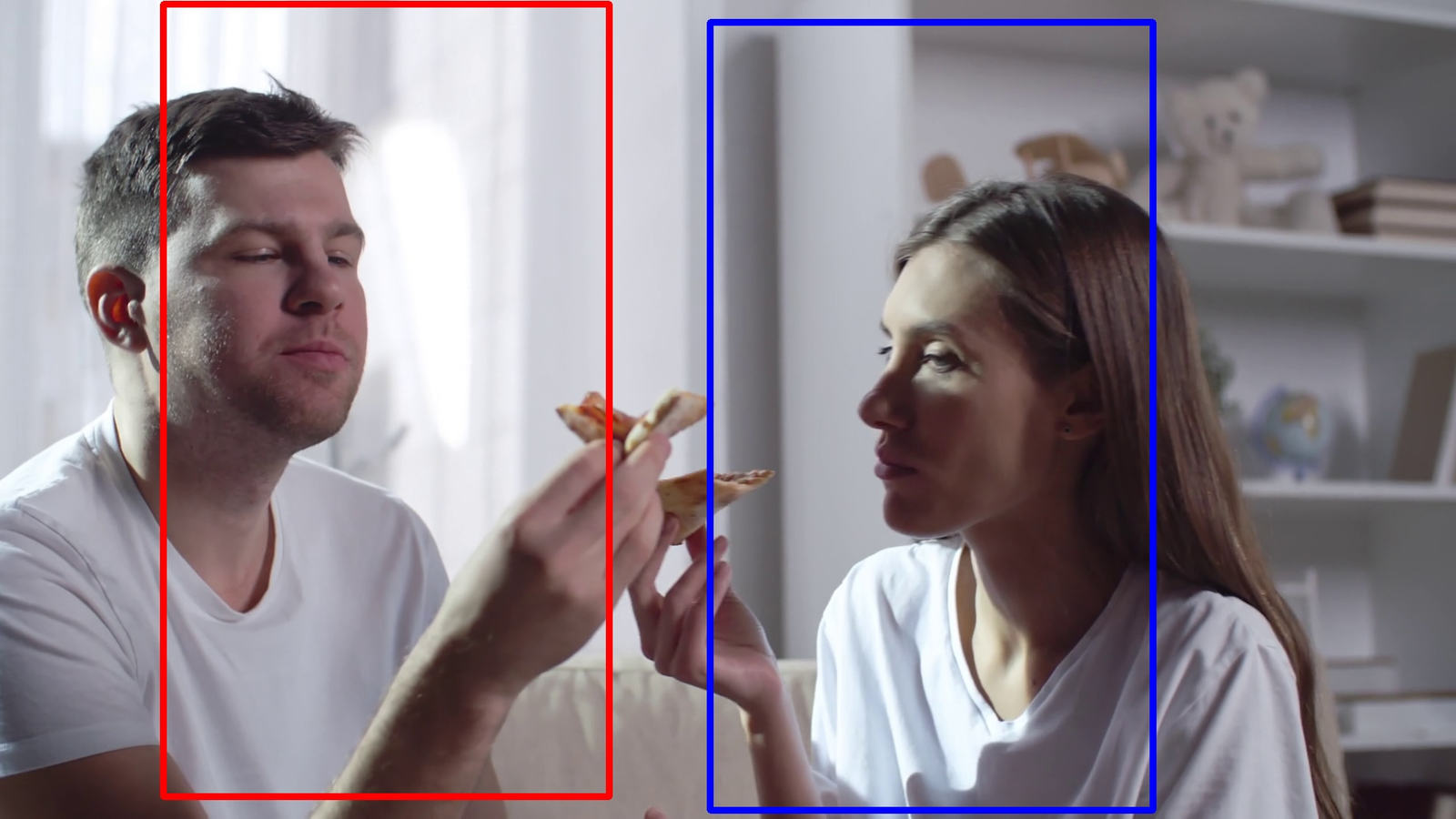} \\
        \end{tabular}
        \caption{Video ID: 4008033}
    \end{subfigure}
    \caption{\textbf{Qualitative comparison on MiraData \cite{ju2024miradata}.}
Rows correspond to CenterCrop (top), Gemini 3 Flash (middle), and
BayesCrop (bottom). BayesCrop provides more varied viewpoints while
maintaining stable framing around relevant content over time.}
    \label{fig:main_qualitative_comparison}
\end{figure*}

\subsection{Evaluation on the Multi-Target Long Video Benchmark}

\begin{table}[t]
\centering
\caption{
\textbf{Evaluation on the Multi-Target Long Video Benchmark} using VBench
and a three-way human preference study (180 total votes).
BayesCrop uses $S=3$ temporal segments for periodic view switching.
}
\label{tab:long_video_vbench}
\resizebox{\columnwidth}{!}{
\begin{tabular}{lcccc}
\toprule
\textbf{Method}
& \textbf{Aesthetic} $\uparrow$
& \textbf{Temp. Smooth.} $\uparrow$
& \textbf{Text-Video Align.} $\uparrow$
& \textbf{Preference} $\uparrow$ \\
\midrule
CenterCrop
& 54.2
& 96.5
& 24.0
& 33.3\% \\

Gemini 3 Flash
& 53.8
& \textbf{96.6}
& 23.4
& 27.2\% \\

\midrule
\textbf{BayesCrop + Periodic Agent}
& \textbf{54.7}
& 96.4
& \textbf{24.2}
& \textbf{39.4\%} \\
\bottomrule
\end{tabular}
}
\end{table}

Table~\ref{tab:long_video_vbench} evaluates BayesCrop on long-form multi-target videos. The Periodic Agent alternates between the two BayesCrop trajectories over fixed temporal segments, providing a lightweight single-output reframing strategy. BayesCrop achieves the
highest Aesthetic Quality (54.7) and Text-Video Alignment (24.2), while maintaining Temporal Smoothness comparable to the baselines. We further report results combining BayesCrop with MLLM-based view planning in the Supp.

The small variation in Temporal Smoothness (96.4--96.6) indicates that all methods produce visually plausible and temporally stable outputs. CenterCrop naturally inherits the original camera trajectory, while Gemini 3 Flash often remains focused on a single target and therefore
achieves the highest smoothness score. In contrast, BayesCrop introduces greater viewpoint variation while preserving comparable temporal stability.

We further conduct a three-way human preference study with 9 participants over the same 20 videos. Across 180 votes, BayesCrop receives the highest preference rate of 39.4\%, compared with 33.3\% for CenterCrop and 27.2\% for Gemini 3 Flash, suggesting a modest preference for its balance between important-content preservation and
viewpoint diversity.

Fig.~\ref{fig:main_qualitative_comparison} provides qualitative examples on MiraData. Compared with CenterCrop and direct Gemini cropping, BayesCrop produces more varied viewpoints across time while maintaining stable framing around semantically relevant regions. These examples complement the quantitative results in Table~\ref{tab:long_video_vbench} by illustrating how the proposed multi-view trajectories provide greater reframing diversity without introducing noticeable temporal instability. We further show more visual results including video samples in the Supp.

\begin{table}[t]
\centering
\caption{
\textbf{GMM sampling analysis on LIVE-YT-VC++.} Best-of-$N$ measures set-level reference coverage, Avg. IoU measures average candidate quality against the single reference, and Intra/Cross IoU characterize the compactness within each mode and separation between the two modes.
}
\label{tab:gmm_sampling}
\resizebox{\columnwidth}{!}{
\begin{tabular}{lcccc}
\toprule
\textbf{Candidates}
& \textbf{Best-of-$N$} $\uparrow$
& \textbf{Avg. IoU} $\uparrow$
& \textbf{Intra IoU}
& \textbf{Cross IoU} $\downarrow$ \\
\midrule
Means (2)   & 68.4 & \textbf{53.0} & --     & 42.6 \\
Samples (2) & 66.5 & 51.6 & --     & \textbf{41.2} \\
Samples (6) & 70.9 & 51.3 & 85.3 & 41.3 \\
Samples (10)& \textbf{72.5} & 51.4 & 85.5 & 41.3 \\
\bottomrule
\end{tabular}
}
\end{table}

\subsection{GMM Sampling Analysis}

Table~\ref{tab:gmm_sampling} evaluates whether the predicted GMM provides useful candidates beyond its two representative means. Increasing the sampling budget improves Best-of-$N$ IoU from 66.5 to 72.5, while Avg. IoU remains stable at 51-52 despite averaging samples from both modes against a single reference. This indicates that the coverage gain does not come from adding arbitrary low-quality boxes.

The diversity metrics further reveal a structured distribution: Intra-mode IoU remains high ($\sim$85), while Cross-mode IoU stays much lower ($\sim$41). Thus, samples remain concentrated around a consistent reframing hypothesis within each component, while the two components capture spatially distinct alternatives. Overall, the results suggest that the learned variance enables meaningful local exploration around complementary crop modes, improving reference coverage beyond the two GMM means.

\section{Conclusions}
This work introduces BayesCrop, a probabilistic multi-crop framework for efficient video reframing. By combining probabilistic multi-crop prediction with compressed-domain anchor-and-propagate tracking, BayesCrop preserves diverse reframing choices while achieving 112 FPS at the default point. Experiments on LIVE-YT-MC, LIVE-YT-VC++, and MiraData demonstrate a strong balance among localization accuracy, multi-crop coverage, visual quality, and efficiency, supporting practical real-time and multi-view video reframing.

{
    \small
    \bibliographystyle{abbrvnat}
    \bibliography{main}

@String(ICIP = {IEEE Int. Conf. Image Process.})

@String(ICIP  = {ICIP})

@inproceedings{huang2024vbench,
  title={Vbench: Comprehensive benchmark suite for video generative models},
  author={Huang, Ziqi and He, Yinan and Yu, Jiashuo and Zhang, Fan and Si, Chenyang and Jiang, Yuming and Zhang, Yuanhan and Wu, Tianxing and Jin, Qingyang and Chanpaisit, Nattapol and others},
  booktitle={Proceedings of the IEEE/CVF Conference on Computer Vision and Pattern Recognition},
  pages={21807--21818},
  year={2024}
}

@article{ju2024miradata,
  title={Miradata: A large-scale video dataset with long durations and structured captions},
  author={Ju, Xuan and Gao, Yiming and Zhang, Zhaoyang and Yuan, Ziyang and Wang, Xintao and Zeng, Ailing and Xiong, Yu and Xu, Qiang and Shan, Ying},
  journal={Advances in Neural Information Processing Systems},
  volume={37},
  pages={48955--48970},
  year={2024}
}

@inproceedings{apostolidis2021fast,
  title={A fast smart-cropping method and dataset for video retargeting},
  author={Apostolidis, K. and Mezaris, V.},
  booktitle={2021 IEEE International Conference on Image Processing (ICIP)},
  pages={2618--2622},
  year={2021}
}

@inproceedings{gu2024context,
  title={Context-Guided Spatio-Temporal Video Grounding},
  author={Gu, Xin and Fan, Heng and Huang, Yan and Luo, Tiejian and Zhang, Libo},
  booktitle={Proc. of the IEEE Conference on Comput. Vision Pattern Recog.},
  pages={18330--18339},
  year={2024}
}

@article{jin2022embracing,
  title={Embracing consistency: A one-stage approach for spatio-temporal video grounding},
  author={Jin, Yang and Yuan, Zehuan and Mu, Yadong and others},
  journal={Advances in Neural Information Processing Systems},
  volume={35},
  pages={29192--29204},
  year={2022}
}

@article{schuhmann2022laion,
  title={LAION-5B: An open large-scale dataset for training next generation image-text models},
  author={Schuhmann, Christoph and Beaumont, Romain and Vencu, Richard and Gordon, Cade and Wightman, Ross and Cherti, Mehdi and Coombes, Theo and Katta, Aarush and Mullis, Clayton and Wortsman, Mitchell and others},
  journal={Advances in Neural Information Processing Systems},
  volume={35},
  pages={25278--25294},
  year={2022}
}

@inproceedings{gu2025dynamic,
  title={Dynamic Beauty is Easy to Find: A Large-Scale Composition-Aware Dataset and an End-to-End Framework for Video Reframing},
  author={Gu, Sitian and Pan, Zhiyu and Hong, Chaoyi and Liu, Chengxin and Cao, Zhiguo},
  booktitle={Proceedings of the 33rd ACM International Conference on Multimedia},
  pages={7776--7784},
  year={2025}
}

@inproceedings{lee2024subjective,
  title={Subjective Portrait Region Cropping On Landscape Video Study},
  author={Lee, Cheng-Han and Mandal, Maniratnam and Birkbeck, Neil and Wang, Yilin and Adsumilli, Balu and Bovik, Alan C},
  booktitle={2024 IEEE International Conference on Image Processing (ICIP)},
  pages={48--54},
  year={2024},
  organization={IEEE}
}

@inproceedings{wang2019youtube,
  title={YouTube UGC dataset for video compression research},
  author={Wang, Y. and Inguva, S. and Adsumilli, B.},
  booktitle={2019 IEEE 21st International Workshop on Multimedia Signal Processing (MMSP)},
  pages={1--5},
  year={2019}
}

@inproceedings{ying2021patch,
  title={Patch-VQ:'Patching Up'the video quality problem},
  author={Ying, Z. and Mandal, M. and Ghadiyaram, D. and Bovik, A.},
  booktitle={Proc. of the IEEE Conference on Comput. Vision Pattern Recog.},
  pages={14019--14029},
  year={2021}
}

@article{zhu2021horizontal,
  title={Horizontal-to-vertical video conversion},
  author={Zhu, Tun and Zhang, Daoxin and Hu, Yao and Wang, Tianran and Jiang, Xiaolong and Zhu, Jianke and Li, Jiawei},
  journal={IEEE Transactions on Multimedia},
  volume={24},
  pages={3036--3048},
  year={2021},
  publisher={IEEE}
}

@article{cao2024reframe,
  title={Reframe anything: Llm agent for open world video reframing},
  author={Cao, Jiawang and Wu, Yongliang and Chi, Weiheng and Zhu, Wenbo and Su, Ziyue and Wu, Jay},
  journal={arXiv preprint arXiv:2403.06070},
  year={2024}
}

@article{chen2003visual,
  title={A visual attention model for adapting images on small displays},
  author={Chen, Li-Qun and Xie, Xing and Fan, Xin and Ma, Wei-Ying and Zhang, Hong-Jiang and Zhou, He-Qin},
  journal={Multimedia systems},
  volume={9},
  number={4},
  pages={353--364},
  year={2003},
  publisher={Springer}
}

@inproceedings{santella2006gaze,
  title={Gaze-based interaction for semi-automatic photo cropping},
  author={Santella, Anthony and Agrawala, Maneesh and DeCarlo, Doug and Salesin, David and Cohen, Michael},
  booktitle={Proceedings of the SIGCHI conference on Human Factors in computing systems},
  pages={771--780},
  year={2006}
}

@inproceedings{chen2017learning,
  title={Learning to compose with professional photographs on the web},
  author={Chen, Yi-Ling and Klopp, Jan and Sun, Min and Chien, Shao-Yi and Ma, Kwan-Liu},
  booktitle={Proceedings of the 25th ACM international conference on Multimedia},
  pages={37--45},
  year={2017}
}

@inproceedings{wang2017deep,
  title={Deep cropping via attention box prediction and aesthetics assessment},
  author={Wang, Wenguan and Shen, Jianbing},
  booktitle={Proceedings of the IEEE international conference on computer vision},
  pages={2186--2194},
  year={2017}
}

@article{zeng2020grid,
  title={Grid anchor based image cropping: A new benchmark and an efficient model},
  author={Zeng, Hui and Li, Lida and Cao, Zisheng and Zhang, Lei},
  journal={IEEE Transactions on Pattern Analysis and Machine Intelligence},
  volume={44},
  number={3},
  pages={1304--1319},
  year={2020},
  publisher={IEEE}
}

@article{guo2018automatic,
  title={Automatic image cropping for visual aesthetic enhancement using deep neural networks and cascaded regression},
  author={Guo, Guanjun and Wang, Hanzi and Shen, Chunhua and Yan, Yan and Liao, Hong-Yuan Mark},
  journal={IEEE Transactions on Multimedia},
  volume={20},
  number={8},
  pages={2073--2085},
  year={2018},
  publisher={IEEE}
}

@inproceedings{hong2021composing,
  title={Composing photos like a photographer},
  author={Hong, Chaoyi and Du, Shuaiyuan and Xian, Ke and Lu, Hao and Cao, Zhiguo and Zhong, Weicai},
  booktitle={Proceedings of the IEEE/CVF Conference on Computer Vision and Pattern Recognition},
  pages={7057--7066},
  year={2021}
}

@inproceedings{li2020learning,
  title={Learning to learn cropping models for different aspect ratio requirements},
  author={Li, Debang and Zhang, Junge and Huang, Kaiqi},
  booktitle={Proceedings of the IEEE/CVF Conference on Computer Vision and Pattern Recognition},
  pages={12685--12694},
  year={2020}
}

@inproceedings{pan2021transview,
  title={Transview: Inside, outside, and across the cropping view boundaries},
  author={Pan, Zhiyu and Cao, Zhiguo and Wang, Kewei and Lu, Hao and Zhong, Weicai},
  booktitle={Proceedings of the IEEE/CVF International Conference on Computer Vision},
  pages={4218--4227},
  year={2021}
}

@inproceedings{wang2023image,
  title={Image cropping with spatial-aware feature and rank consistency},
  author={Wang, Chao and Niu, Li and Zhang, Bo and Zhang, Liqing},
  booktitle={Proceedings of the IEEE/CVF Conference on Computer Vision and Pattern Recognition},
  pages={10052--10061},
  year={2023}
}

@article{zhang2020emotion,
  title={Emotion attention-aware collaborative deep reinforcement learning for image cropping},
  author={Zhang, Xiaoyan and Li, Zhuopeng and Jiang, Jianmin},
  journal={IEEE Transactions on Multimedia},
  volume={23},
  pages={2545--2560},
  year={2020},
  publisher={IEEE}
}

@article{wang2023internvid,
  title={Internvid: A large-scale video-text dataset for multimodal understanding and generation},
  author={Wang, Yi and He, Yinan and Li, Yizhuo and Li, Kunchang and Yu, Jiashuo and Ma, Xin and Li, Xinhao and Chen, Guo and Chen, Xinyuan and Wang, Yaohui and others},
  journal={arXiv preprint arXiv:2307.06942},
  year={2023}
}

@inproceedings{carion2020end,
  title={End-to-end object detection with transformers},
  author={Carion, Nicolas and Massa, Francisco and Synnaeve, Gabriel and Usunier, Nicolas and Kirillov, Alexander and Zagoruyko, Sergey},
  booktitle={European conference on computer vision},
  pages={213--229},
  year={2020},
  organization={Springer}
}

@article{oquab2023dinov2,
  title={Dinov2: Learning robust visual features without supervision},
  author={Oquab, Maxime and Darcet, Timoth{\'e}e and Moutakanni, Th{\'e}o and Vo, Huy and Szafraniec, Marc and Khalidov, Vasil and Fernandez, Pierre and Haziza, Daniel and Massa, Francisco and El-Nouby, Alaaeldin and others},
  journal={arXiv preprint arXiv:2304.07193},
  year={2023}
}

@article{lee2026subjective,
  title={Subjective Portrait Region Cropping in Landscape Videos with Temporal Annotation Smoothing},
  author={Lee, Cheng-Han and Mandal, Maniratnam and Birkbeck, Neil and Wang, Yilin and Adsumilli, Balu and Bovik, Alan C},
  journal={arXiv preprint arXiv:2604.24947},
  year={2026}
}

@misc{google_gemini_3_flash_2026,
  author       = {{Google}},
  title        = {Gemini 3 Flash},
  month        = aug,
  year         = {2026},
  note         = {Large language model},
  url          = {https://google.com},
  version      = {3 Flash}
}
}
\clearpage
\setcounter{page}{1}

\section{Appendix: Implementation Details}

\subsection{Data Curation}
\subsubsection{\textbf{Sliding Window Process for Dual-View GMM Data Generation }}
To construct a robust dataset for dual-view probabilistic modeling, we employ a sliding window strategy that aggregates temporal information around designated anchor frames. For each anchor frame, we define a window of size 15 and warp the ground-truth ROIs from adjacent frames to the anchor's coordinate space. This approach allows the model to capture the local distribution of human attention while maintaining spatial consistency. To ensure the integrity of these temporal associations, we first preprocess all video content using the PySceneDetect toolbox; warping is strictly confined to frames belonging to the same scene to prevent cross-cut contamination.

The source videos consist of 180 frames (approximately 6 seconds) with 30 annotated frames. To maintain a consistent window size and avoid boundary truncation at the start or end of a sequence, we constrain the selection of anchor frames. Specifically, we shift the anchor from frame 5 to frame 24, yielding 20 valid anchor frames per video segment. This curation process ensures that every training sample is supported by a complete temporal context, providing the GMM predictor with high-quality, localized human subjective labels.

\subsubsection{\textbf{GMM Data Pre-processing and Initialization}}
We utilize a \textbf{rescue\_bbox} function to ensure all ground-truth ROIs remain within the frame boundaries. To facilitate easier convergence, we enforce a consistent data order: for any dual-view pair, the left-most ROI is assigned to the first component and the right-most to the second. For frames where the GMM contains fewer than two components, we apply a specific padding strategy to maintain a fixed-length target vector. If only one component exists, we set its weight to 1.0 and pad the second component with a weight of 0.0 and a centered, full-frame ROI ($x=960, y=540, r=1080$). If no components are detected, we initialize the primary component as a centered crop ($r=864$) with a small variance ($\sigma=50$) and pad the second as described above. This consistent initialization ensures a stable optimization surface across diverse cinematic scenarios.

\subsection{Model Design}
\begin{figure*}[t]
\begin{center}
 \includegraphics[width=1.0\linewidth]{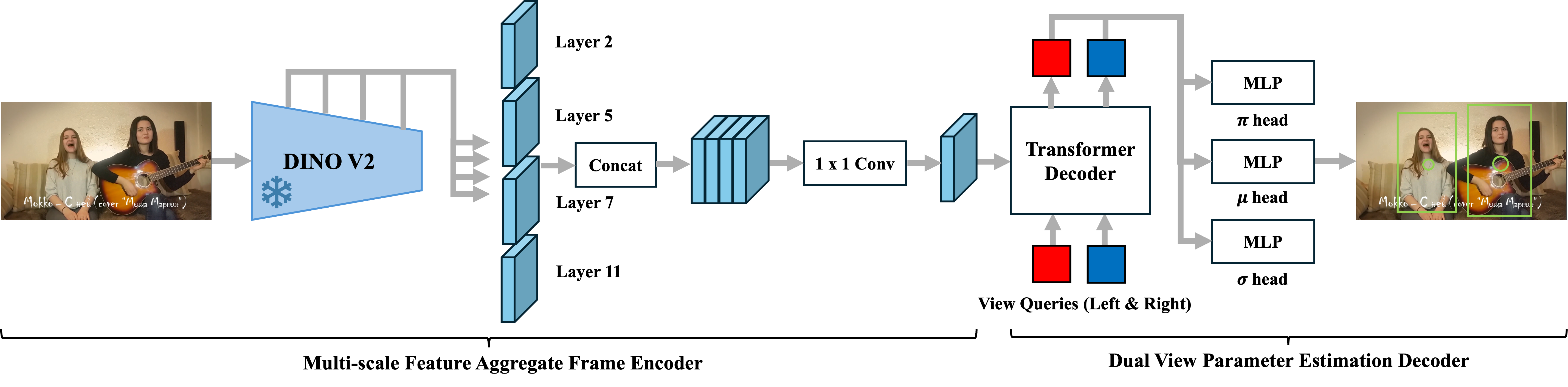}
\end{center}
\caption{Details of Model Design}
\label{fig:long}
\label{fig:onecol}
\label{fig:arch}
\end{figure*}  
As shown in Fig. \ref{fig:arch}, the architecture of the Probabilistic Multi-Crop Predictor utilizes a hierarchical feature extraction and probabilistic prediction pipeline designed for multi-focal video reframing. The system begins with a frozen DINOv2-ViT-S/14 backbone, which processes a 224 $\times$ 224 input frame to extract rich semantic descriptors from layers 2, 5, 8, and 11. These multi-scale features are concatenated along the channel dimension, and a 1x1 convolutional layer is applied to reduce the channel size to 1/4 of the original dimension, creating a compact feature representation. This latent information, accompanied by position embeddings to preserve spatial context, is then processed by a Transformer Decoder equipped with two learnable query tokens—one specifically designated for the left view and the other for the right view. These tokens interact with the encoded image features via cross-attention to attend to different narrative subjects. Finally, the decoder outputs are fed into a prediction head that generates the parameters for a spatially isotropic Gaussian Mixture Model (GMM), providing the mixing coefficients ($\pi$), spatial means ($\mu$), and sigmas ($\sigma$) for $(x, y, r)$ coordinates. This design allows the model to explicitly decouple and localize two distinct regions of interest while maintaining a unified probabilistic framework.

\subsection{End-to-end Multi-objective Learning}

\subsubsection{Detailed Loss of Mixture Weights}

For a predicted probability $\hat{p}_i \in [0,1]$ and a soft target
$y_i \in [0,1]$, the loss is defined as
\begin{equation}
\mathcal{L}_{\pi}
=
-\frac{\alpha}{N}
\sum_{i=1}^{N}
(1-p_{t,i})^{\gamma}
\left[
y_i \log(\hat{p}_i)
+
(1-y_i)\log(1-\hat{p}_i)
\right].
\end{equation}

Here, $p_{t,i}$ represents the agreement between the prediction and the target:
\begin{equation}
p_{t,i}
=
\hat{p}_i y_i
+
(1-\hat{p}_i)(1-y_i).
\end{equation}

The Focal Binary Cross Entropy (BCE) loss adapts to soft human consensus
($y_i \in [0,1]$) by measuring the agreement $p_{t,i}$ between the model
prediction and the human annotation. With $\gamma=2.0$, the modulating factor
$(1-p_{t,i})^{\gamma}$ reduces the contribution of samples for which the
prediction already agrees well with the target, while assigning greater
weight to samples with larger disagreement. We use $\alpha=0.25$ as the balancing factor.

\subsubsection{Detailed Loss of Gaussian Means}

This loss supervises the crop center and scale parameters $(x,y,r)$ of
the predicted Gaussian components. Let $\mu_i=(x_i,y_i,r_i)$ denote the
ground-truth mean vector and $\hat{\mu}_i$ its prediction. The loss is
defined as
\begin{equation}
\mathcal{L}_{\mu}
=
\frac{1}{N}
\sum_{i=1}^{N}
\frac{
|\hat{\mu}_i-\mu_i|^{\gamma}
}{
r_i+\epsilon
}
(\hat{\mu}_i-\mu_i)^2,
\end{equation}
where the absolute value, exponentiation, and squared-error operations are
applied element-wise over the $(x,y,r)$ dimensions.

The Focal Scale MSE Loss combines scale-aware normalization with
error-dependent weighting. The factor $1/(r_i+\epsilon)$ normalizes the
regression error according to the target crop scale, reducing the
disproportionate contribution of larger crops. The focal term, with
$\gamma=2.0$, assigns greater weight to dimensions with larger prediction
errors while suppressing the contribution of well-predicted parameters.
This encourages the model to focus on correcting substantial errors in
crop position and scale.

\subsubsection{Detailed Loss of Gaussian Variances}

For a batch of $N$ predicted uncertainty vectors $\hat{\sigma}_i$ and
ground-truth uncertainty vectors $\sigma_i$, the loss is defined as
\begin{equation}
\mathcal{L}_{\sigma}
=
\frac{1}{N}
\sum_{i=1}^{N}
|\hat{\sigma}_i-\sigma_i|^{\gamma}
(\hat{\sigma}_i-\sigma_i)^2,
\end{equation}
where the absolute value, exponentiation, and squared-error operations are
applied element-wise over the uncertainty dimensions.

The Focal $\sigma$-Loss applies error-dependent weighting to the prediction
of the local uncertainty associated with each Gaussian component. We set
$\gamma=0.5$, such that dimensions with larger uncertainty-prediction errors
receive greater emphasis, while well-predicted dimensions contribute less
to the optimization.

In BayesCrop, $\sigma$ controls the spread of each Gaussian component over
crop position and scale. Smaller values correspond to more concentrated
crop distributions, whereas larger values represent a broader range of
plausible reframing variations. Supervising $\sigma$ therefore enables the
model to capture the local variation observed in the human-derived crop
distribution rather than assigning a fixed or uniformly broad uncertainty
to all reframing modes.

\subsubsection{\textbf{Hyperparameters}}
To balance multi-objective optimization during training, we empirically determined the loss weighting coefficients to align the magnitudes of the distinct gradient signals. The total objective function is regularized by the following hyperparameters: $\lambda_{\pi} = 100$ for the mixing coefficients, $\lambda_{\mu} = 5000$ for the spatial means, and $\lambda_{\sigma} = 10$ for the spatially isotropic covariance parameters. Since the raw values of the covariance $\sigma$ within the $(x, y, r)$ parameter space are numerically small and difficult for the network to resolve, we scale these values by a factor of 10 to provide a more effective gradient signal during backpropagation. 
Furthermore, all target coordinates $(x, y, r)$ are normalized to the range $[0, 1]$ relative to the original frame dimensions to ensure numerical stability.
Our framework is implemented using the PyTorch library, employing DINOv2-ViT-S/14 as the core feature extractor with an input resolution of 224 $\times$ 224. The substantial weight assigned to $\lambda_{\mu}$ reflects the primary importance of precise ROI localization in the 3D coordinate space of $(x, y, r)$. For the compound loss formulation, we set $\lambda_{\text{IoU}} = 100$ to ensure robust geometric overlap, while the distribution-level supervision is governed by $\lambda_{\text{KL}} = 0.1$. All experiments were conducted on a single NVIDIA GeForce RTX 4090 (24 GB VRAM) using the AdamW optimizer with a constant learning rate of 5e-4. To prevent overfitting to the specific aesthetic biases of the training set, we limited training to 5 epochs, which we found sufficient for convergence and strong generalization.

\subsection{Temporal Smoothing Parameters}
\label{sec:temporal_smoothing}

We apply Gaussian temporal smoothing to the predicted crop trajectories,
with different configurations for short-video real-time evaluation and
long-video offline evaluation.

For LIVE-YT-VC++, we use a \emph{causal} Gaussian smoothing
kernel to preserve the online nature of the reframing pipeline and to
evaluate its real-time capability. Specifically, the smoothing window
contains only the current and preceding frames, with a window size of
30 frames and Gaussian standard deviation $\sigma=9$.

For the Multi-Target Long Video Benchmark, we instead use \emph{centered non-causal} Gaussian smoothing on the complete predicted
trajectory. Since this benchmark evaluates the quality of the final
rendered video rather than streaming latency, both past and future frames
can be used for smoothing. We use a window size of 90 frames and Gaussian
standard deviation $\sigma=39$.

These settings reflect the different evaluation goals: causal smoothing
is used for the real-time LIVE-YT-VC++ experiments, whereas non-causal
smoothing provides stronger trajectory stabilization for offline
long-video rendering.

\subsection{Prompt Design for MLLM-Based Reframing}
\label{sec:supp_prompt}

We evaluate Gemini 3 Flash in two settings with different responsibilities.

\textbf{Direct MLLM Baseline.}
As shown in Fig.~\ref{fig:prompt_baseline}, Gemini directly predicts a
$9{:}16$ ROI for each queried frame in the form
$[x_1,y_1,x_2,y_2]$.
Thus, the MLLM is responsible for both semantic subject selection and
spatial crop localization.
The predicted coordinates are then used to construct the reframed video.

\textbf{BayesCrop + MLLM-based View Switch Agent.}
In our extension (Fig.~\ref{fig:prompt_ours}), BayesCrop first
generates two temporally coherent crop trajectories.
Gemini is not asked to predict coordinates; instead, it selects between
the two candidate views and outputs temporal segments specified by
\texttt{start\_frame}, \texttt{end\_frame}, and \texttt{box\_index}.
This separates spatial localization from high-level shot planning:
BayesCrop determines \emph{where to crop}, while the MLLM determines
\emph{when to switch views}.

The MLLM planner is an optional downstream extension and is not required
by the core BayesCrop pipeline.

\begin{figure}[h]
    \centering
    \begin{lstlisting}[language=Python, backgroundcolor=\color{backgray}]
prompt = """
Role: Professional AI Video Editor.
Task: Create a dynamic 9:16 vertical crop plan for this video.

Instructions:
1. For each frame, provide the 'best_roi' coordinates that focus on the 
   most important and best aesthetic part of the frame.
2. Use pixel coordinates [x1, y1, x2, y2], where x1 and x2 are 
   ranging from 0 to 1920, and y1 and y2 are ranging from 0 to 1080.

Output Format: The output should be a JSON list of objects.
Schema: [{"frame_index": int, "best_roi": [x1, y1, x2, y2]}]
"""
    \end{lstlisting}
    \caption{Baseline prompt for frame-wise ROI prediction.}
    \label{fig:prompt_baseline}
\end{figure}

\begin{figure}[h]
    \centering
    \begin{lstlisting}[language=Python, backgroundcolor=\color{backblue}]
prompt = f"""
You are a professional video editor. I have a video with {total_frames} frames 
and bboxes for every frame in this json file: {json_path}.

Based on the video and the bboxes, create an edit plan for a 9:16 vertical crop.
You MUST vary the selection between the bboxes at least 2-3 times
to create perspective shifts (cuts) that follow the story.

OUTPUT FORMAT:
Return ONLY a JSON list of segments.
Example:
[
    {{"start_frame": 0, "end_frame": 100, "box_index": 0}},
    {{"start_frame": 101, "end_frame": 250, "box_index": 1}}
]
"""
    \end{lstlisting}
    \caption{The prompt used in our method for cinematic segment planning.}
    \label{fig:prompt_ours}
\end{figure}

\subsection{Evaluation Metrics}
\subsubsection{\textbf{Bounding-Box Accuracy (mIoU and $\pi$-mIoU)}}

We evaluate spatial localization accuracy using Intersection over Union
(IoU) between predicted and reference reframing boxes. For a conventional
single-output prediction, mean IoU (mIoU) is computed as
\begin{equation}
\mathrm{mIoU}
=
\frac{1}{n}
\sum_{i=1}^{n}
\mathrm{IoU}
\left(
B_i^{pred}, B_i^{gt}
\right),
\end{equation}
where $n$ denotes the number of evaluated frames.

For the $K=2$ probabilistic predictor on LIVE-YT-MC, we additionally
define mixture-weighted IoU ($\pi$-mIoU) as
\begin{equation}
\pi\text{-mIoU}
=
\frac{1}{n}
\sum_{i=1}^{n}
\sum_{k=1}^{2}
\pi_{i,k}^{gt}
\,
\mathrm{IoU}
\left(
B_{i,k}^{pred},
B_{i,k}^{gt}
\right),
\end{equation}
where $B_{i,k}^{pred}$ and $B_{i,k}^{gt}$ denote the predicted and
ground-truth representative crops of the $k$-th GMM component,
respectively, and $\pi_{i,k}^{gt}$ is its ground-truth mixture weight.
This metric evaluates the localization accuracy of both reframing modes
while assigning greater importance to modes that receive higher mass in
the human-opinion distribution.

For single-output evaluation on LIVE-YT-VC++, we select the predicted
mode with the larger predicted mixture weight,
\begin{equation}
k_i^{*}
=
\arg\max_{k\in\{1,2\}}\hat{\pi}_{i,k},
\end{equation}
and report the mIoU of $B_{i,k_i^{*}}^{pred}$ against the reference crop.
We refer to this setting as $\hat{\pi}$-based selection.

\subsubsection{\textbf{Mixture-Weight Selection Accuracy ($\pi$ Acc.)}}

To evaluate whether the predicted mixture weights capture the relative
preference between the two reframing modes, we compute Selection Accuracy
($\pi$ Acc.). Because small deviations around equal preference are
unlikely to indicate a meaningful ordering, we use a tolerance
$\epsilon=0.05$ and define
\begin{equation}
\Phi(\pi)=
\begin{cases}
  1,  & \pi > 0.5+\epsilon, \\
 -1,  & \pi < 0.5-\epsilon, \\
  0,  & |\pi-0.5|\leq\epsilon.
\end{cases}
\end{equation}

Selection Accuracy is then
\begin{equation}
\pi\text{ Acc.}
=
\frac{1}{n}
\sum_{i=1}^{n}
\mathbb{I}
\left[
\Phi(\hat{\pi}_i)
=
\Phi(\pi_i^{gt})
\right],
\end{equation}
where $\hat{\pi}_i$ and $\pi_i^{gt}$ denote the predicted and
ground-truth mixture weights of the corresponding component.

\subsubsection{\textbf{Distribution Accuracy ($W_2$)}}

We measure agreement between the predicted and ground-truth crop
distributions using the 2-Wasserstein distance ($W_2$). For two Gaussian
distributions
$\mathcal{N}(\boldsymbol{\mu}_1,\boldsymbol{\Sigma}_1)$ and
$\mathcal{N}(\boldsymbol{\mu}_2,\boldsymbol{\Sigma}_2)$, the squared
2-Wasserstein distance is
\begin{equation}
W_2^2
=
\|\boldsymbol{\mu}_1-\boldsymbol{\mu}_2\|_2^2
+
\mathrm{Tr}
\left(
\boldsymbol{\Sigma}_1
+
\boldsymbol{\Sigma}_2
-
2
\left(
\boldsymbol{\Sigma}_2^{1/2}
\boldsymbol{\Sigma}_1
\boldsymbol{\Sigma}_2^{1/2}
\right)^{1/2}
\right).
\end{equation}

Because BayesCrop uses diagonal covariance with spatially isotropic
$(x,y)$ variance and an independent crop-scale variance, the
component-wise Gaussian distance can be computed efficiently. Moreover,
our consistent left-to-right ordering establishes a fixed correspondence
between predicted and ground-truth GMM components. We therefore report a
mixture-weighted component-wise Wasserstein distance:
\begin{equation}
W_2^{\mathrm{weighted}}
=
\sum_{k=1}^{2}
\pi_k^{gt}
W_2
\left(
\mathcal{N}_k^{pred},
\mathcal{N}_k^{gt}
\right),
\end{equation}
where $\pi_k^{gt}$ denotes the ground-truth mixture weight of the
$k$-th component.

We report $W_2(x,y)$ using only the crop-center dimensions and
$W_2(x,y,r)$ using both crop center and crop scale. Lower values indicate
better component-wise agreement between the predicted and human-derived
crop distributions.

\subsubsection{\textbf{Best-of-$N$ Reference Coverage}}

To evaluate whether a set of predicted crops contains a plausible crop
consistent with the subjective reference, we report Best-of-$N$ IoU.
Given a candidate set
$\mathcal{C}_i=\{B_{i,1},\ldots,B_{i,N}\}$ for frame $i$, we compute
\begin{equation}
\mathrm{Best\text{-}of\text{-}}N
=
\frac{1}{n}
\sum_{i=1}^{n}
\max_{j\in\{1,\ldots,N\}}
\mathrm{IoU}
\left(
B_{i,j},B_i^{gt}
\right).
\end{equation}

For the two representative GMM means, this metric is reported as
Best-of-2. For probabilistic sampling experiments, $N$ denotes the total
number of sampled crop candidates. Best-of-$N$ measures \emph{set-level
reference coverage} rather than conventional single-output localization
accuracy; therefore, it is reported separately from mIoU.

\subsubsection{\textbf{Average Candidate IoU}}

Best-of-$N$ naturally increases as more candidates are considered.
Therefore, we additionally report the average IoU of all candidates to
verify that increased reference coverage is not obtained by introducing
arbitrary low-quality crops:
\begin{equation}
\mathrm{AvgIoU}
=
\frac{1}{n}
\sum_{i=1}^{n}
\frac{1}{N}
\sum_{j=1}^{N}
\mathrm{IoU}
\left(
B_{i,j},B_i^{gt}
\right).
\end{equation}

A stable AvgIoU as $N$ increases indicates that additional samples remain
plausible crop hypotheses even when they do not correspond to the
single reference annotation.

\subsubsection{\textbf{Intra-Mode IoU}}

To measure the local compactness of samples drawn from the same GMM
component, we compute the average pairwise IoU between candidates sampled
from each mode. Let
$\mathcal{C}_{i,k}$ denote the samples drawn from component $k$ for frame
$i$. The intra-mode IoU is
\begin{equation}
\mathrm{IoU}_{\mathrm{intra}}
=
\frac{1}{n}
\sum_{i=1}^{n}
\frac{1}{2}
\sum_{k=1}^{2}
\frac{1}{|\mathcal{P}_{i,k}|}
\sum_{(a,b)\in\mathcal{P}_{i,k}}
\mathrm{IoU}
\left(
B_{i,k}^{a},
B_{i,k}^{b}
\right),
\end{equation}
where $\mathcal{P}_{i,k}$ contains all evaluated pairs of samples from
the same component. Higher Intra-mode IoU indicates that samples remain
locally coherent around a common reframing mode.

\subsubsection{\textbf{Cross-Mode IoU}}

To measure separation between the two predicted reframing modes, we
compute the average IoU between samples drawn from different GMM
components:
\begin{equation}
\mathrm{IoU}_{\mathrm{cross}}
=
\frac{1}{n}
\sum_{i=1}^{n}
\frac{1}
{|\mathcal{C}_{i,1}||\mathcal{C}_{i,2}|}
\sum_{a\in\mathcal{C}_{i,1}}
\sum_{b\in\mathcal{C}_{i,2}}
\mathrm{IoU}
\left(
B_{i,1}^{a},
B_{i,2}^{b}
\right).
\end{equation}

Lower Cross-mode IoU indicates stronger spatial separation between the
two crop modes. In combination, high Intra-mode IoU and substantially
lower Cross-mode IoU indicate locally coherent yet spatially distinct
reframing hypotheses.

\subsubsection{\textbf{Frame-wise Aesthetic Quality
(VBench~\cite{huang2024vbench})}}

We use the Aesthetic Quality metric from VBench to evaluate the visual
appeal of the reframed video frames. The metric employs the LAION
aesthetic predictor trained on human aesthetic preferences
\cite{schuhmann2022laion}, capturing aspects of visual composition and
overall perceptual quality. Higher values indicate better aesthetic
quality.

\subsubsection{\textbf{Temporal Smoothness
(VBench~\cite{huang2024vbench})}}

We use the Temporal Smoothness metric from VBench to quantify temporal
consistency of the reframed videos. The metric measures the degree to
which adjacent frames maintain visually smooth transitions without
abrupt temporal changes. Higher values indicate greater temporal
stability.

\subsubsection{\textbf{Text--Video Alignment
(VBench~\cite{huang2024vbench})}}

We use the VBench Text--Video Alignment metric, which evaluates semantic
consistency between the video and its corresponding text description
using ViCLIP~\cite{wang2023internvid}. Higher scores indicate stronger
alignment between the reframed visual content and the semantic content
described by the caption.

\subsubsection{\textbf{Human Preference}}

We additionally evaluate subjective reframing quality using a three-way
human preference study. We randomly select 20 videos from the 114 videos
in the Multi-Target Long Video Benchmark for this evaluation.
Participants are shown reframed outputs produced by the compared methods
for the same source video and are asked to select the output they prefer
based on preservation of important visual content and the quality of the
presented viewpoints.

For each method, the preference rate is computed from its received votes
relative to the total number of valid comparisons. Higher preference
indicates that viewers more frequently favor the corresponding reframing
result.

\subsubsection{\textbf{Inference Efficiency}}

We evaluate inference efficiency in frames per second (FPS) on a
workstation equipped with an NVIDIA RTX 4090 GPU (24 GB VRAM).
Throughput is measured using the end-to-end video reframing pipeline,
including neural crop prediction on anchor frames, compressed-domain
motion propagation on intermediate frames, boundary correction, and
temporal smoothing. The reported FPS therefore reflects system-level
processing throughput rather than predictor-only inference speed.

\section{Appendix: LIVE-YT-MC Quality Evaluation and Data Analysis}
\subsection{Quality Evaluation}
\begin{table}[t]
\centering
\caption{Comparison of Dataset Quality across LIVE-YT benchmarks. We report the VBench Aesthetic Quality (Aes.) and Temporal Smoothness (Temp.) scores to evaluate the baseline characteristics of each corpus.}
\label{tab:dataset_quality_comparison}
\begin{tabular}{lcc}
\toprule
\textbf{Dataset} & \textbf{Aes.} $\uparrow$ & \textbf{Temp.} $\uparrow$ \\
\midrule
\textbf{LIVE-YT-VC} \cite{lee2024subjective} & \underline{46.7} & 90.4 \\
\textbf{LIVE-YT-VC++} \cite{lee2024subjective} & \textbf{46.8} & \underline{92.8} \\
\midrule
\textbf{LIVE-YT-MC} (Left View) & 45.3 & \textbf{94.8} \\
\textbf{LIVE-YT-MC} (Right View) & 45.3 & \textbf{94.8} \\
\bottomrule
\end{tabular}
\end{table}
Table \ref{tab:dataset_quality_comparison} compares the dataset quality across three LIVE-YT benchmarks using VBench metrics. LIVE-YT-VC++ achieves the highest aesthetic score (46.8) and an improvement in temporal smoothness (92.8) by introducing a frame distance-aware weighting and an intra-frame temporal filter algorithm mechanism. This strategy prevents warping errors from distant reference frames and incorporates explicit frame smoothing to ensure perceptually seamless trajectories.

In contrast, LIVE-YT-MC does not utilize distance-aware weighting, leading to a slight decrease in aesthetic quality (45.3) due to minor warping noise, though the aesthetic score remains competitive. Crucially, while LIVE-YT-VC and LIVE-YT-VC++ are limited to single-view representations, LIVE-YT-MC is the first to support dual-view annotation, providing a more comprehensive capture of multi-focal narratives. Despite the slight drop in aesthetic score, LIVE-YT-MC achieves the superior temporal smoothness score of 94.8. This is attributed to our multi-focal approach that separates the narrative into distinct Left and Right views; by reducing the spatial degrees of freedom for each individual box, the model generates more stable tracking paths across the sequence.

\subsection{Box Area Analysis}
\begin{figure}[t]
\begin{center}
 \includegraphics[width=1.0\linewidth]{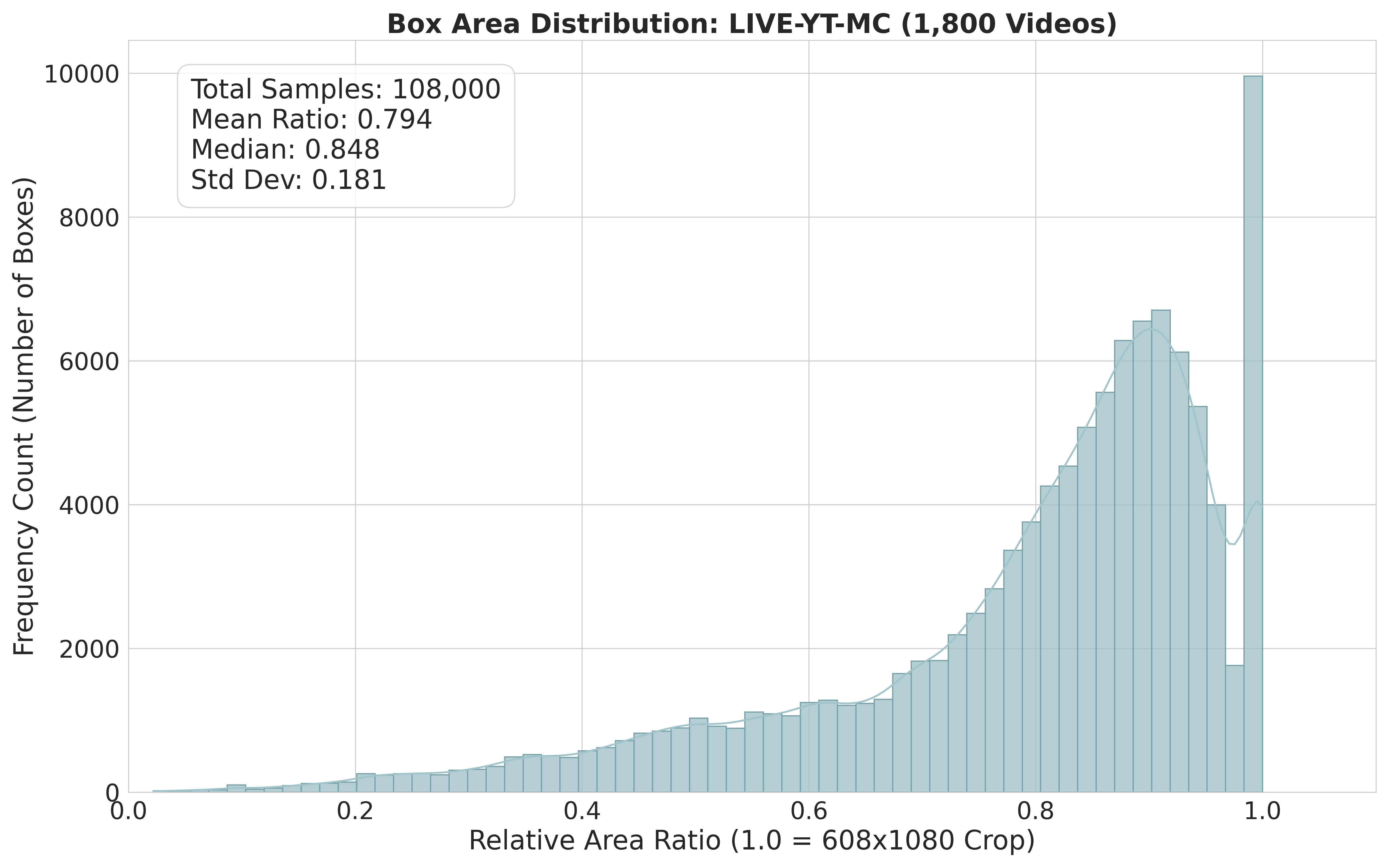}
\end{center}
\caption{Box area analysis of LIVE-YT-MC}
\label{fig:long}
\label{fig:onecol}
\label{fig_area}
\end{figure}  
Fig. \ref{fig_area} illustrates the distribution of relative box areas across the LIVE-YT-MC dataset, comprising 108,000 samples from 1,800 videos. The area ratios, normalized against a standard 9:16 crop ($608 \times 1080$), exhibit a high mean ratio of 0.794 and a median of 0.848, indicating that the majority of ground-truth ROIs occupy a significant portion of the target frame. The distribution is notably right-skewed, with a prominent peak near 1.0, suggesting that human annotators frequently prioritize large, inclusive crops to capture complex cinematic scenes. However, the standard deviation of 0.181 and the presence of smaller ratios (down to 0.2) demonstrate the dataset's diversity, encompassing a wide range of focal scales from tight close-ups to expansive wide shots. This variance underscores the necessity for a flexible reframing model capable of adapting to diverse aesthetic and narrative requirements.

\subsection{Box Overlap Analysis}
\begin{figure}[t]
\begin{center}
 \includegraphics[width=1.0\linewidth]{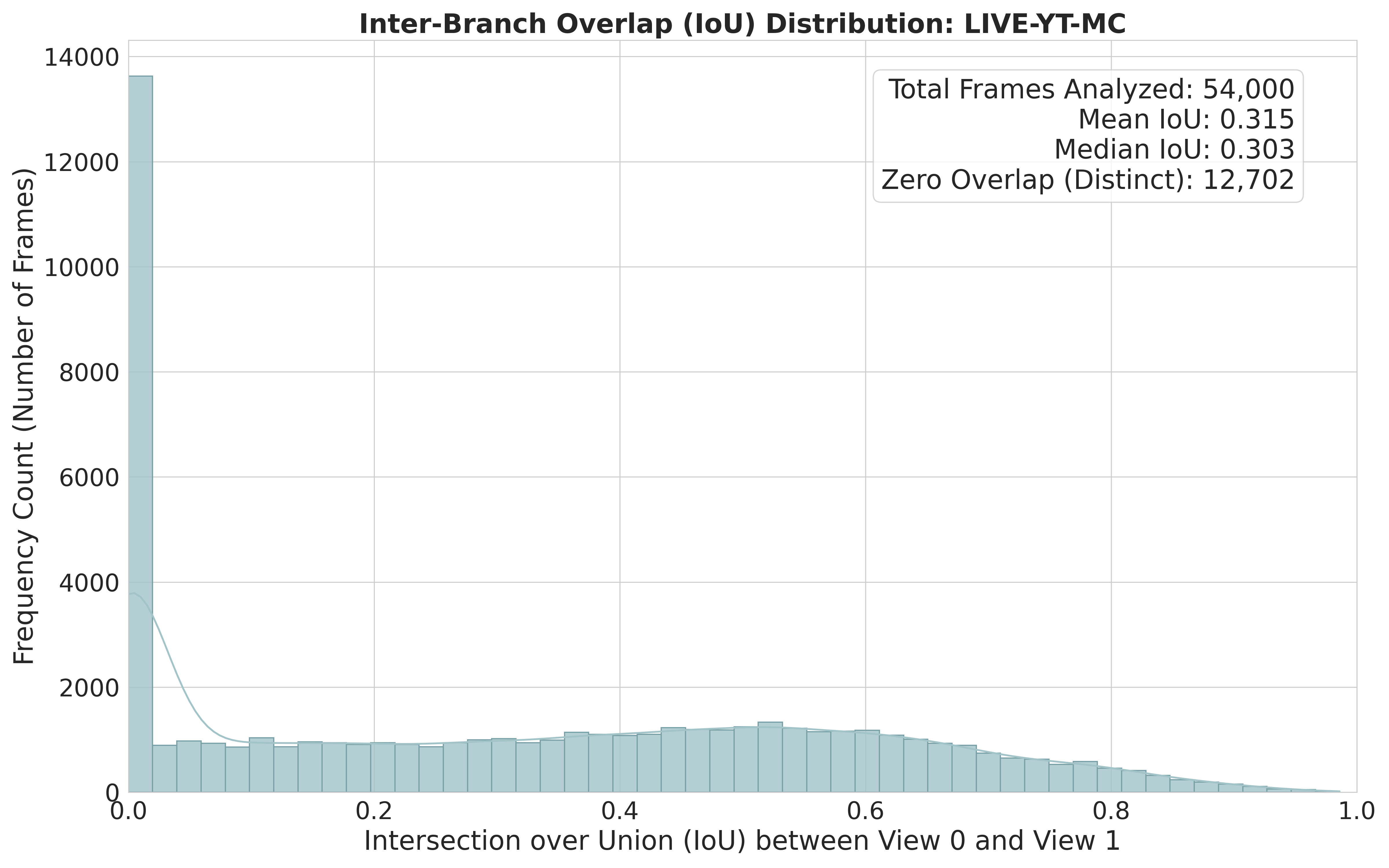}
\end{center}
\caption{Box overlap analysis of LIVE-YT-MC}
\label{fig:long}
\label{fig:onecol}
\label{fig:overlap}
\end{figure}  
Fig. \ref{fig:overlap} presents the Inter-Branch Overlap (IoU) distribution for the LIVE-YT-MC dataset, analyzing the geometric relationship between the two primary narrative views (View 0 and View 1) across 54,000 frames. The distribution reveals a mean IoU of 0.315 and a median of 0.303, indicating that while the two crops often share some visual context, they primarily focus on distinct spatial regions. Notably, 12,702 frames (approximately 23.5\%) exhibit zero overlap, signifying entirely disjoint subjects or viewpoints within the same cinematic scene. This high frequency of distinct, non-overlapping ROIs justifies our use of a multi-focal GMM framework, as it confirms that single-crop regression models are insufficient for capturing the diverse, multi-target narratives prevalent in professional cinematic content.

\section{Appendix: Evaluation on RetargetVid Benchmark}

\begin{table}[t]
\centering
\caption{
Quantitative comparison on the RetargetVid~\cite{apostolidis2021fast} benchmark. Deterministic baselines are evaluated using standard single-output mIoU. For BayesCrop, we additionally report $\hat{\pi}$-based single-view selection and Best-of-2 IoU, where the latter measures set-level reference coverage over the two predicted crop hypotheses. Best-of-2 IoU is therefore not directly comparable to conventional single-output mIoU. All outputs are standardized to the benchmark setting using a fixed $1{:}3$ aspect ratio while preserving the original frame height.
}
\label{tab:sota_iou_only}
\begin{tabular}{lc}
\toprule
\textbf{Method} & \textbf{IoU} $\uparrow$ \\
\midrule
AutoFlip & 50.8 \\
SmartVidCrop (w/o Clustering) \cite{apostolidis2021fast} & 46.1 \\
SmartVidCrop \cite{apostolidis2021fast} & 49.9 \\
FEVR \cite{gu2025dynamic} & \underline{54.9} \\
\midrule
\textbf{BayesCrop} (Left View, $R=5$) & 43.1 \\
\textbf{BayesCrop} (Right View, $R=5$) & 45.8 \\
\textbf{BayesCrop} ($\hat{\pi}$-selected, $R=5$) & 46.3 \\
\textbf{BayesCrop} (Best-of-2, $R=5$) & \textbf{61.4} \\
\bottomrule
\end{tabular}
\end{table}

Table~\ref{tab:sota_iou_only} evaluates BayesCrop on RetargetVid. FEVR achieves the strongest single-output result with 54.9 mIoU, while $\hat{\pi}$-based BayesCrop selection reaches 46.3 mIoU. SmartVidCrop obtains 49.9 mIoU, with its saliency model trained using RetargetVid data. In contrast, BayesCrop is trained from human-derived probabilistic annotations on LIVE-YT-MC, while FEVR benefits from large-scale MLLM-generated pseudo-labels and video-level training.

When both BayesCrop hypotheses are retained, Best-of-2 IoU increases substantially to 61.4. This indicates that the model often generates a candidate consistent with the reference, even when the predicted mixture weights do not rank that candidate first. The gap between $\hat{\pi}$-based selection and Best-of-2 coverage therefore suggests that the multi-crop representation transfers more robustly than single-view preference ranking. Differences in resolution and target aspect ratio between RetargetVid and our training setting may further contribute to this gap.

Best-of-2 IoU is reported as a set-level reference coverage metric and is not directly comparable to conventional single-output mIoU.

\section{Appendix: Optional MLLM-Based View Switch Agent}
\label{sec:mllm_switch}
\begin{table}[t]
\centering
\caption{
Comparison of BayesCrop view-switching strategies on the Multi-Target
Long Video Benchmark. Both methods use the same two BayesCrop trajectories
and differ only in the temporal view-selection policy.
}
\label{tab:mllm_switch}
\resizebox{\linewidth}{!}{
\begin{tabular}{lccc}
\toprule
\textbf{View-Switch Strategy}
& \textbf{Aesthetic} $\uparrow$
& \textbf{Temp. Smooth.} $\uparrow$
& \textbf{Text--Video Align.} $\uparrow$ \\
\midrule
Periodic View Switch Agent
& 54.7 & 96.4 & \textbf{24.2} \\
MLLM-Based View Switch Agent
& 54.7 & 96.4 & 24.0 \\
\bottomrule
\end{tabular}}
\end{table}
As shown in Table~\ref{tab:mllm_switch}, the MLLM-based planner achieves
nearly identical Aesthetic Quality and Temporal Smoothness to the
Periodic View Switch Agent, while slightly reducing Text--Video Alignment
from 24.2 to 24.0. These results suggest that the additional benefit of
semantic view planning is limited under the current configuration.

We note, however, that the effectiveness of the MLLM-based planner depends
on both prompt design and the video-understanding capability of the
underlying MLLM. Different prompts, stronger video-capable models, or more
explicit temporal reasoning may lead to different view-selection behavior
and potentially improved performance.

Because the Periodic View Switch Agent introduces negligible overhead,
requires no prompt engineering, and provides comparable performance in our
evaluation, we adopt it as the default strategy in the main paper. The
MLLM-based planner is therefore presented as an optional downstream
extension rather than a core component of BayesCrop.

\section{Appendix: More Visual Results on the Multi-Target Long Video Benchmark}

In Fig.~\ref{fig:main_qualitative_comparison_part1}, we provide additional
qualitative results on representative sequences from the MiraData
\cite{ju2024miradata} dataset. We compare CenterCrop, the direct Gemini 3
Flash cropping baseline, and BayesCrop combined with the optional
MLLM-based View Switch Agent described in Sec.~\ref{sec:mllm_switch}.

For these visualizations, BayesCrop first generates two temporally coherent
crop trajectories corresponding to alternative plausible reframing views.
Gemini 3 Flash then acts only as a high-level temporal planner, selecting
which BayesCrop trajectory to use over different temporal segments. It does
not directly predict or modify the crop coordinates. The resulting
mixed-view examples illustrate how the dual-trajectory representation can
support content-adaptive view switching between different subjects or
regions of interest while retaining the spatial predictions produced by
BayesCrop.

We emphasize that the MLLM-based View Switch Agent is an optional downstream
extension rather than part of the default BayesCrop pipeline. The main paper
uses the lightweight Periodic View Switch Agent for quantitative evaluation
and real-time operation, while these supplementary examples demonstrate the
flexibility of BayesCrop's dual-trajectory representation for higher-level
semantic shot planning.

For a more comprehensive qualitative evaluation across diverse scenes and
temporal transitions, we provide additional video demonstrations in our project repository: \textcolor{blue}{https://github.com/steven413d/BayesCrop}.

\begin{figure*}[t]
    \centering
    \def\fw{0.15\textwidth} 
    \setlength{\tabcolsep}{1pt} 
    
    \begin{subfigure}{\textwidth}
        \centering
        \begin{tabular}{ccccccc}
             & \textbf{3s} & \textbf{6s} & \textbf{9s} & \textbf{12s} & \textbf{15s} & \textbf{18s} \\
            \rotatebox{90}{\footnotesize Center Crop} &
            \includegraphics[width=\fw]{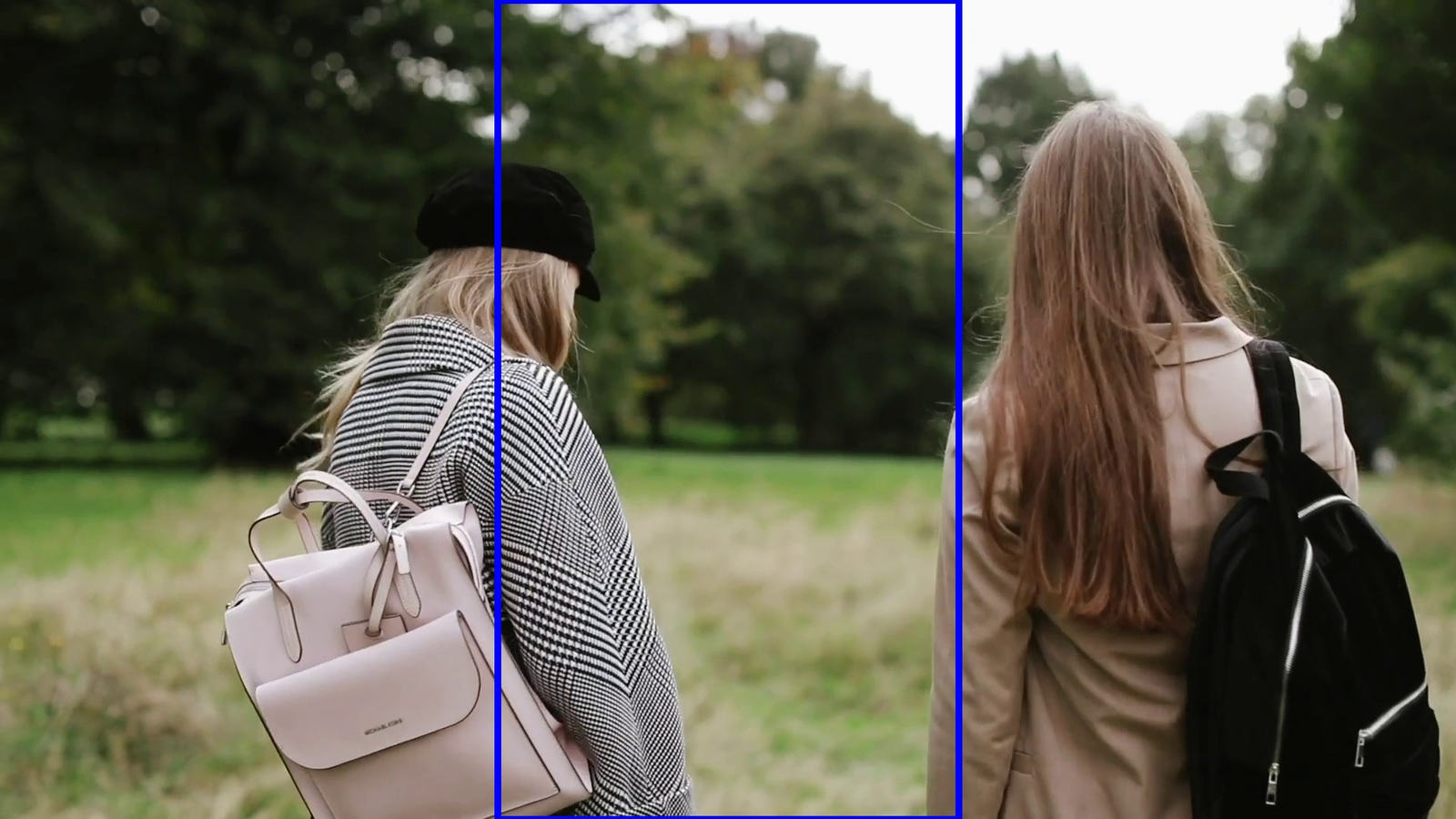} &
            \includegraphics[width=\fw]{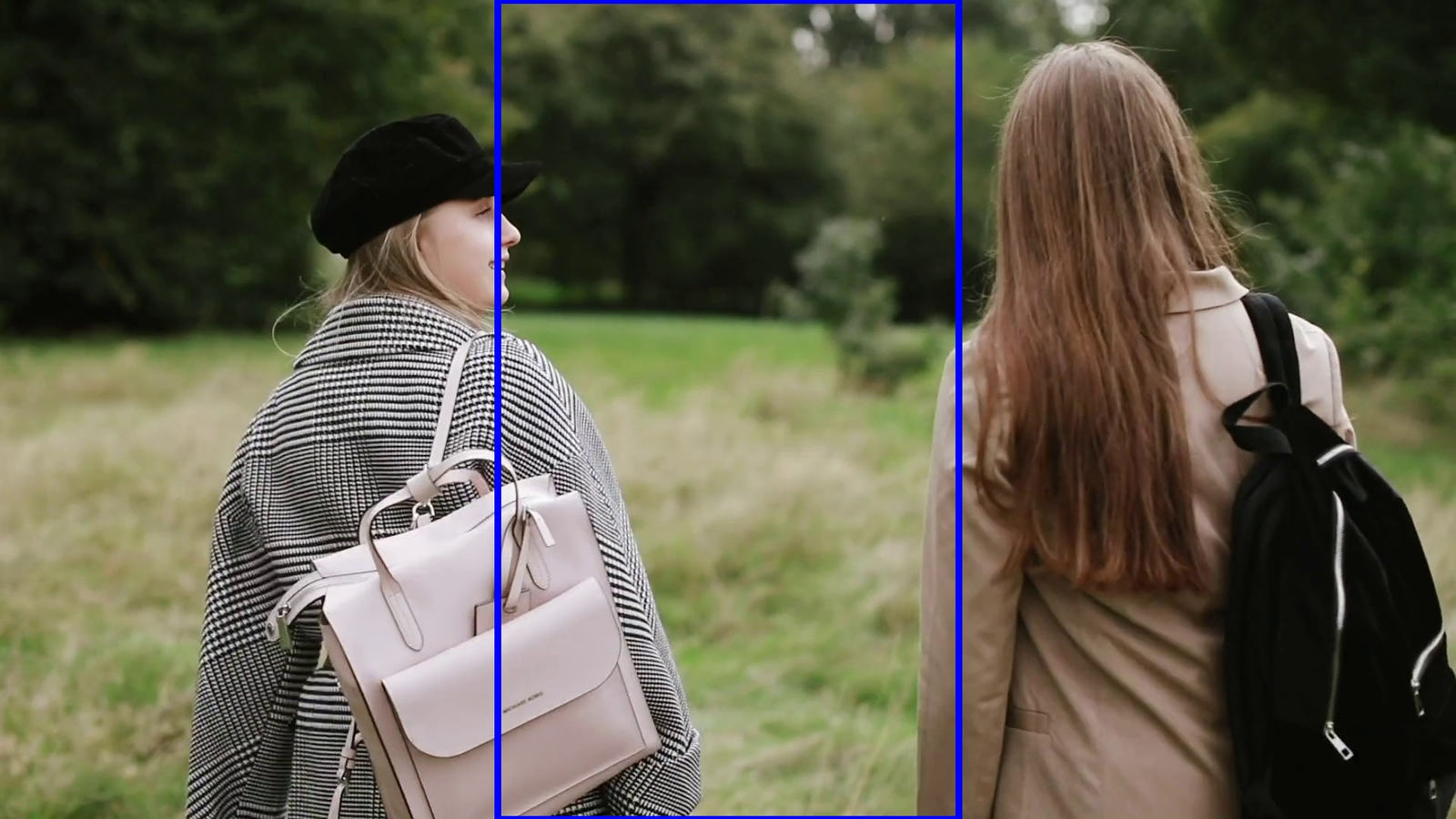} &
            \includegraphics[width=\fw]{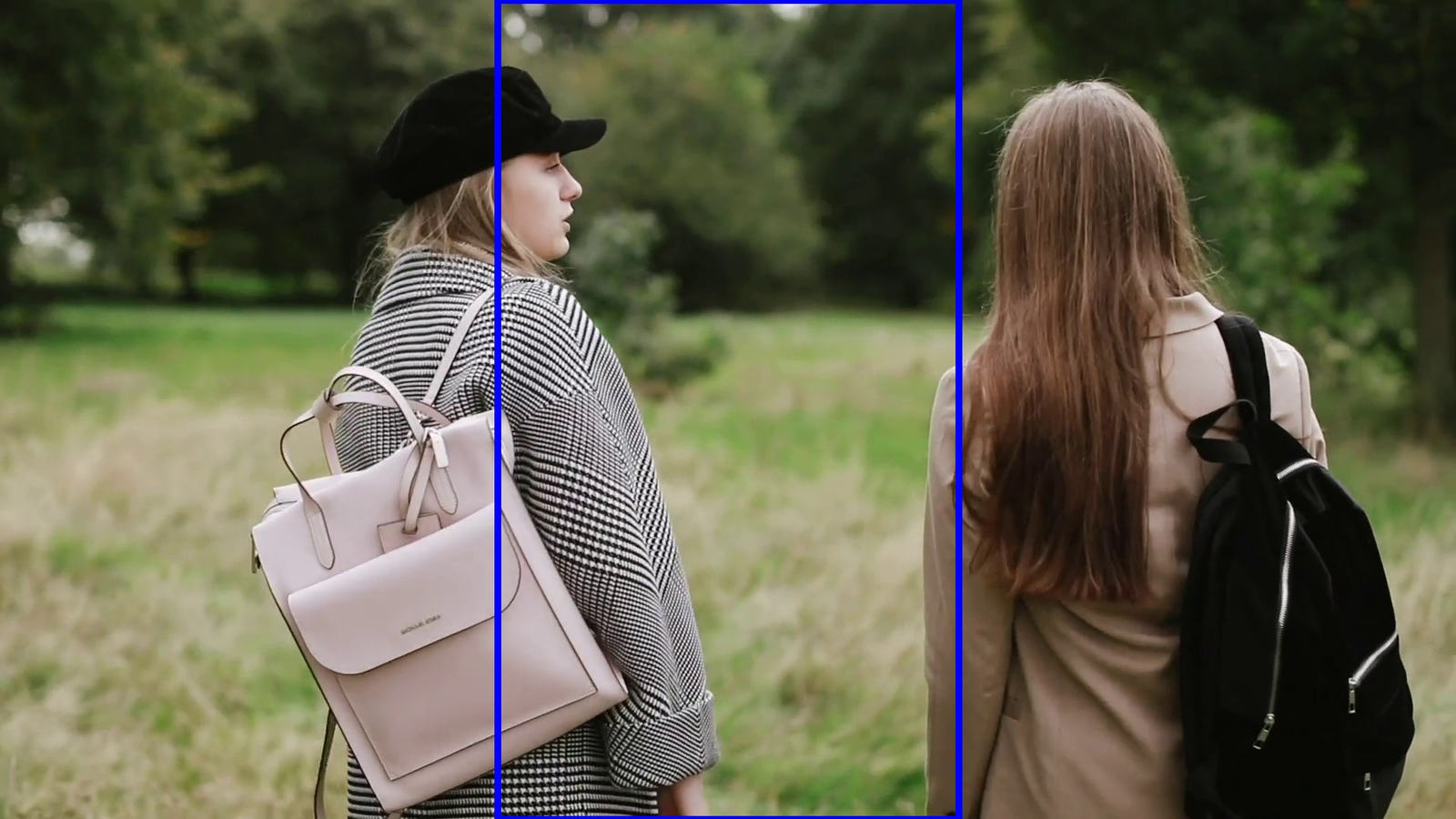} &
            \includegraphics[width=\fw]{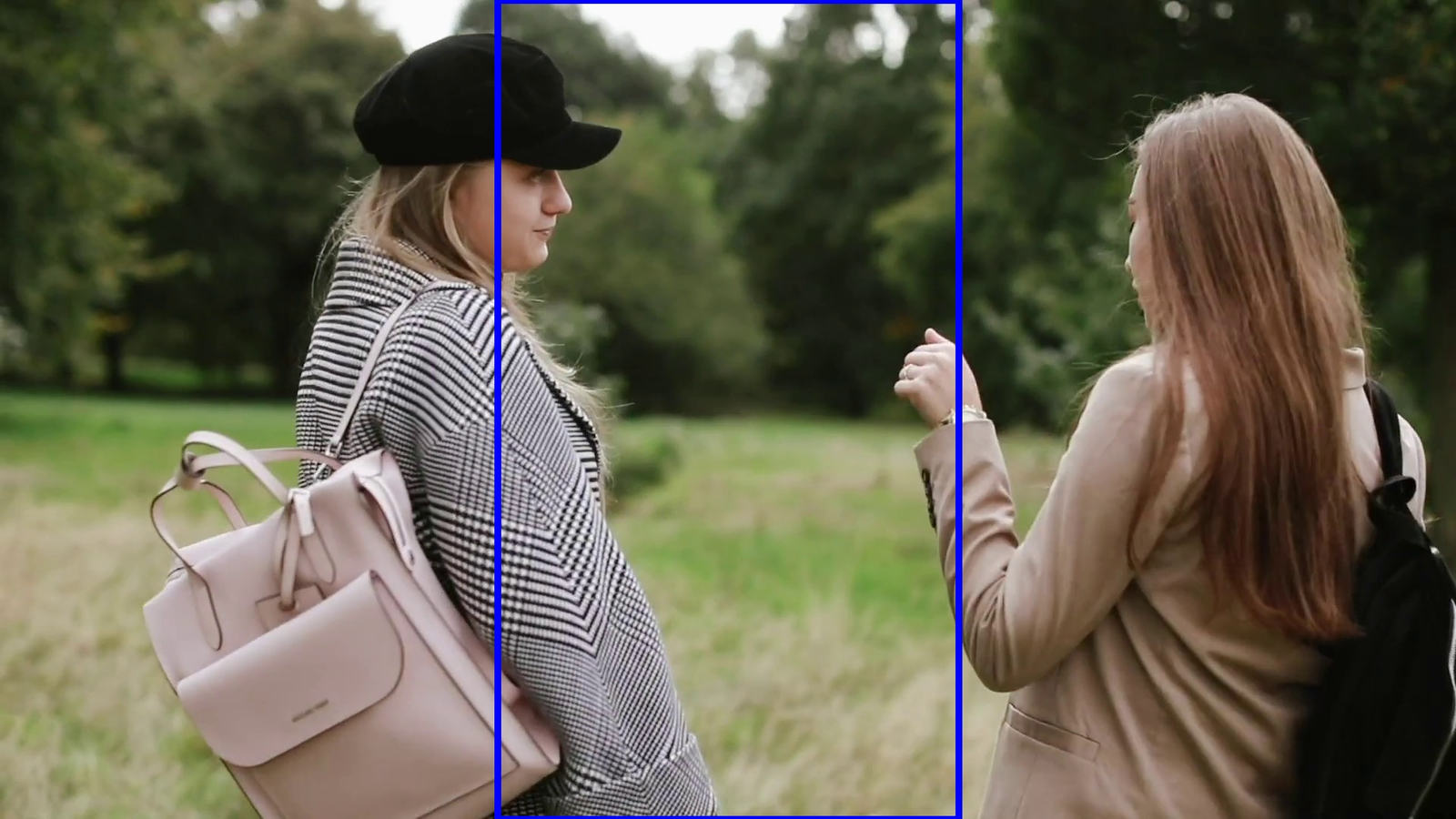} &
            \includegraphics[width=\fw]{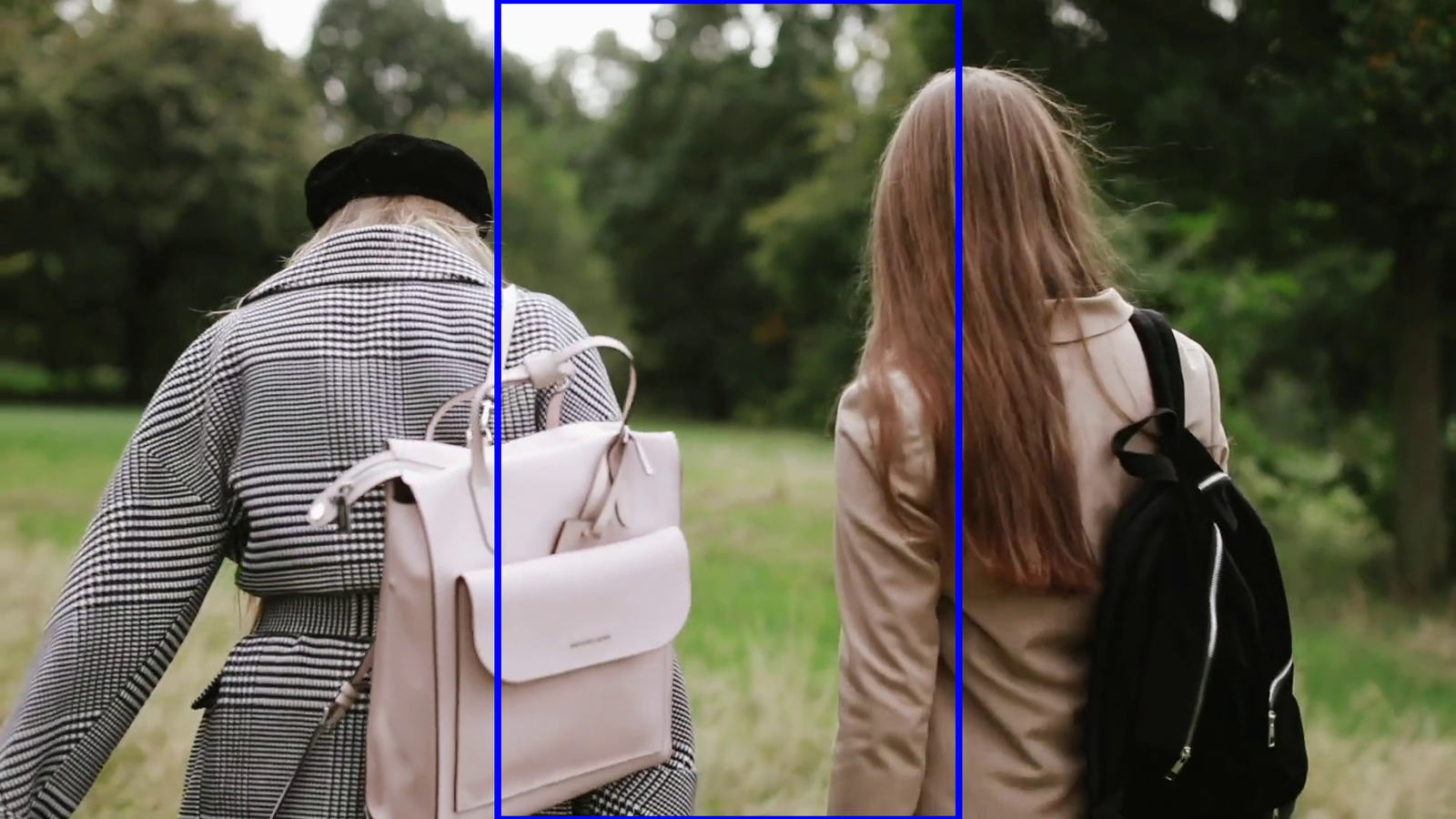} &
            \includegraphics[width=\fw]{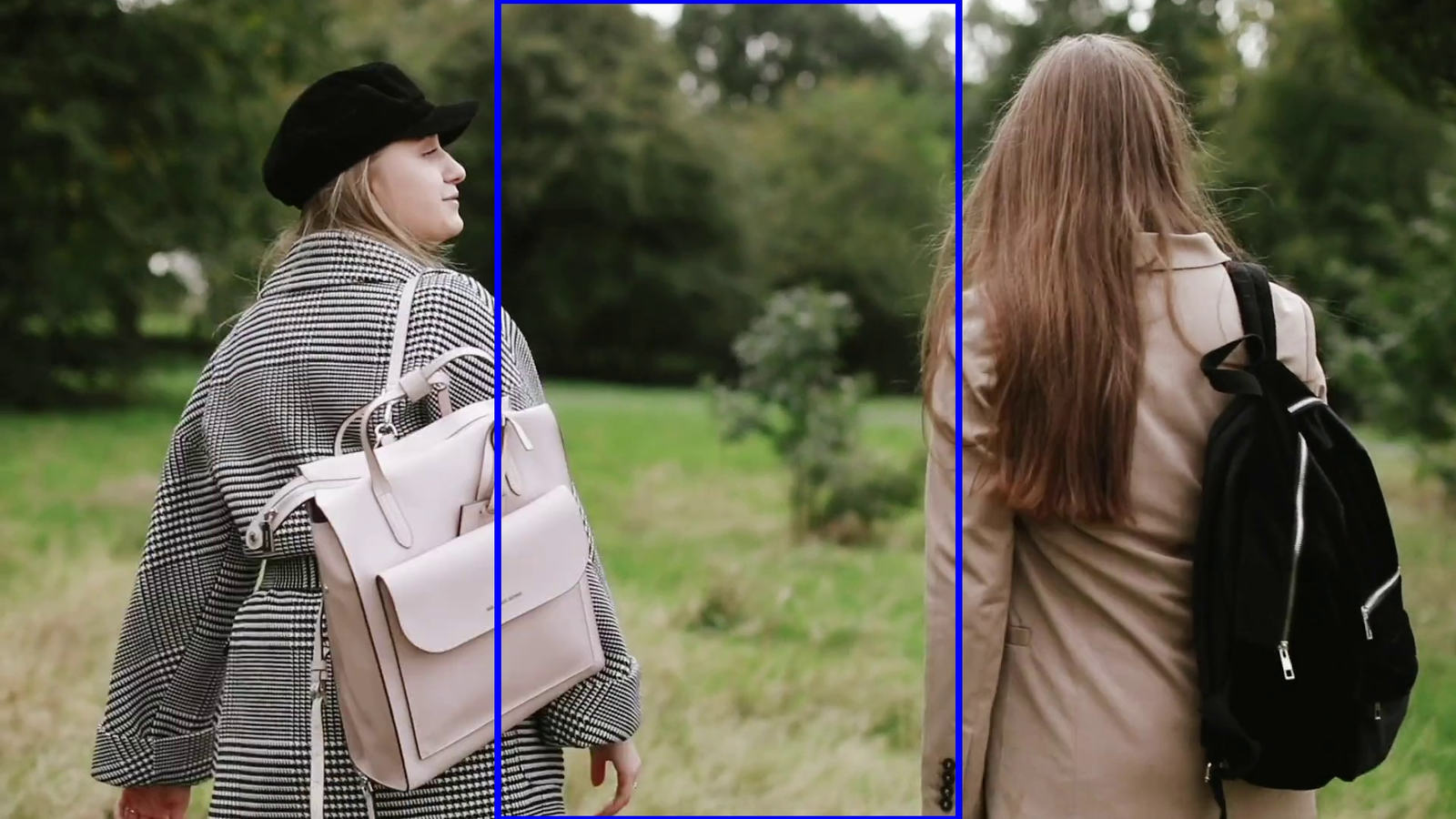} \\
            \rotatebox{90}{\footnotesize Gemini-Base} &
            \includegraphics[width=\fw]{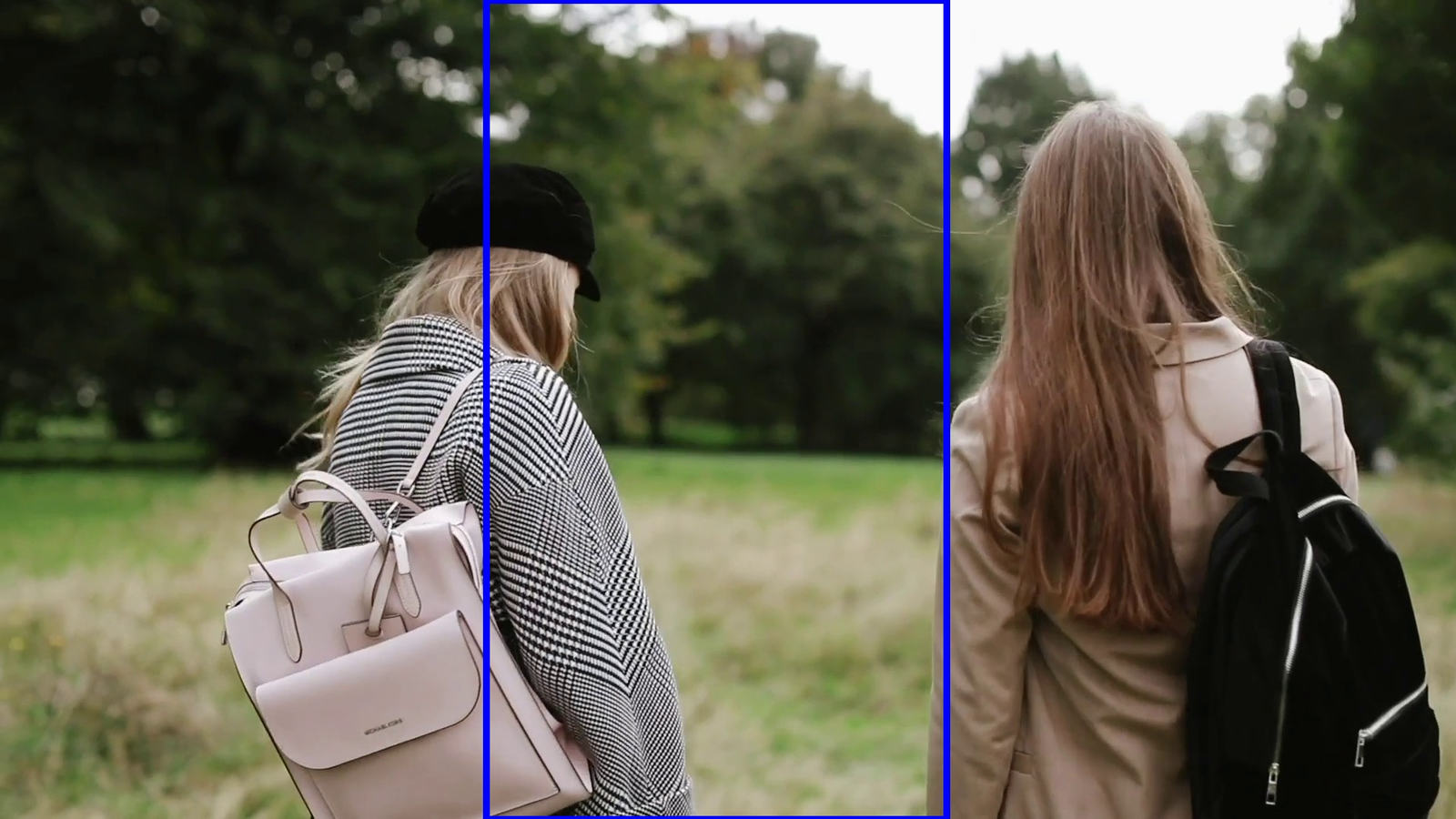} &
            \includegraphics[width=\fw]{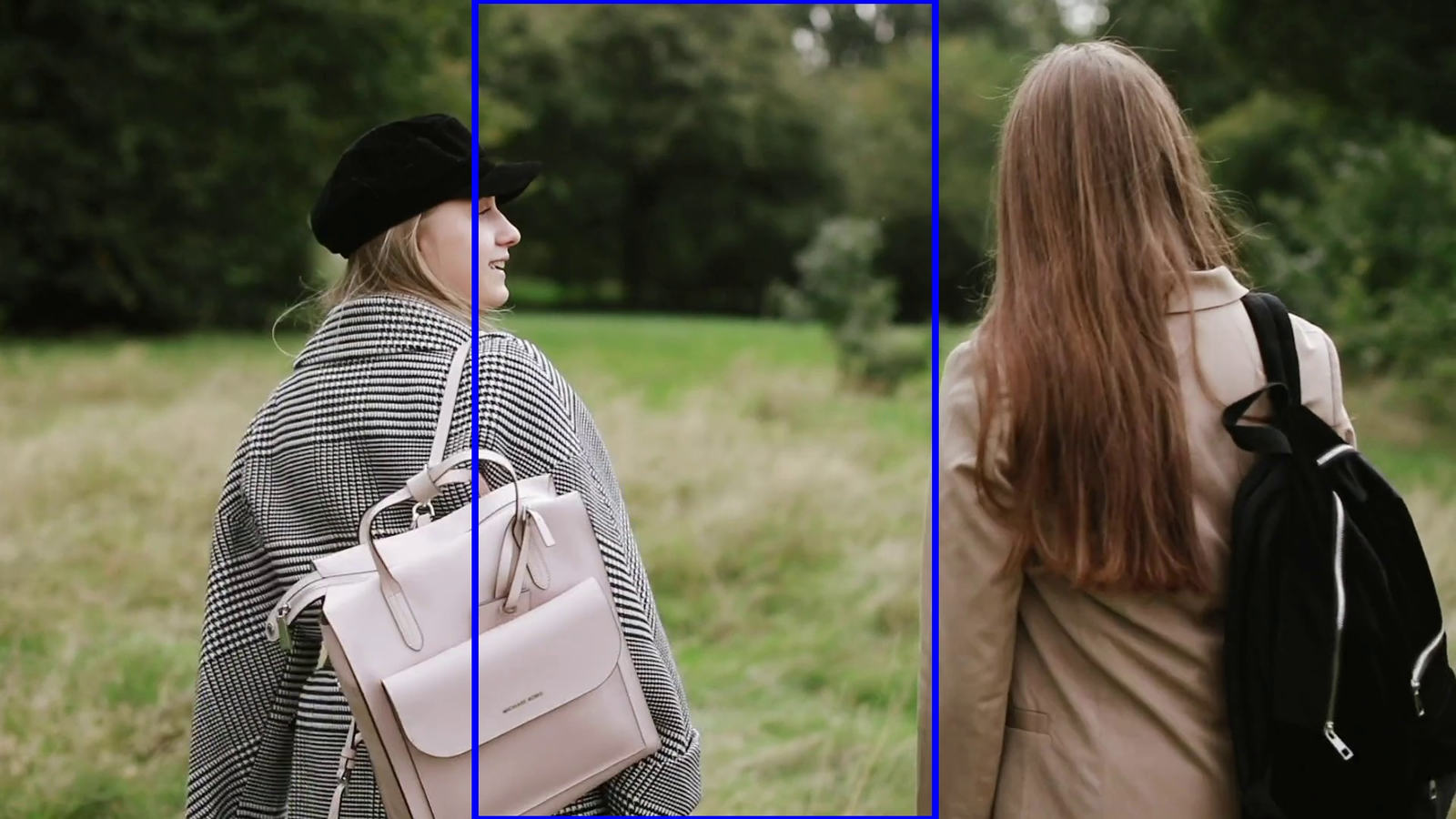} &
            \includegraphics[width=\fw]{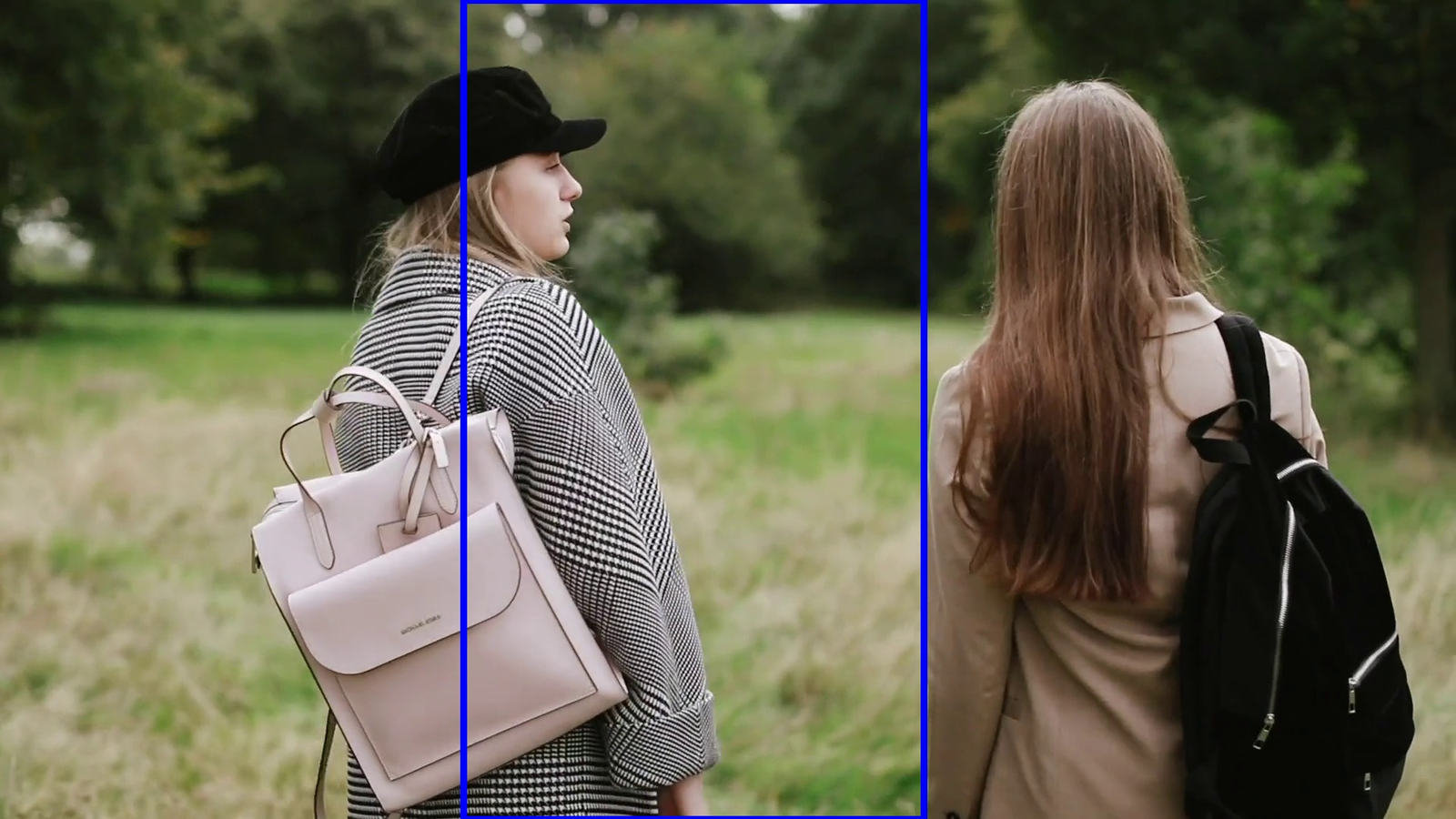} &
            \includegraphics[width=\fw]{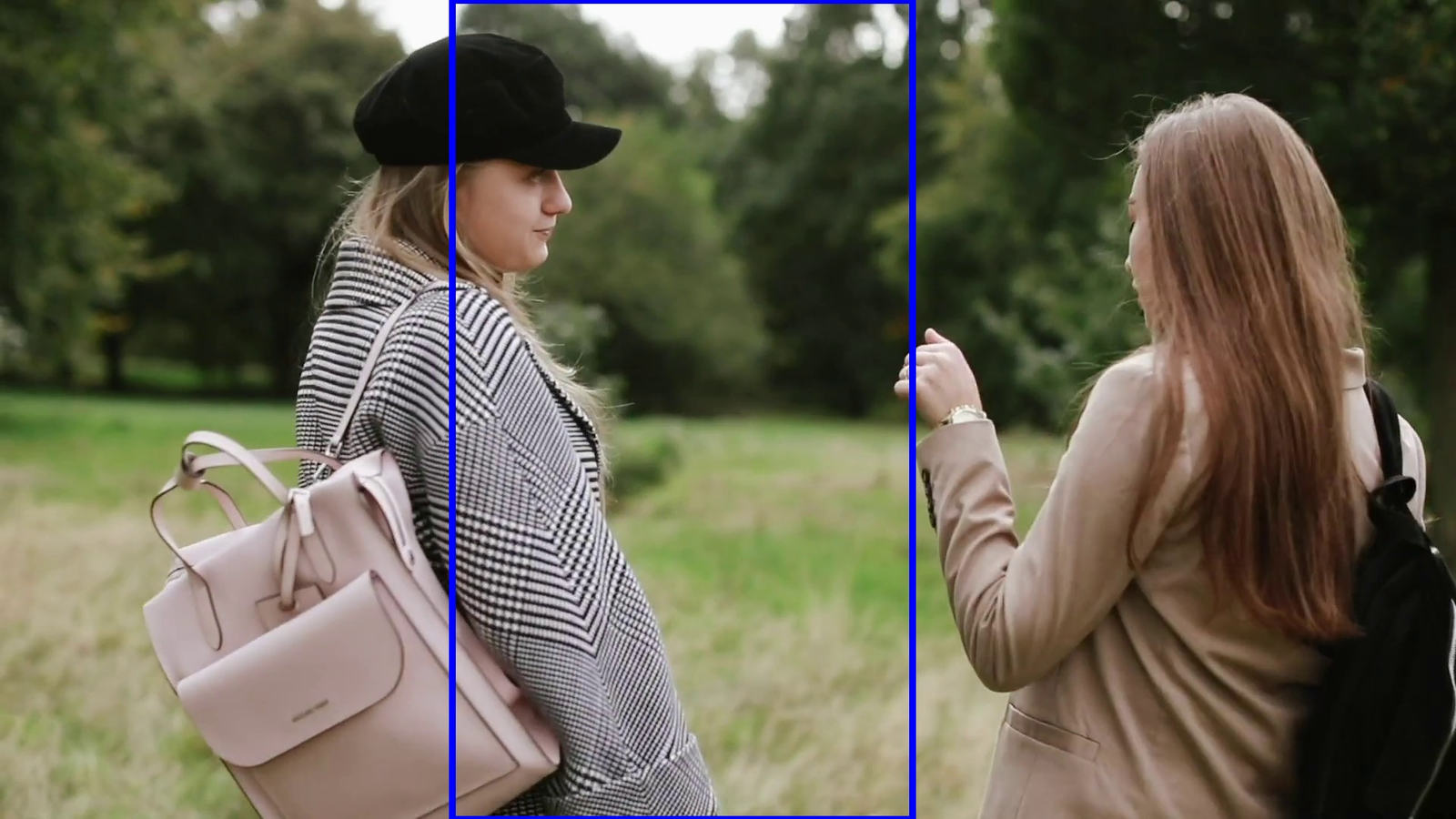} &
            \includegraphics[width=\fw]{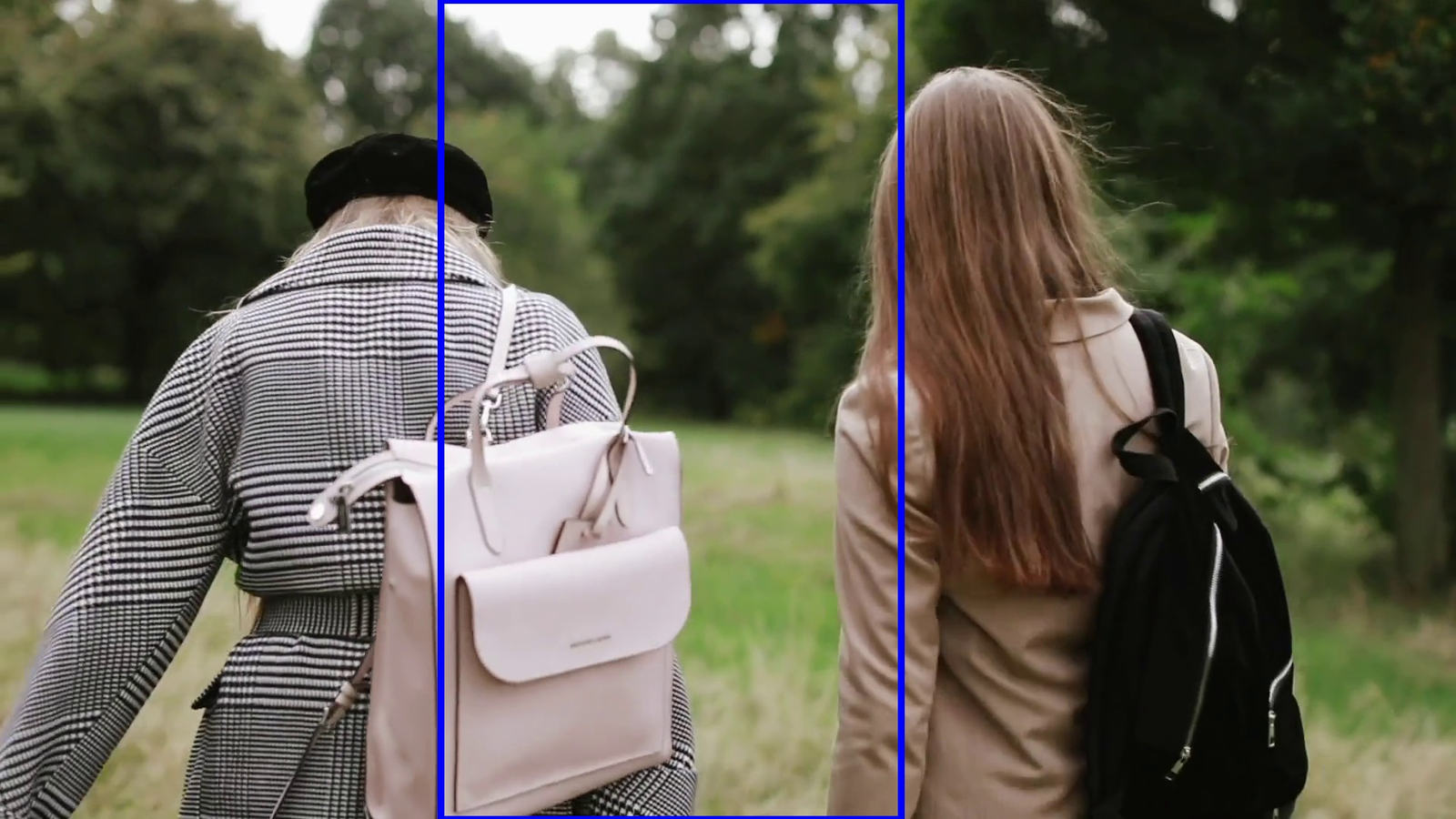} &
            \includegraphics[width=\fw]{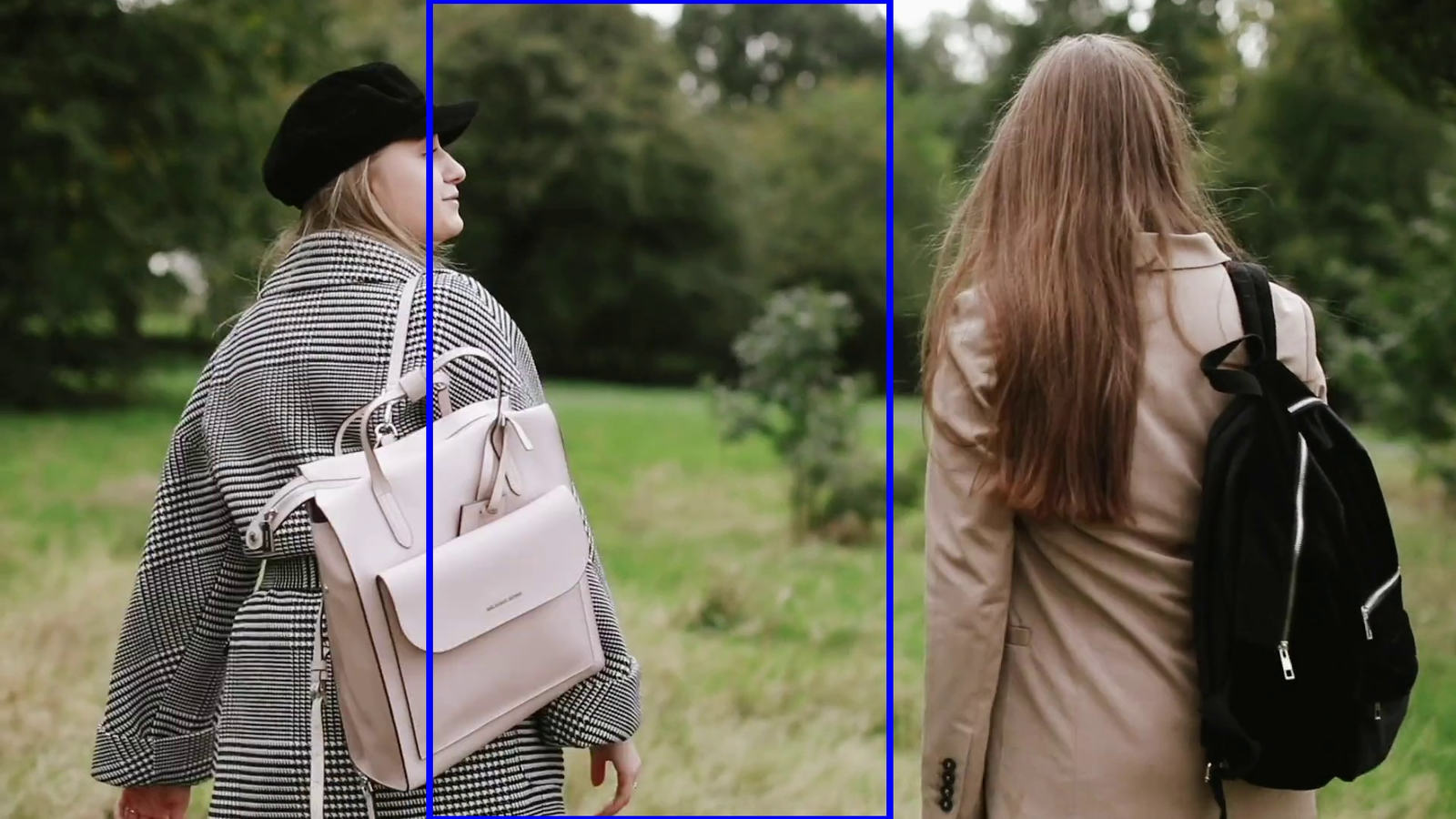} \\
            \rotatebox{90}{\footnotesize Ours} &
            \includegraphics[width=\fw]{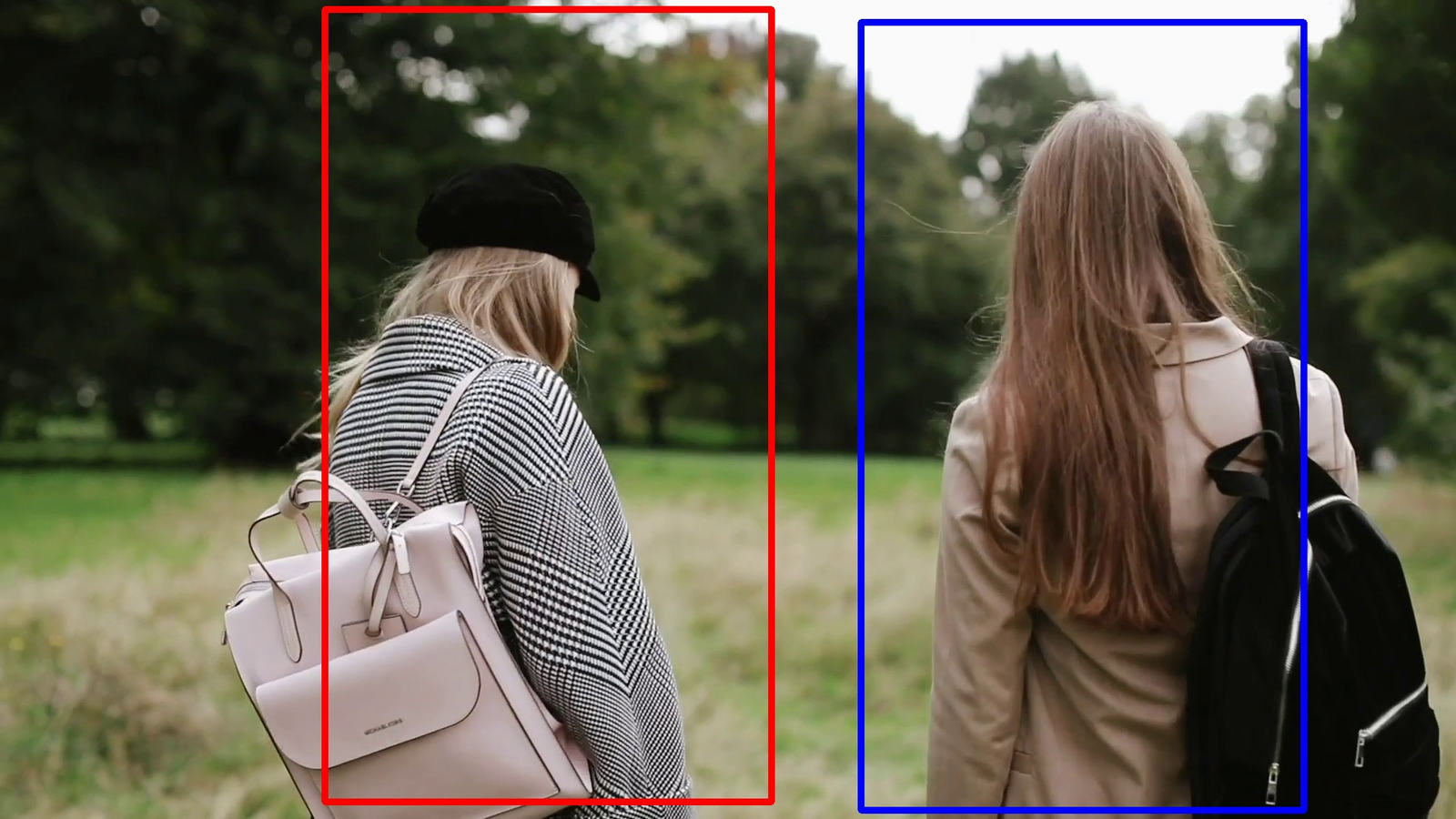} &
            \includegraphics[width=\fw]{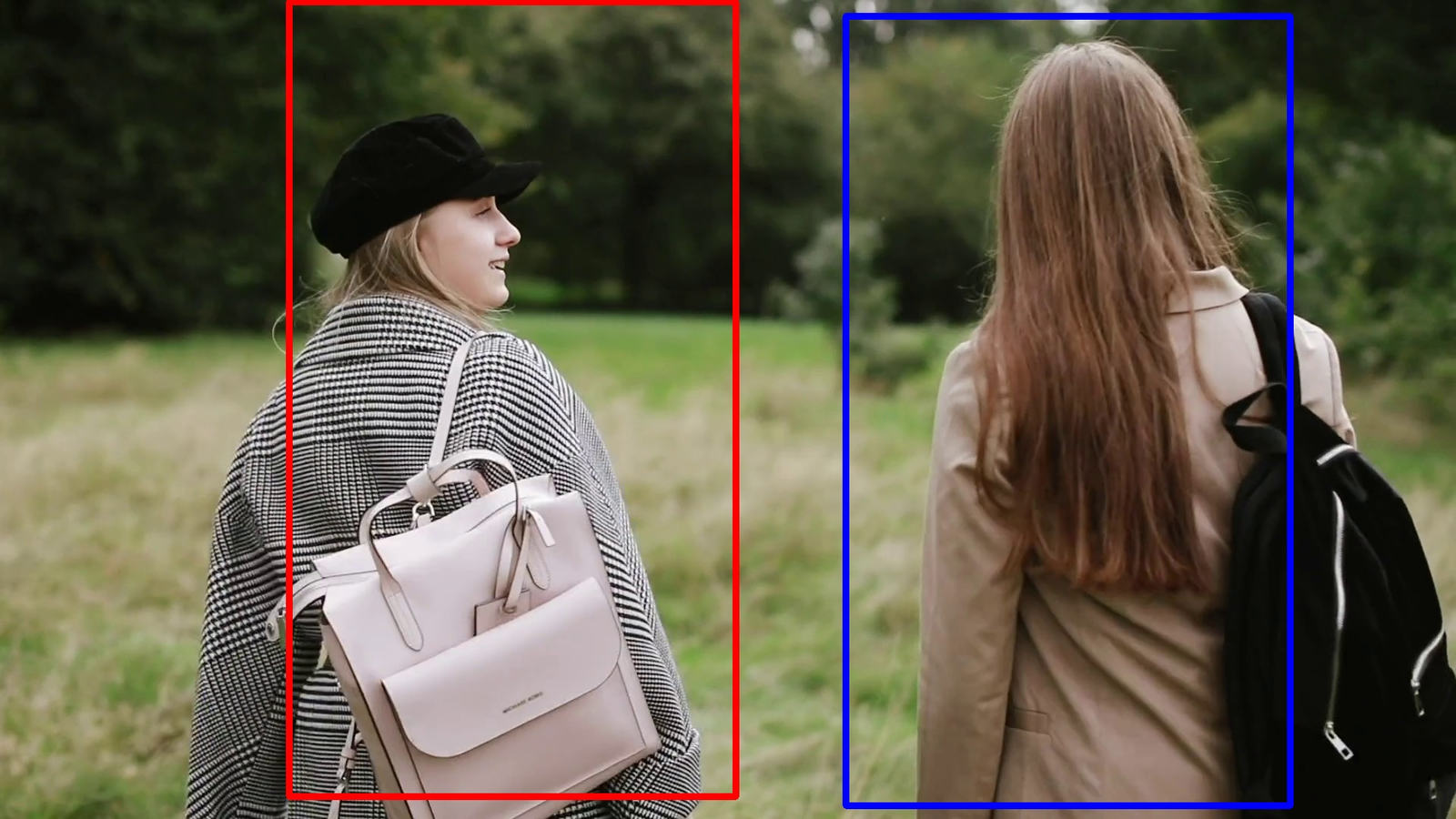} &
            \includegraphics[width=\fw]{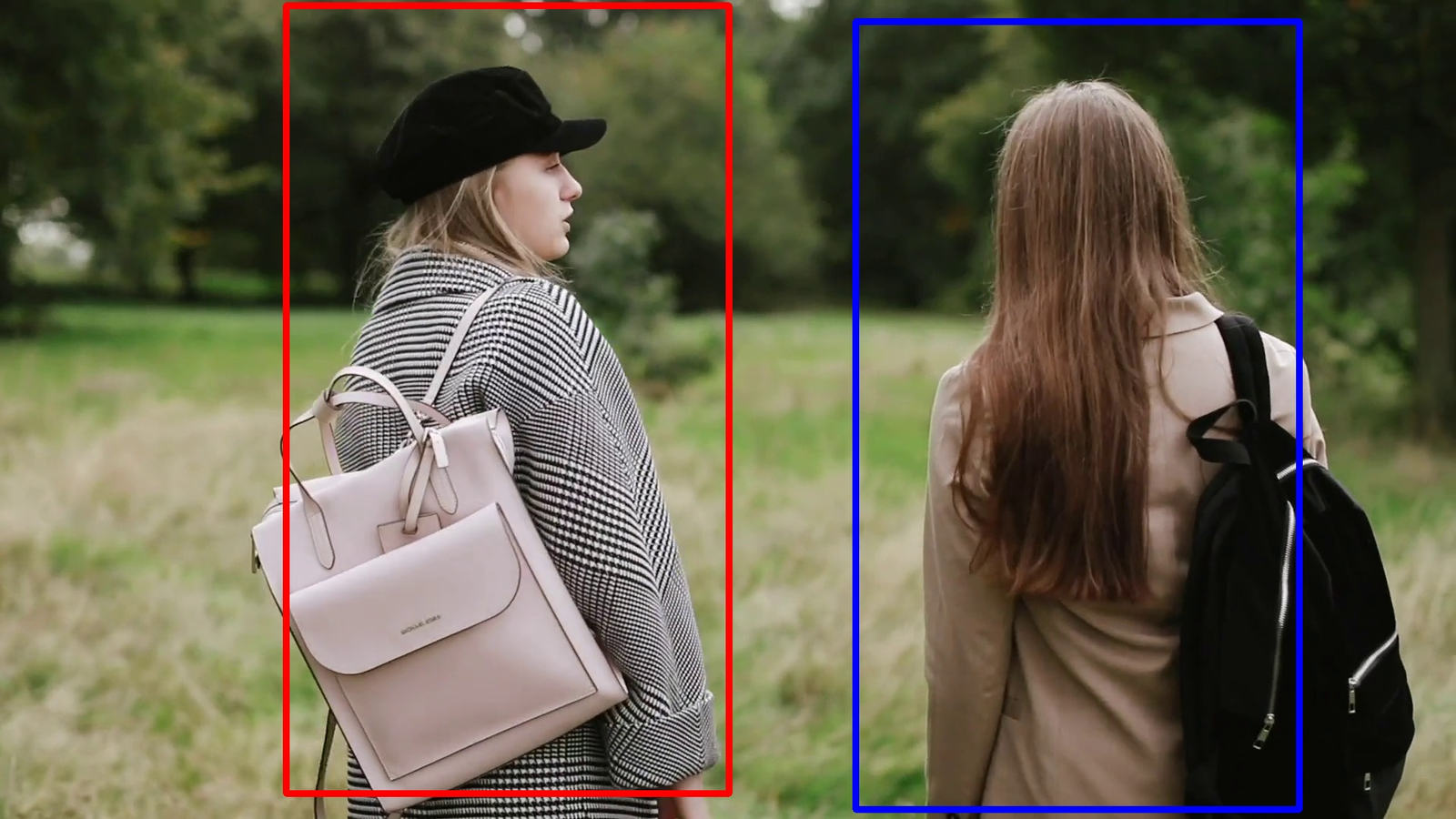} &
            \includegraphics[width=\fw]{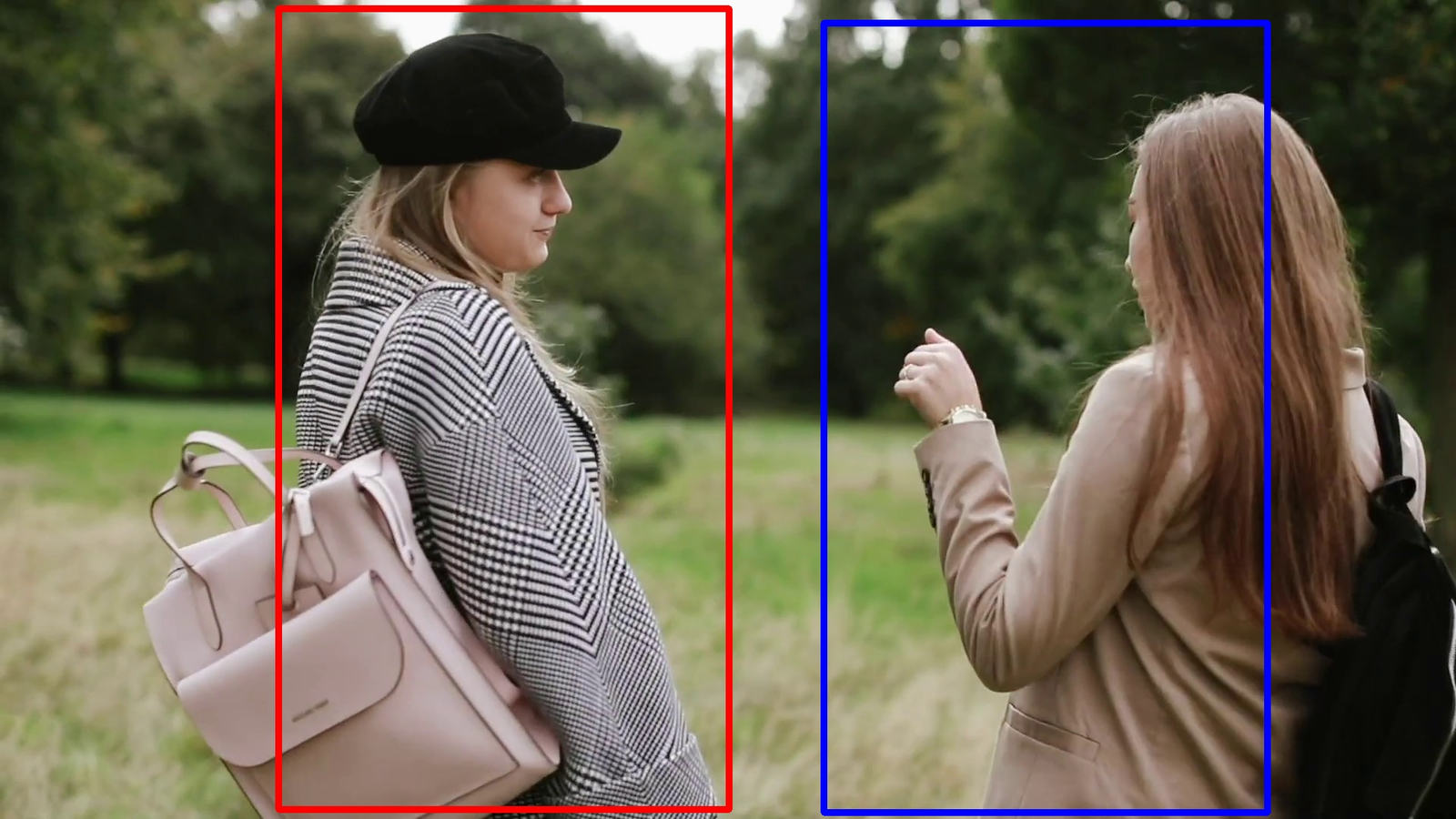} &
            \includegraphics[width=\fw]{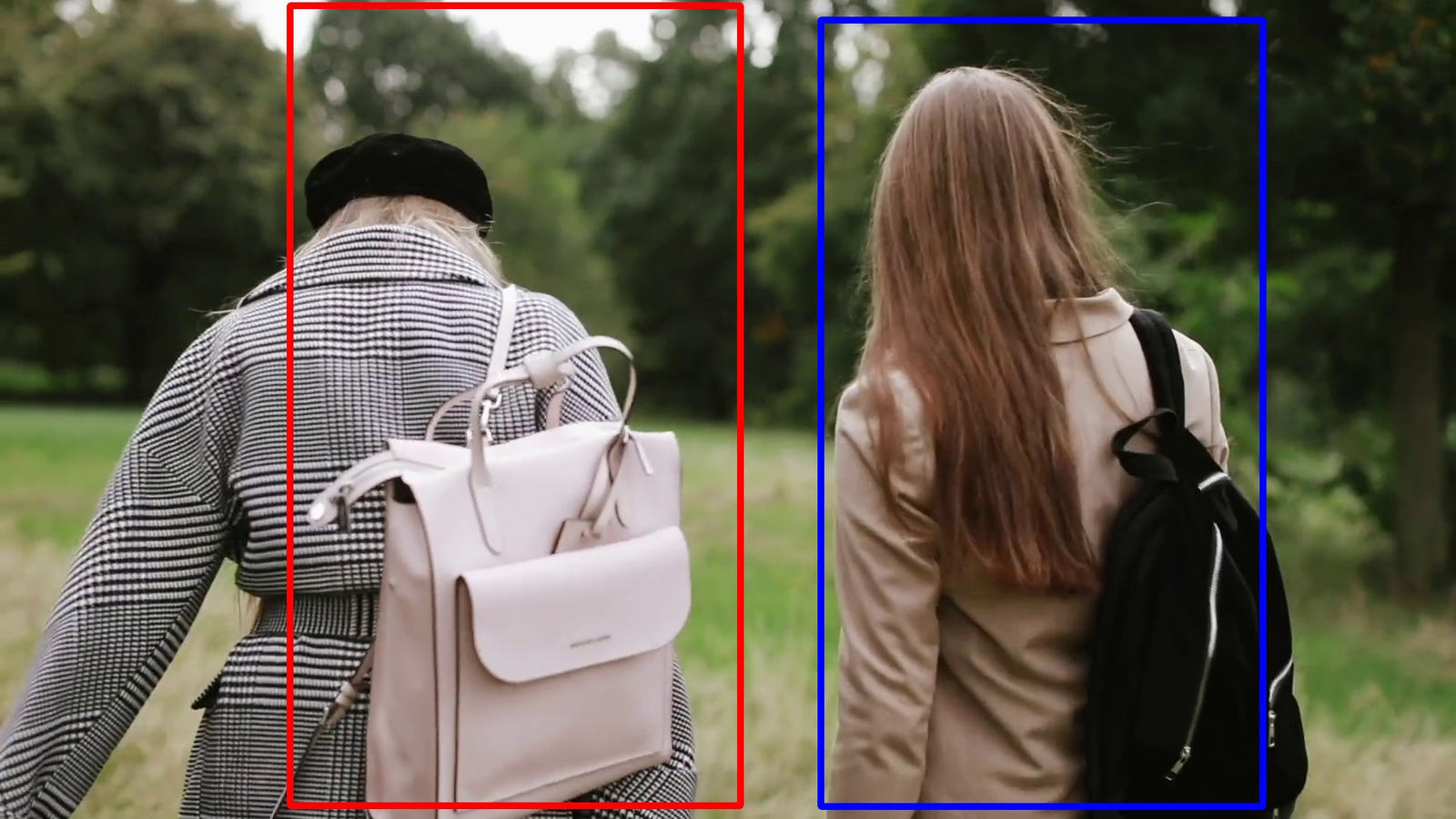} &
            \includegraphics[width=\fw]{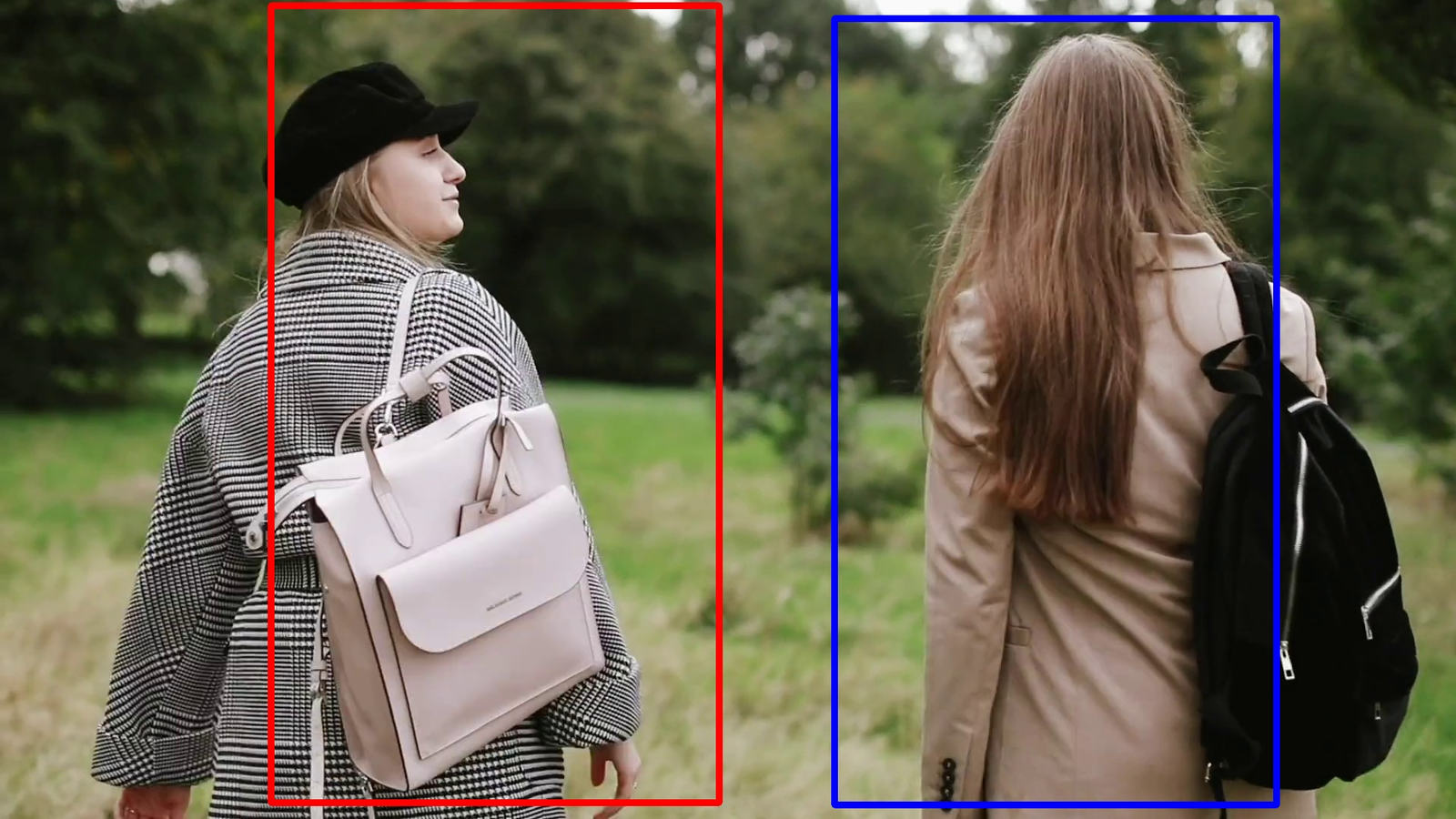} \\
            \rotatebox{90}{\footnotesize Ours + MLLM-based Agent} &
            \includegraphics[width=\fw]{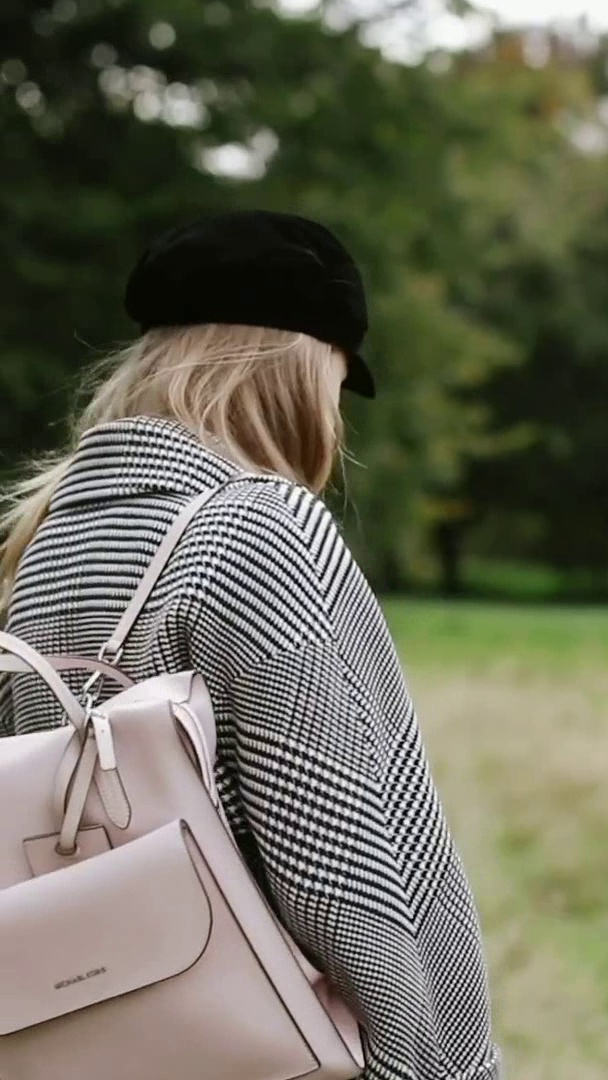} &
            \includegraphics[width=\fw]{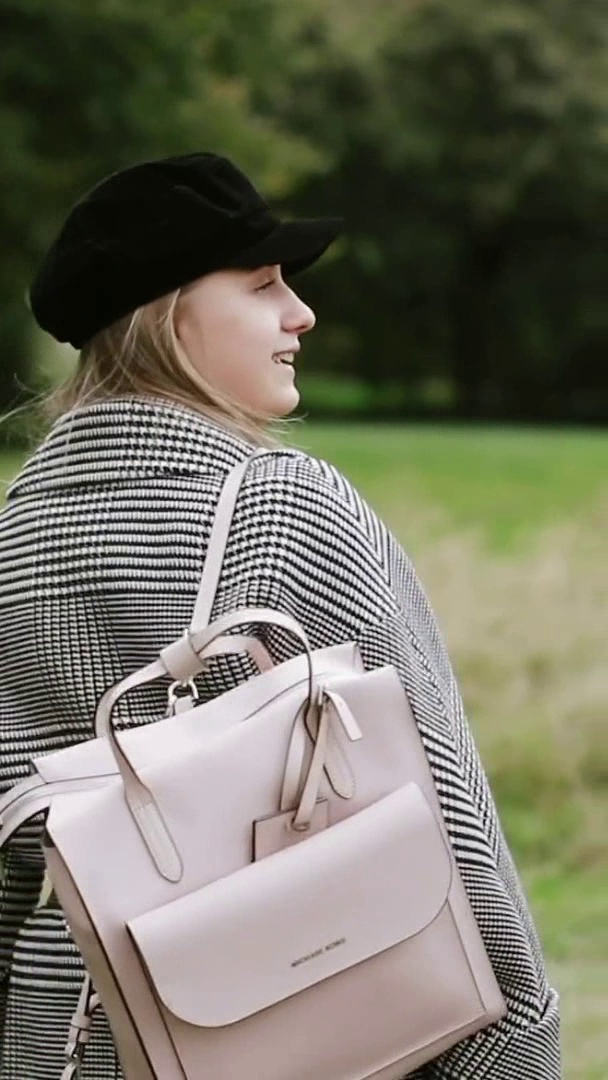} &
            \includegraphics[width=\fw]{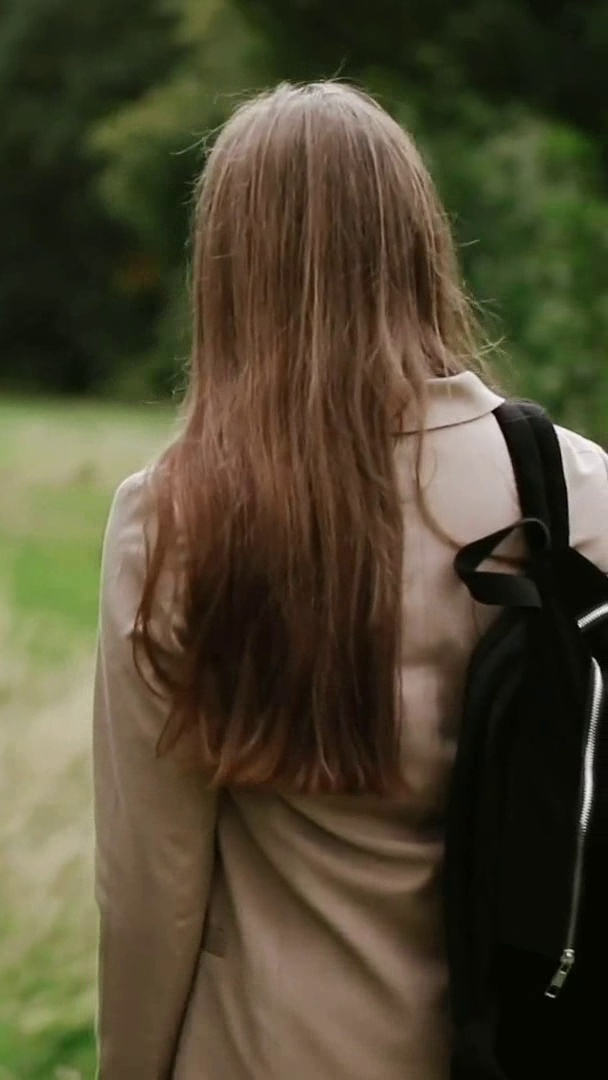} &
            \includegraphics[width=\fw]{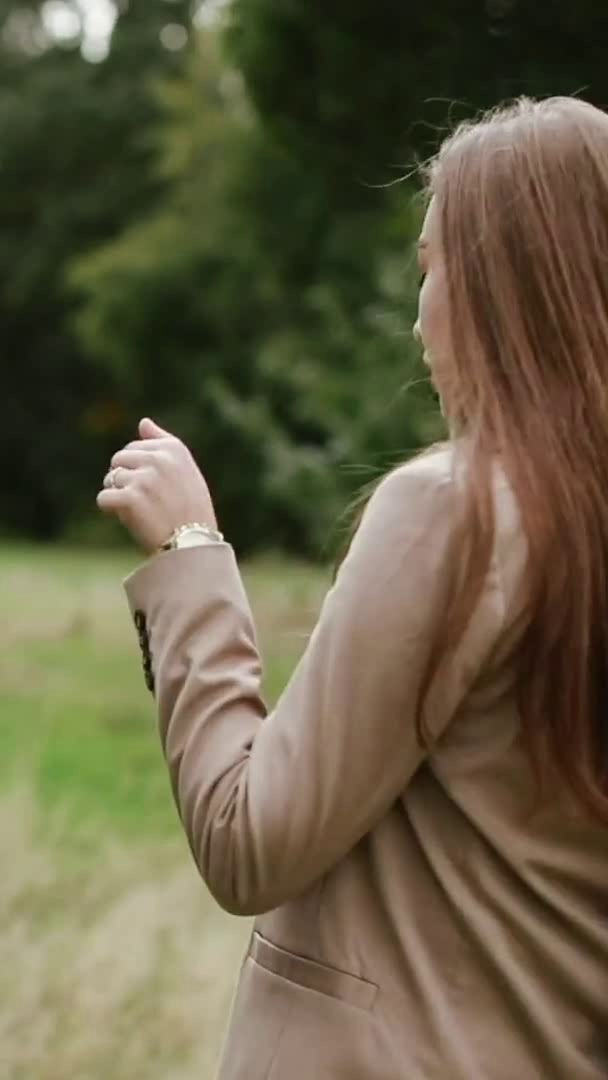} &
            \includegraphics[width=\fw]{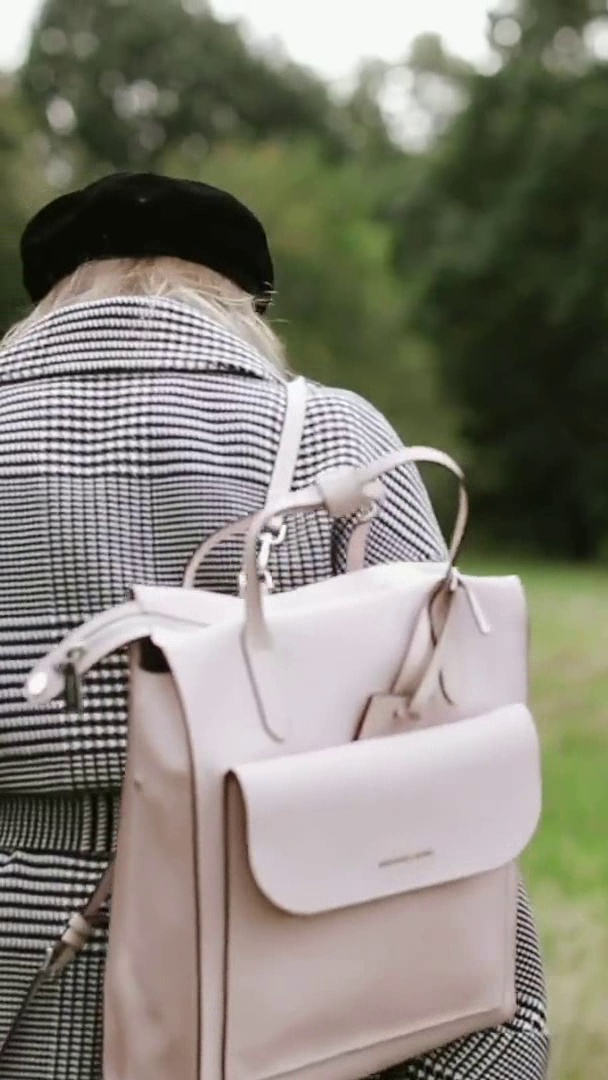} &
            \includegraphics[width=\fw]{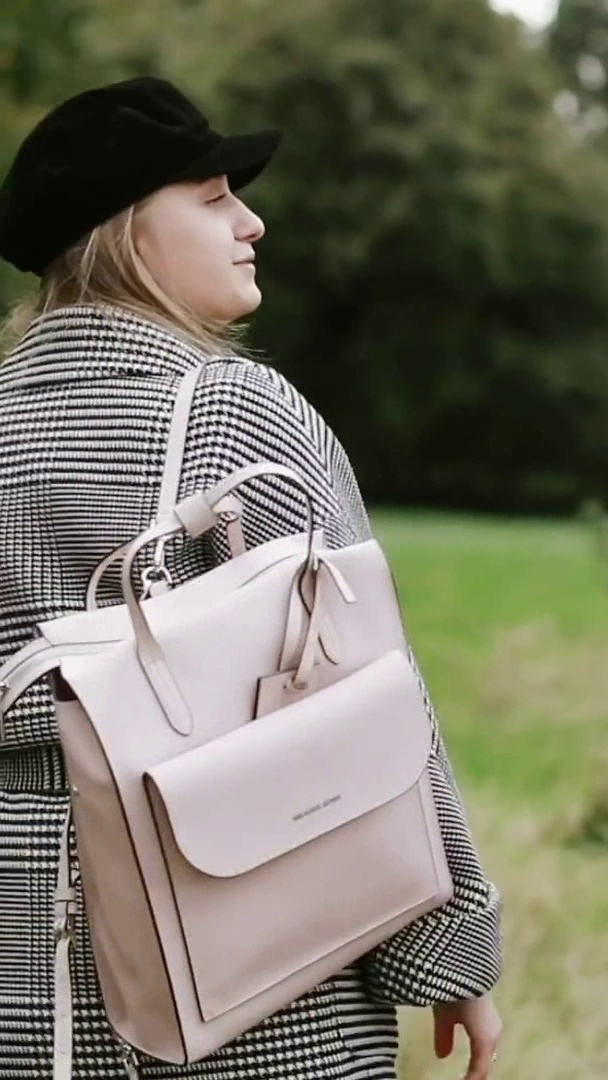} \\
        \end{tabular}
        \caption{Video ID: 4013457}
    \end{subfigure}
    \vspace{-10pt}

    \begin{subfigure}{\textwidth}
        \centering
        \begin{tabular}{ccccccc}
             & \textbf{3s} & \textbf{6s} & \textbf{9s} & \textbf{11s} & \textbf{14s} & \textbf{17s} \\
            \rotatebox{90}{\footnotesize Center Crop} &
            \includegraphics[width=\fw]{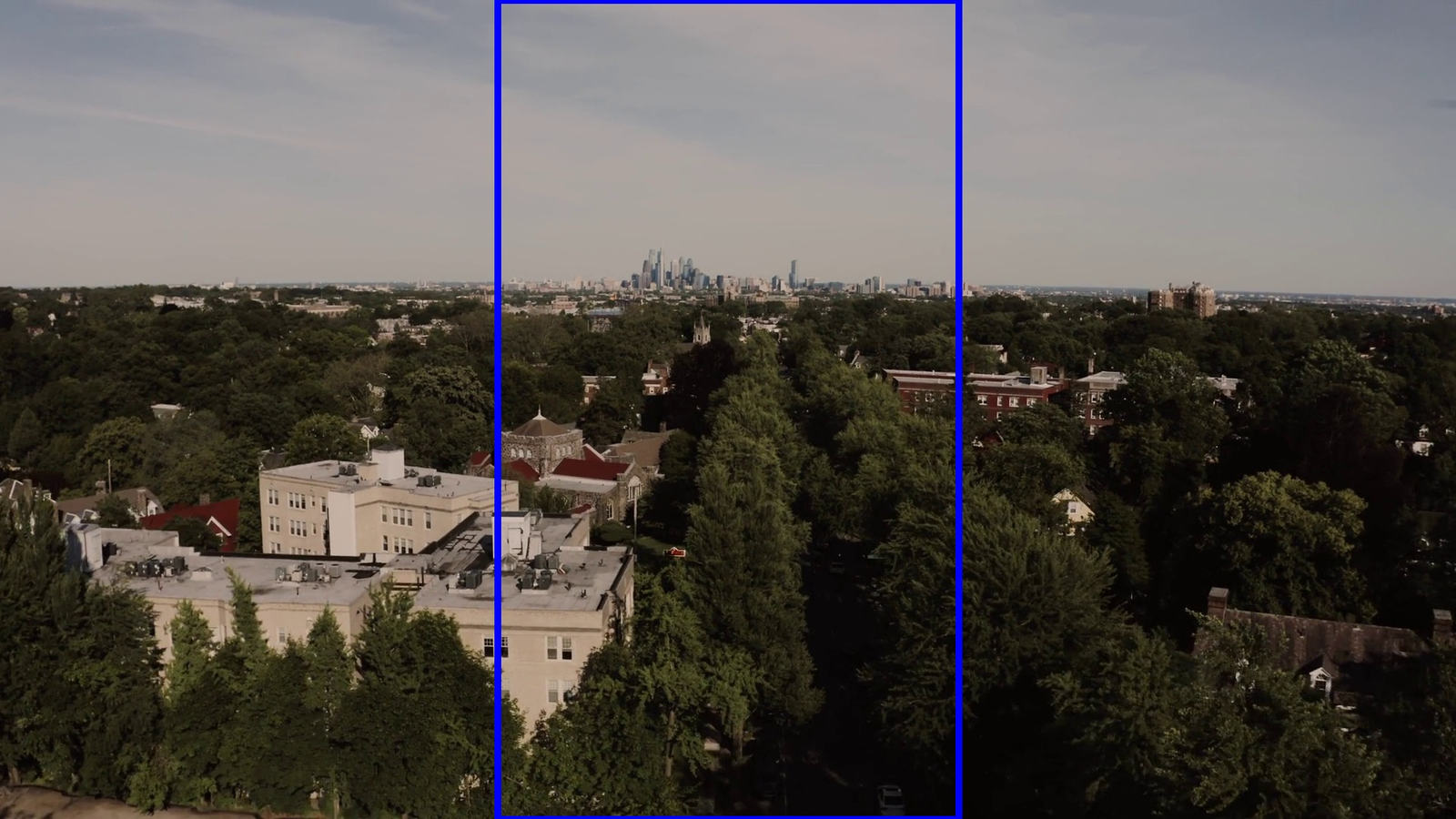} &
            \includegraphics[width=\fw]{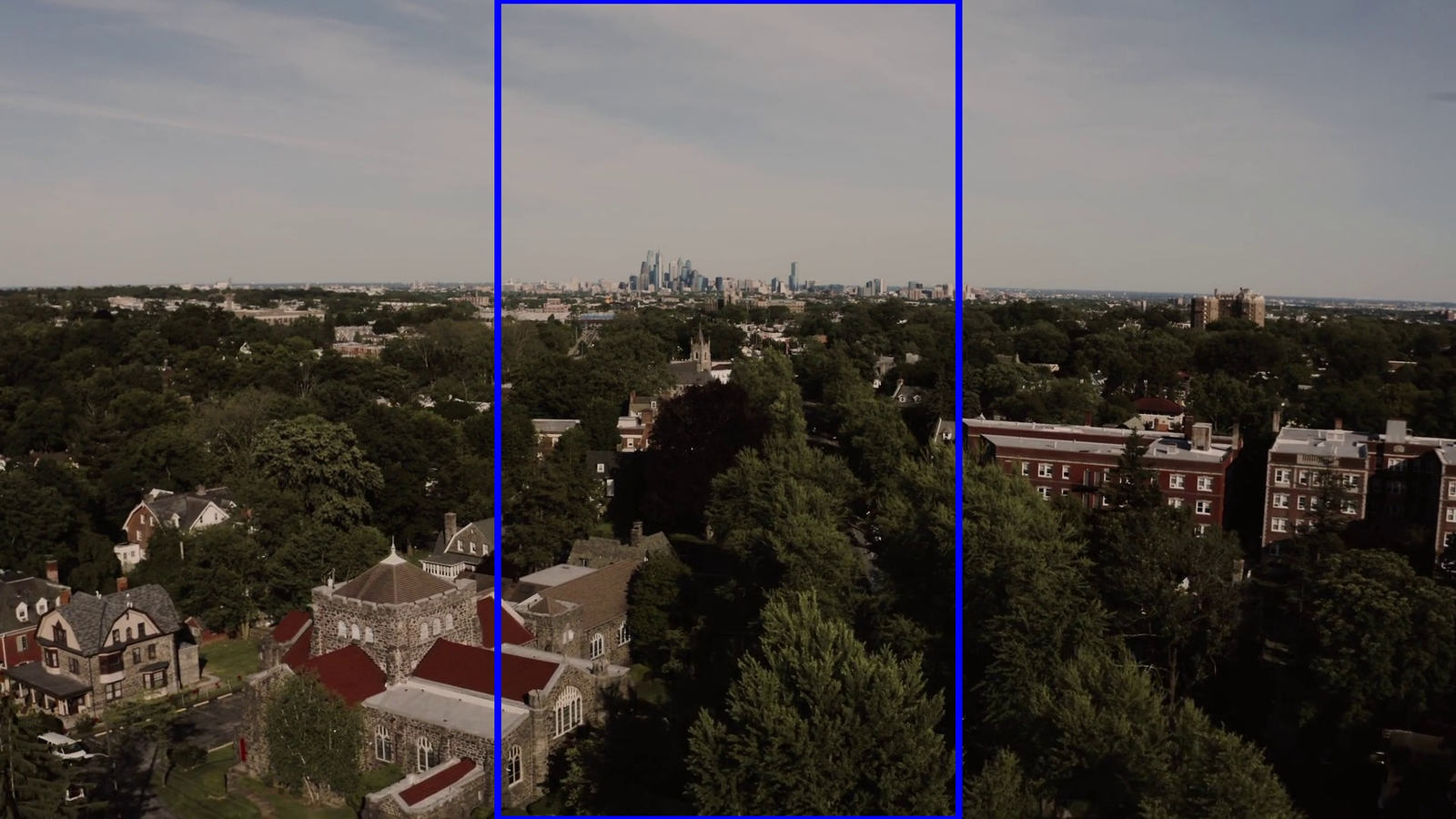} &
            \includegraphics[width=\fw]{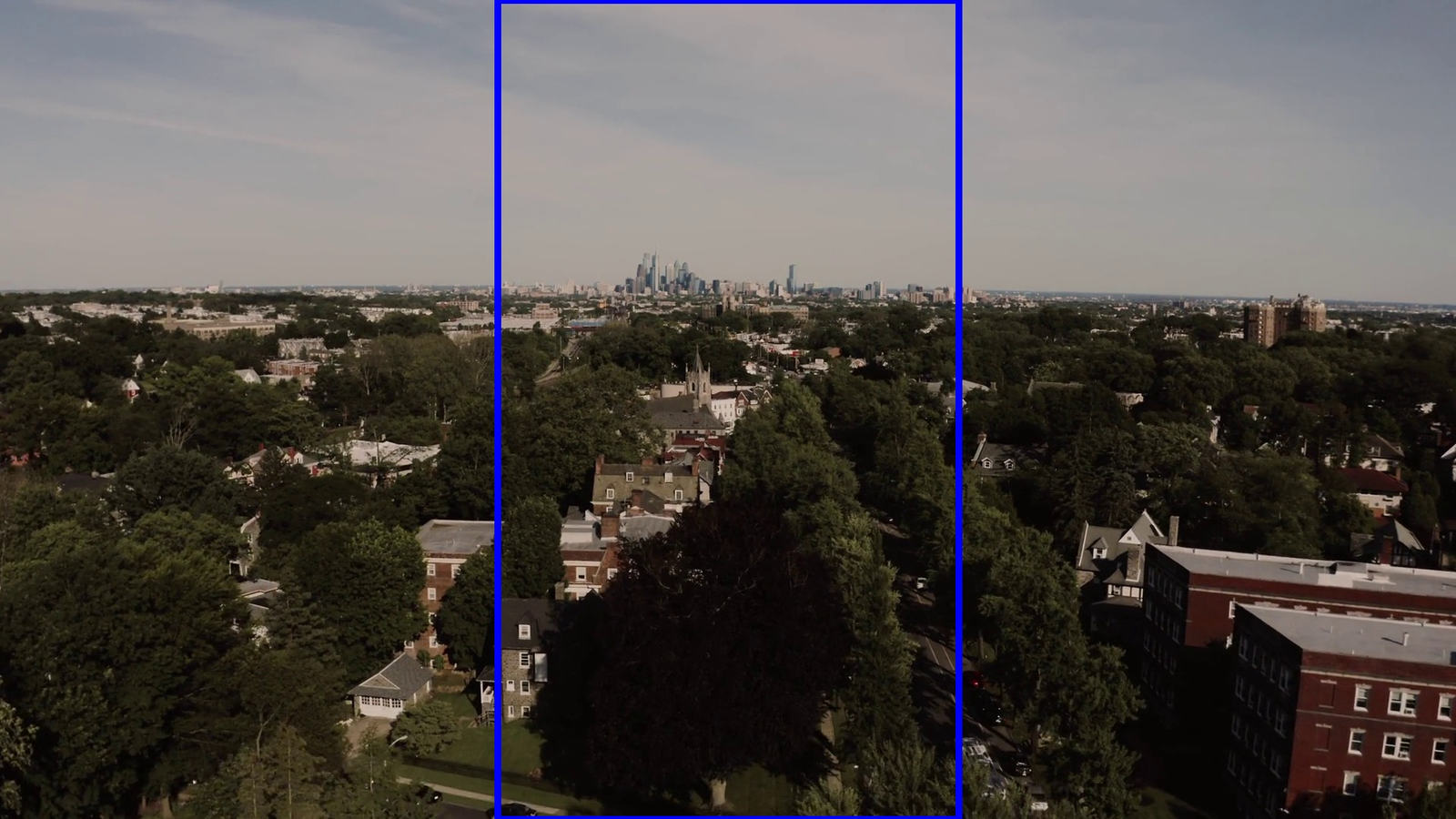} &
            \includegraphics[width=\fw]{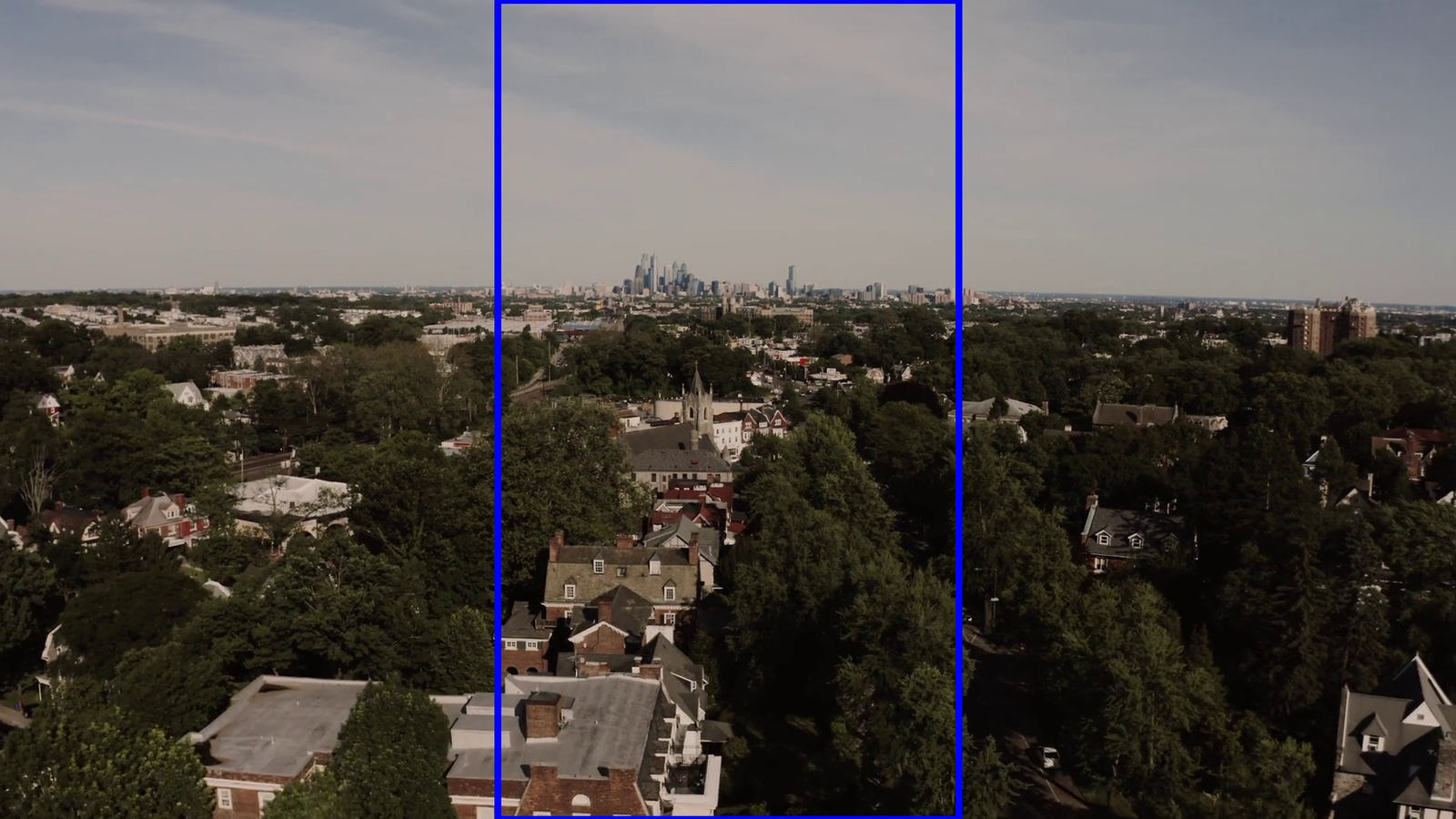} &
            \includegraphics[width=\fw]{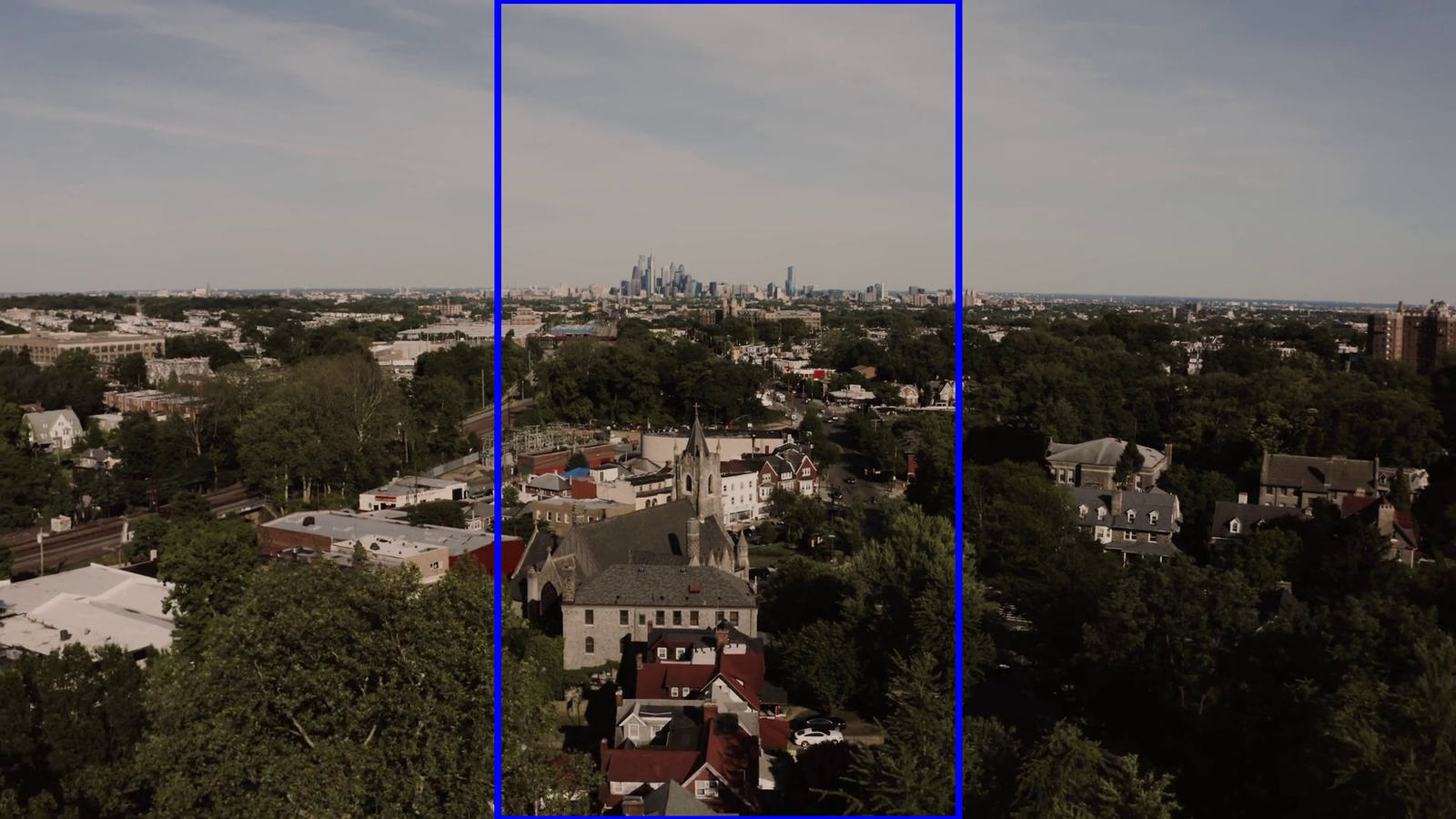} &
            \includegraphics[width=\fw]{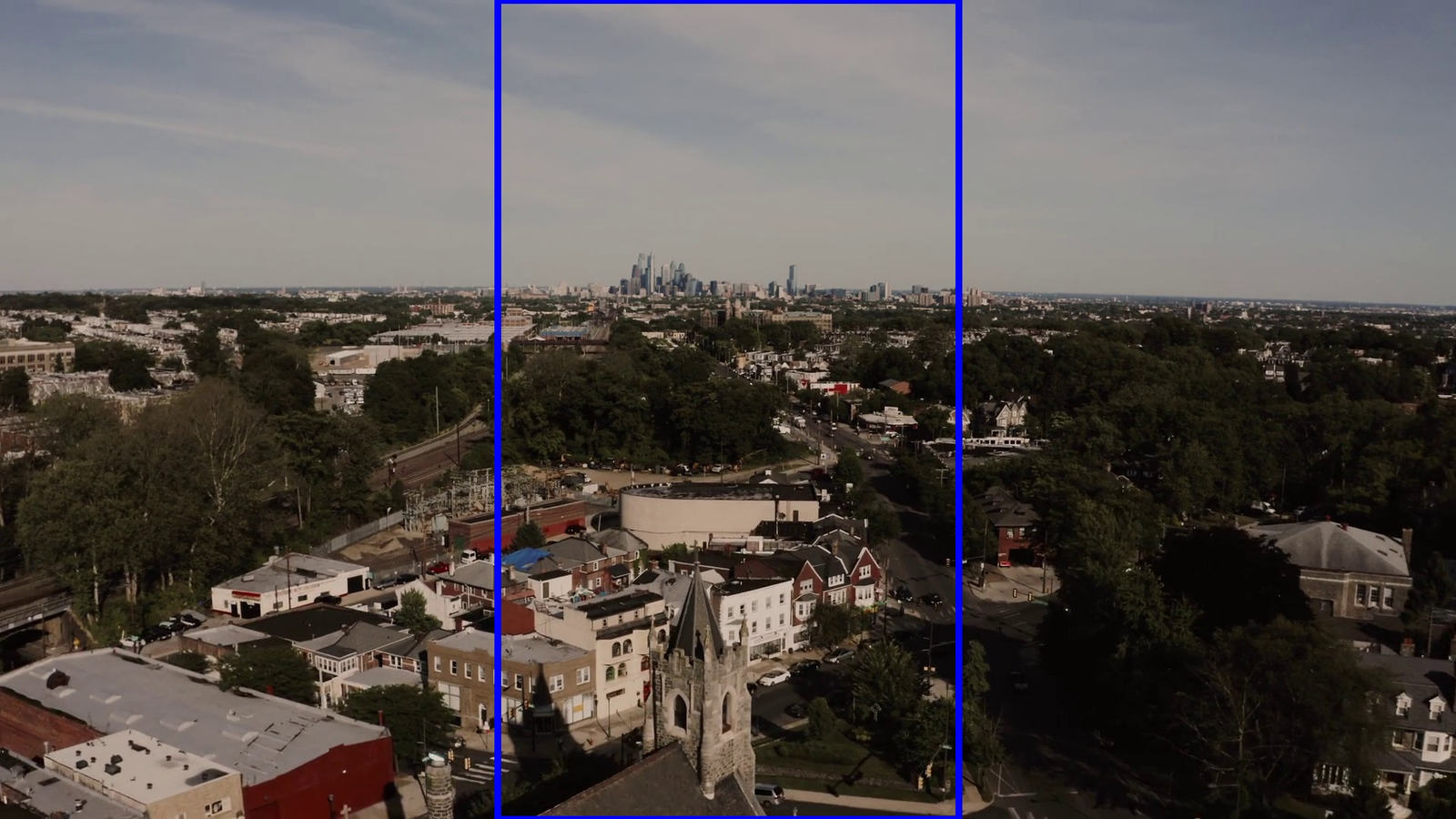} \\
            \rotatebox{90}{\footnotesize Gemini-Base} &
            \includegraphics[width=\fw]{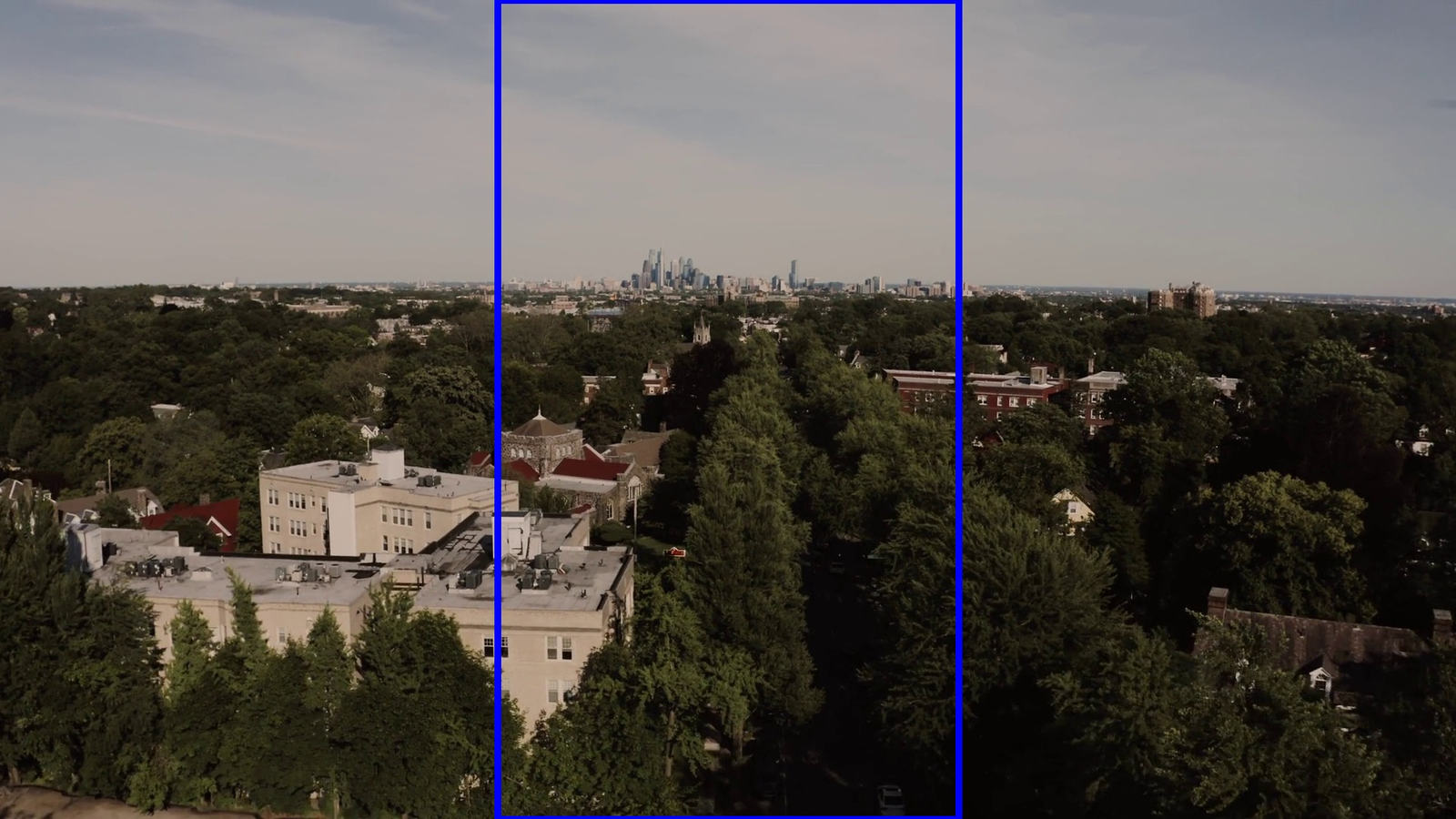} &
            \includegraphics[width=\fw]{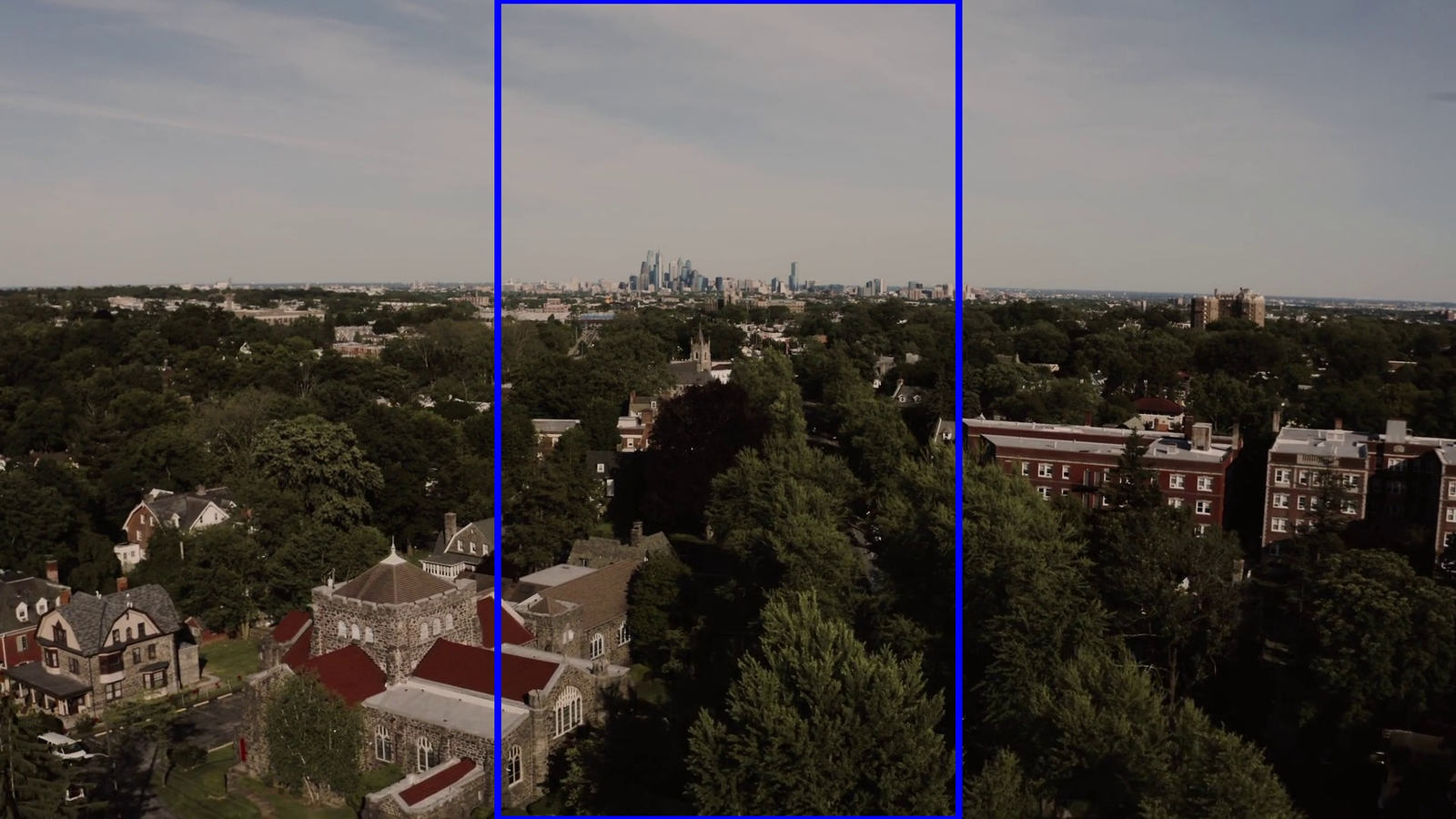} &
            \includegraphics[width=\fw]{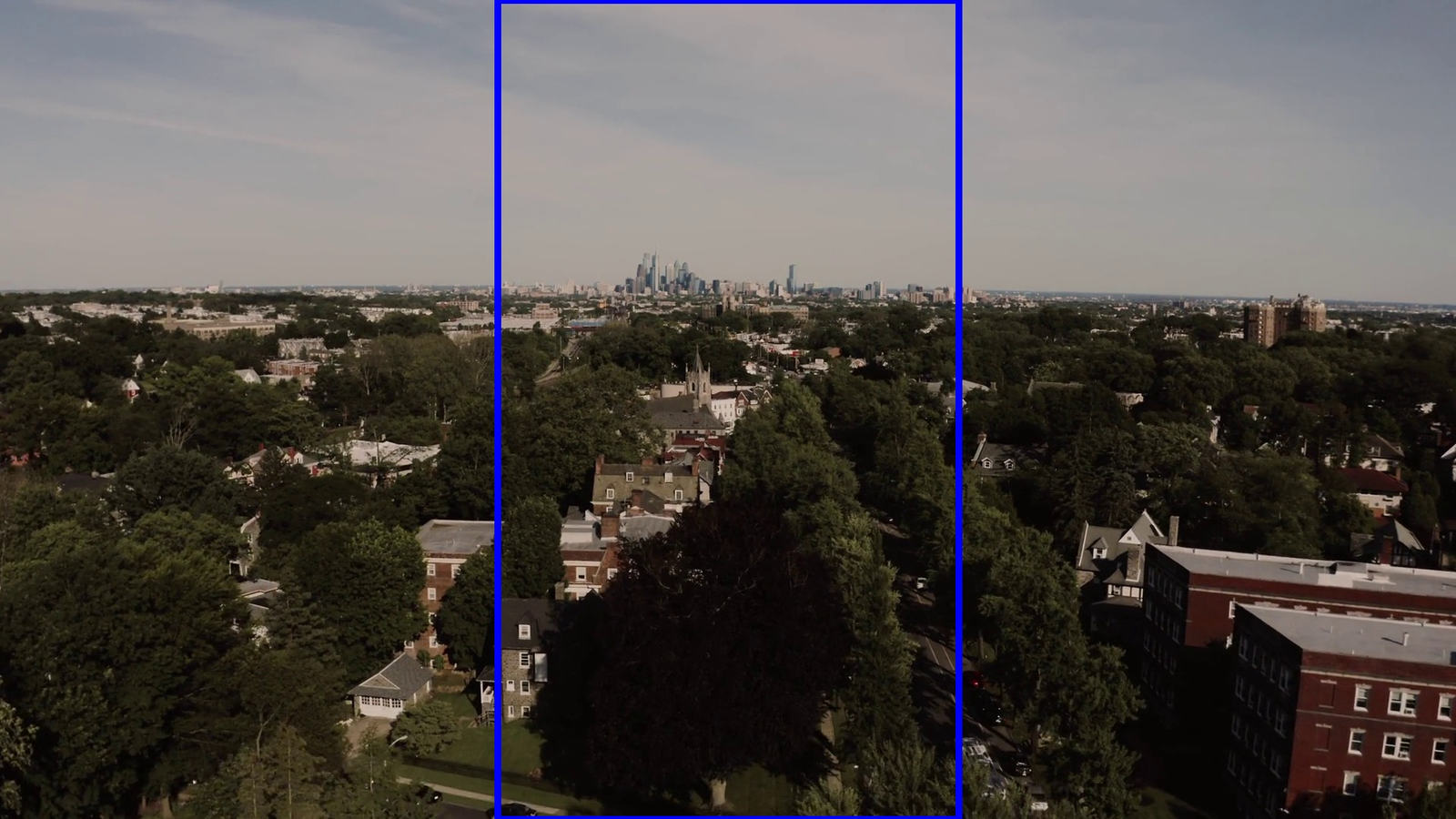} &
            \includegraphics[width=\fw]{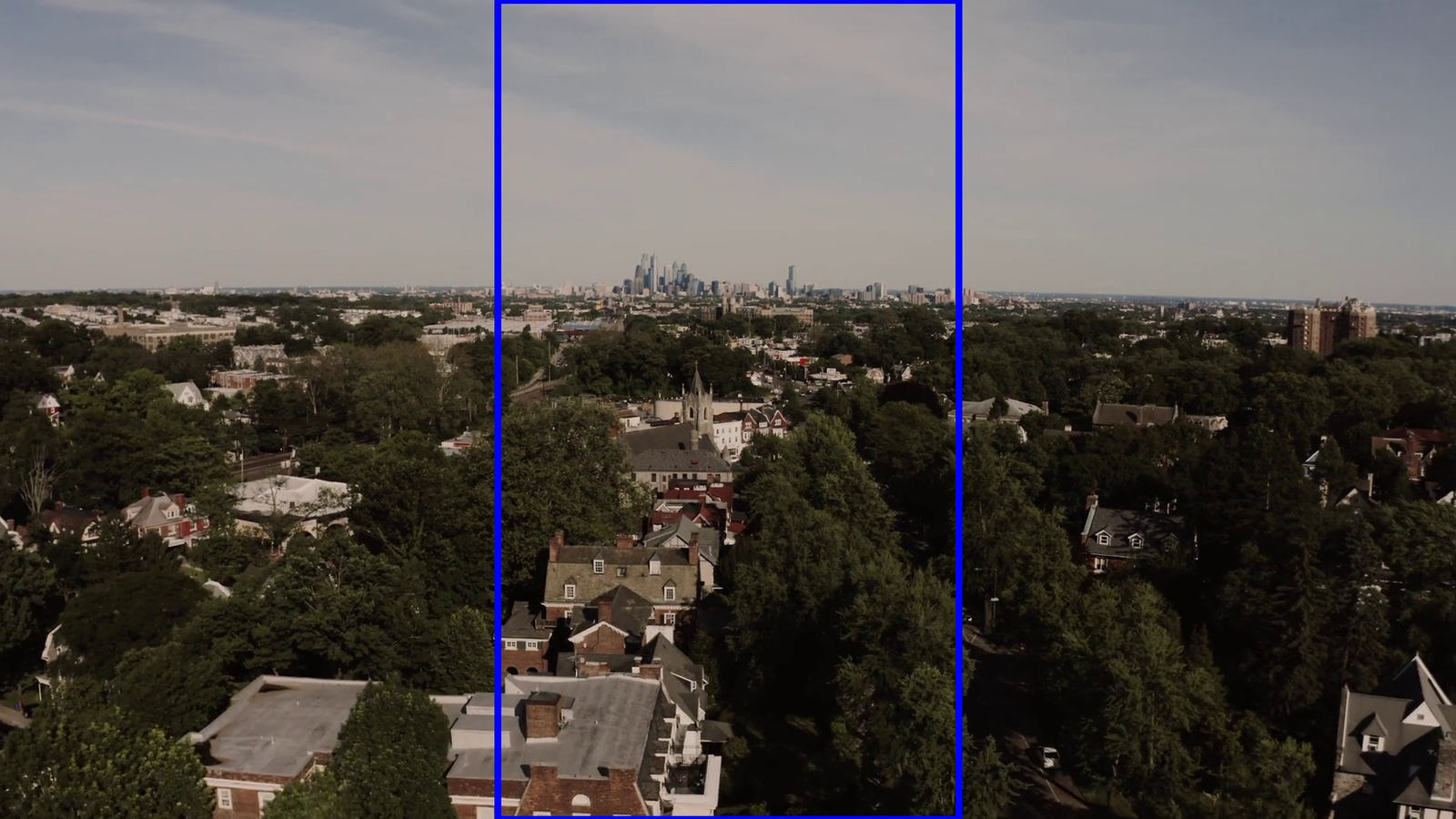} &
            \includegraphics[width=\fw]{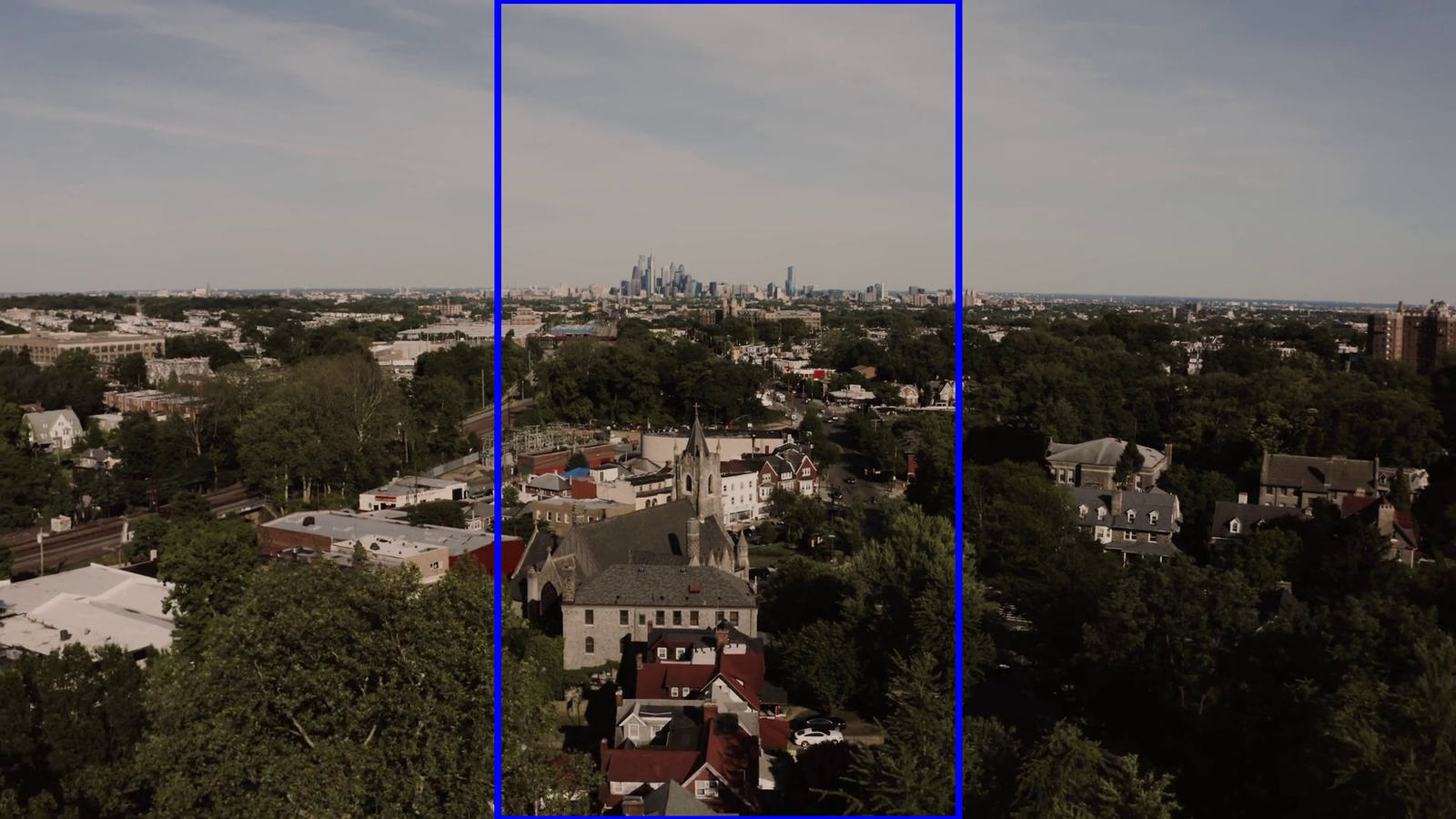} &
            \includegraphics[width=\fw]{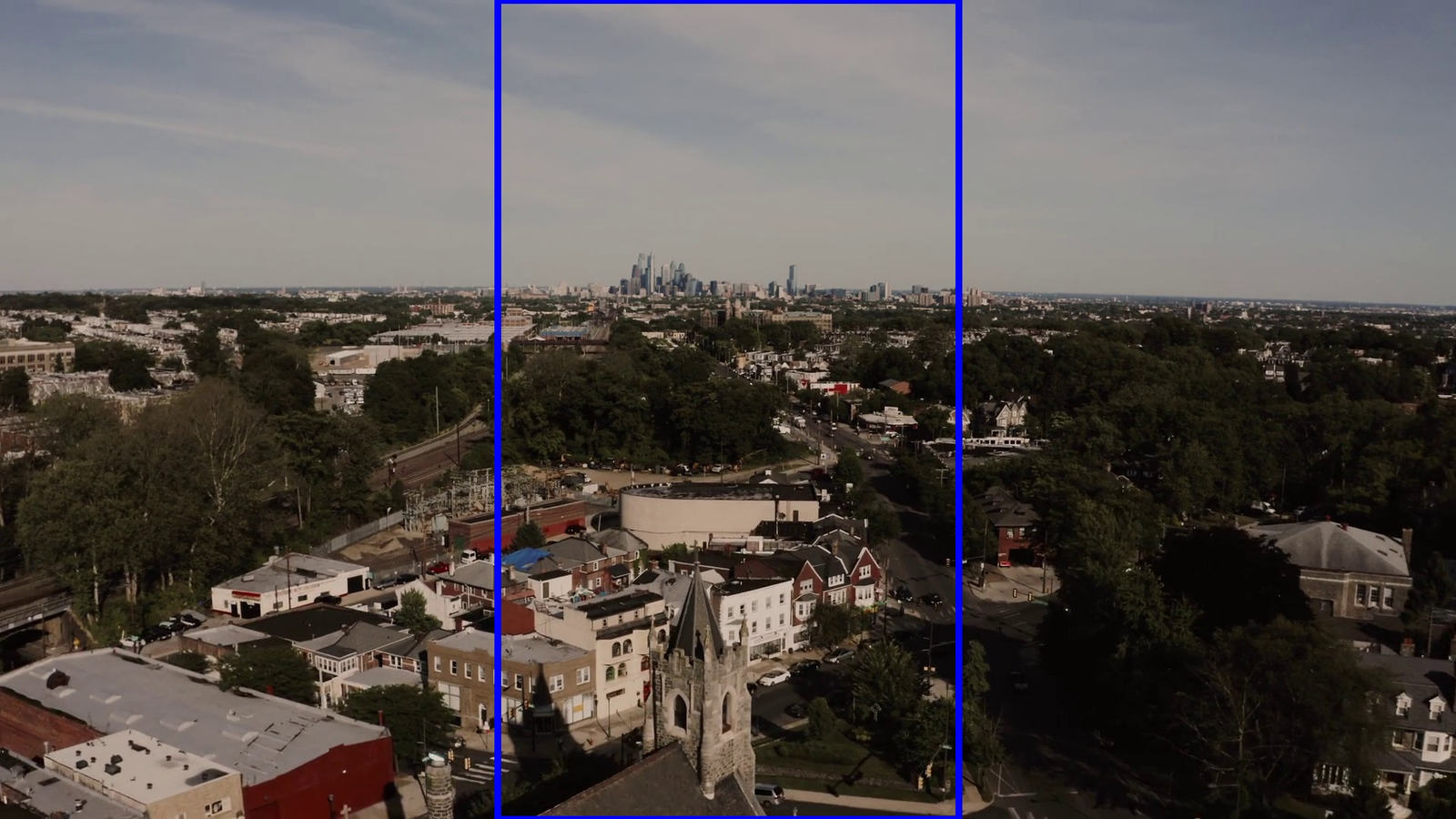} \\
            \rotatebox{90}{\footnotesize Ours} &
            \includegraphics[width=\fw]{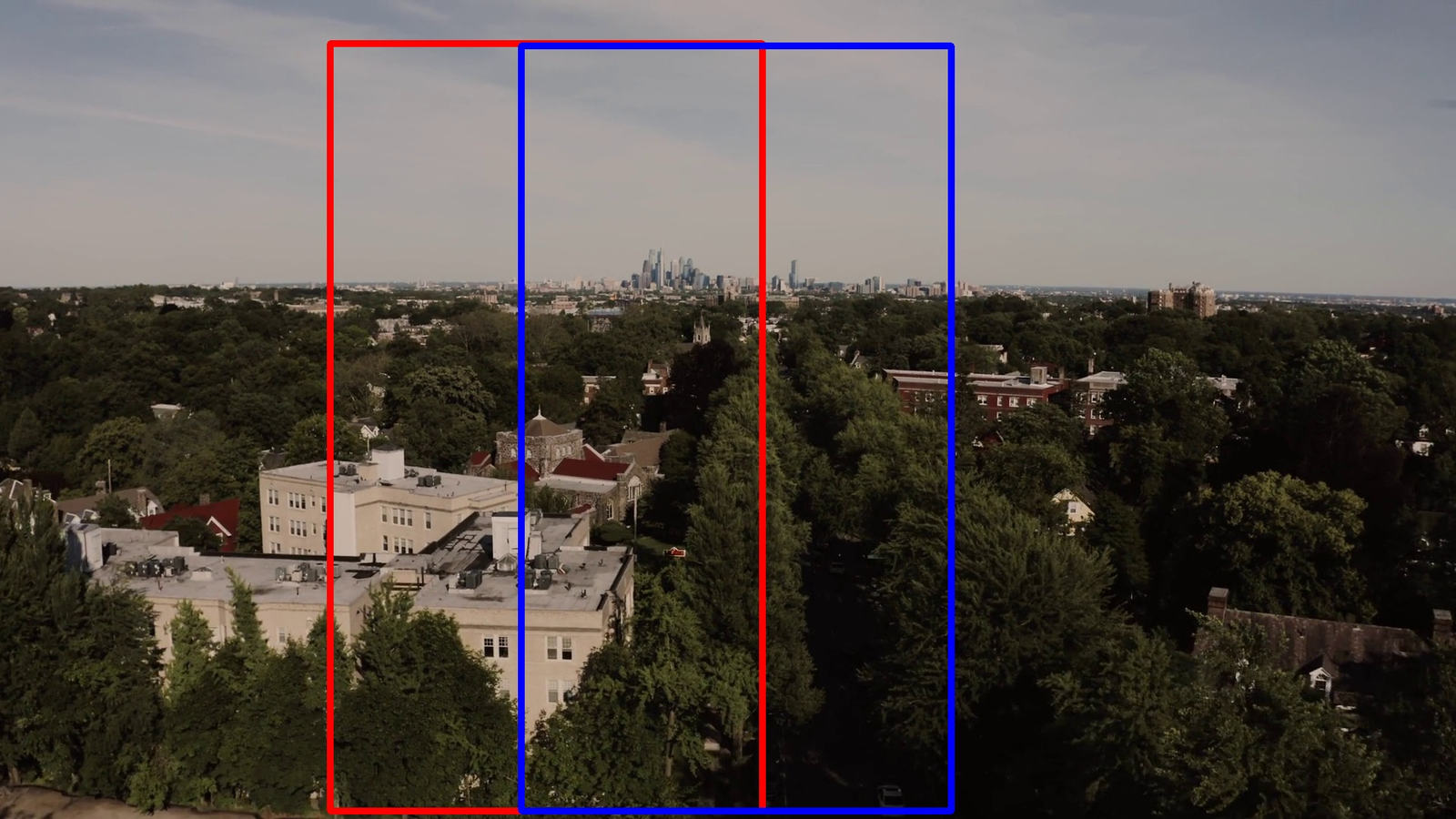} &
            \includegraphics[width=\fw]{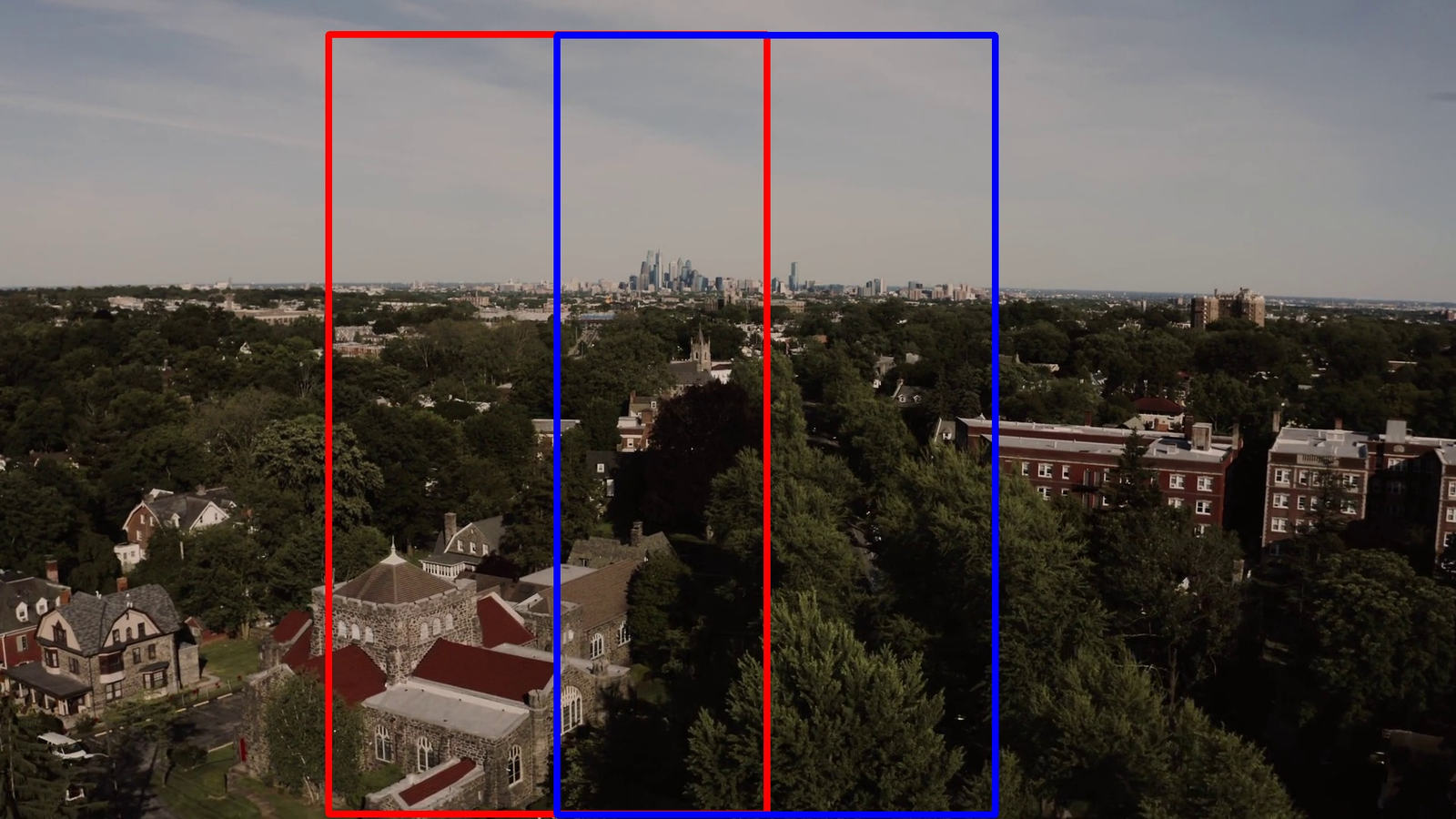} &
            \includegraphics[width=\fw]{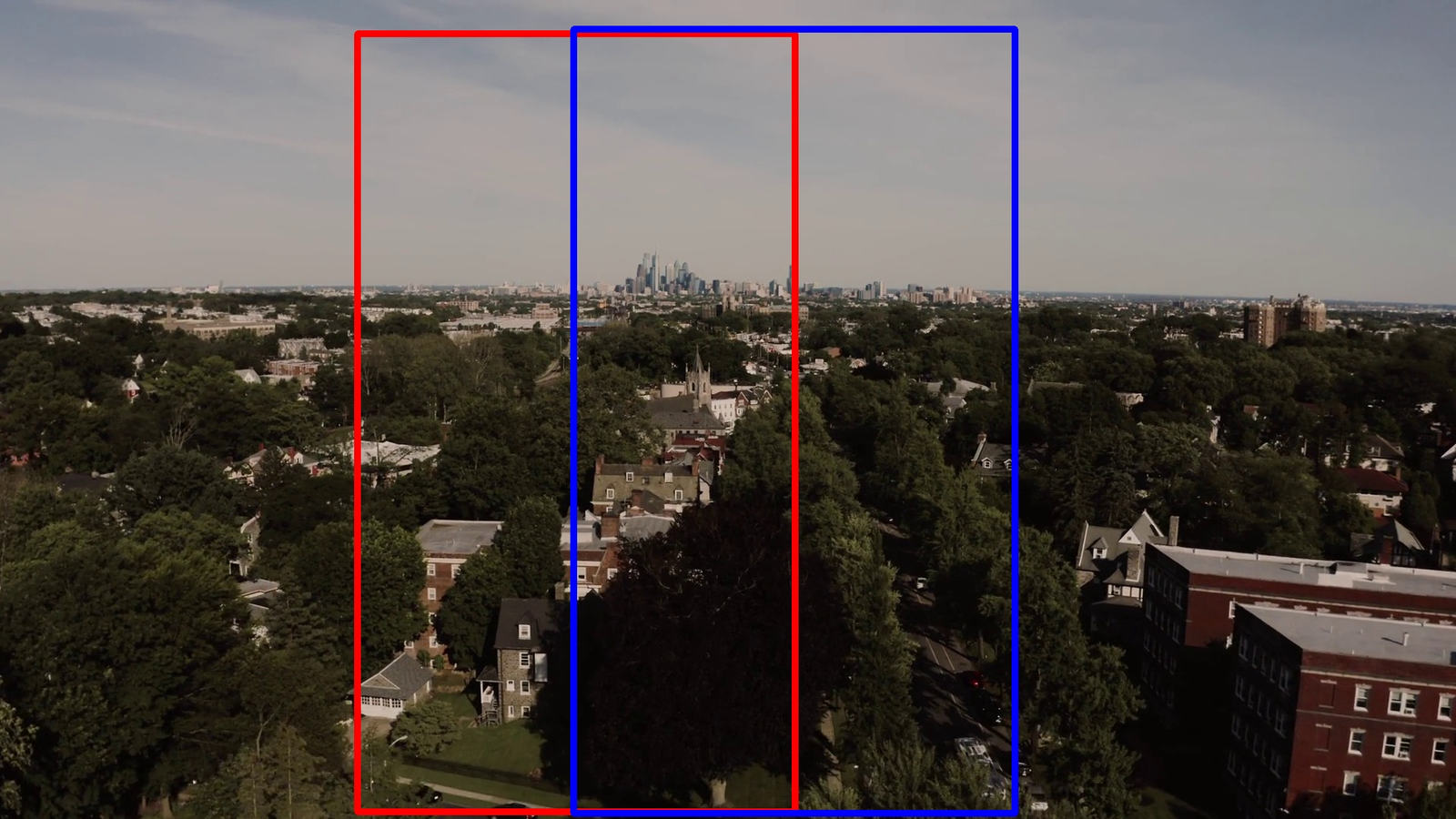} &
            \includegraphics[width=\fw]{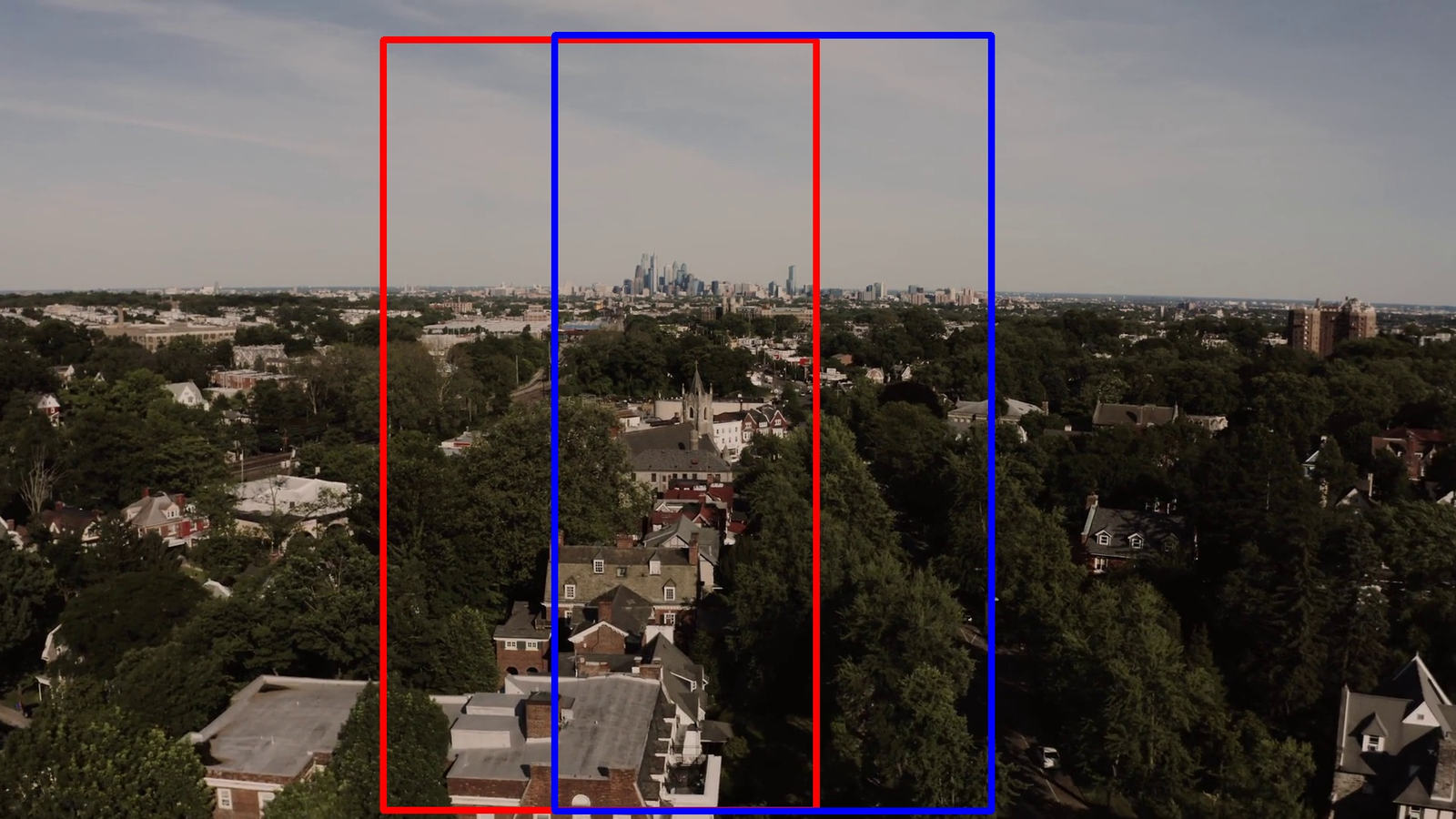} &
            \includegraphics[width=\fw]{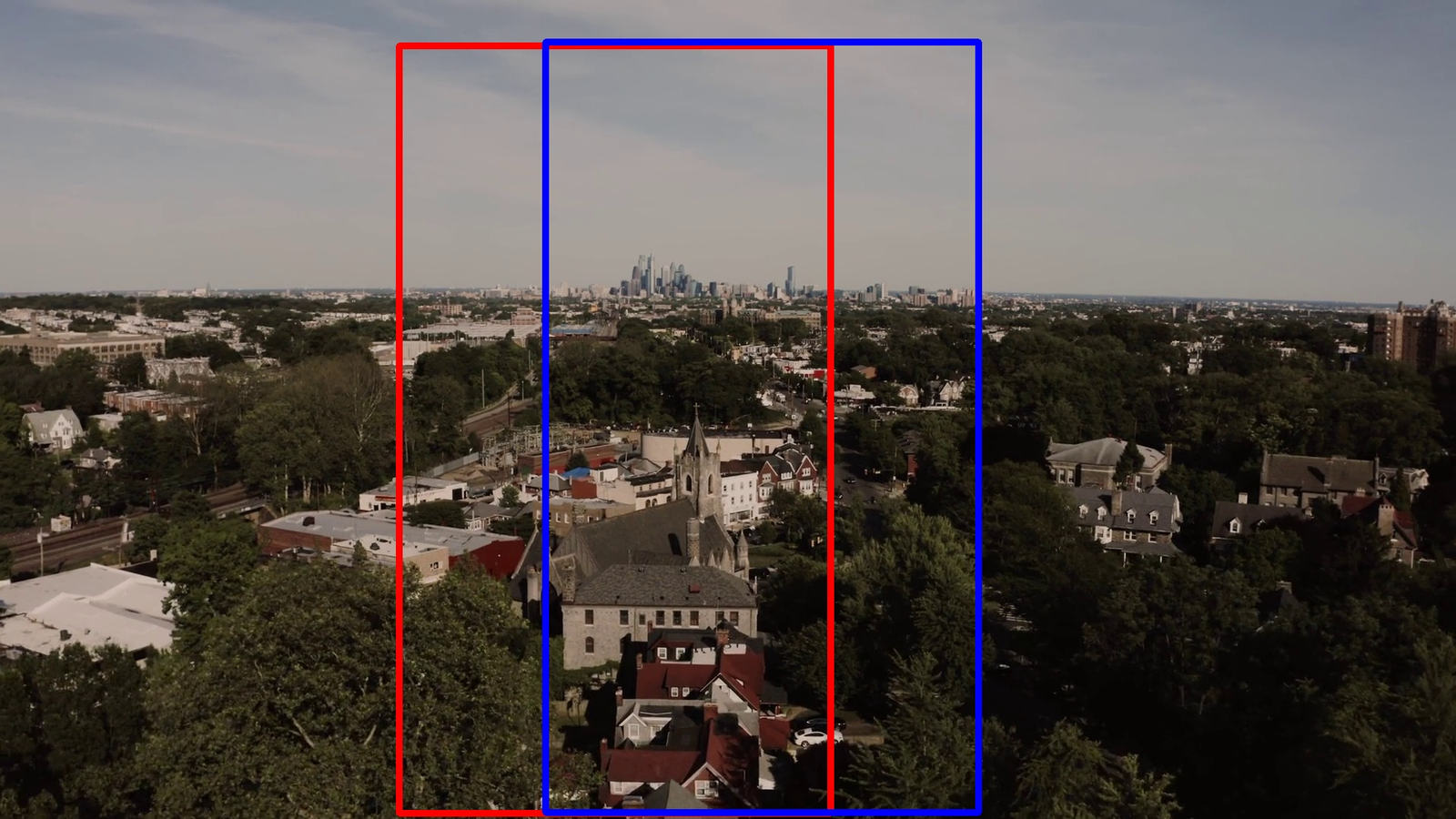} &
            \includegraphics[width=\fw]{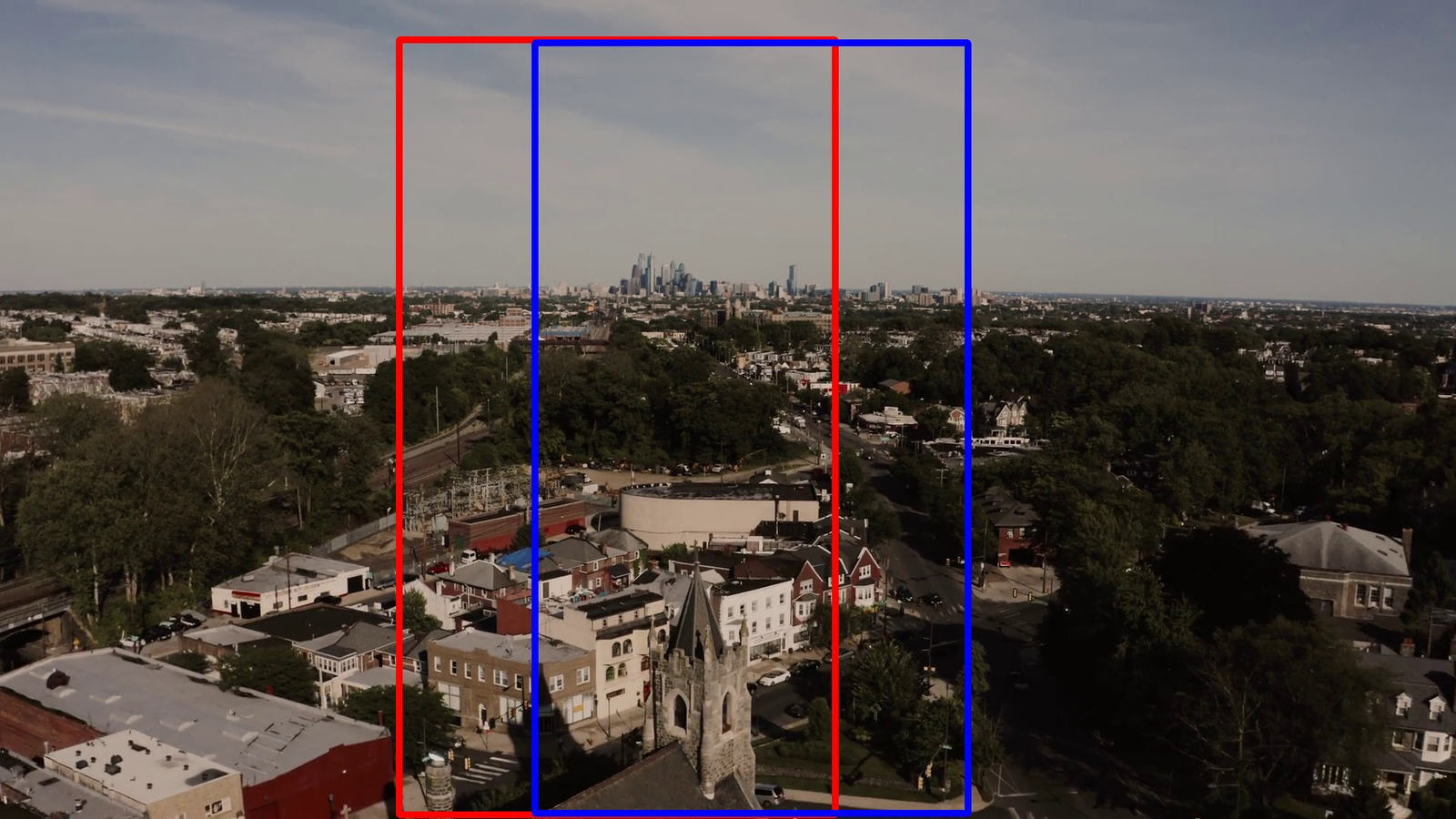} \\
            \rotatebox{90}{\footnotesize Ours + MLLM-based Agent} &
            \includegraphics[width=\fw]{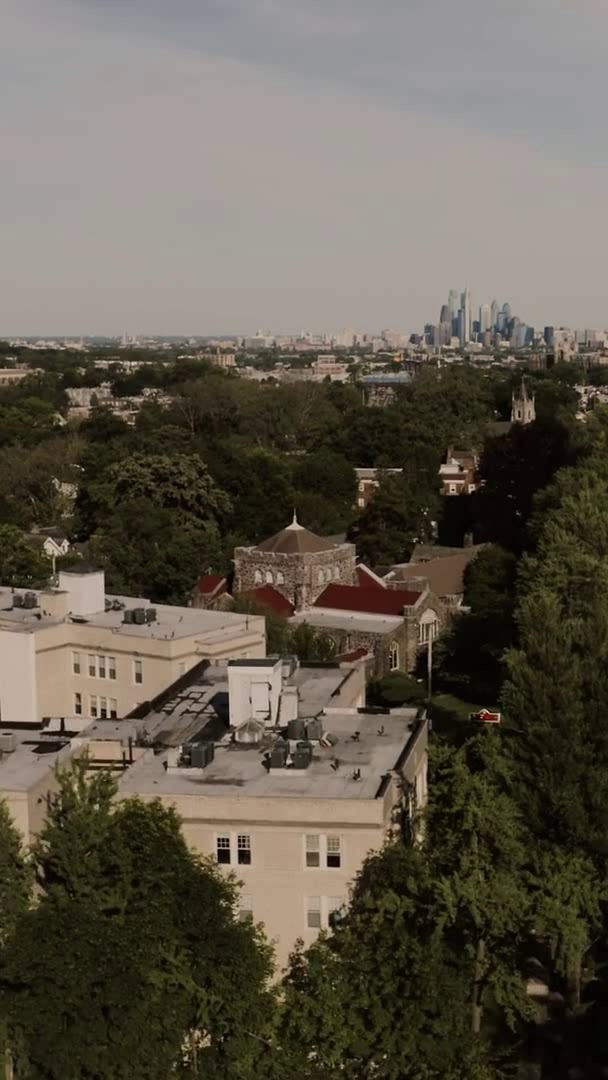} &
            \includegraphics[width=\fw]{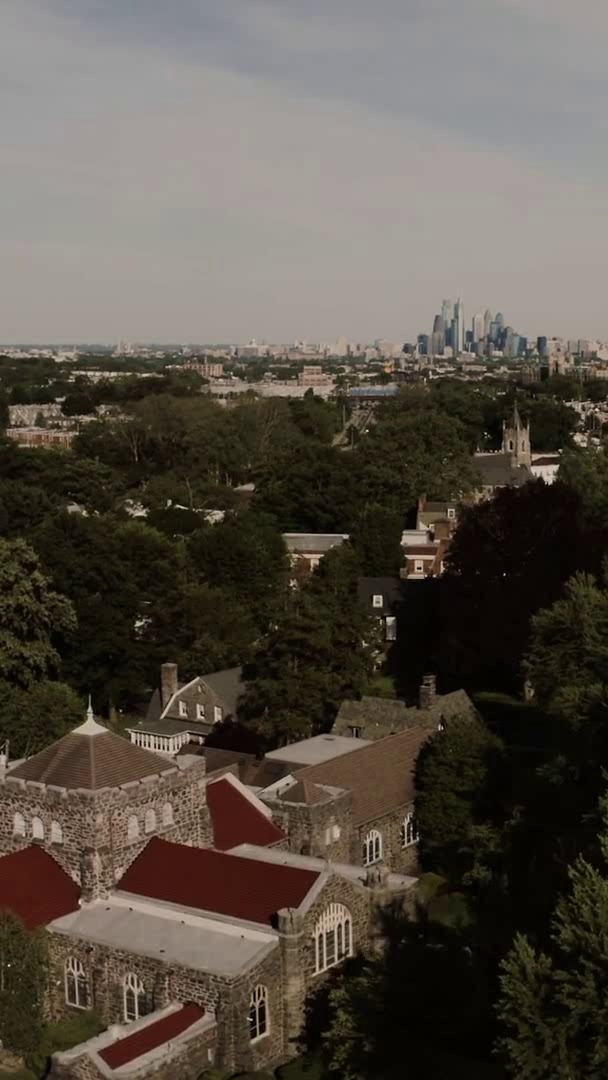} &
            \includegraphics[width=\fw]{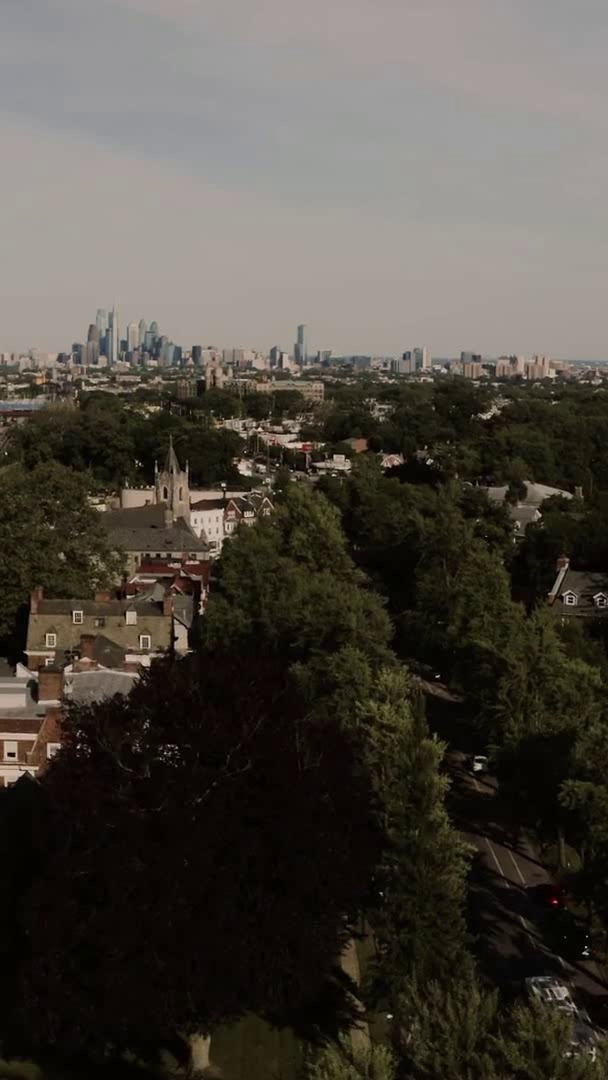} &
            \includegraphics[width=\fw]{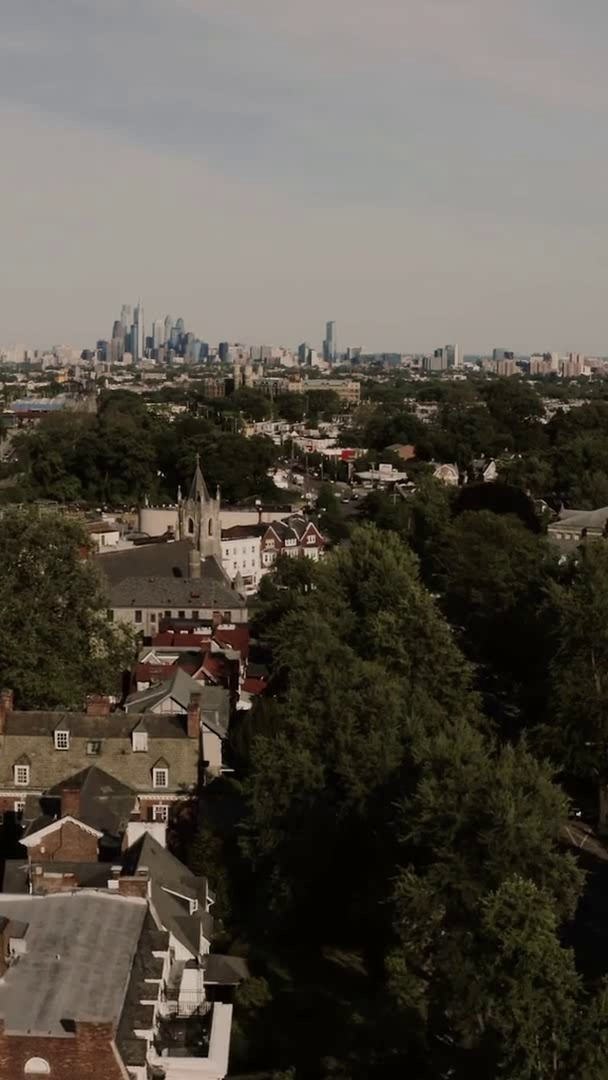} &
            \includegraphics[width=\fw]{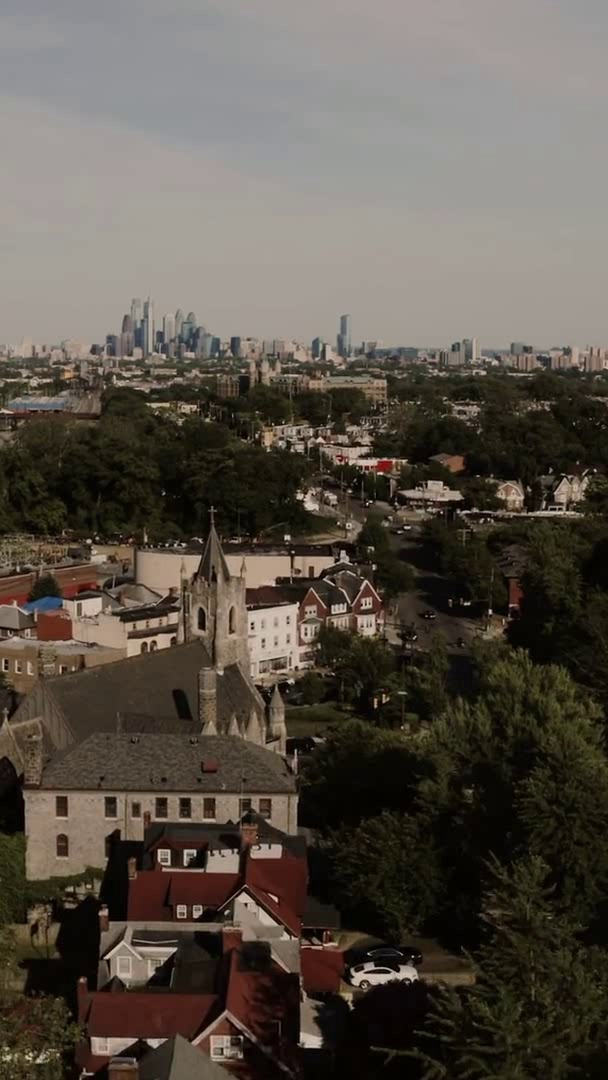} &
            \includegraphics[width=\fw]{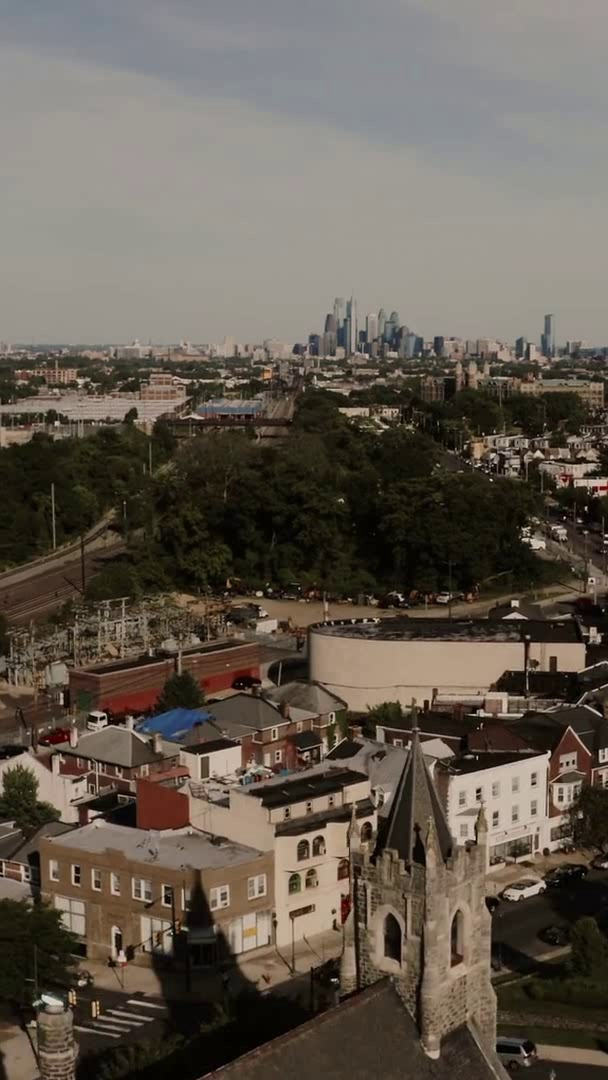} \\
        \end{tabular}
        \caption{Video ID: 4031240}
    \end{subfigure}
    
    \vspace{-10pt}
    \caption{\textbf{Qualitative comparison of ROI extraction methods on the MiraData dataset \cite{ju2024miradata} (Part 1 of 3).} For each subfigure (\textit{a}--\textit{b}), representing individual video samples stacked vertically, the rows correspond to the \textbf{Center Crop} (first), the \textbf{Baseline} (second), \textbf{Ours} (third), and \textbf{Final Crop} (bottom).}
    \label{fig:main_qualitative_comparison_part1}
\end{figure*}

\begin{figure*}[t]
    \centering
    \def\fw{0.15\textwidth} 
    \setlength{\tabcolsep}{1pt} 
    
    \begin{subfigure}{\textwidth}
        \centering
        \begin{tabular}{ccccccc}
             & \textbf{4s} & \textbf{7s} & \textbf{11s} & \textbf{15s} & \textbf{19s} & \textbf{22s} \\
            \rotatebox{90}{\footnotesize Center Crop} &
            \includegraphics[width=\fw]{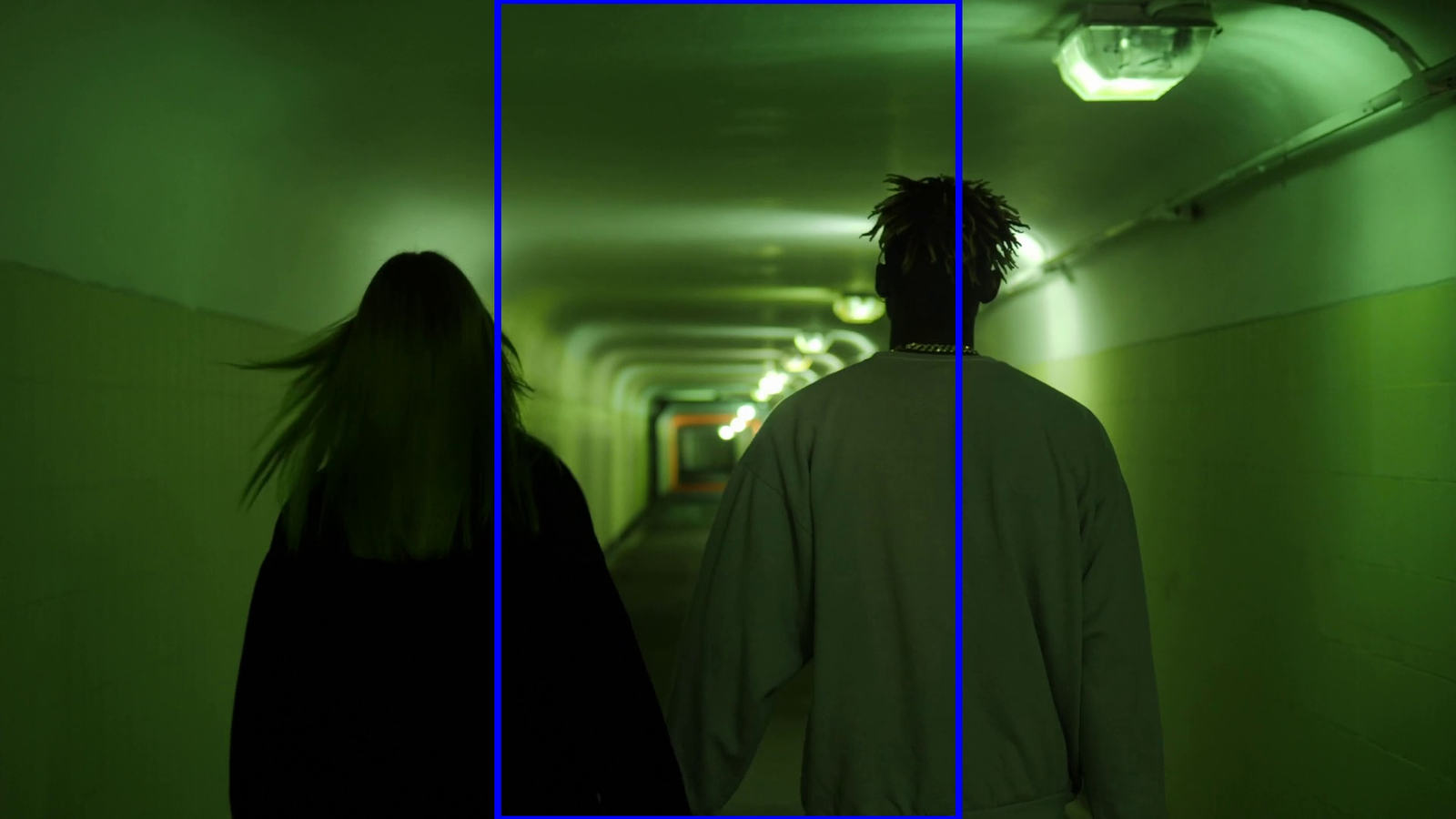} &
            \includegraphics[width=\fw]{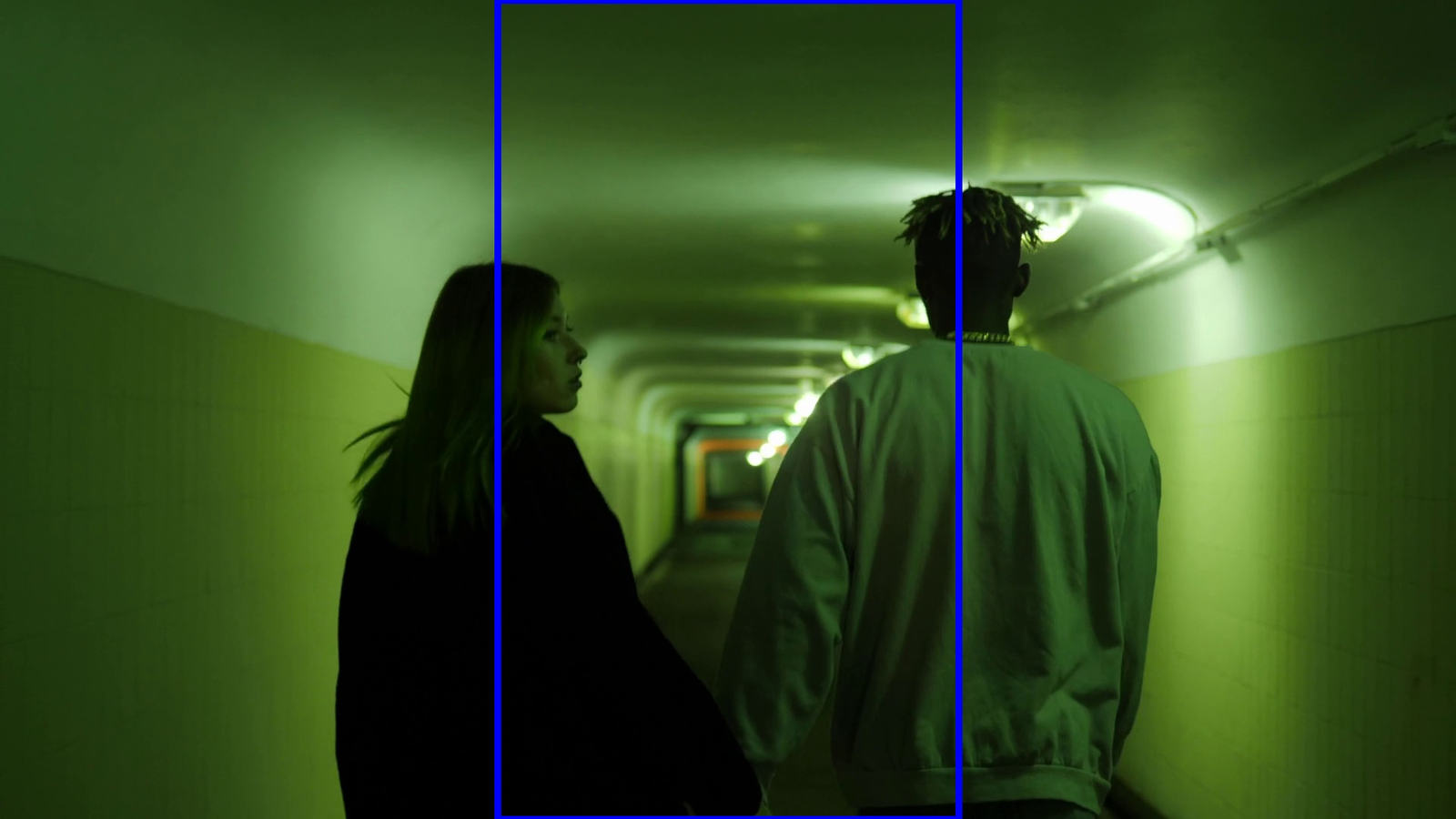} &
            \includegraphics[width=\fw]{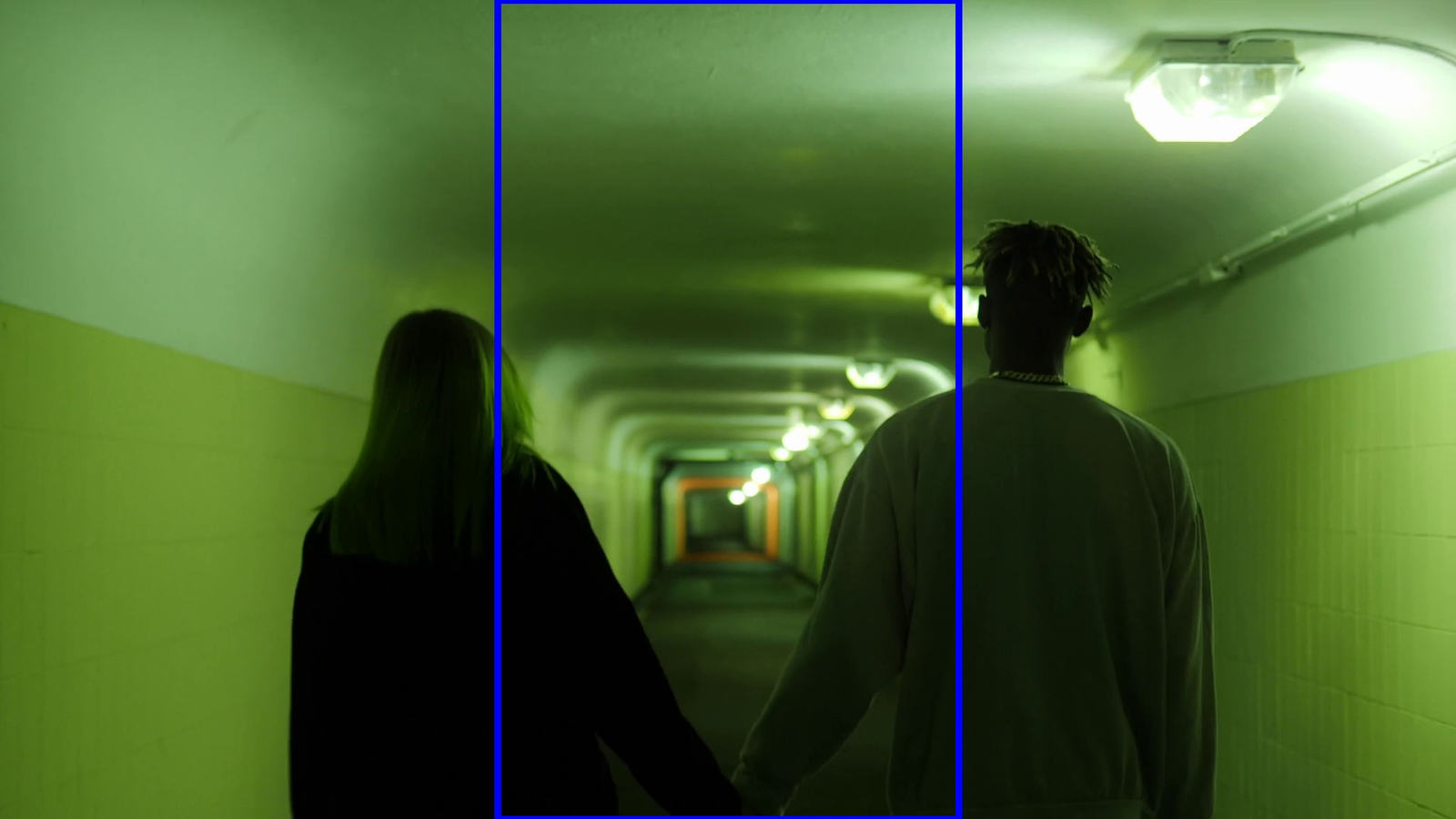} &
            \includegraphics[width=\fw]{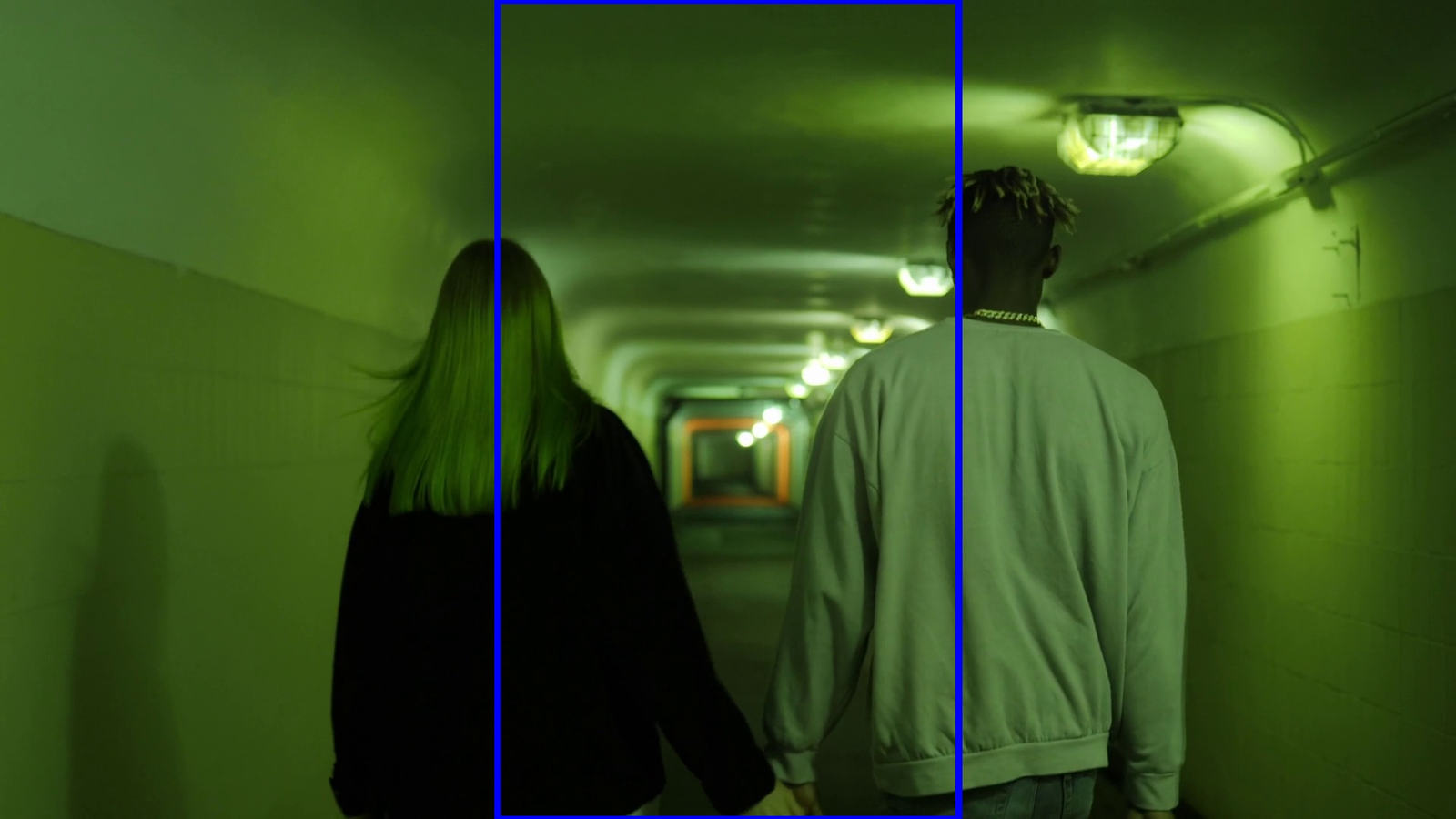} &
            \includegraphics[width=\fw]{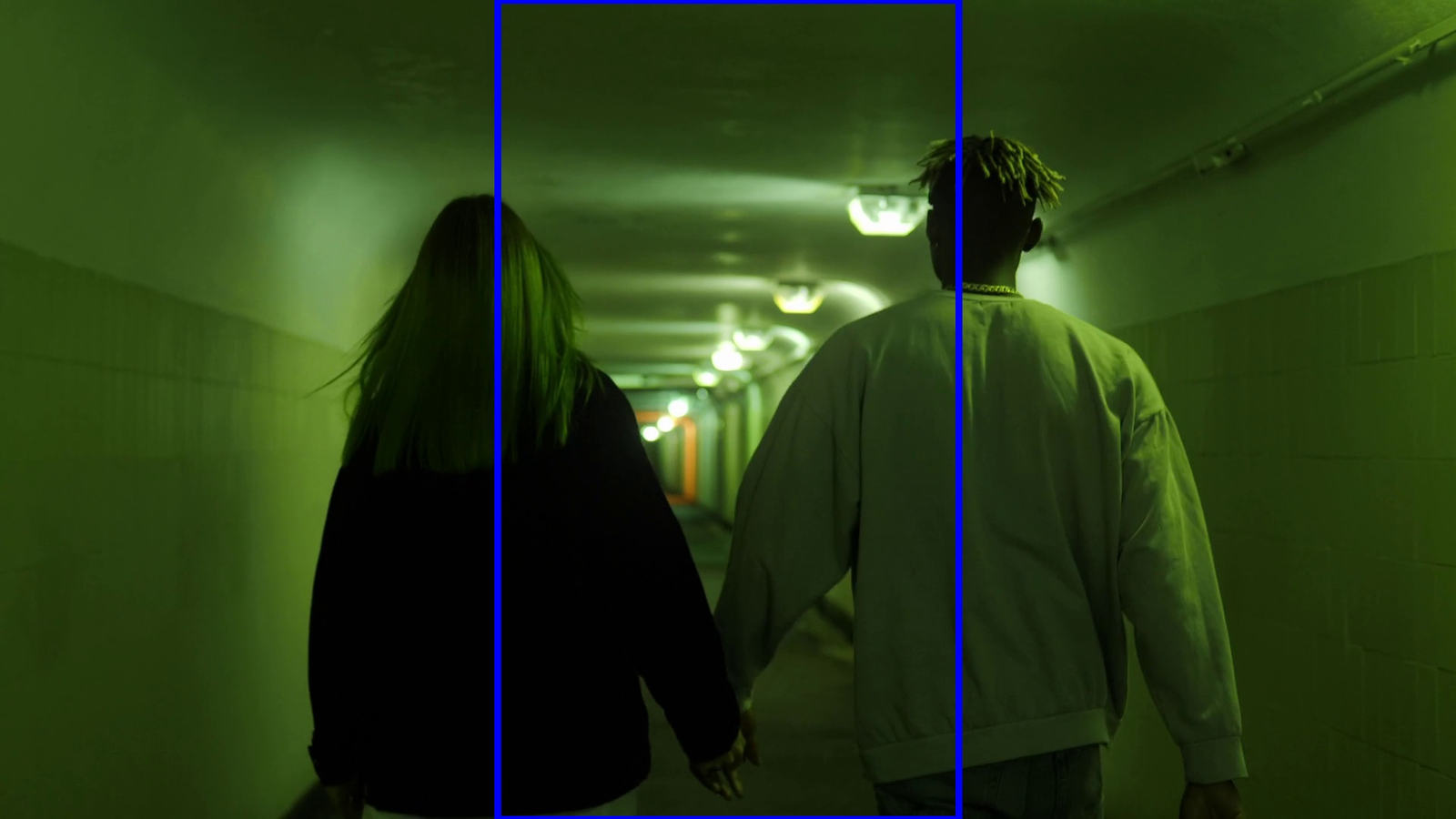} &
            \includegraphics[width=\fw]{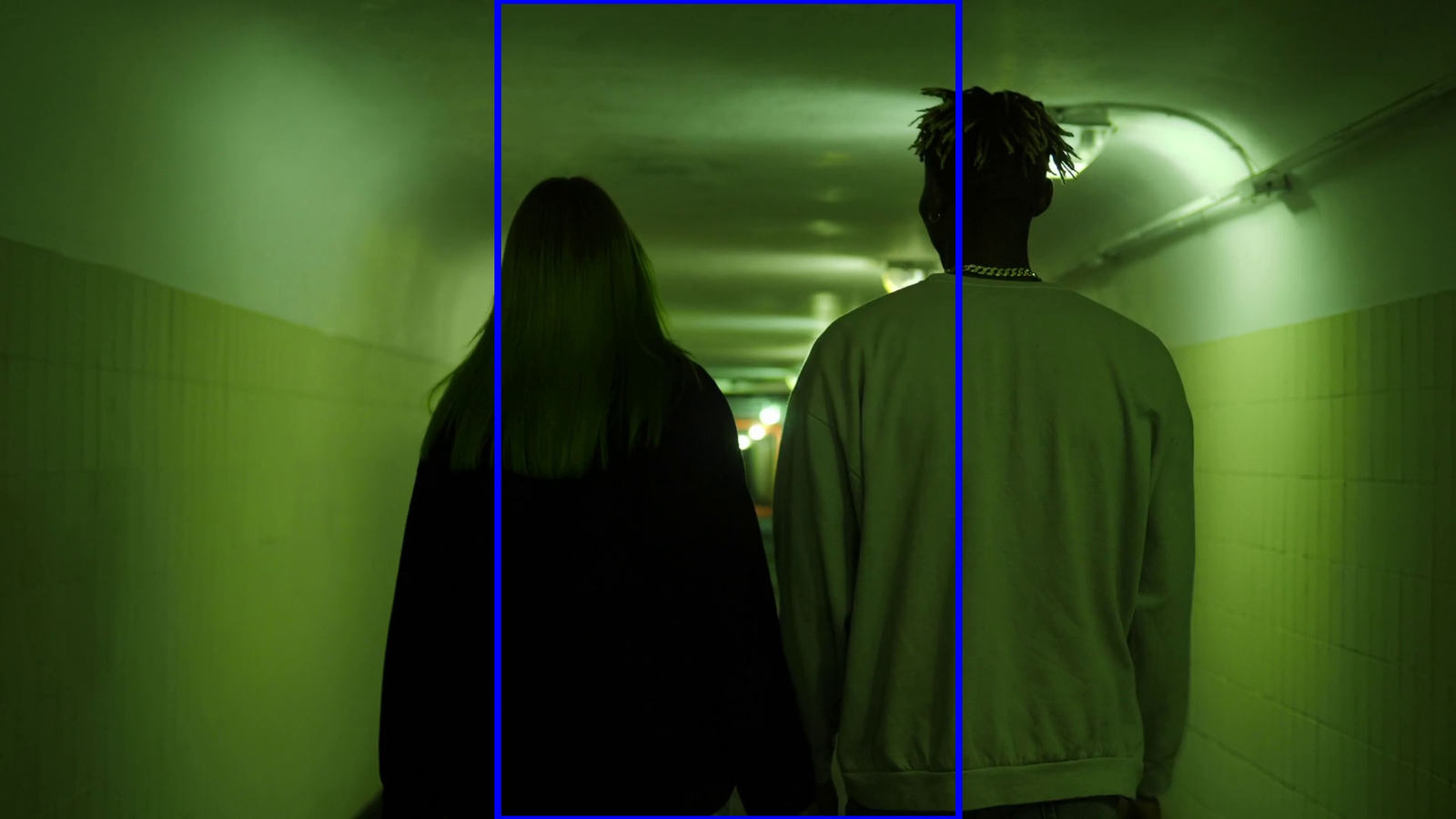} \\
            \rotatebox{90}{\footnotesize Gemini-Base} &
            \includegraphics[width=\fw]{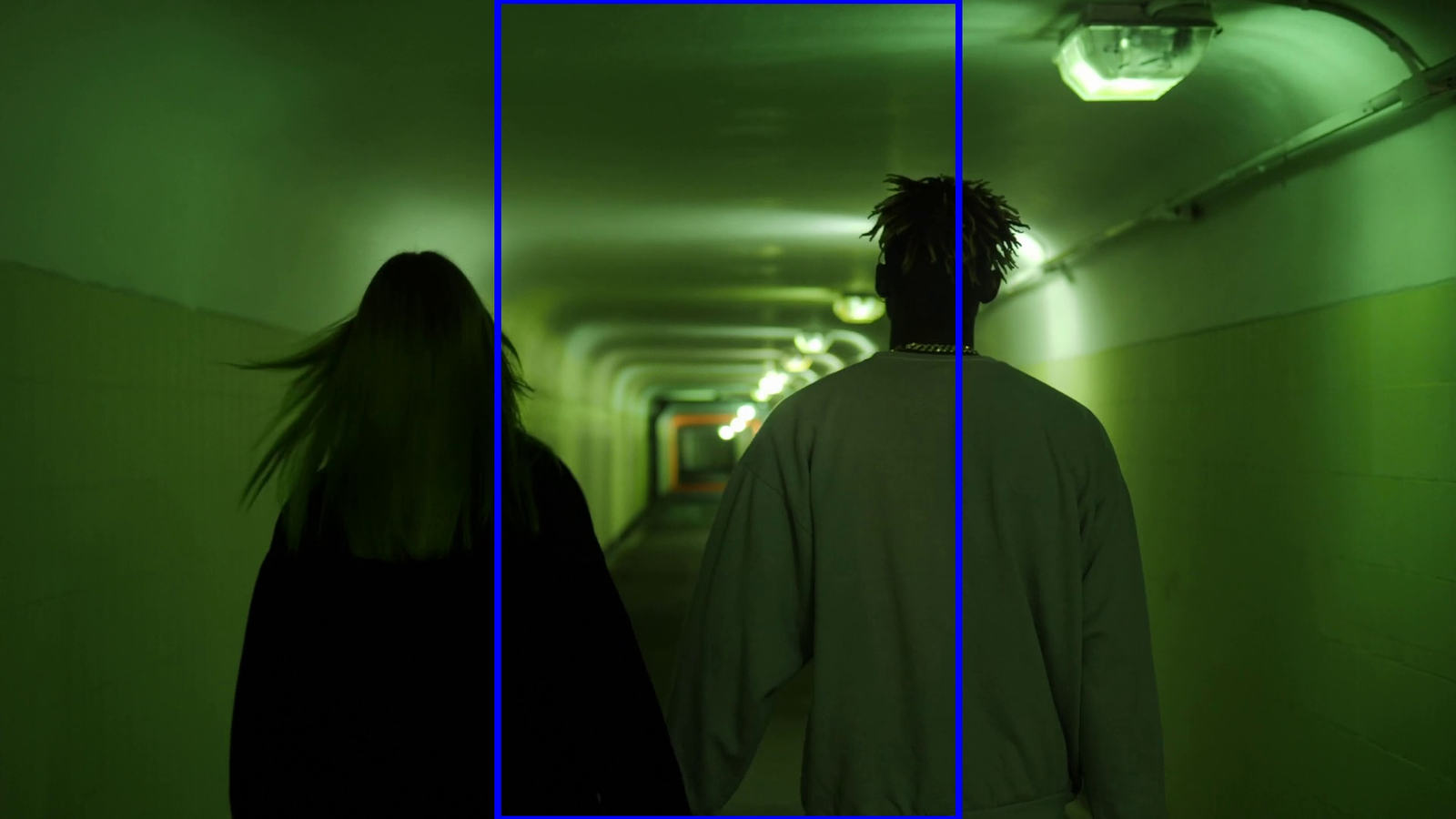} &
            \includegraphics[width=\fw]{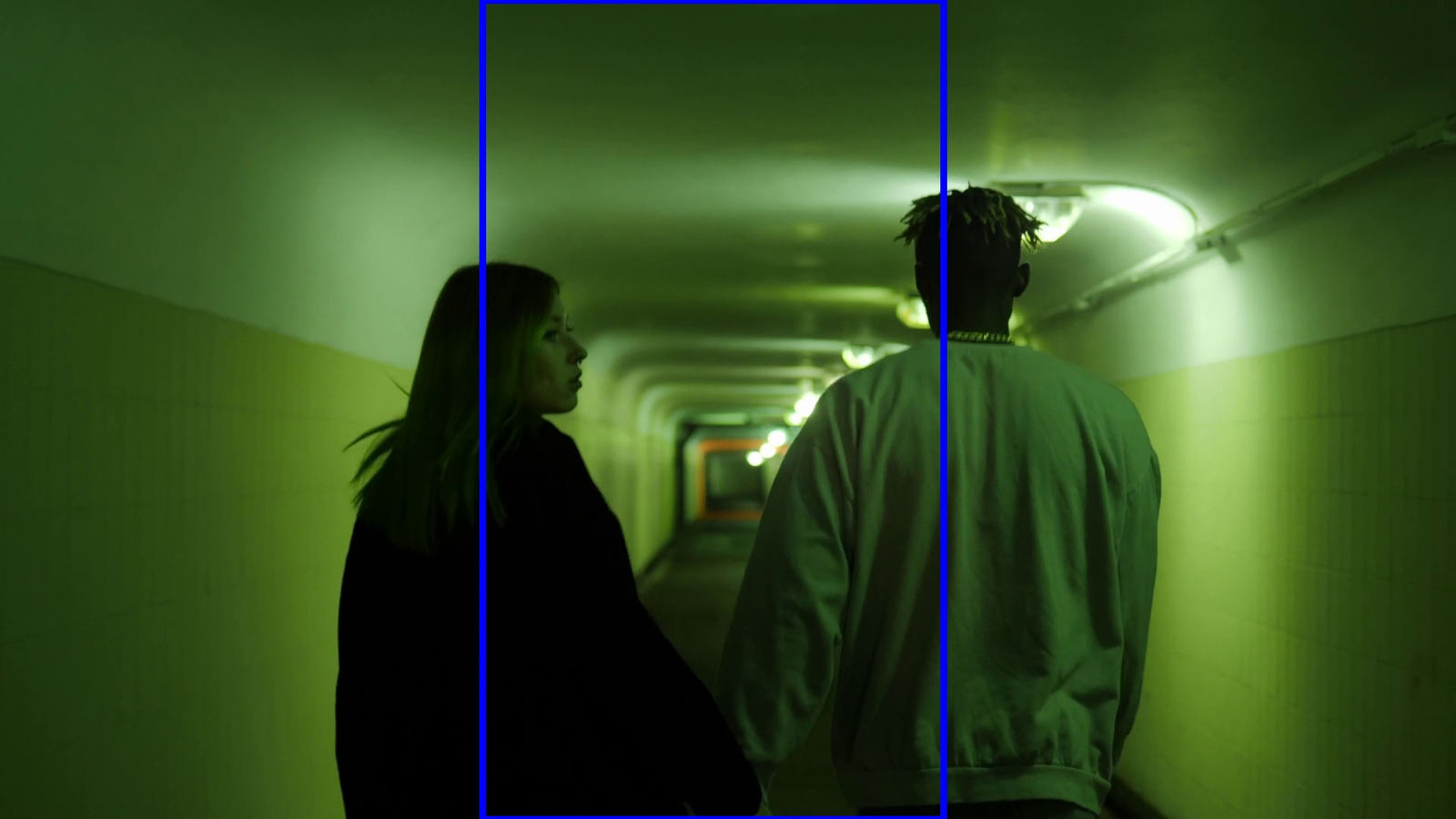} &
            \includegraphics[width=\fw]{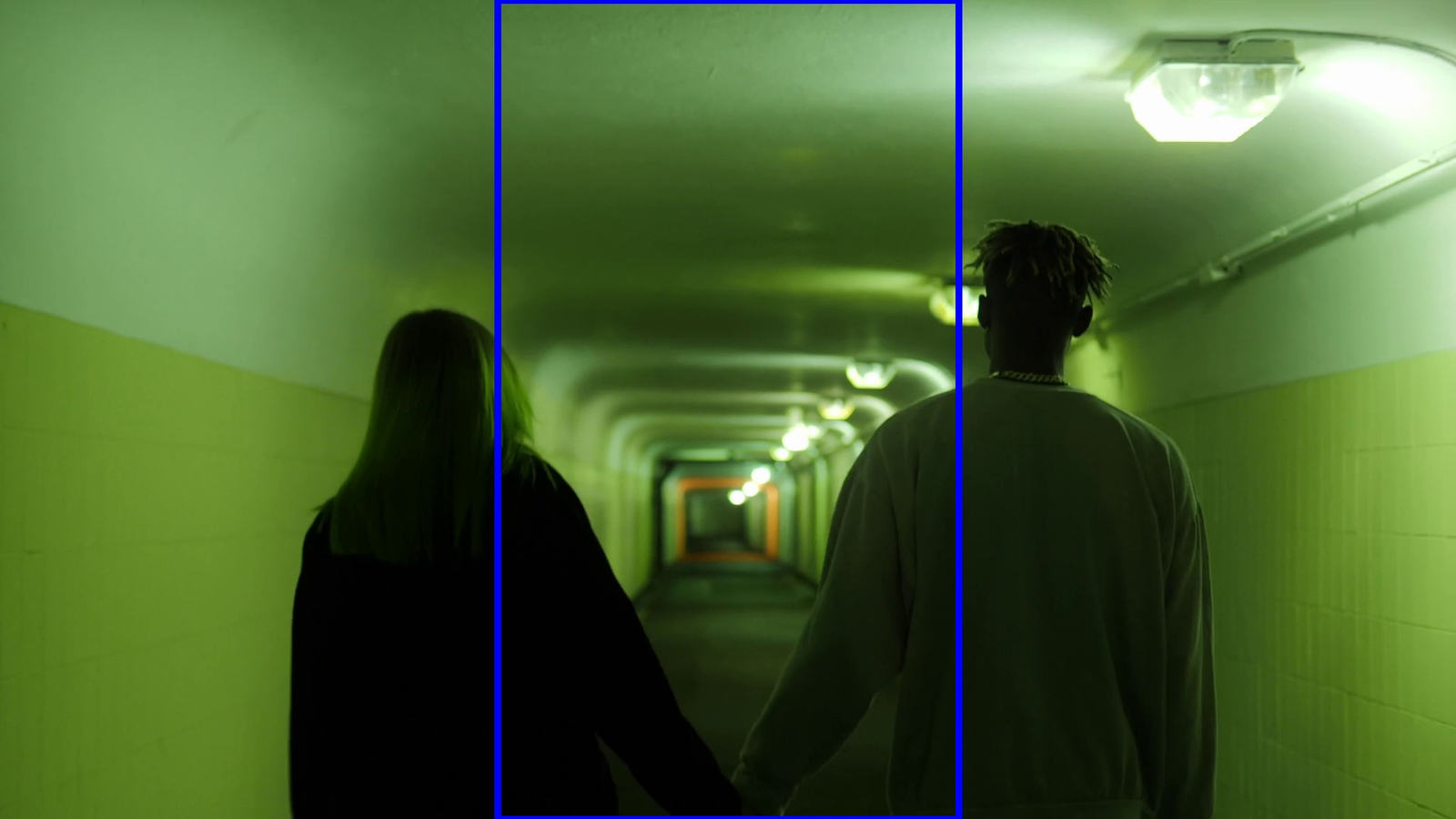} &
            \includegraphics[width=\fw]{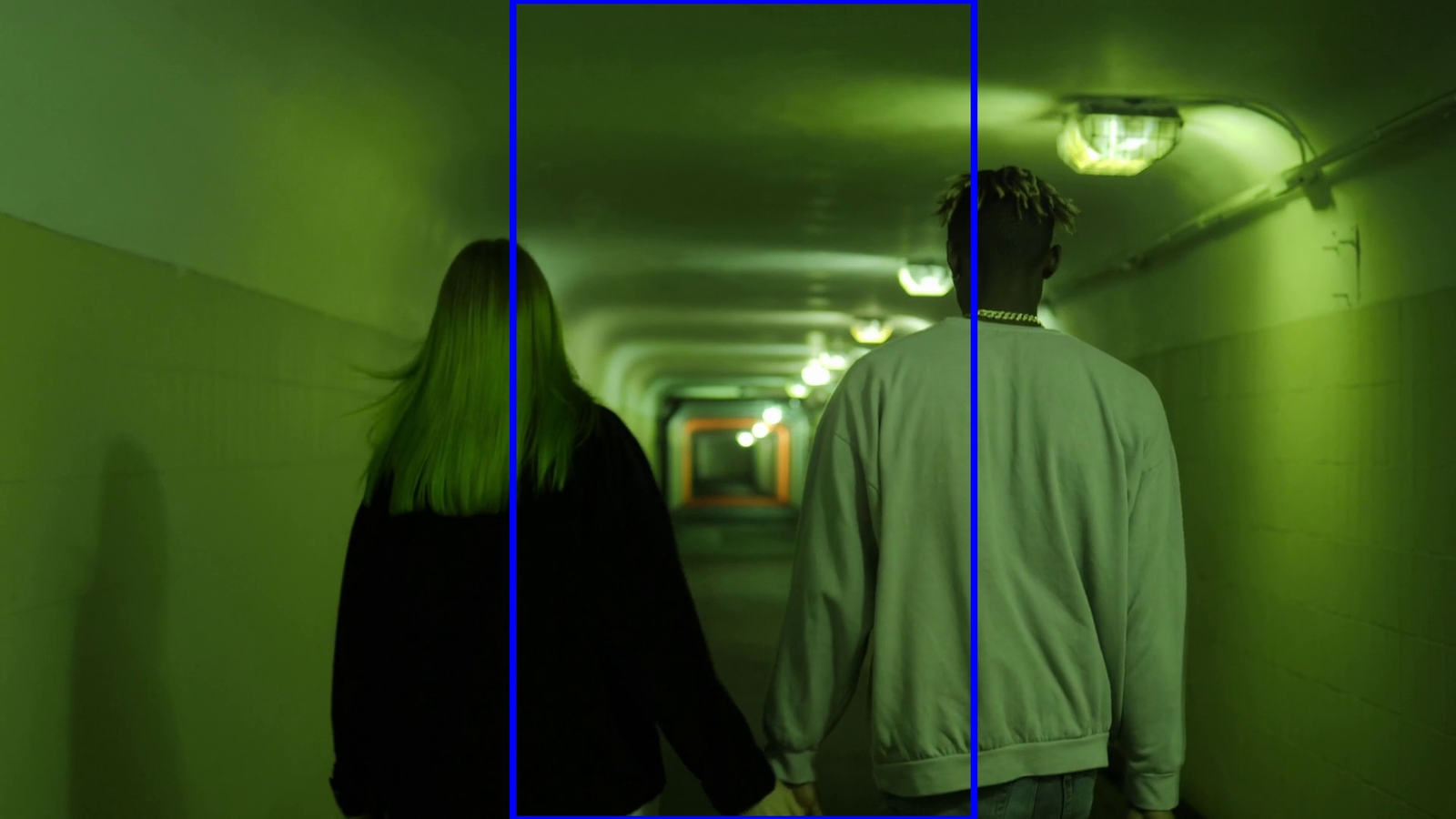} &
            \includegraphics[width=\fw]{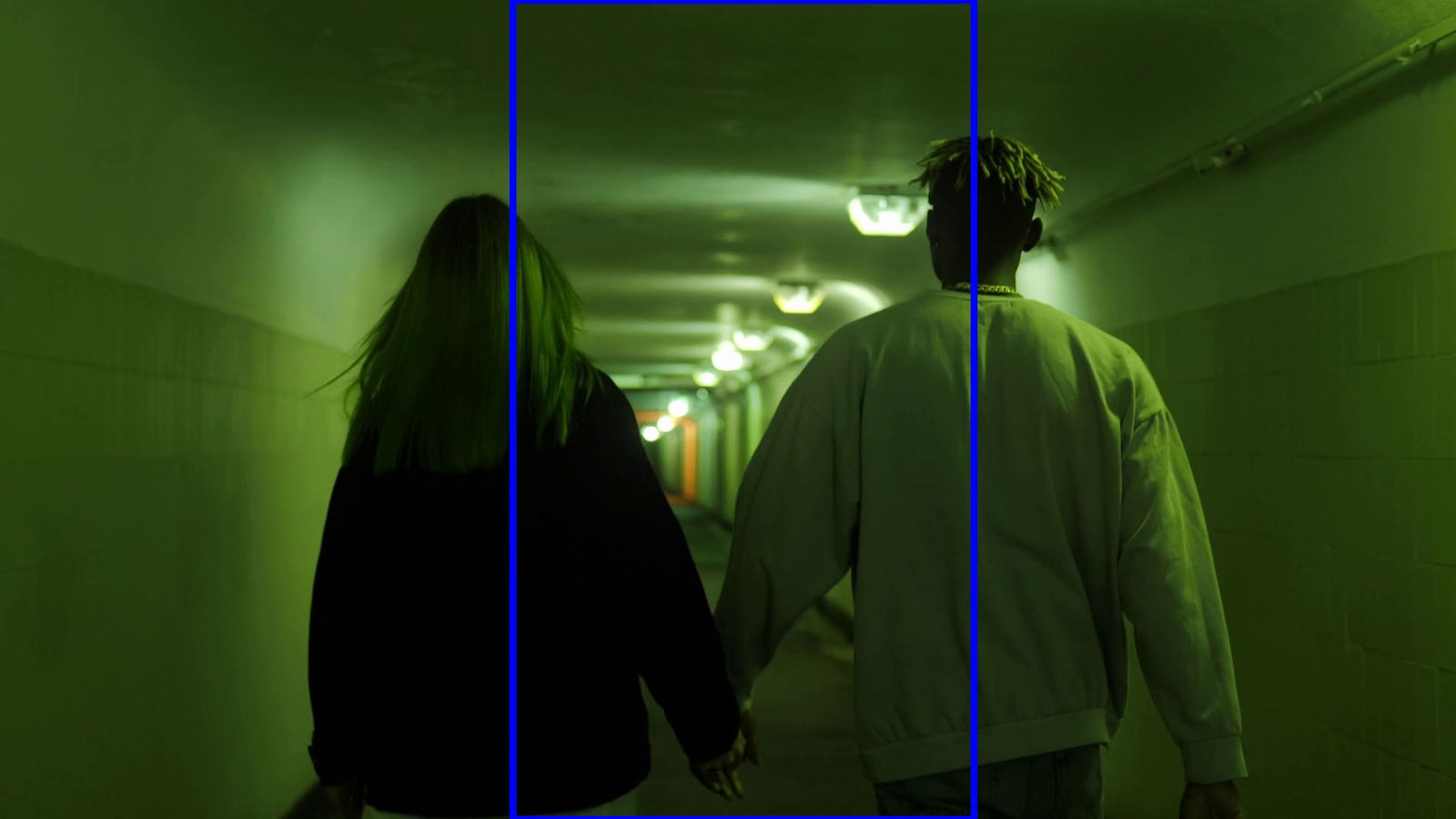} &
            \includegraphics[width=\fw]{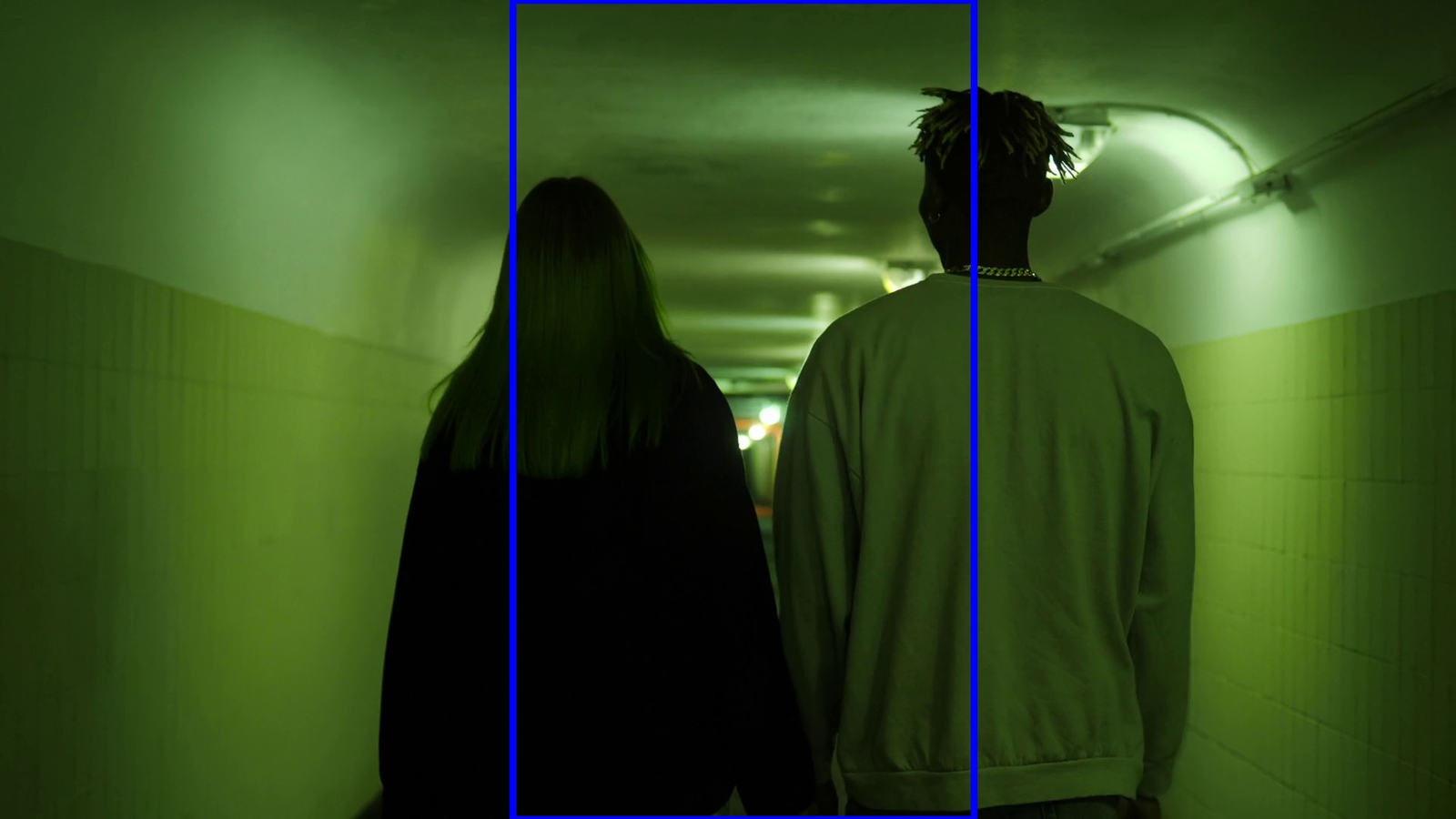} \\
            \rotatebox{90}{\footnotesize Ours} &
            \includegraphics[width=\fw]{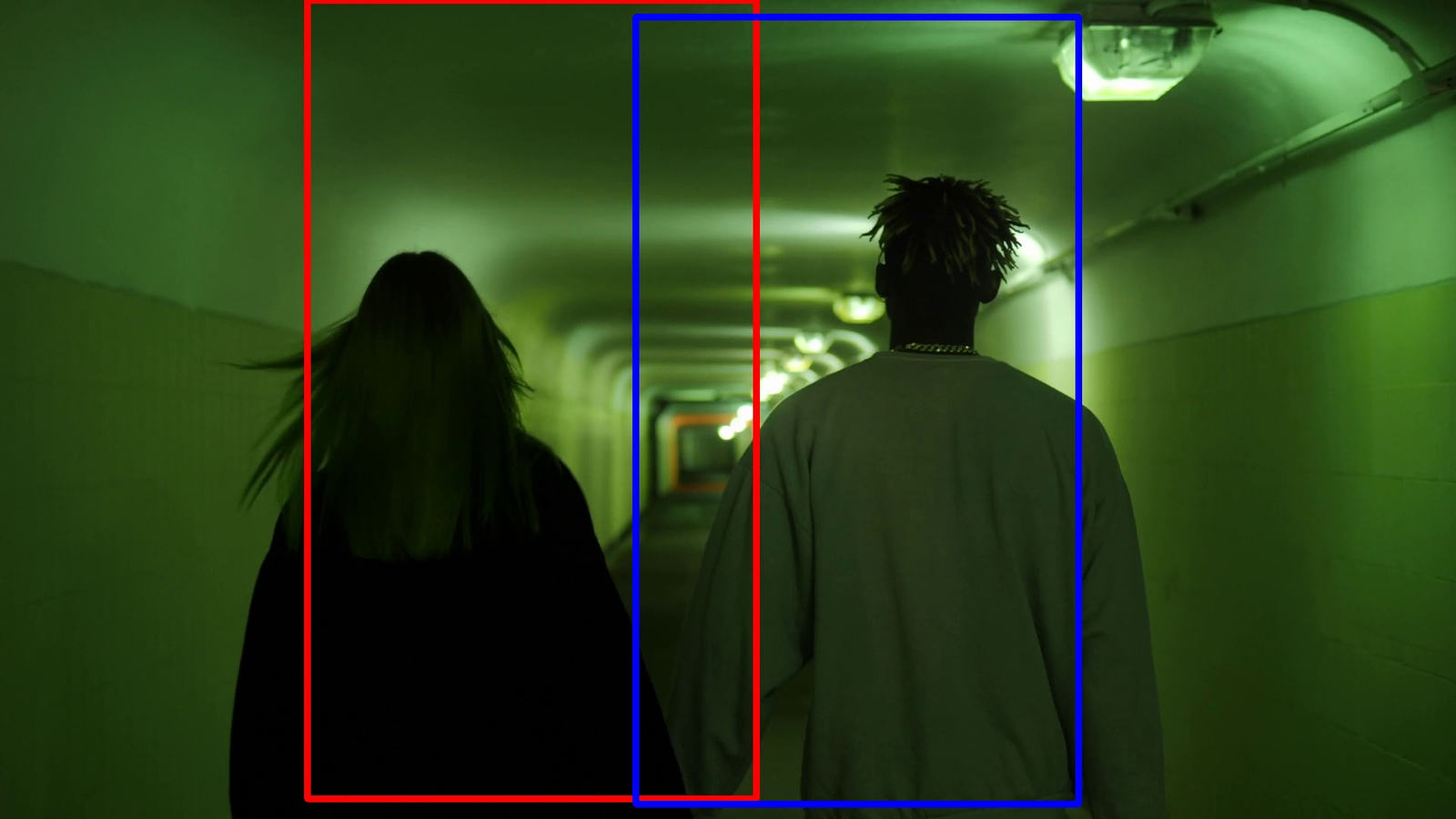} &
            \includegraphics[width=\fw]{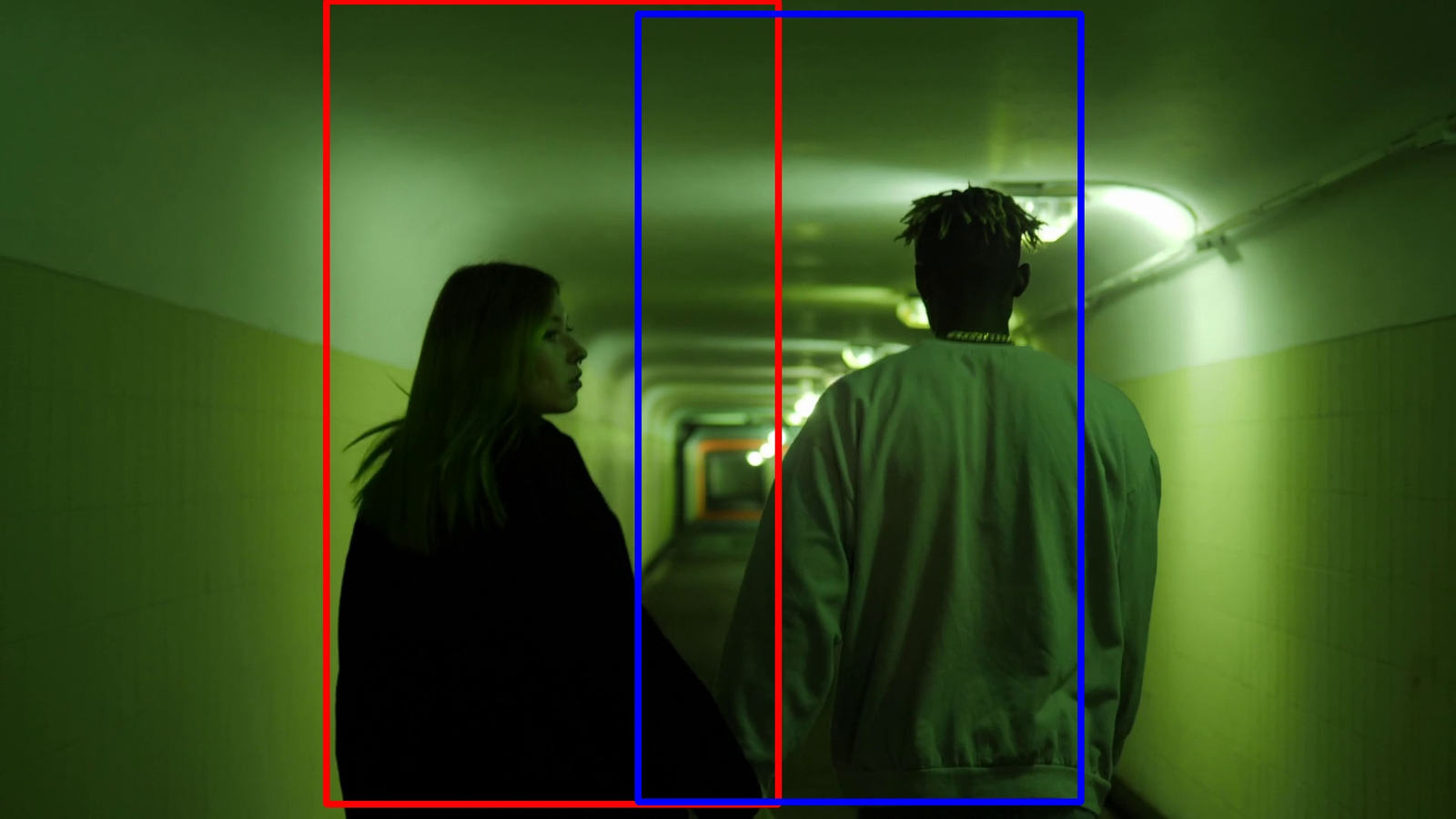} &
            \includegraphics[width=\fw]{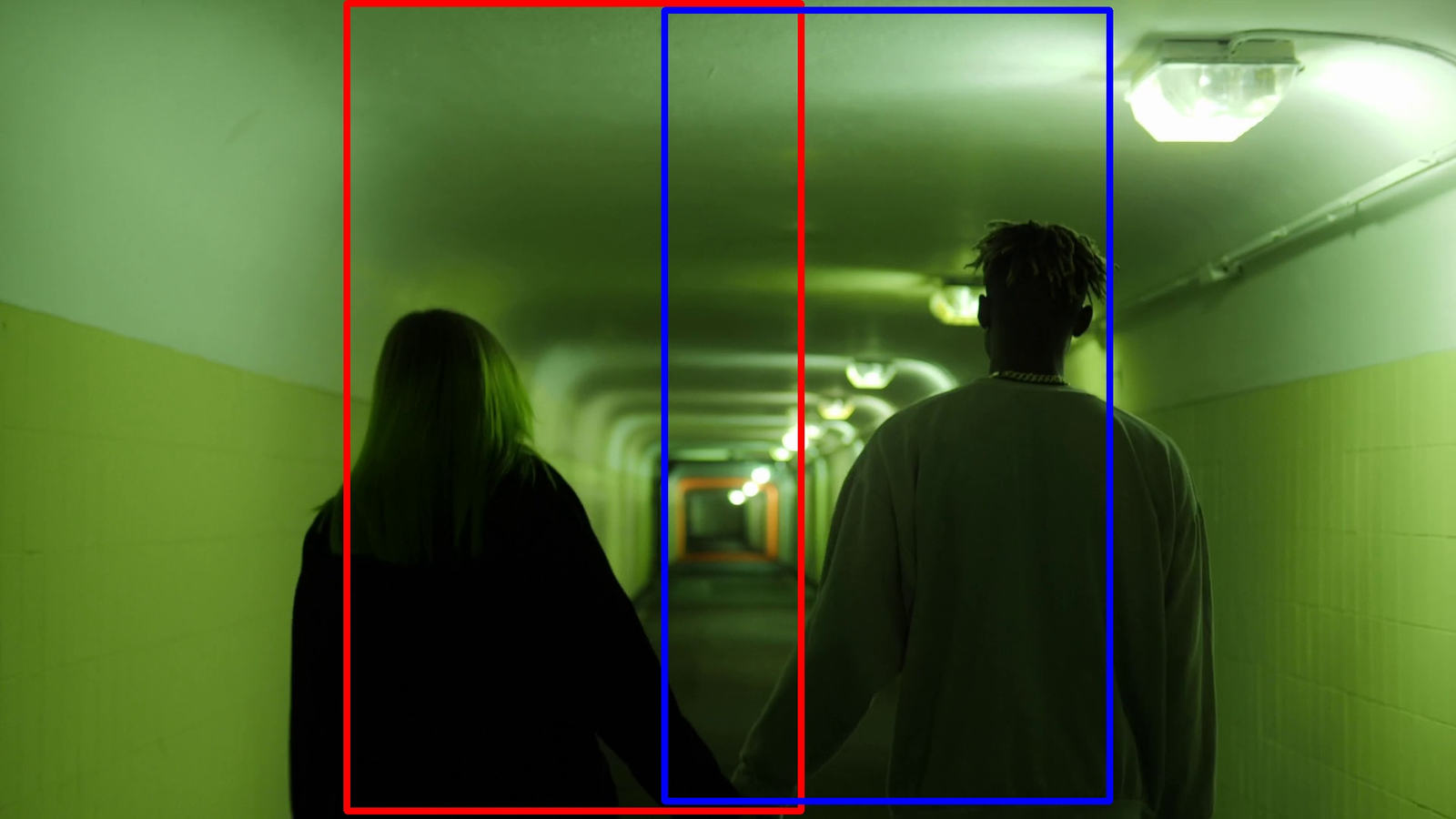} &
            \includegraphics[width=\fw]{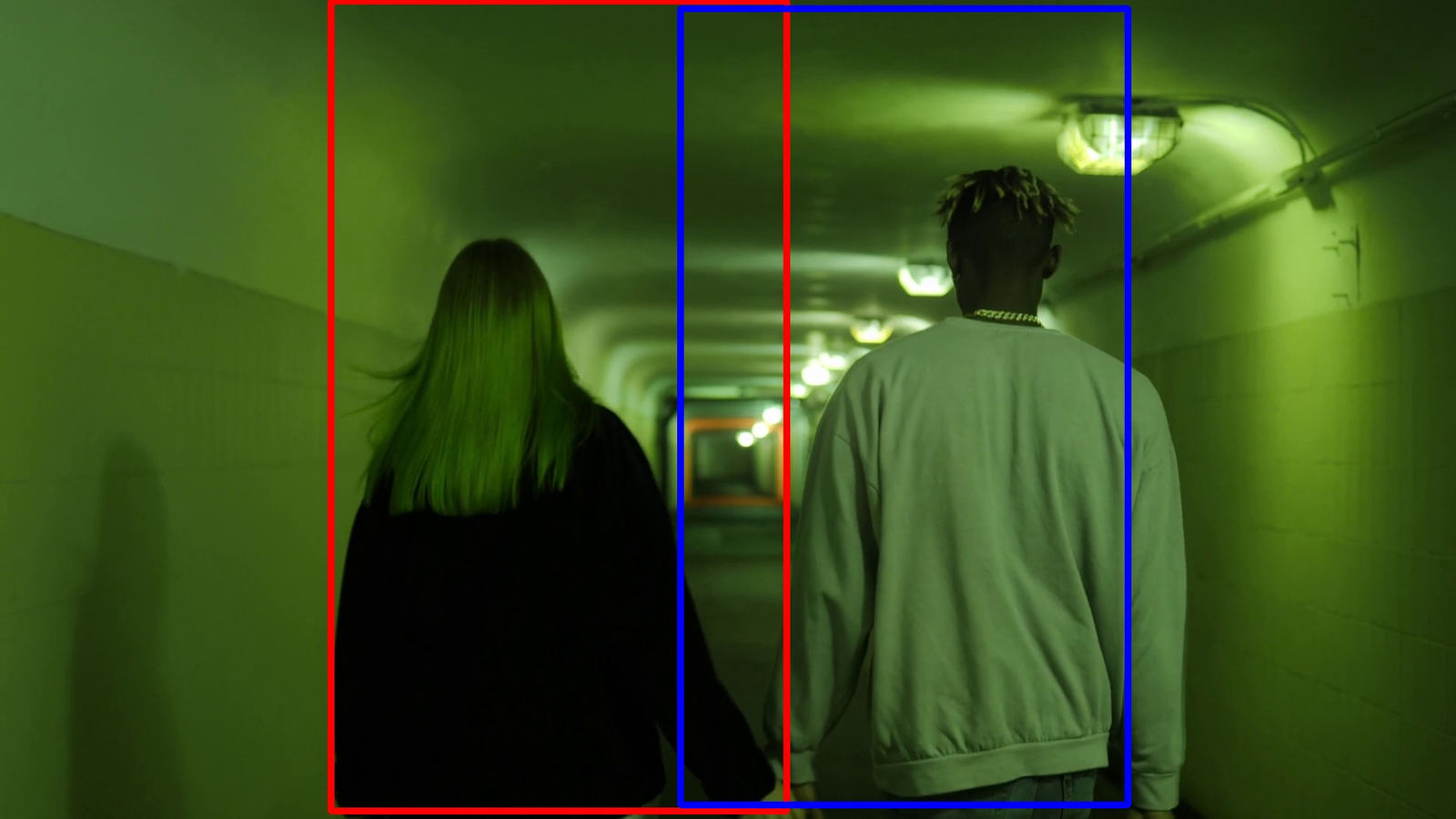} &
            \includegraphics[width=\fw]{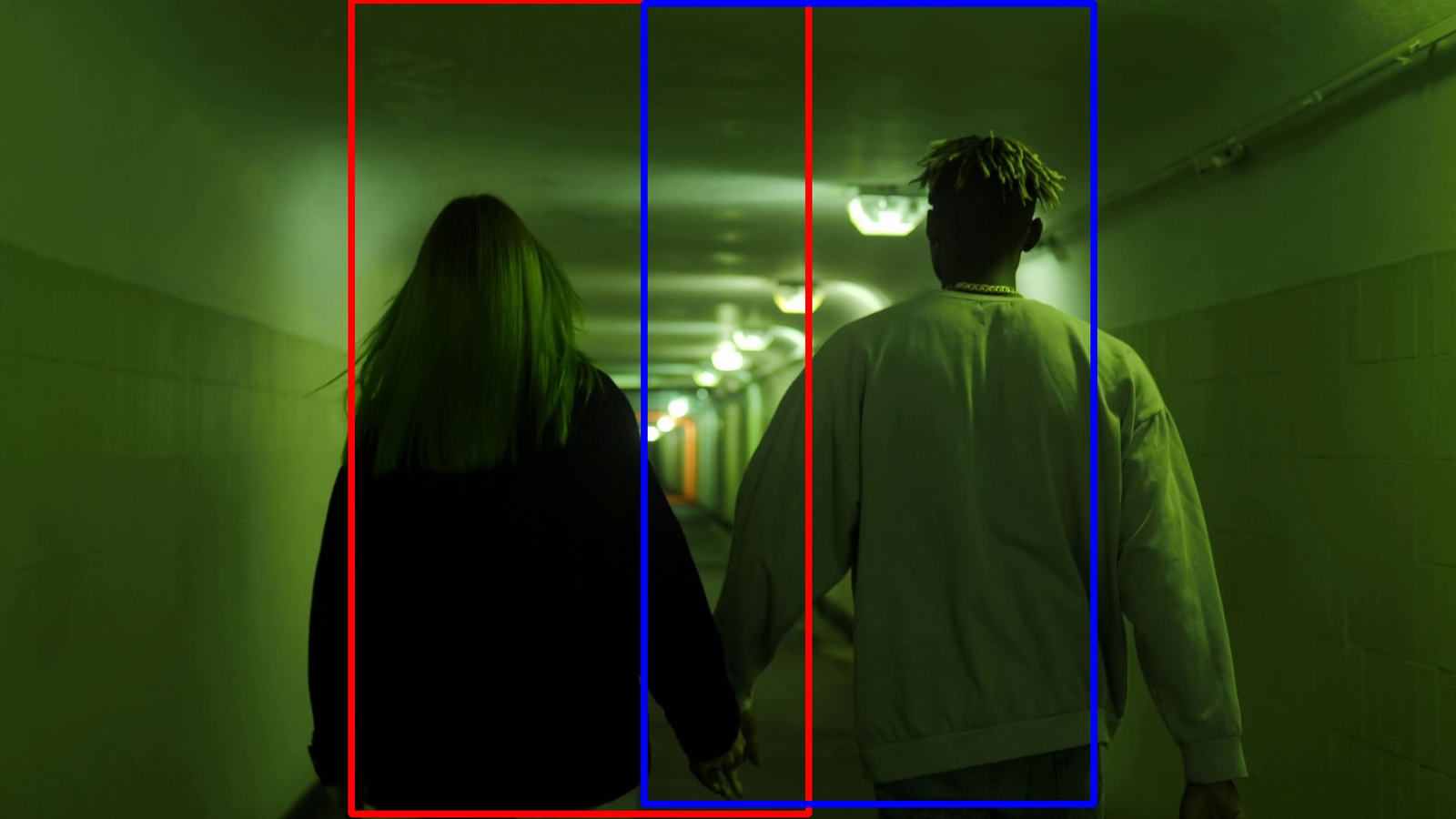} &
            \includegraphics[width=\fw]{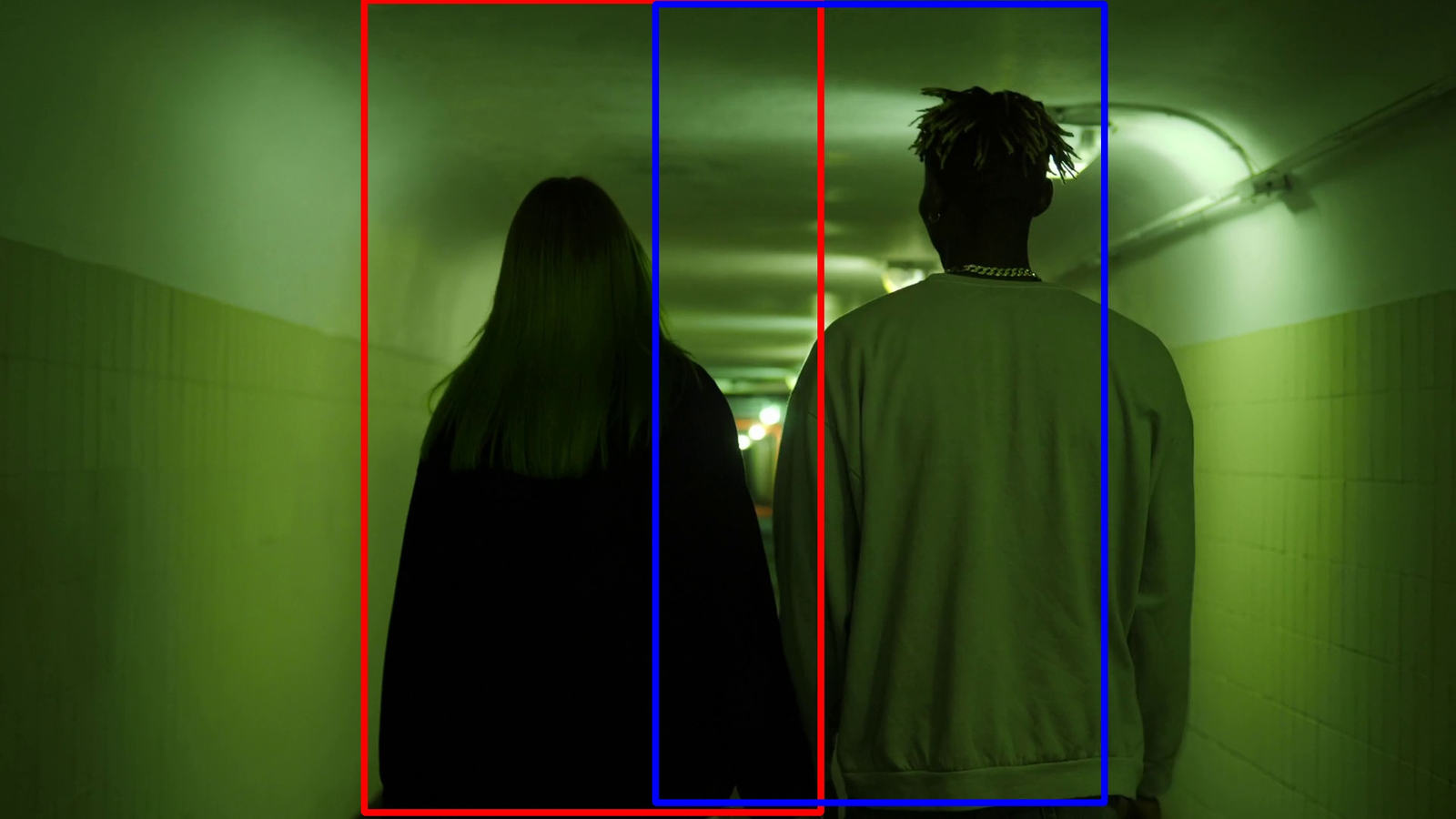} \\
            \rotatebox{90}{\footnotesize Ours + MLLM-based Agent} &
            \includegraphics[width=\fw]{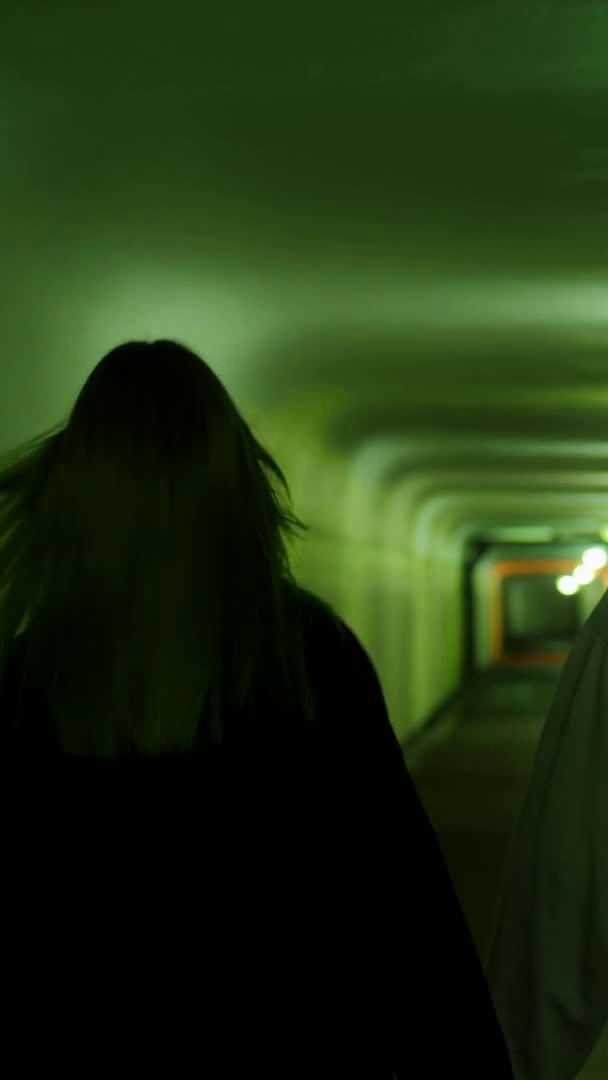} &
            \includegraphics[width=\fw]{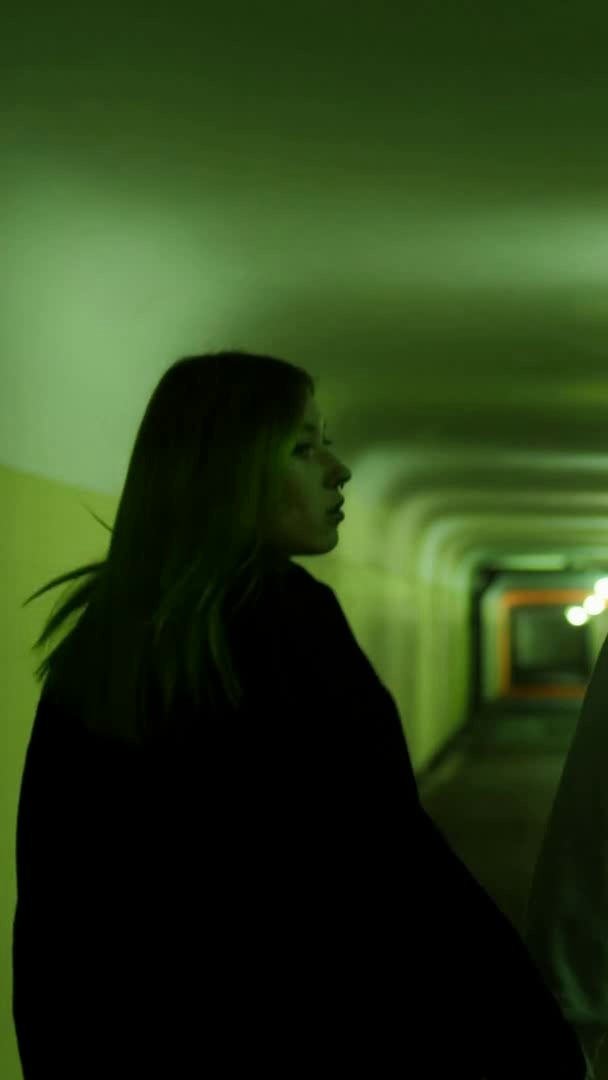} &
            \includegraphics[width=\fw]{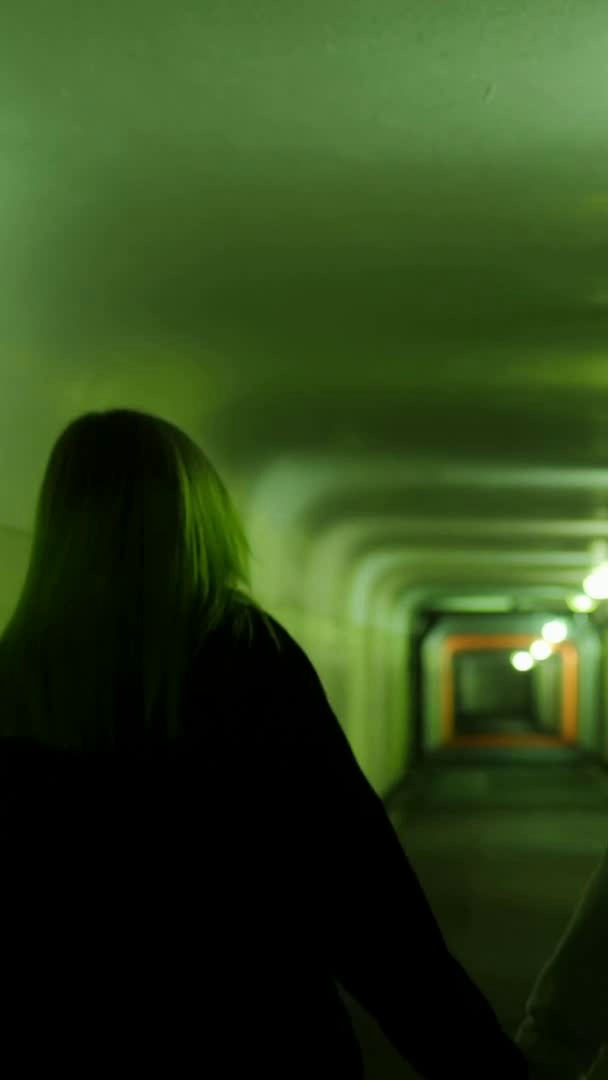} &
            \includegraphics[width=\fw]{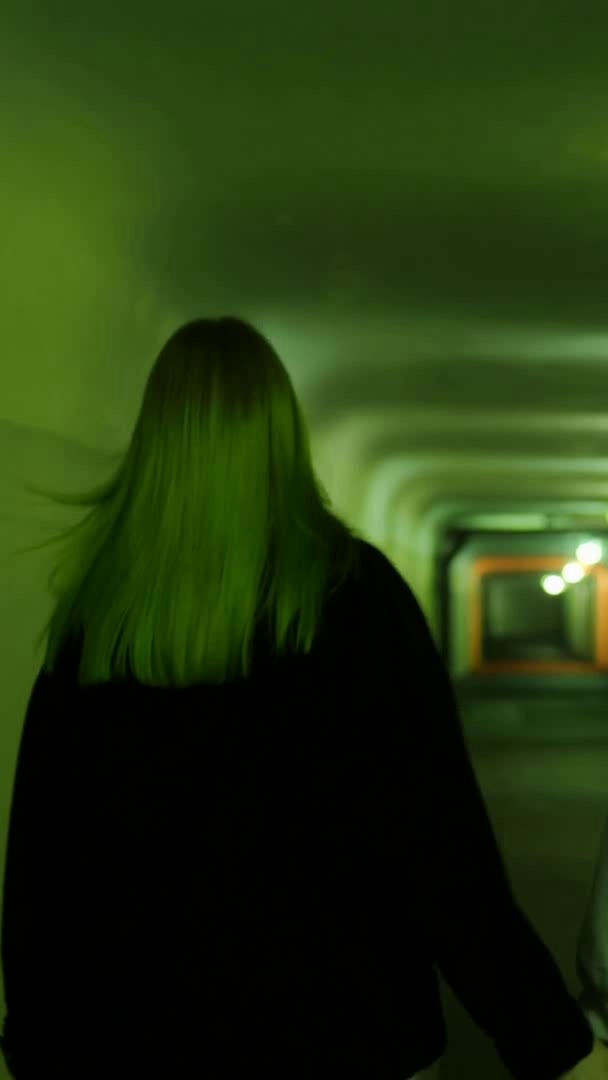} &
            \includegraphics[width=\fw]{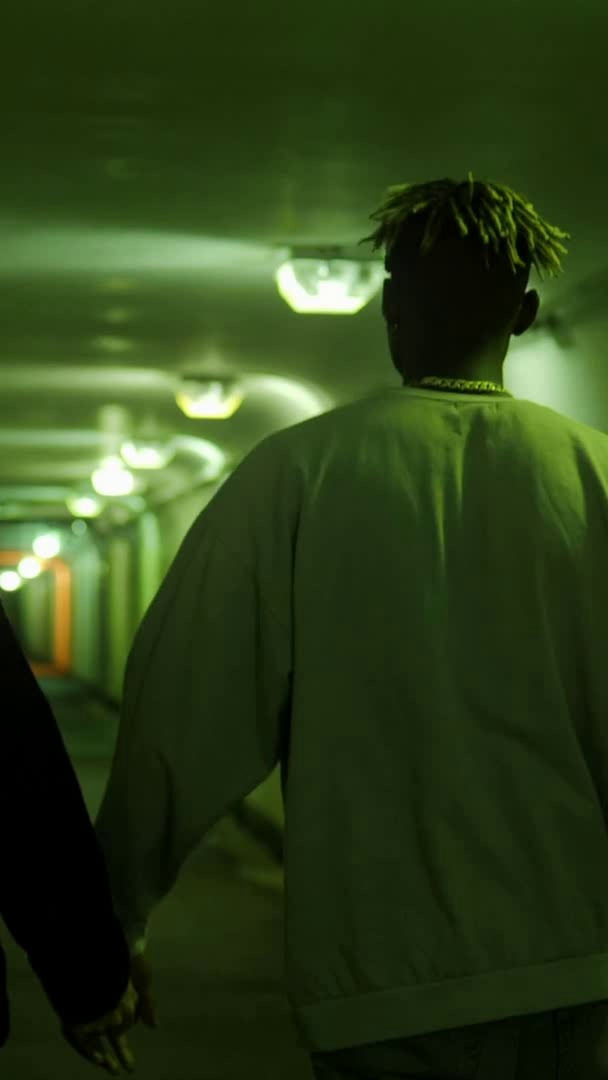} &
            \includegraphics[width=\fw]{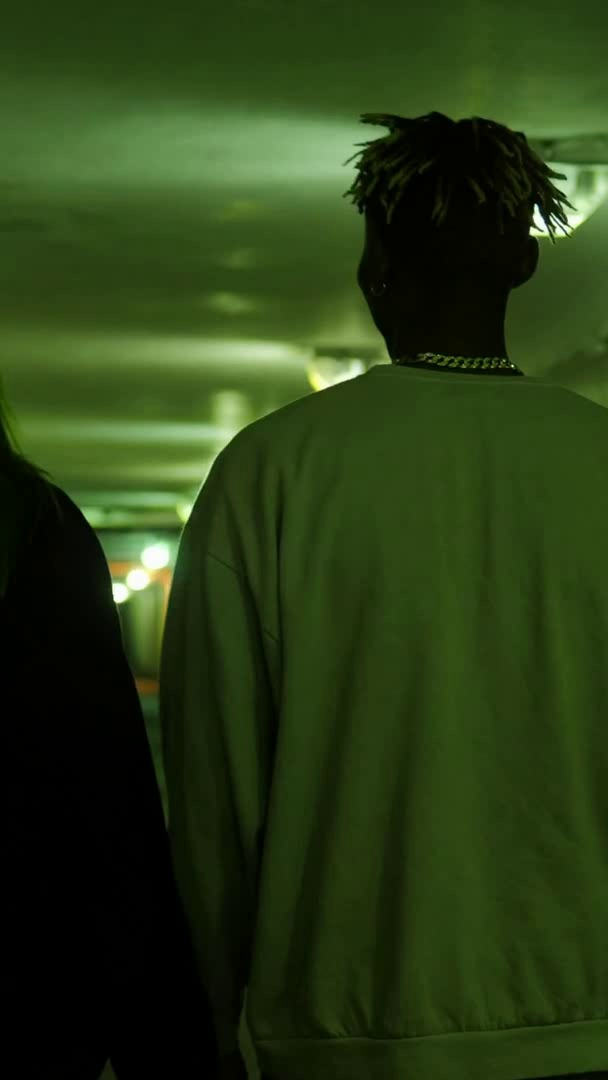} \\
        \end{tabular}
        \caption{Video ID: 4035642}
    \end{subfigure}
    \vspace{-10pt}
    
    \begin{subfigure}{\textwidth}
        \centering
        \begin{tabular}{ccccccc}
             & \textbf{4s} & \textbf{8s} & \textbf{12s} & \textbf{15s} & \textbf{19s} & \textbf{23s} \\
            \rotatebox{90}{\footnotesize Center Crop} &
            \includegraphics[width=\fw]{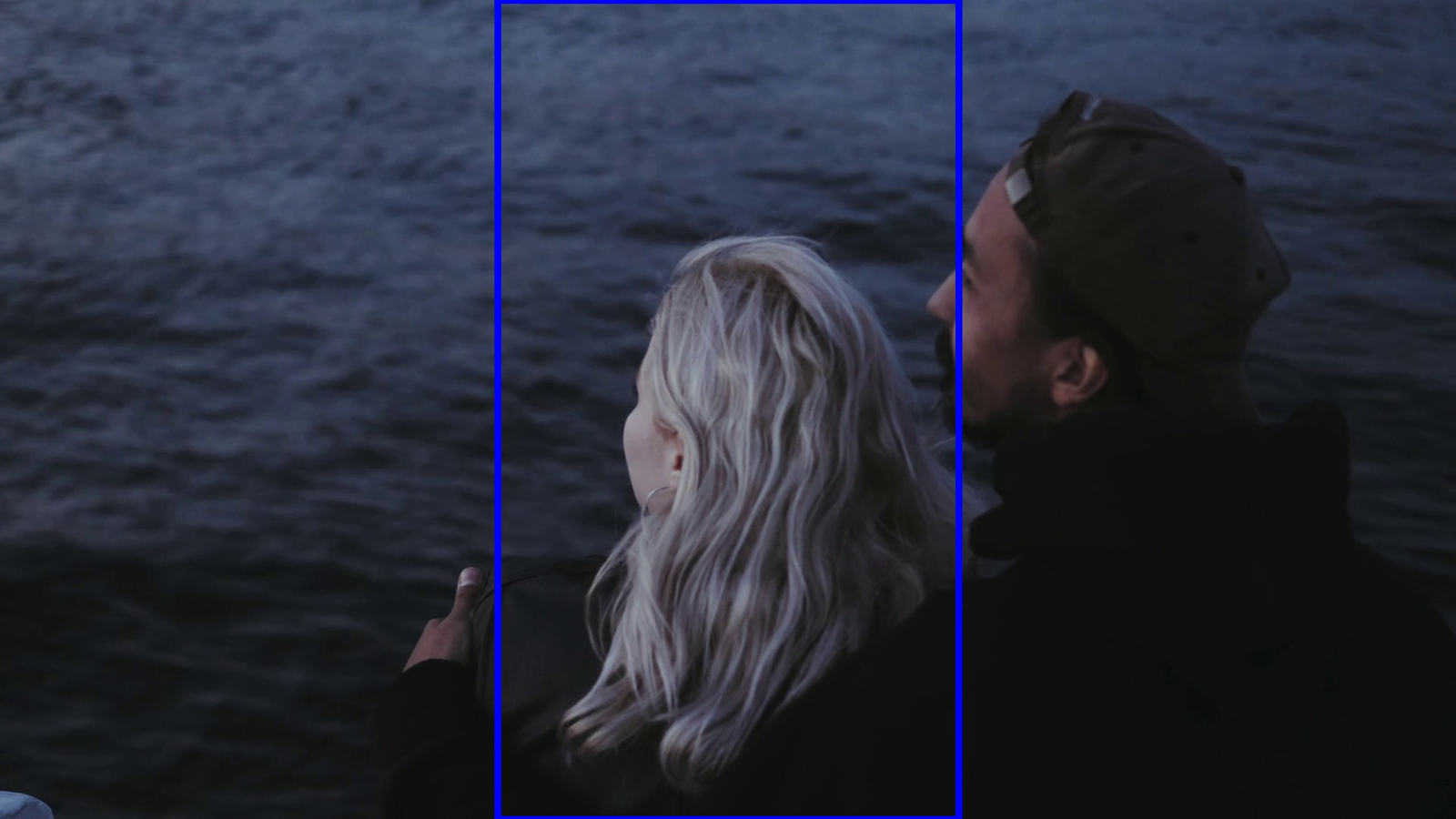} &
            \includegraphics[width=\fw]{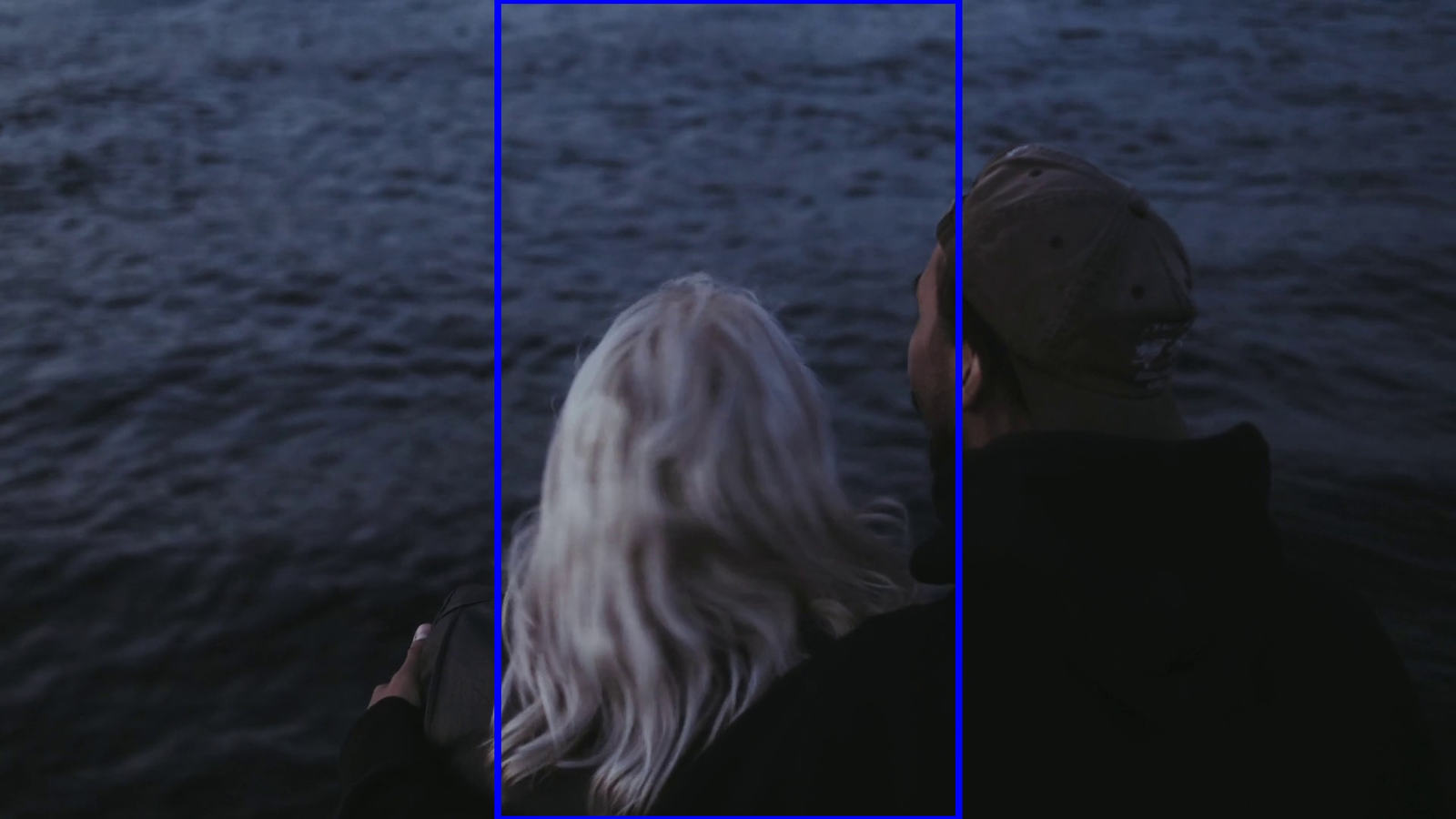} &
            \includegraphics[width=\fw]{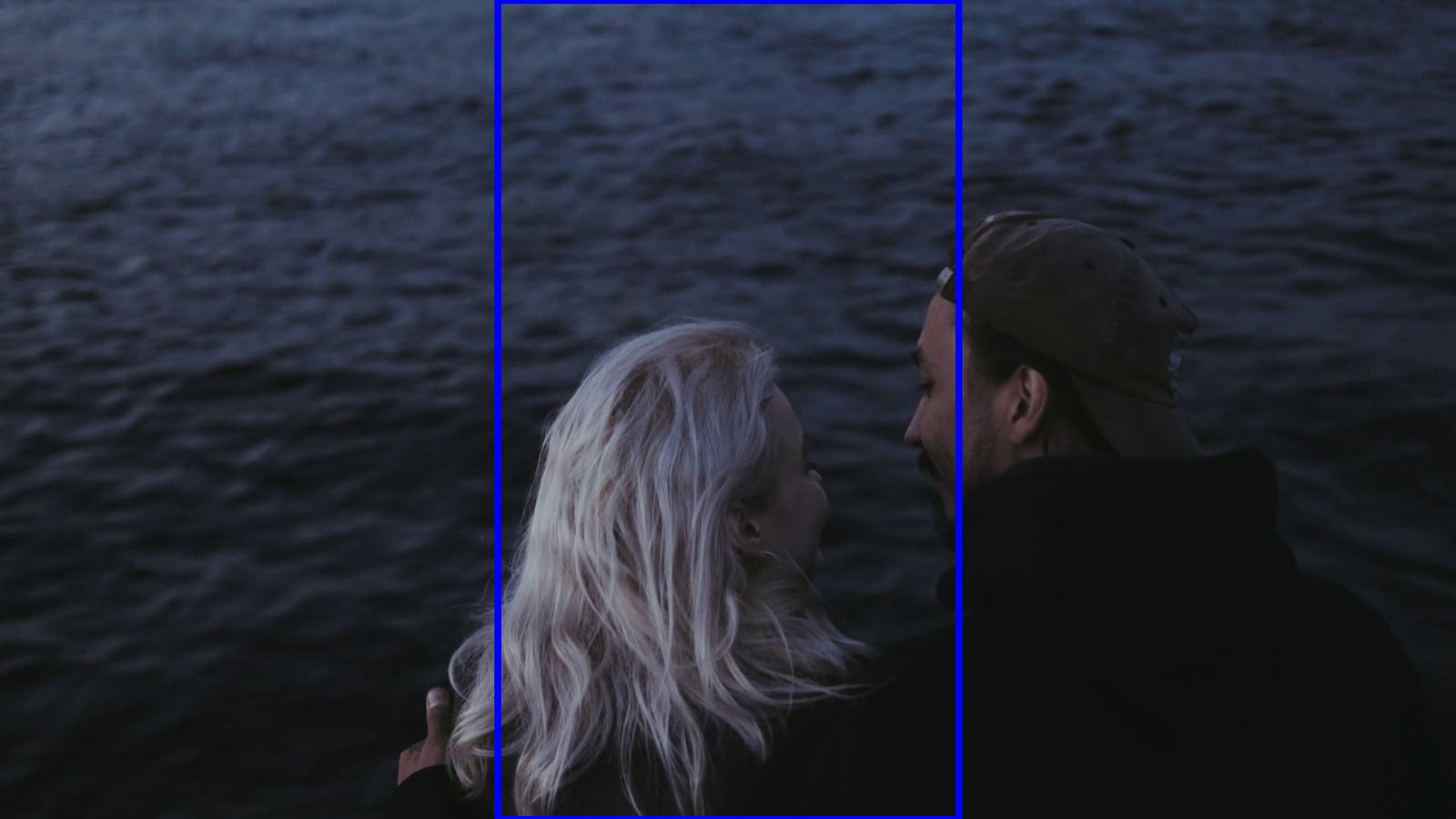} &
            \includegraphics[width=\fw]{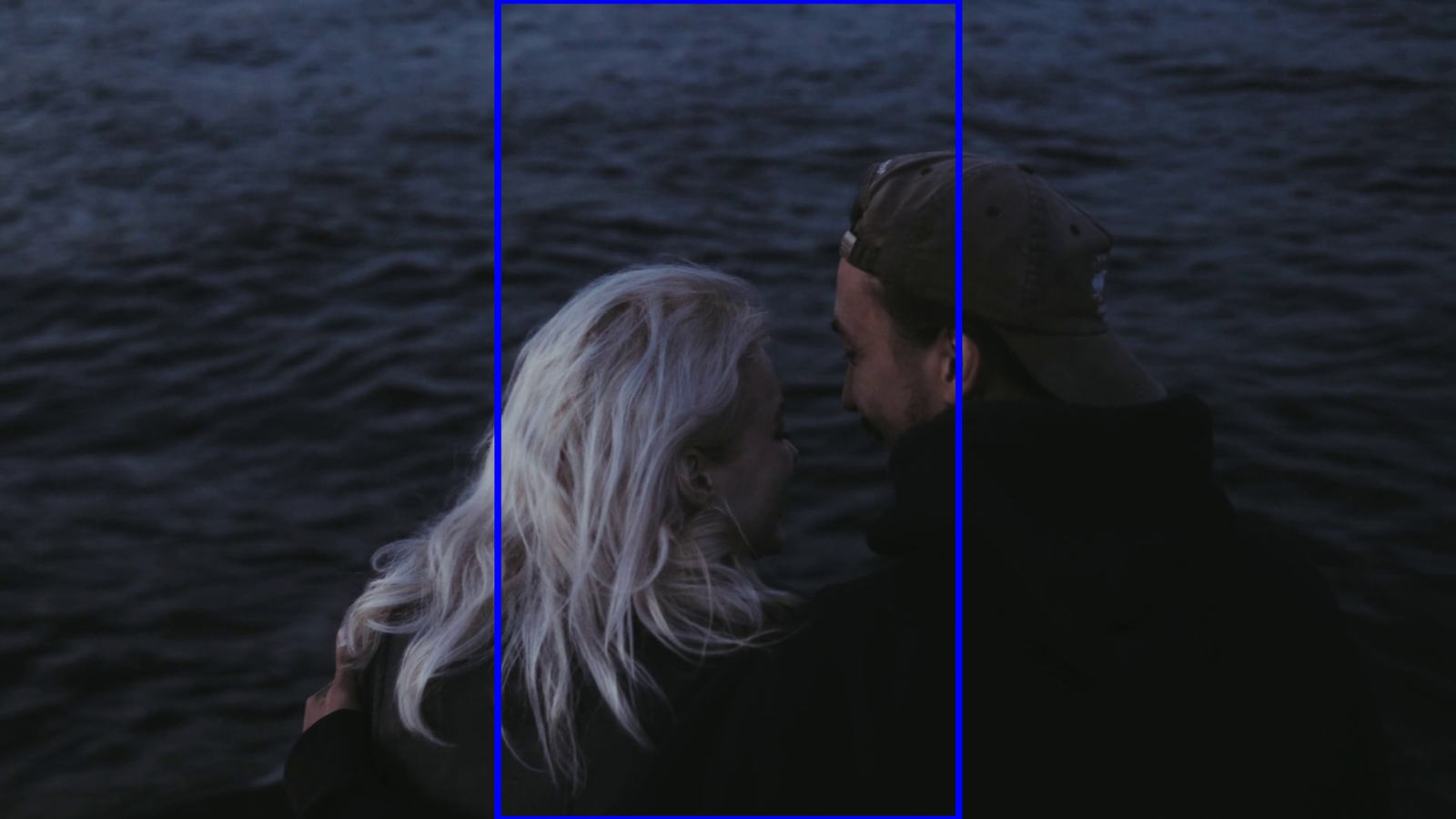} &
            \includegraphics[width=\fw]{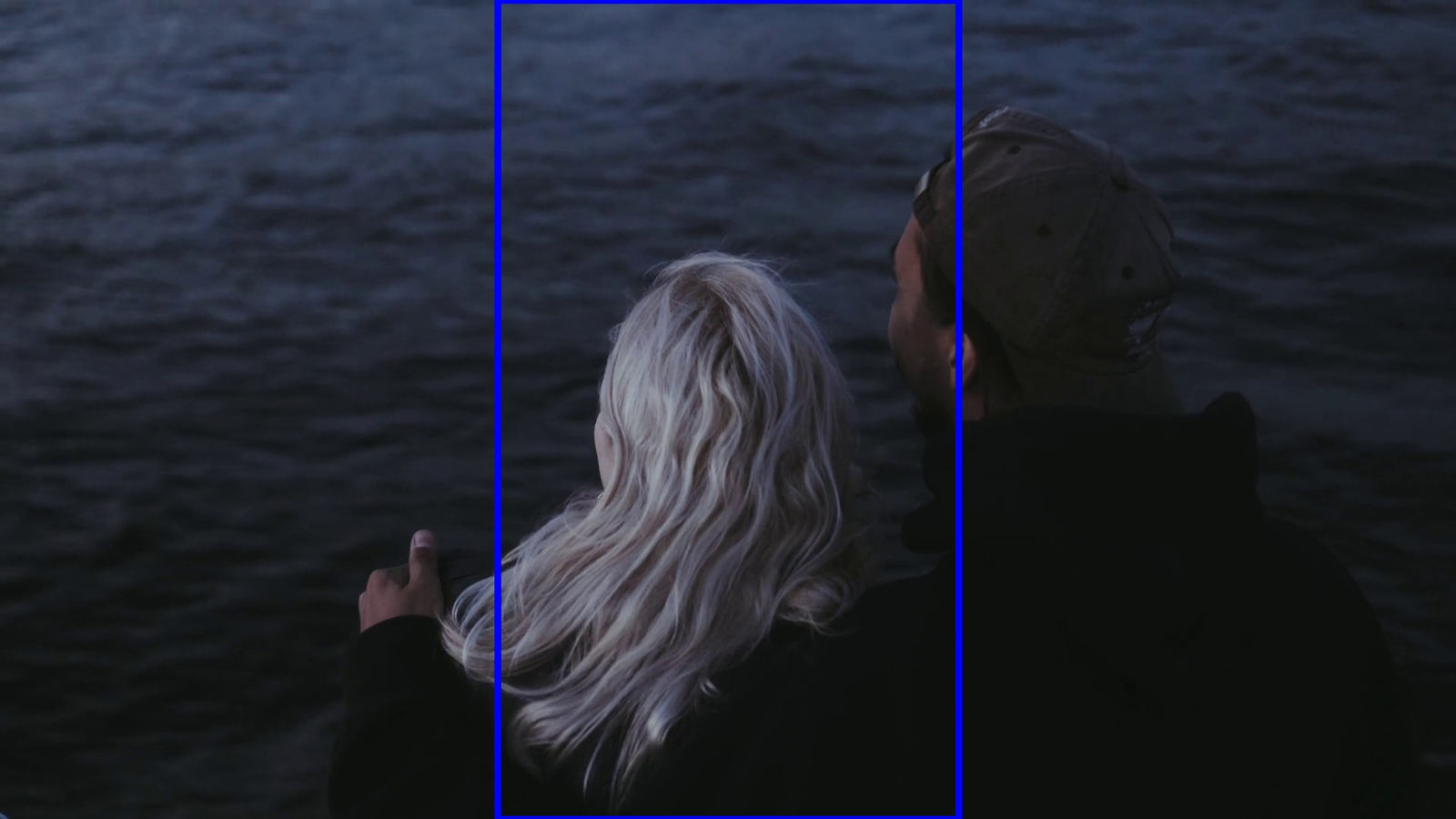} &
            \includegraphics[width=\fw]{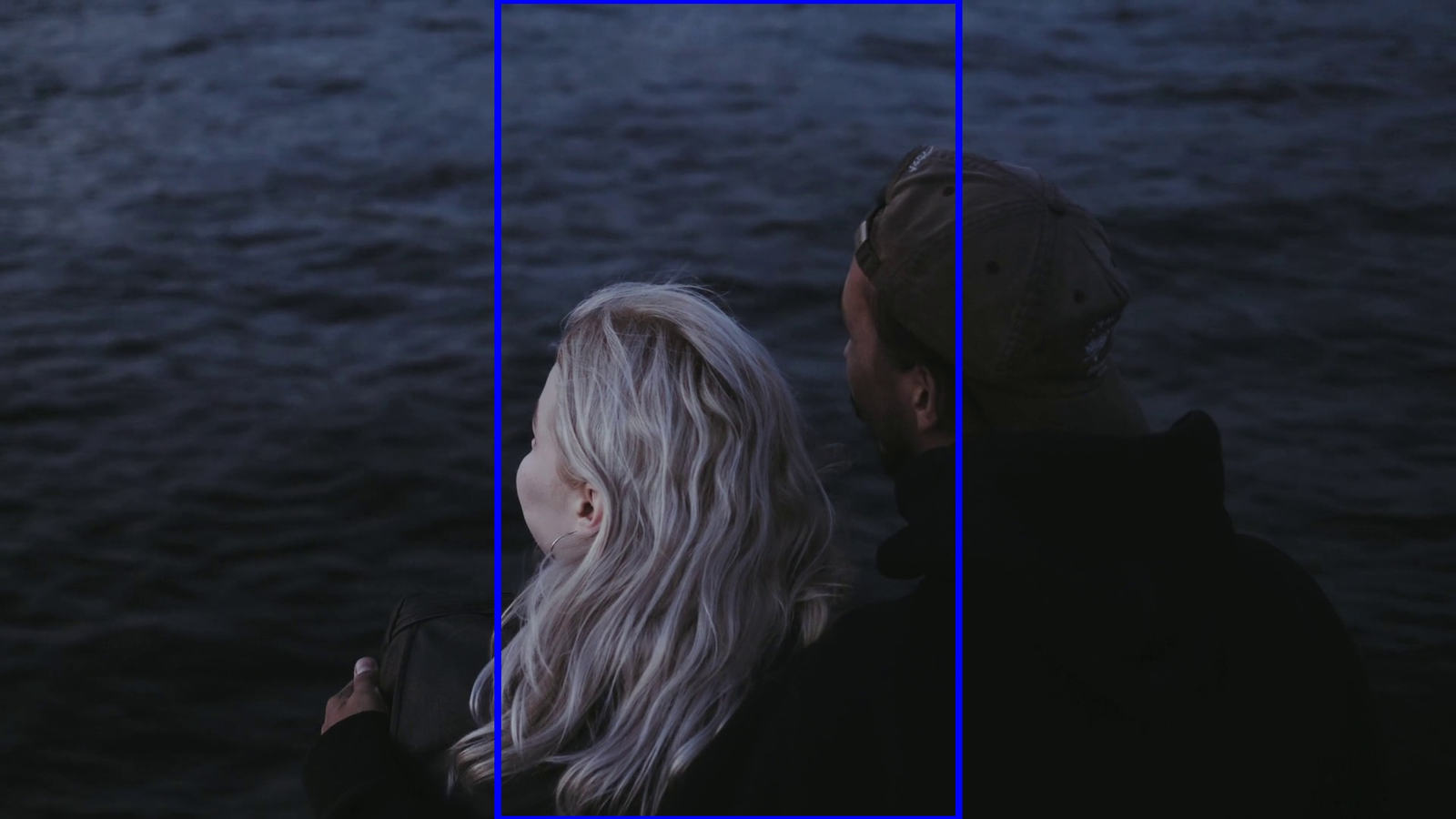} \\
            \rotatebox{90}{\footnotesize Gemini-Base} &
            \includegraphics[width=\fw]{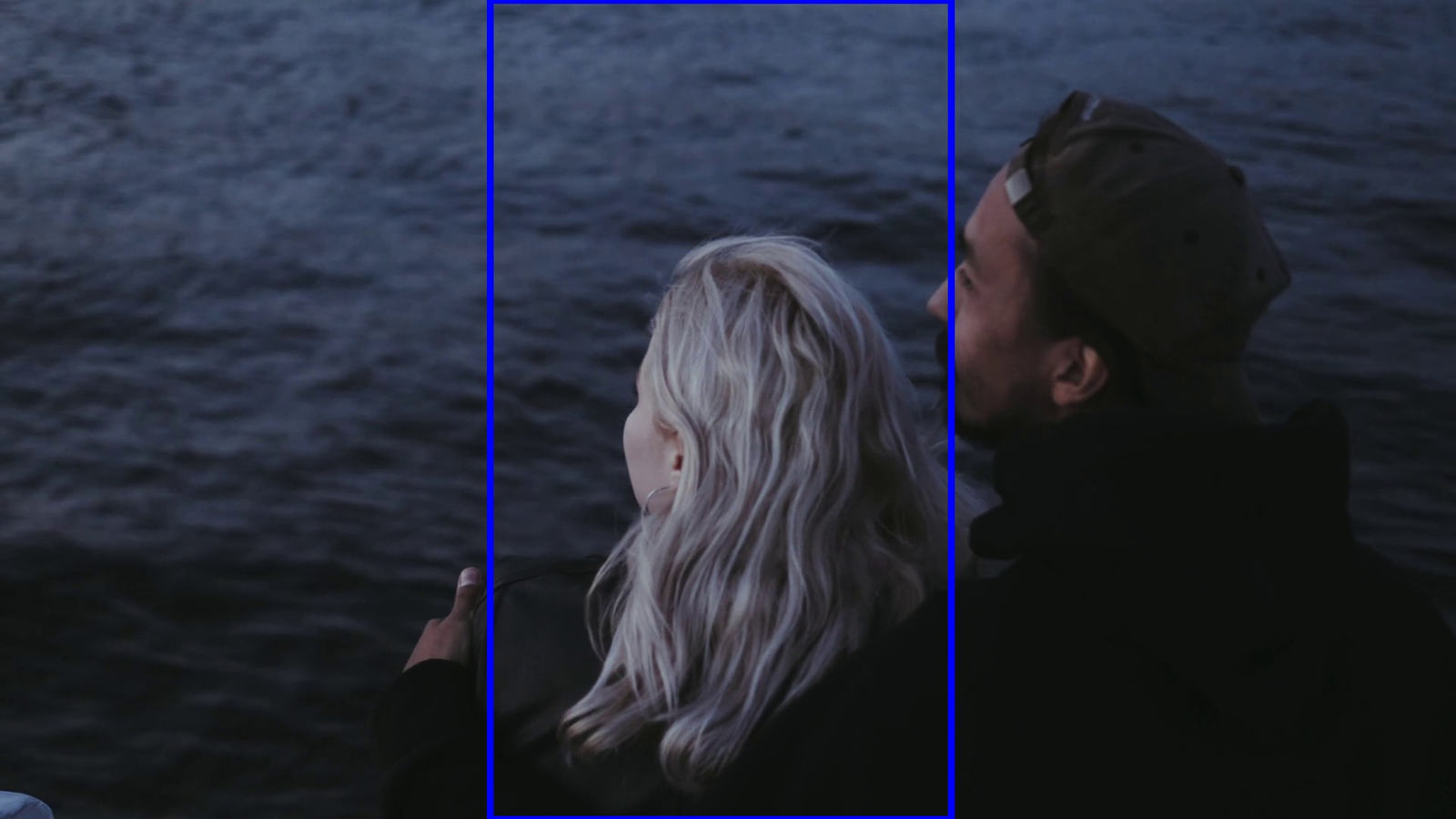} &
            \includegraphics[width=\fw]{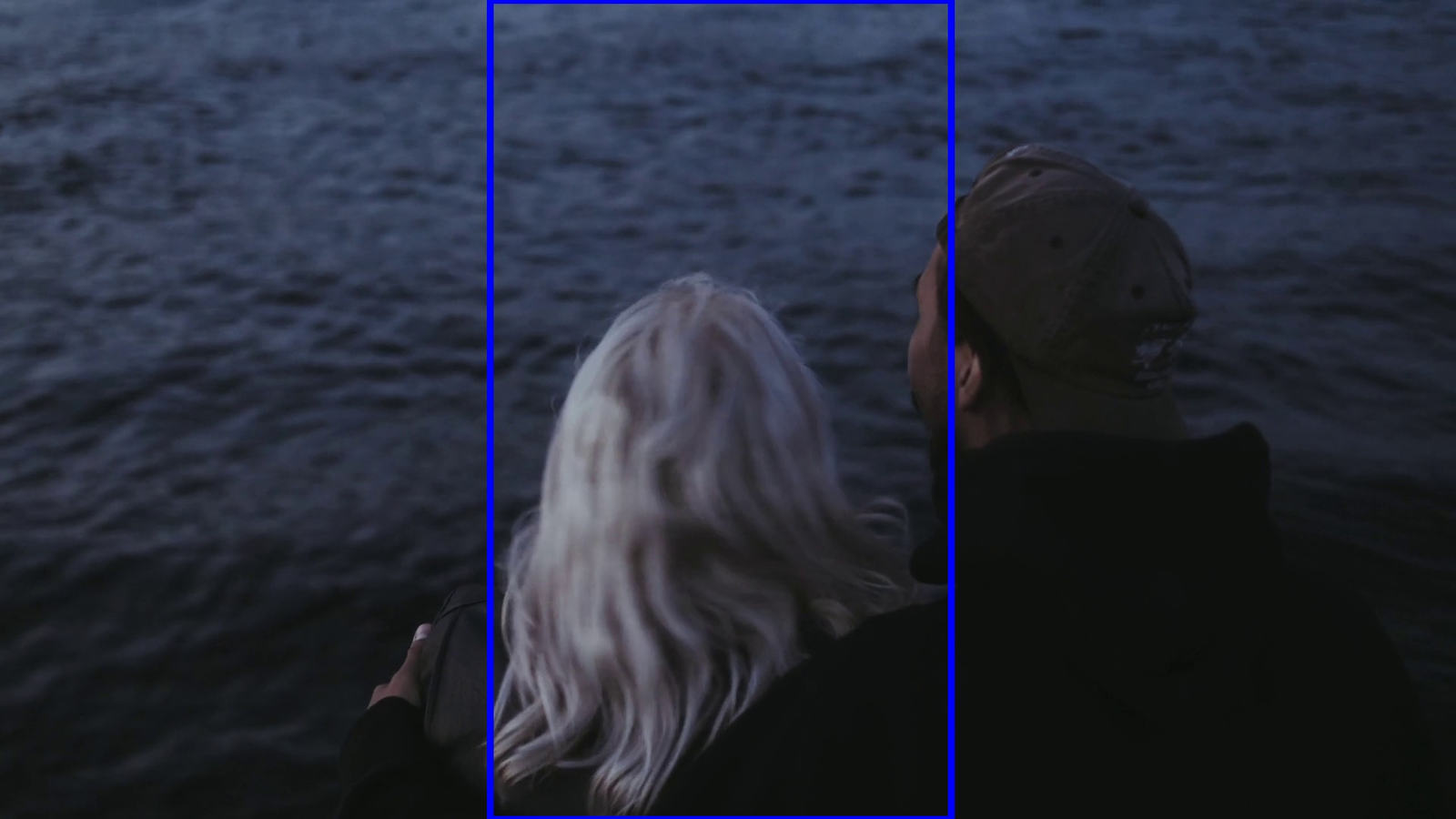} &
            \includegraphics[width=\fw]{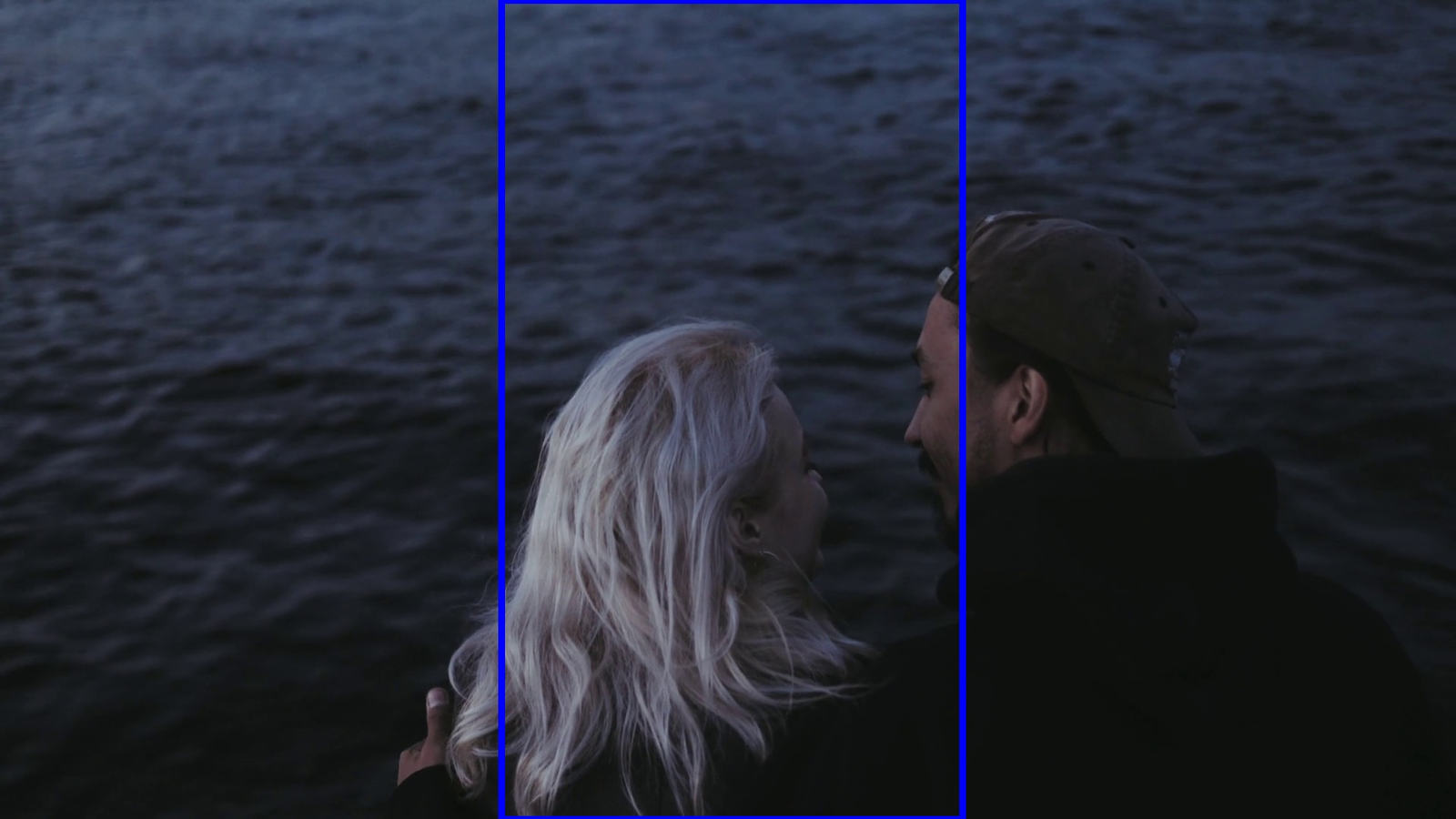} &
            \includegraphics[width=\fw]{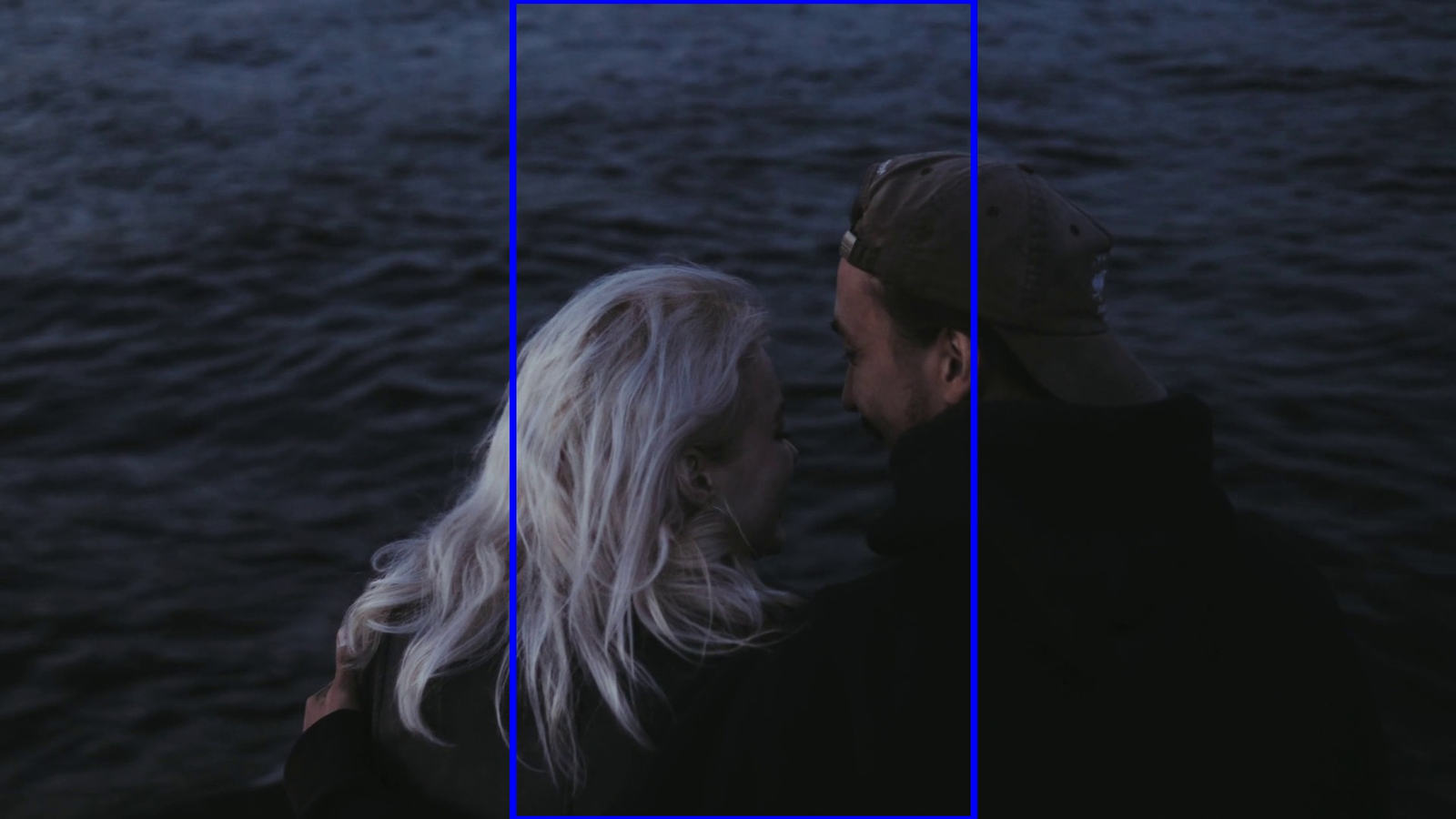} &
            \includegraphics[width=\fw]{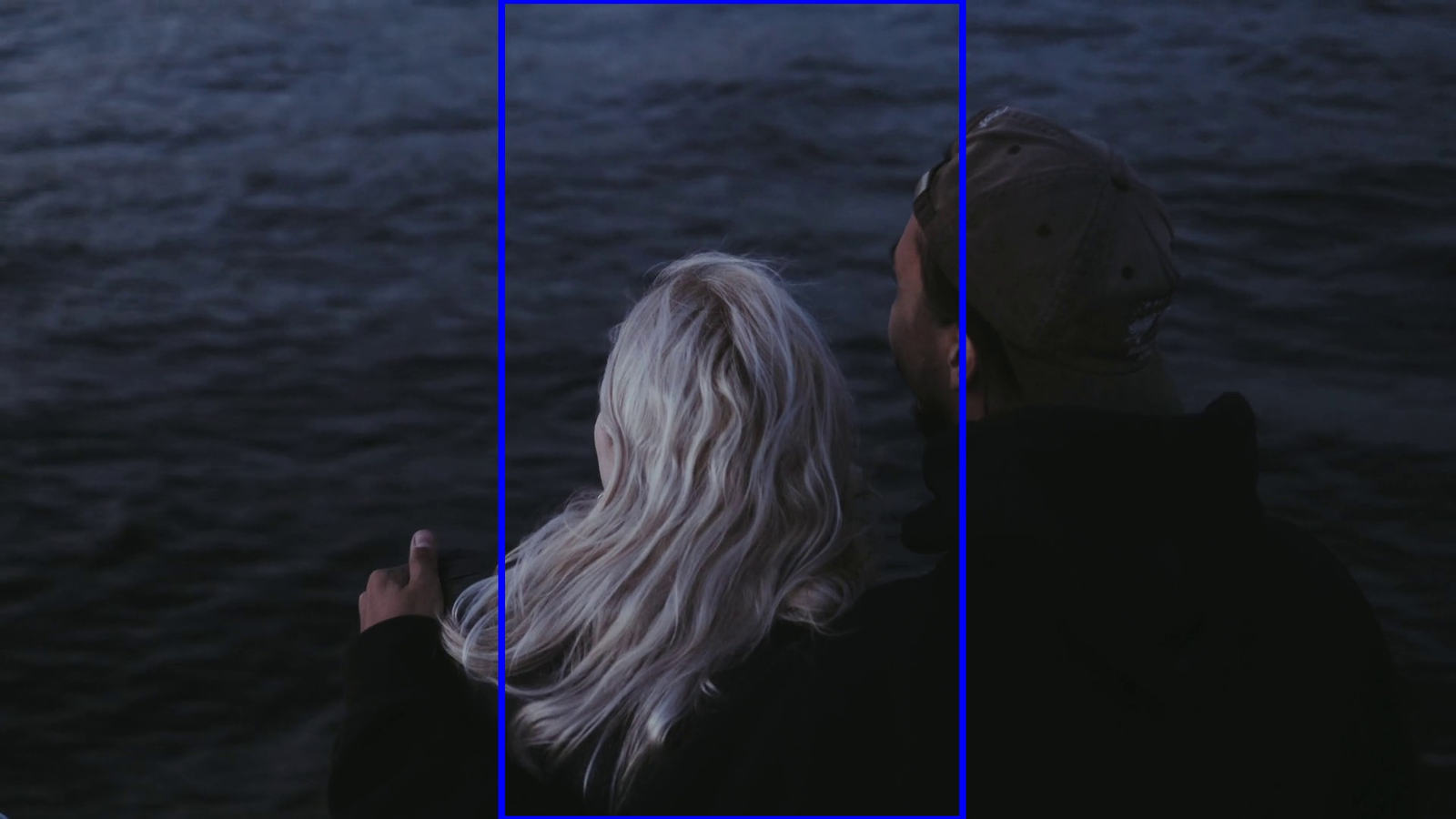} &
            \includegraphics[width=\fw]{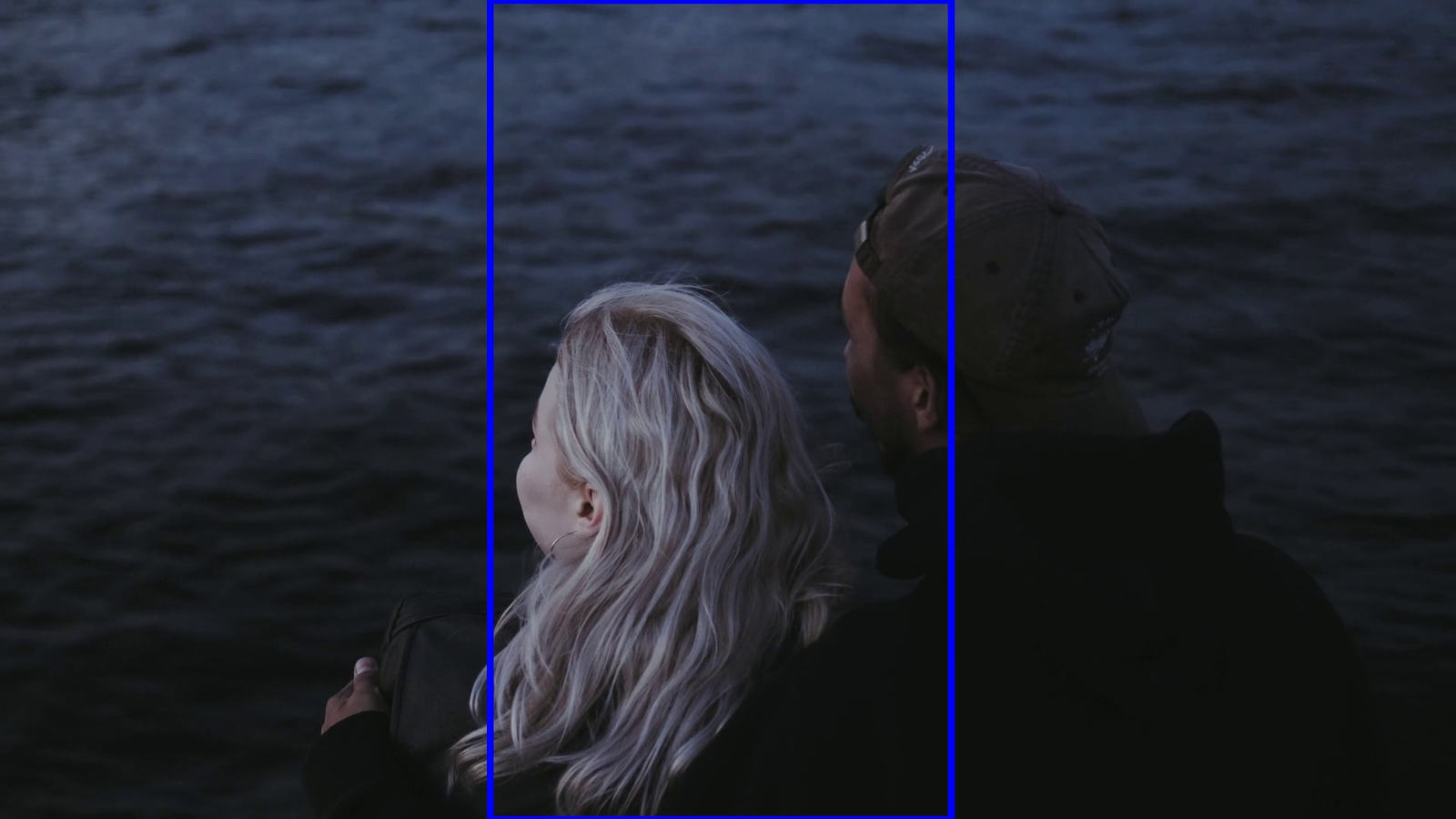} \\
            \rotatebox{90}{\footnotesize Ours} &
            \includegraphics[width=\fw]{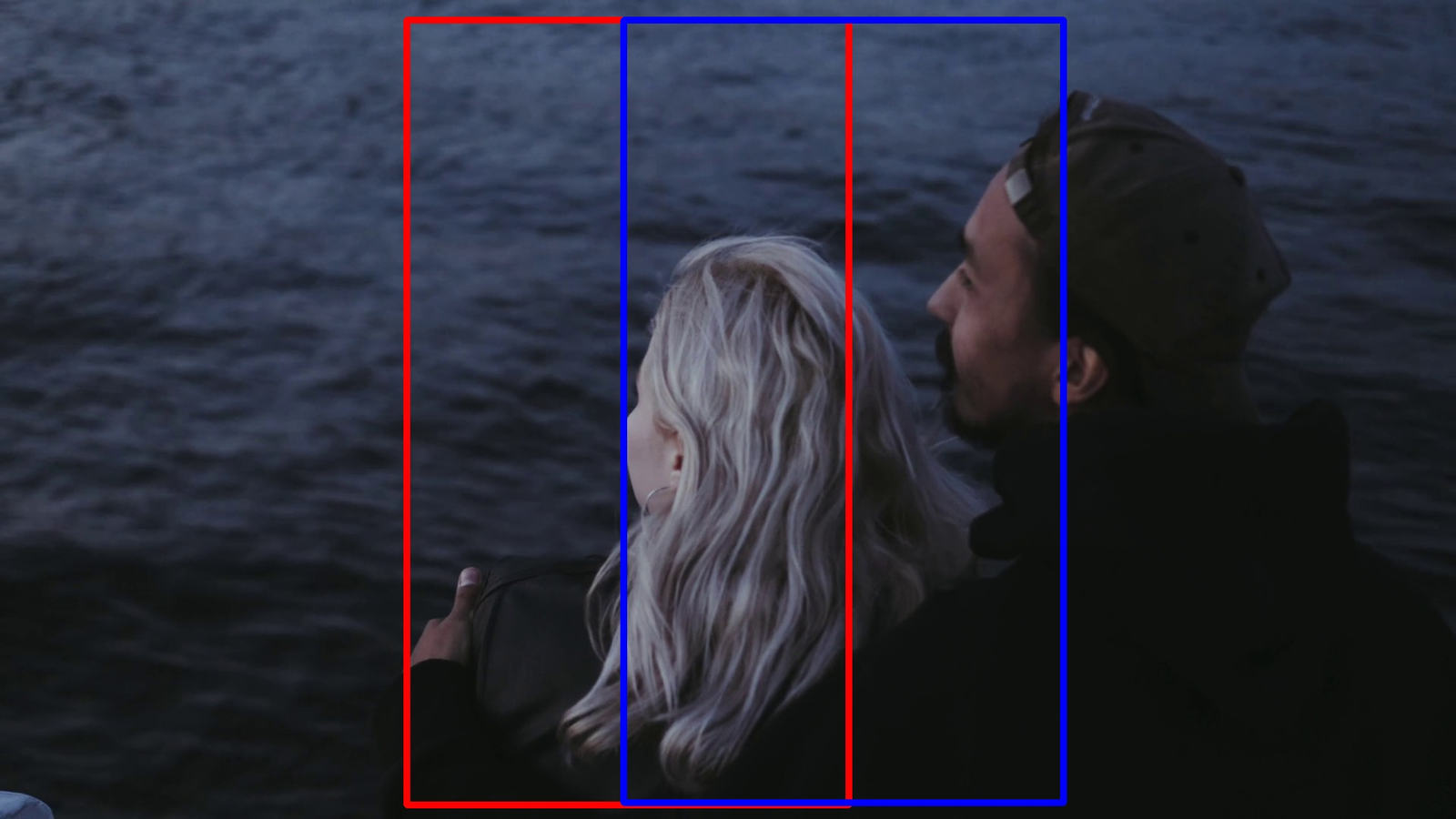} &
            \includegraphics[width=\fw]{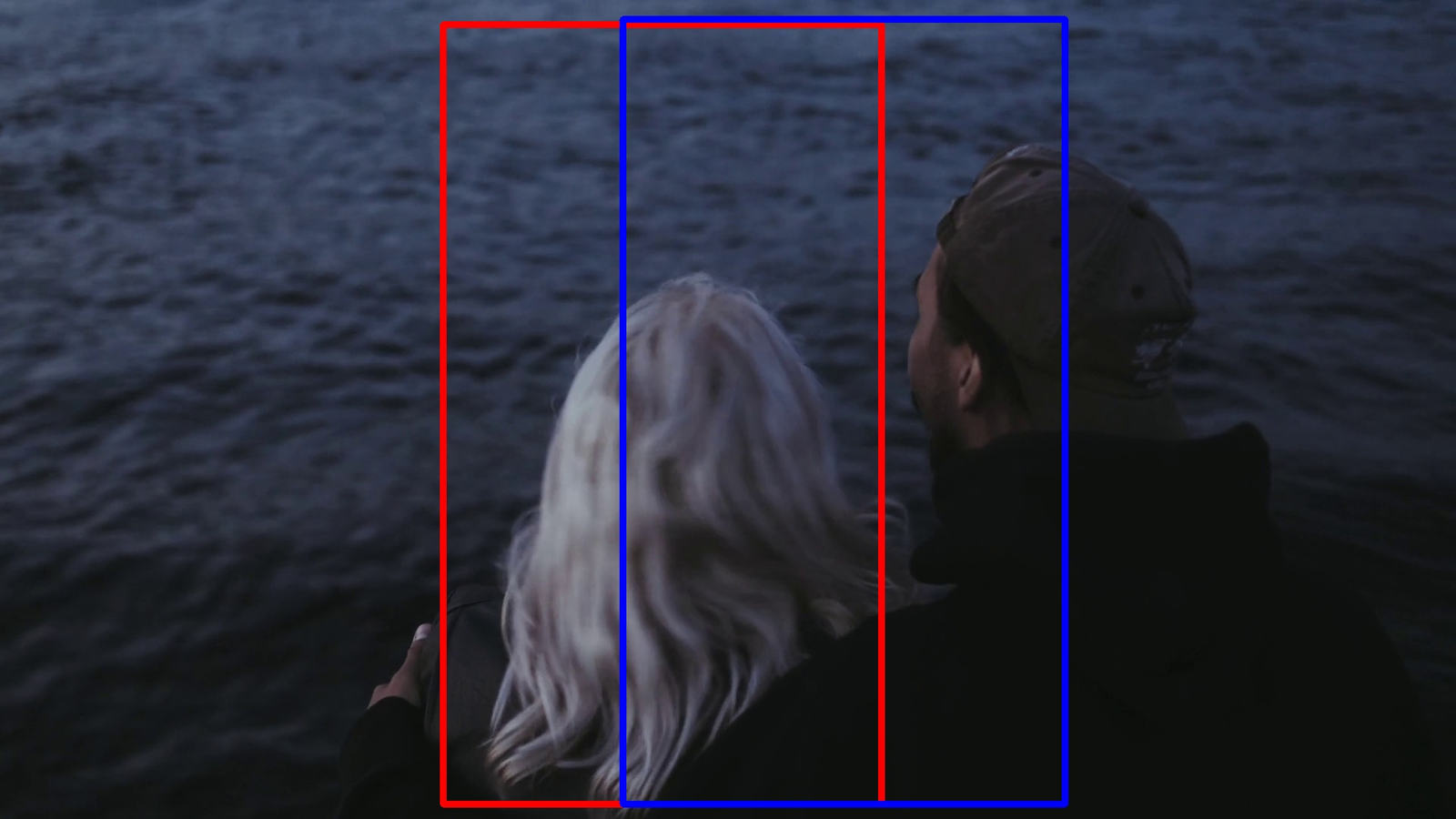} &
            \includegraphics[width=\fw]{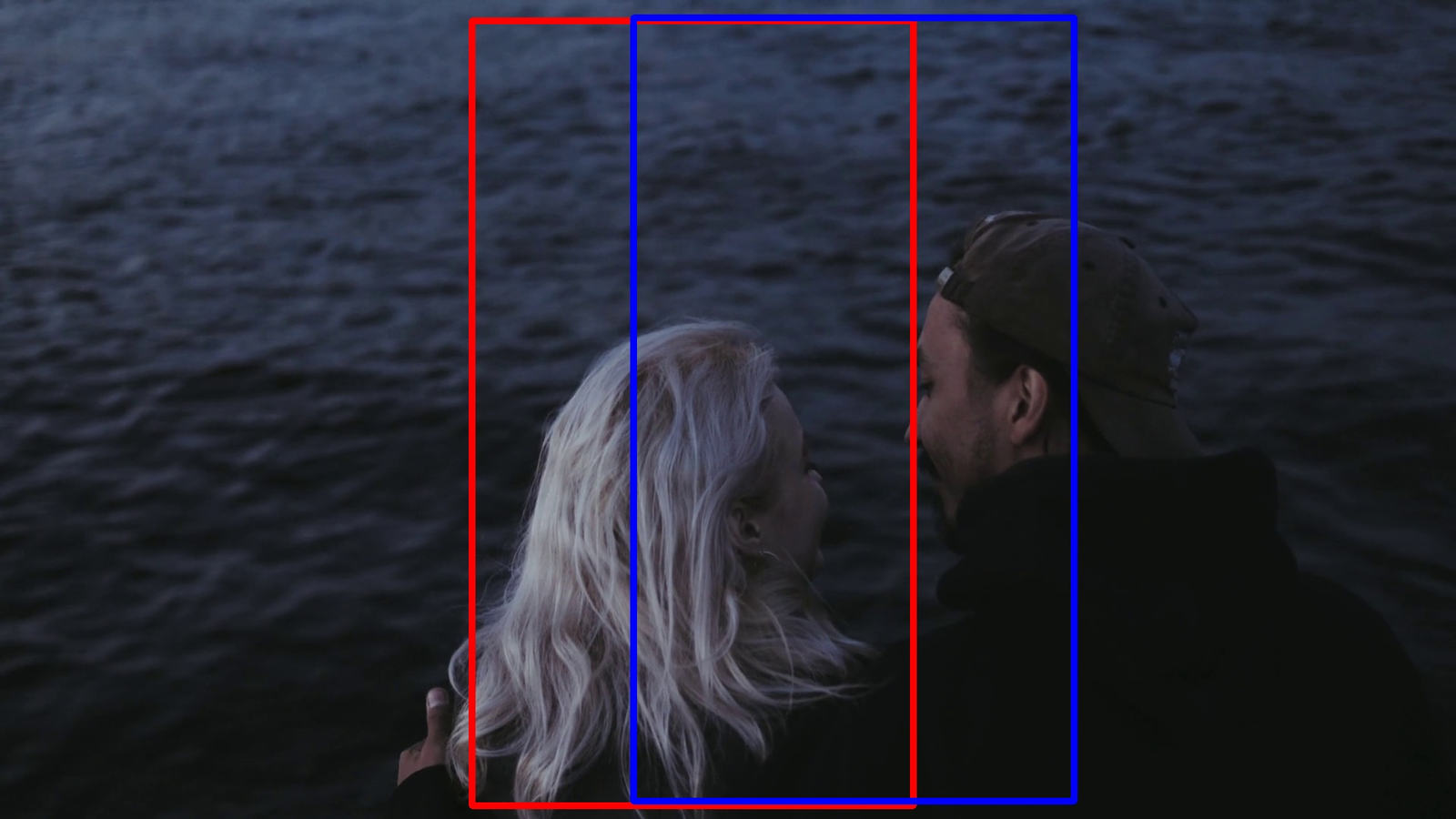} &
            \includegraphics[width=\fw]{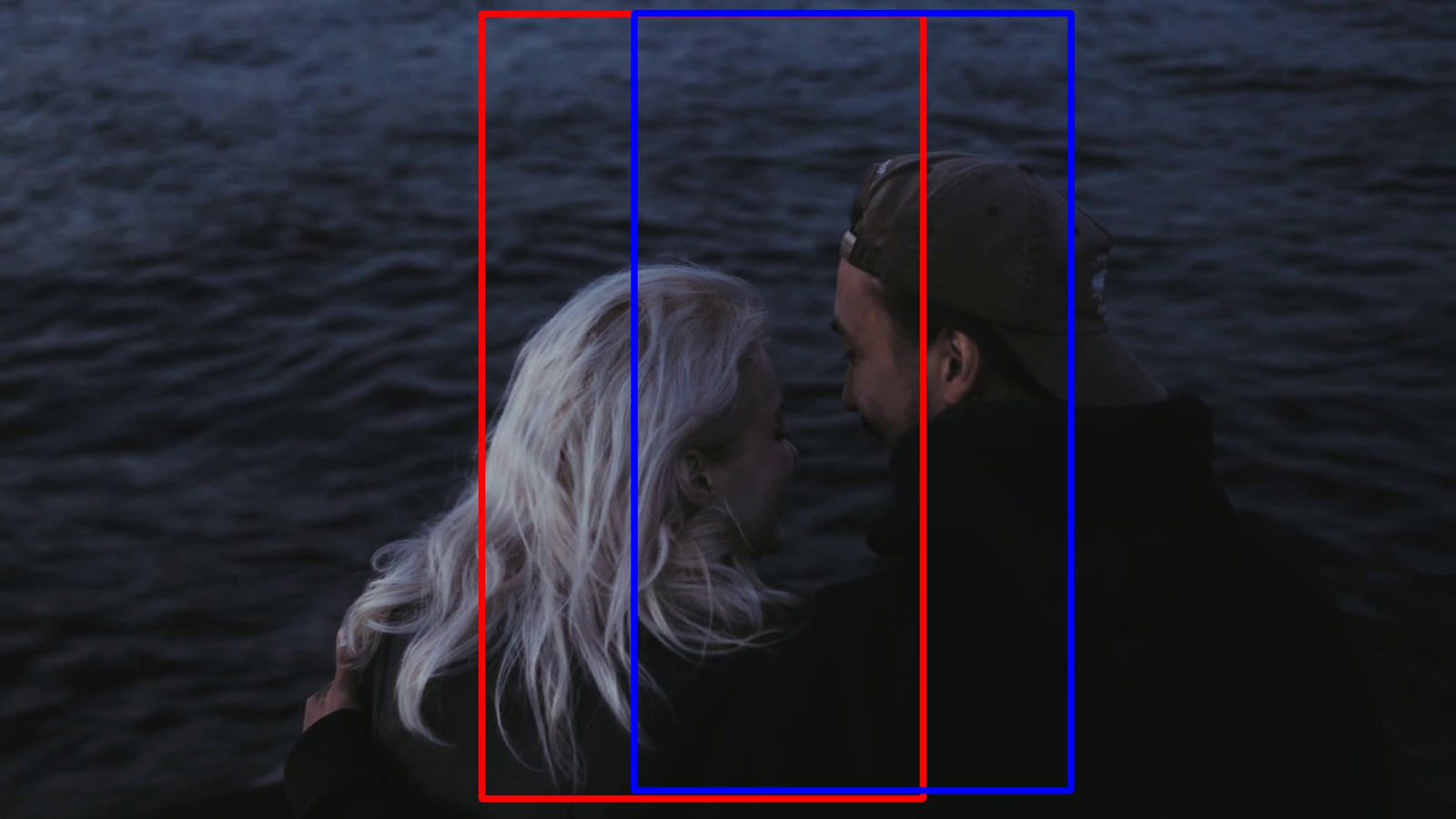} &
            \includegraphics[width=\fw]{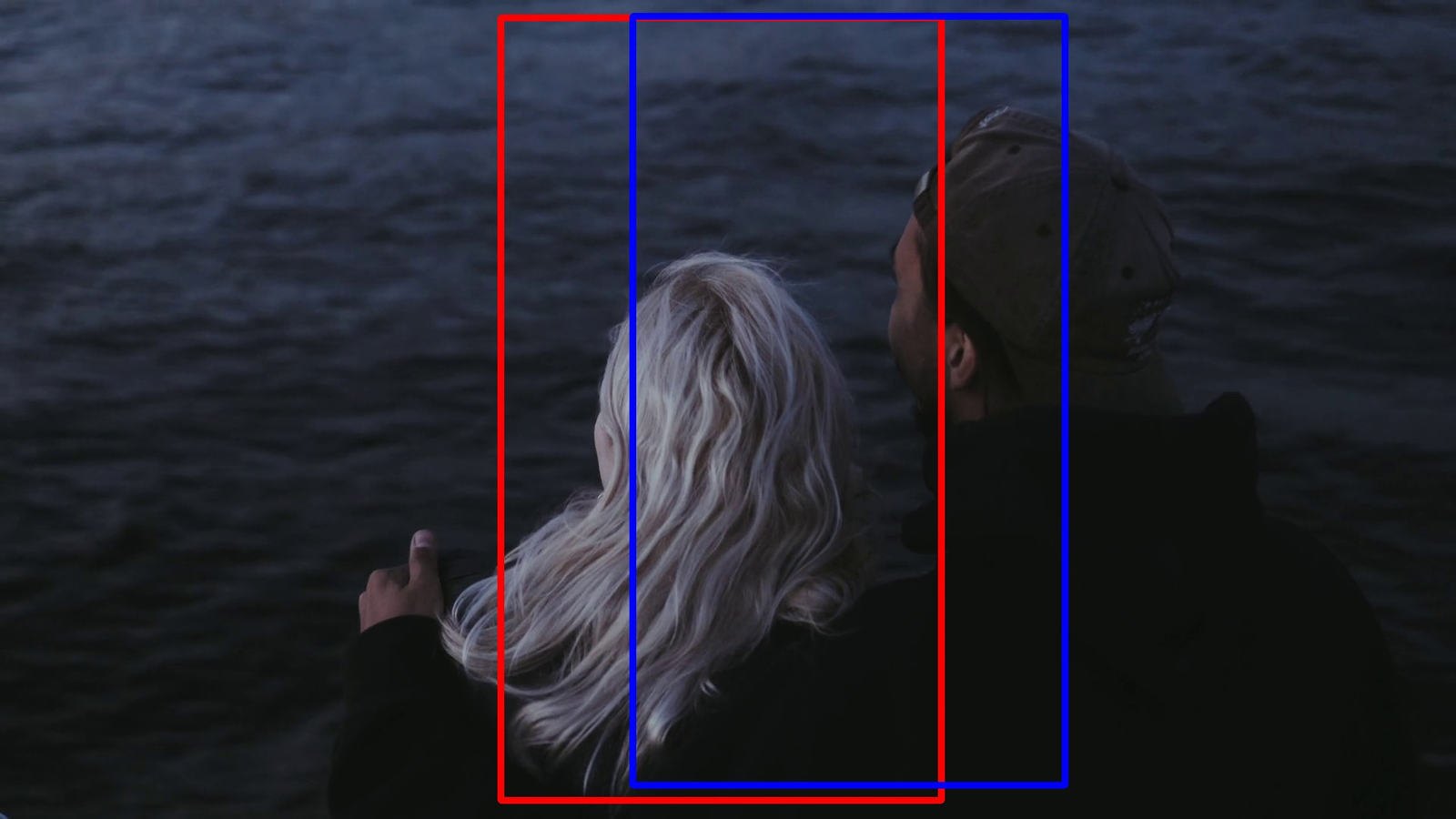} &
            \includegraphics[width=\fw]{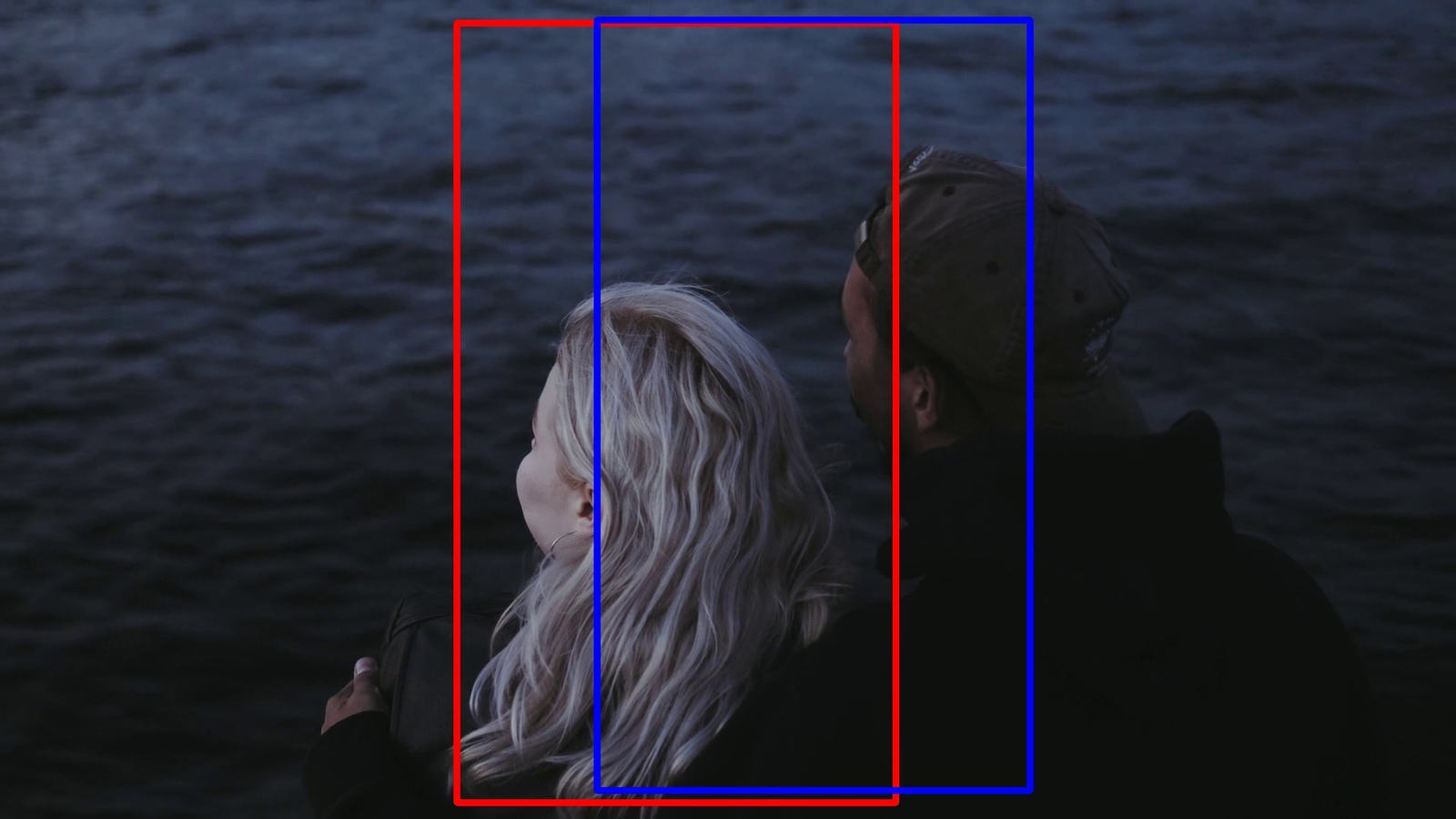} \\
            \rotatebox{90}{\footnotesize Ours + MLLM-based Agent} &
            \includegraphics[width=\fw]{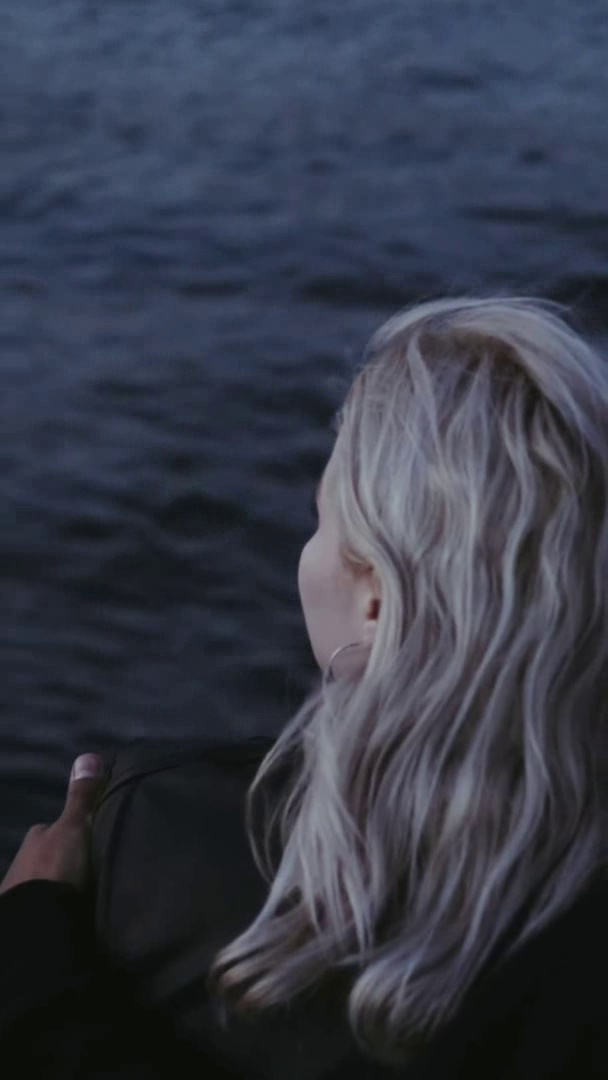} &
            \includegraphics[width=\fw]{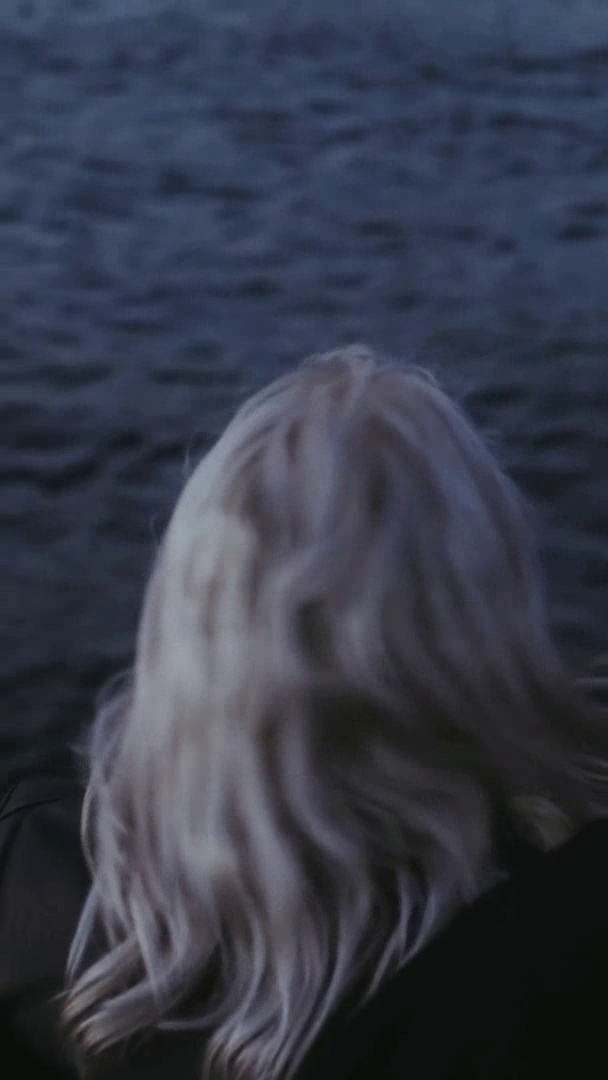} &
            \includegraphics[width=\fw]{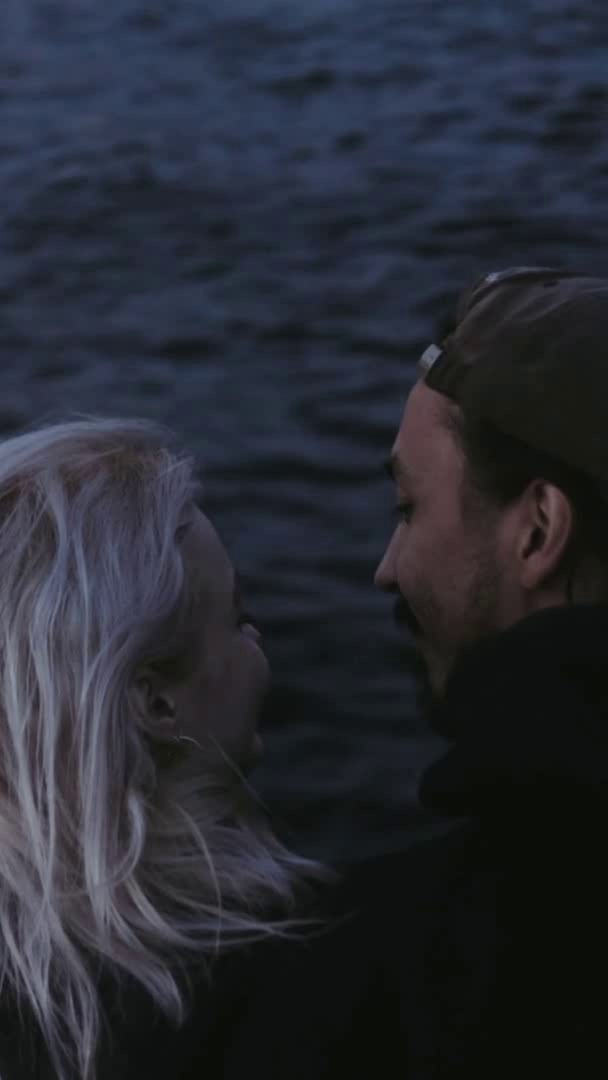} &
            \includegraphics[width=\fw]{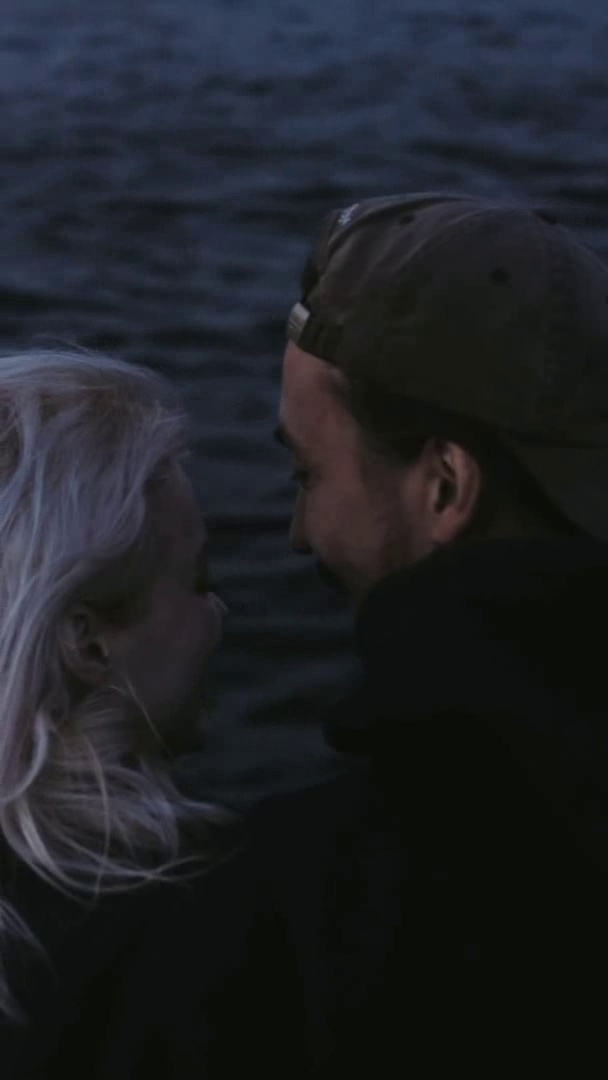} &
            \includegraphics[width=\fw]{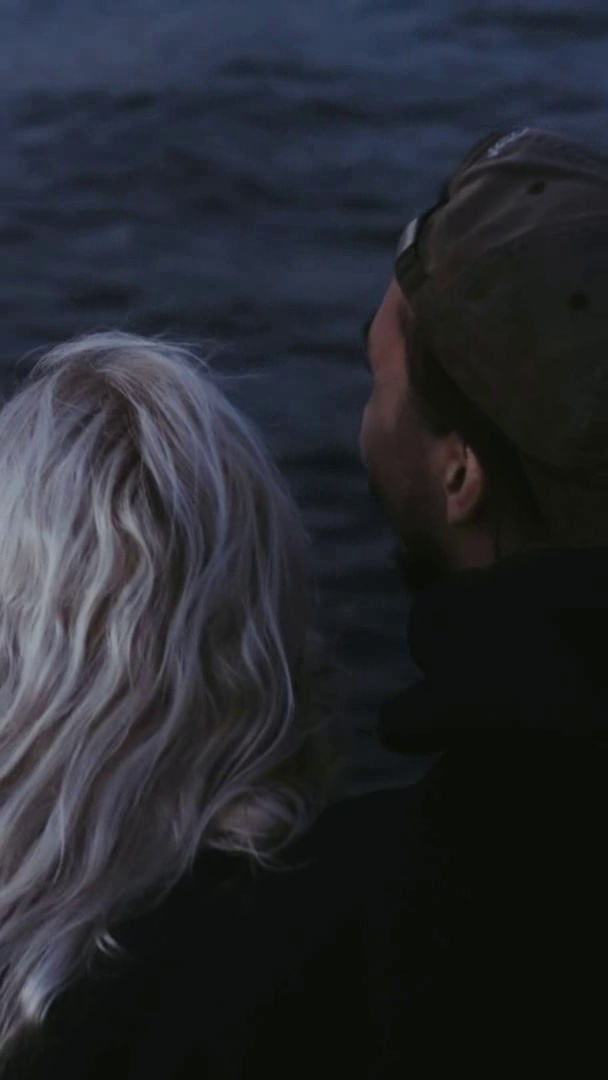} &
            \includegraphics[width=\fw]{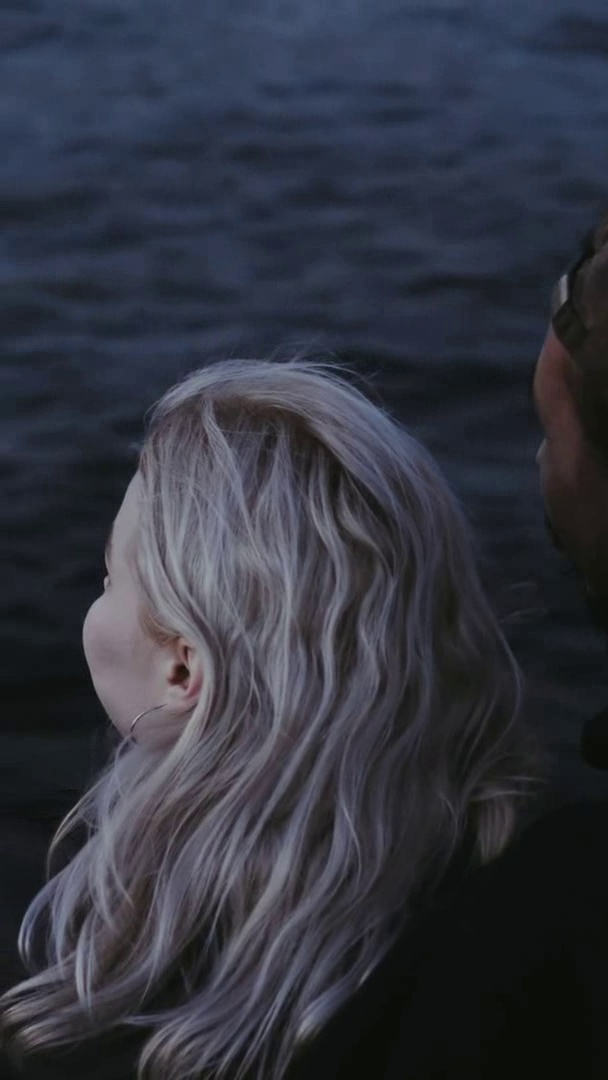} \\
        \end{tabular}
        \caption{Video ID: 4047424}
    \end{subfigure}
    
    \addtocounter{figure}{-1} 
    \vspace{-10pt}
    \caption{\textbf{Qualitative comparison of ROI extraction methods on the MiraData dataset \cite{ju2024miradata} (Part 2 of 3).} Subfigures (\textit{c}--\textit{d}) demonstrate further examples including our \textbf{Final Crop} outputs on the bottom row.}
    \label{fig:main_qualitative_comparison_part2}
\end{figure*}

\begin{figure*}[t]
    \centering
    \def\fw{0.15\textwidth} 
    \setlength{\tabcolsep}{1pt} 

    \begin{subfigure}{\textwidth}
        \centering
        \begin{tabular}{ccccccc}
             & \textbf{3s} & \textbf{6s} & \textbf{9s} & \textbf{13s} & \textbf{16s} & \textbf{19s} \\
            \rotatebox{90}{\footnotesize Center Crop} &
            \includegraphics[width=\fw]{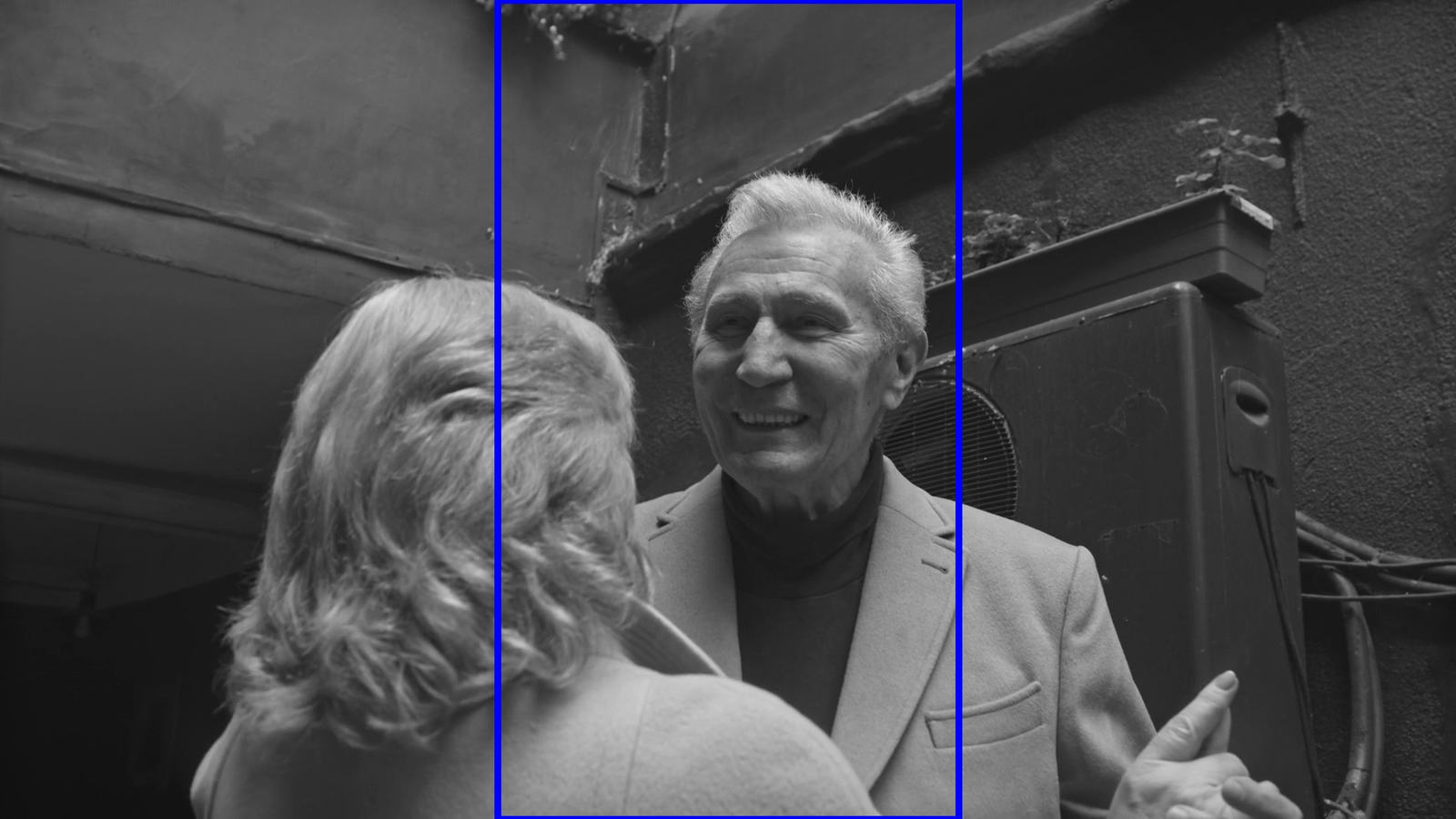} &
            \includegraphics[width=\fw]{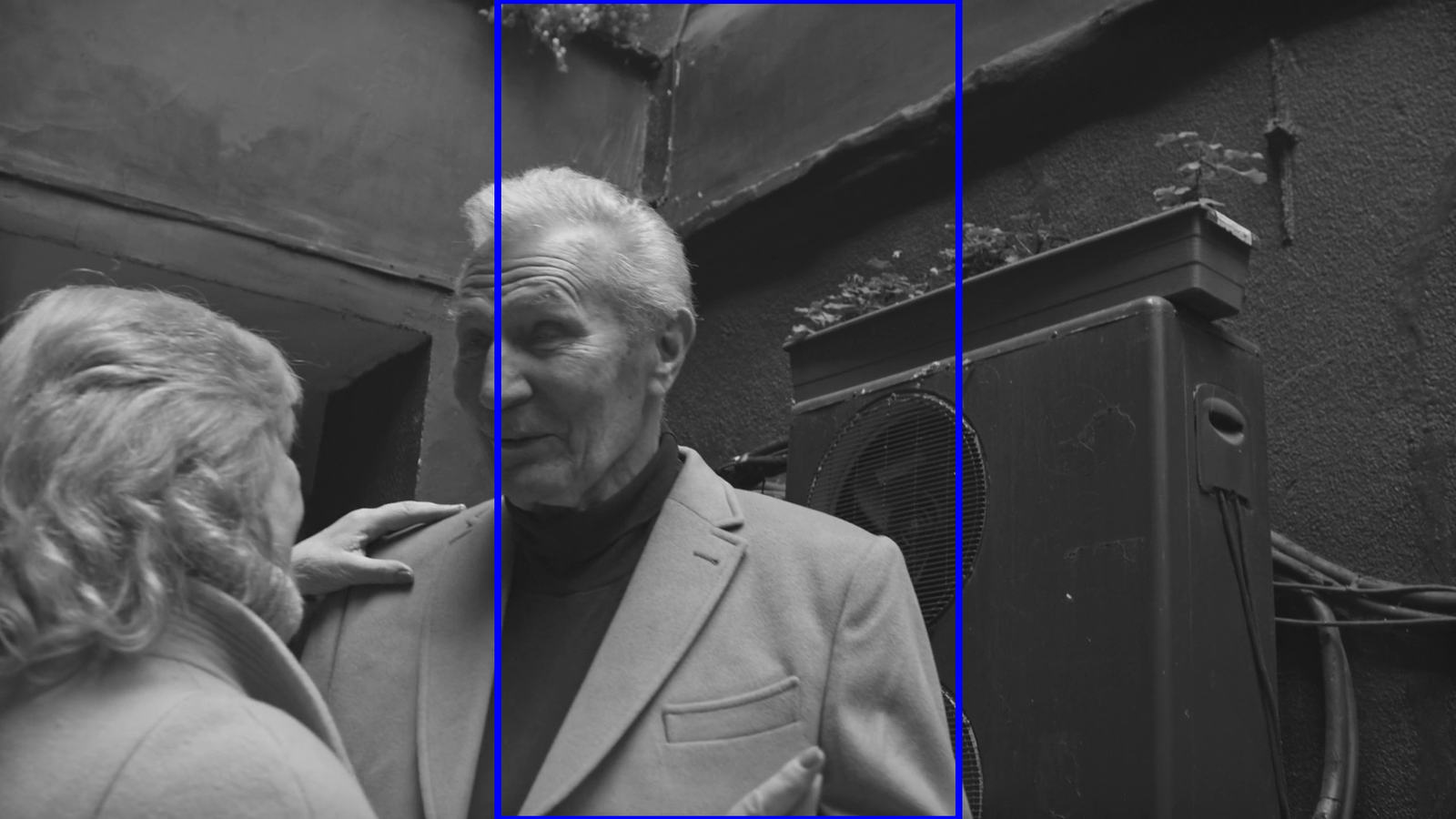} &
            \includegraphics[width=\fw]{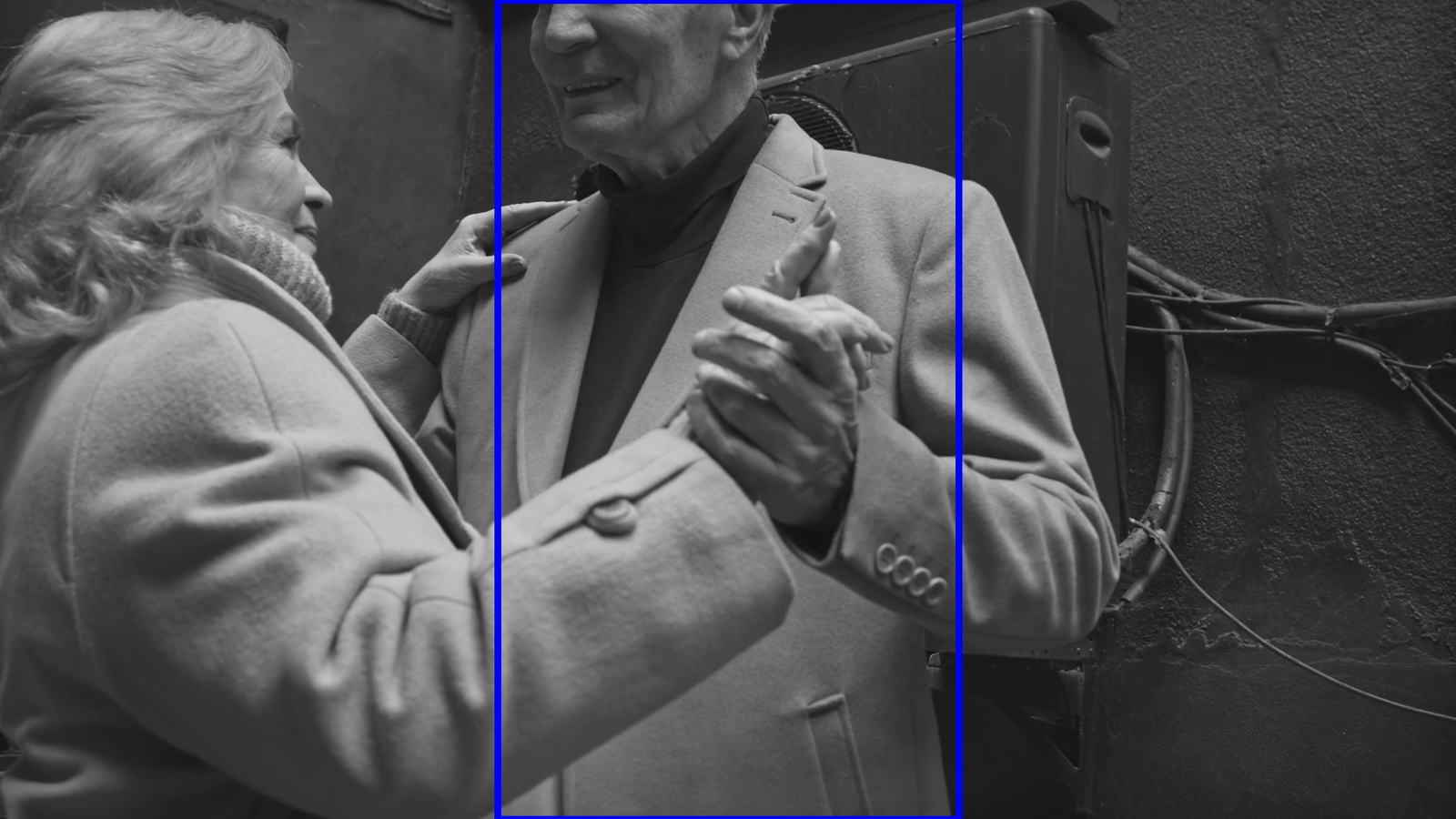} &
            \includegraphics[width=\fw]{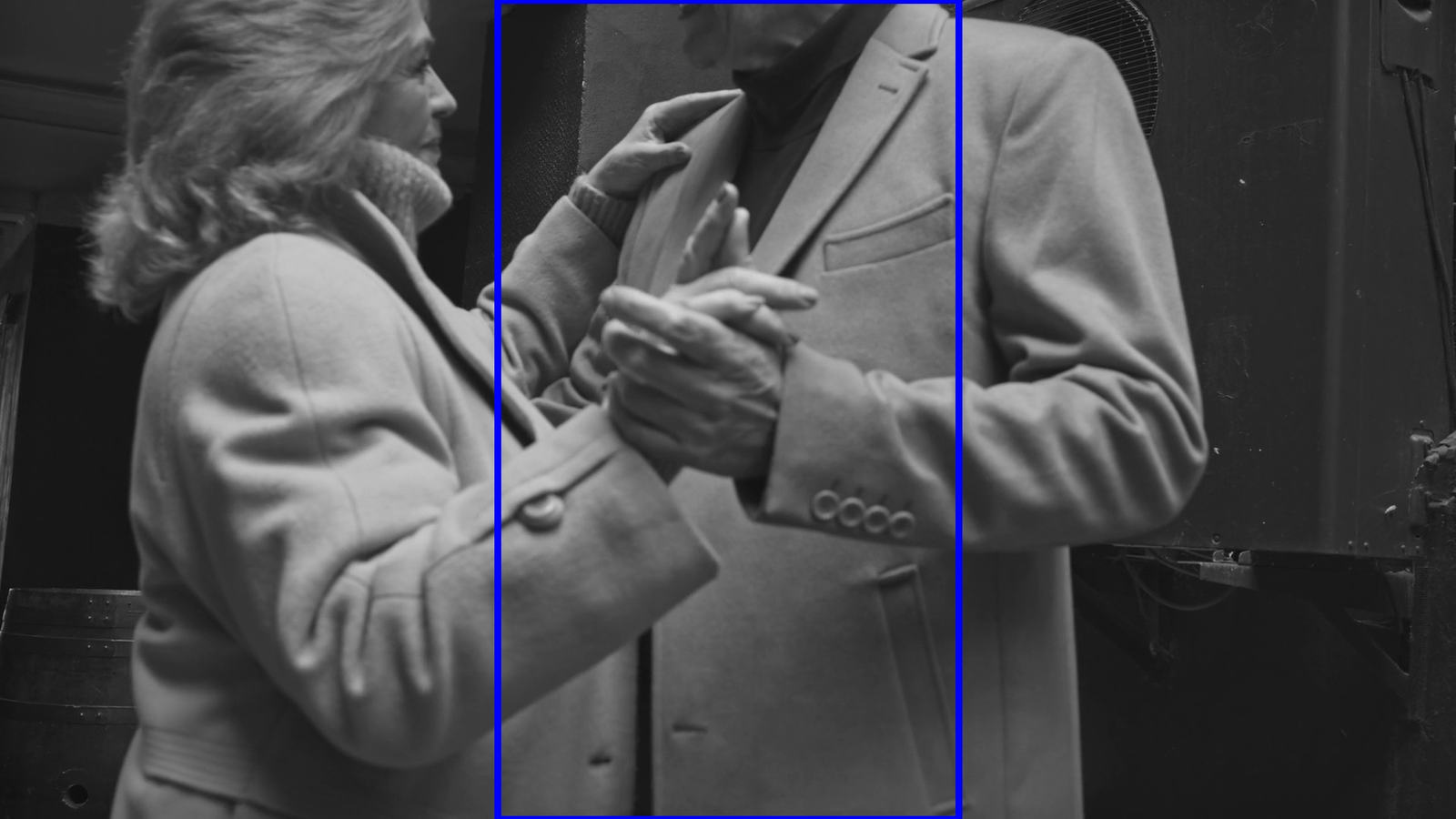} &
            \includegraphics[width=\fw]{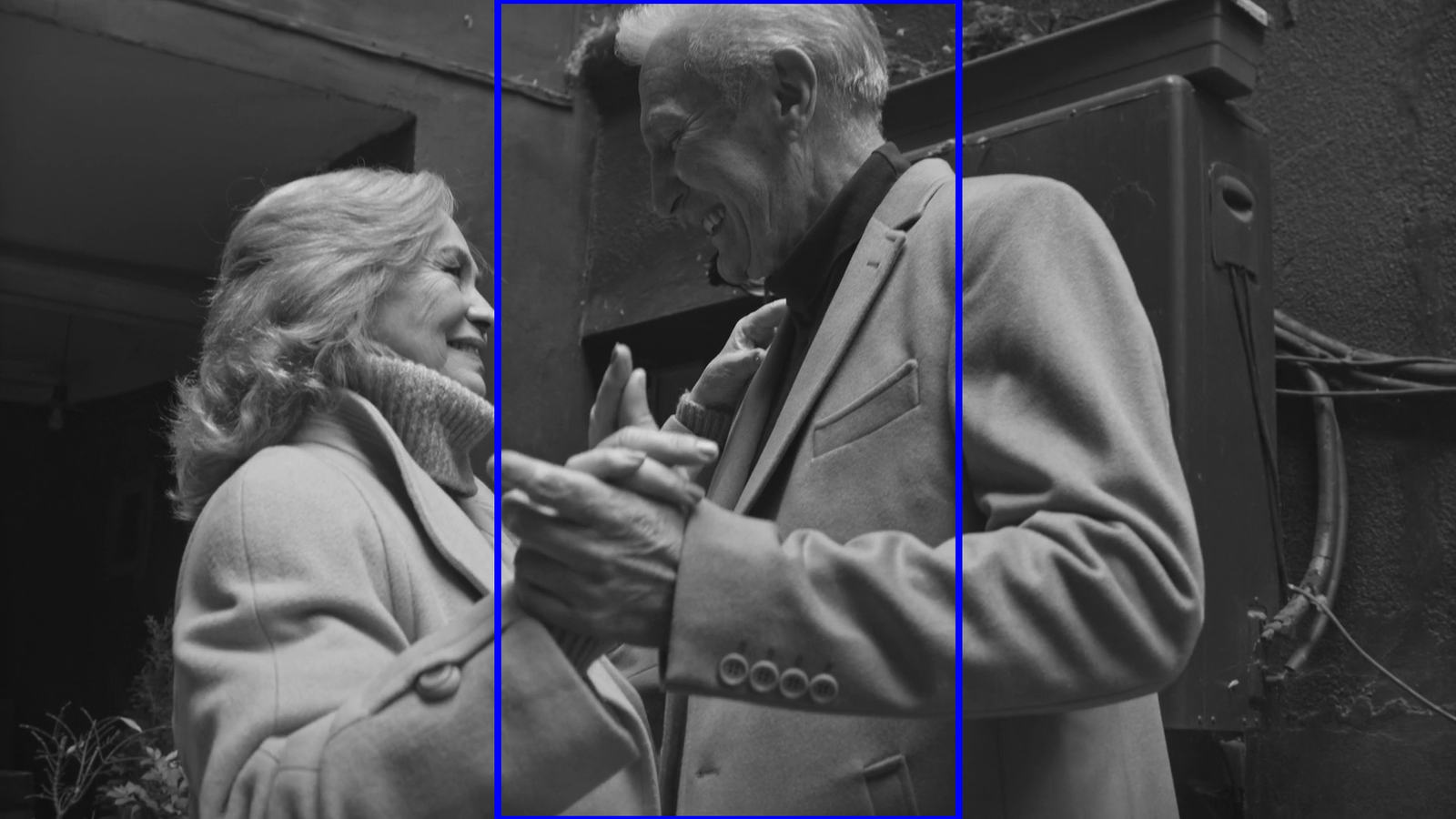} &
            \includegraphics[width=\fw]{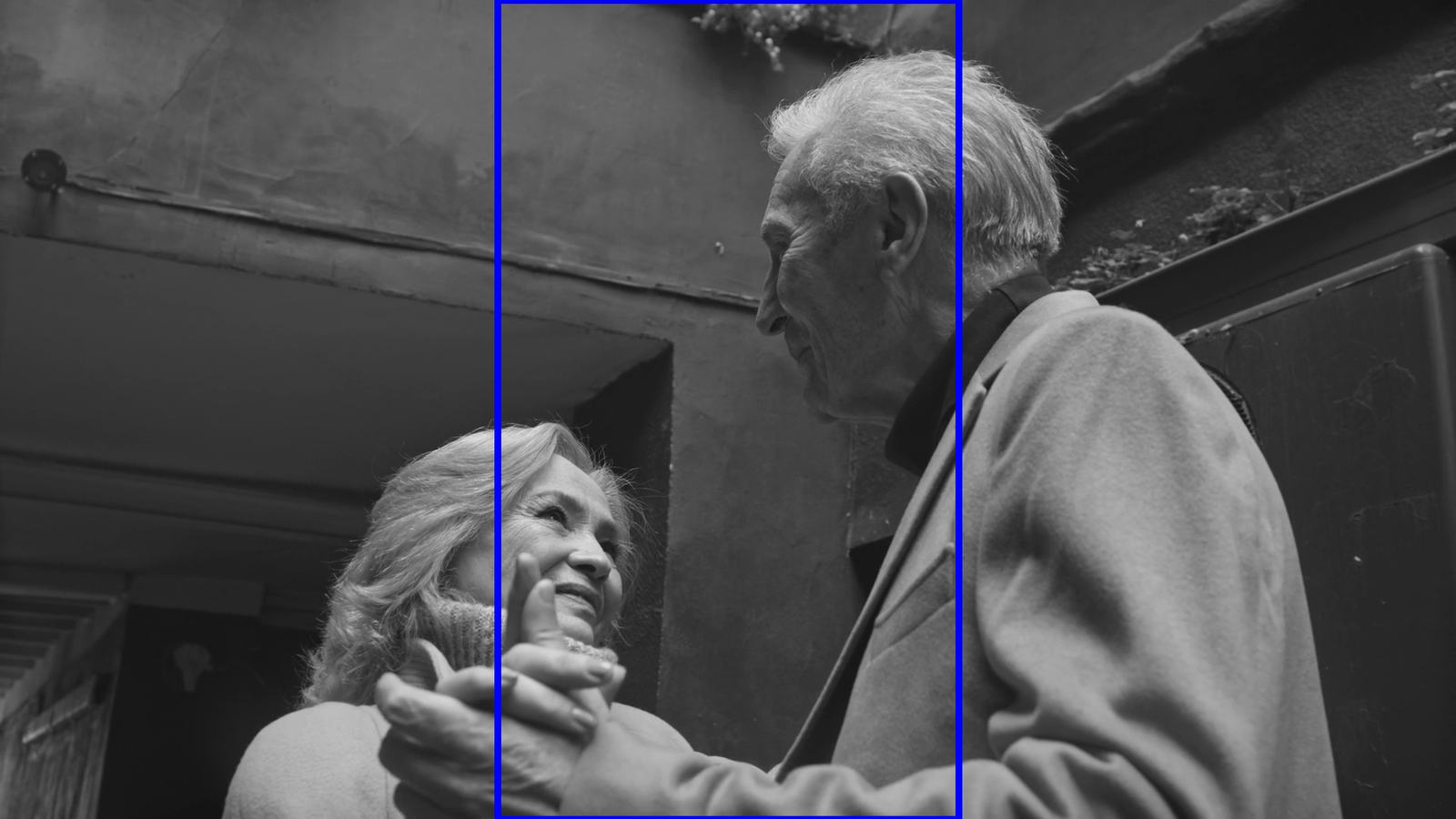} \\
            \rotatebox{90}{\footnotesize Gemini-Base} &
            \includegraphics[width=\fw]{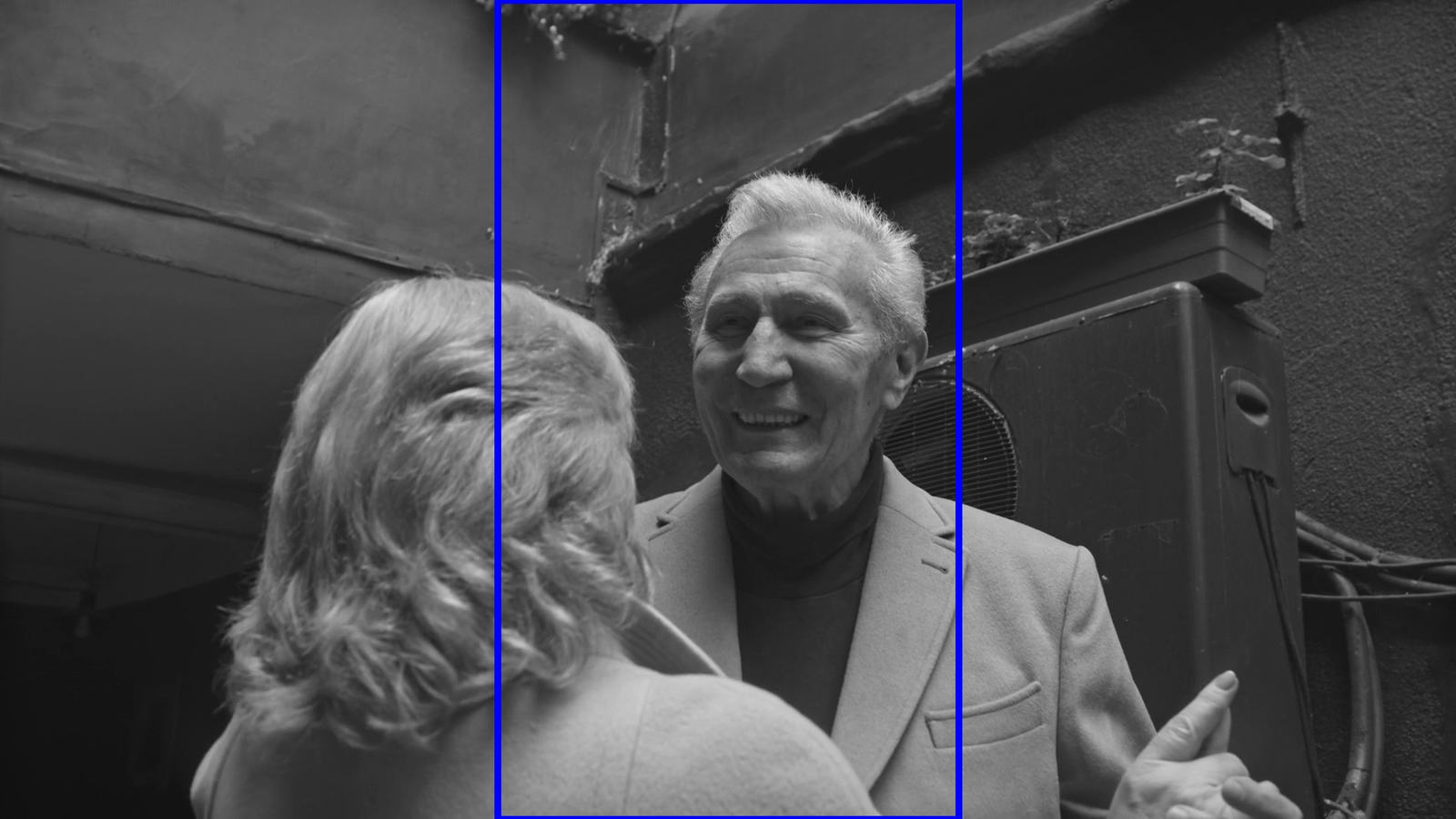} &
            \includegraphics[width=\fw]{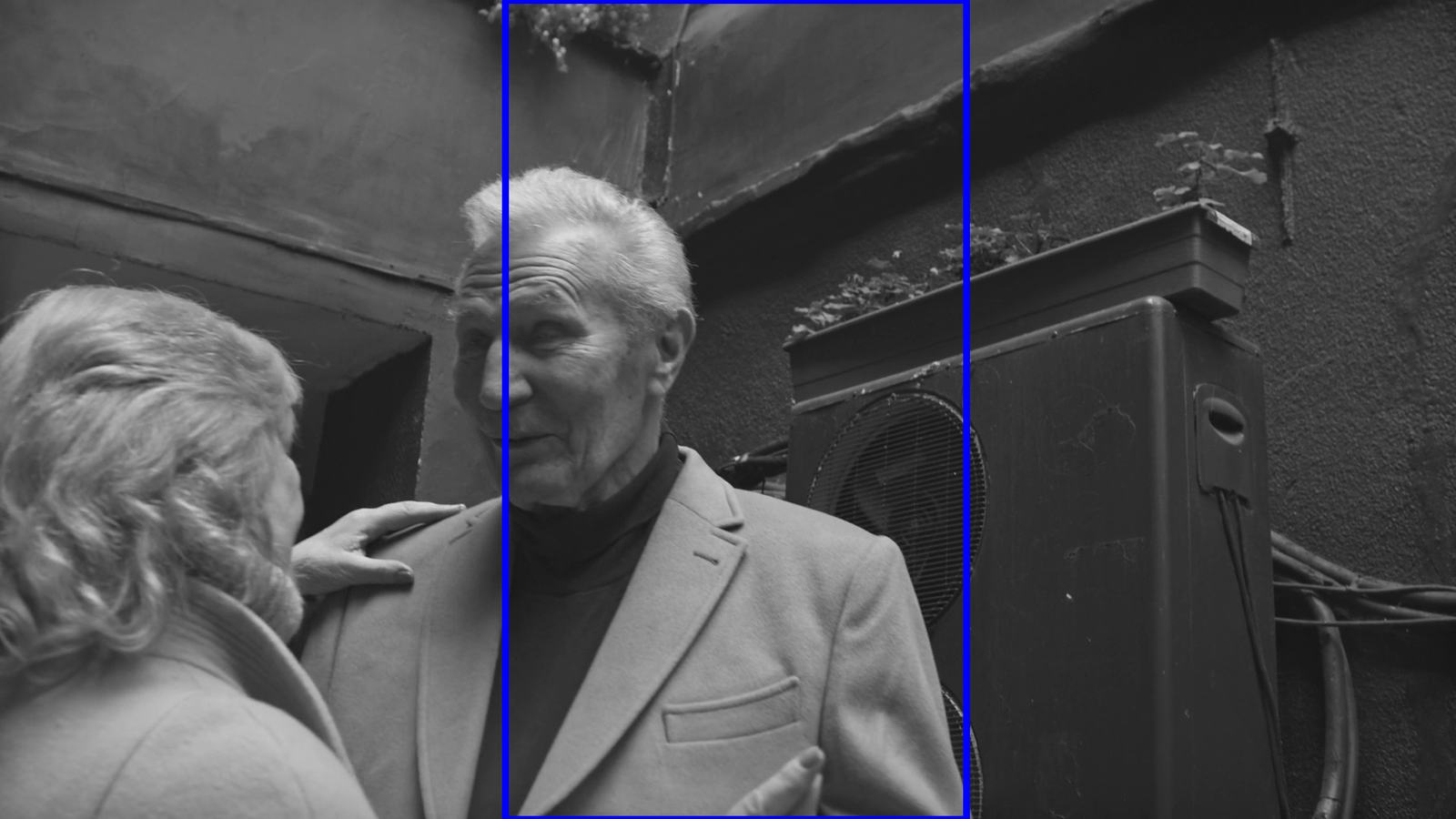} &
            \includegraphics[width=\fw]{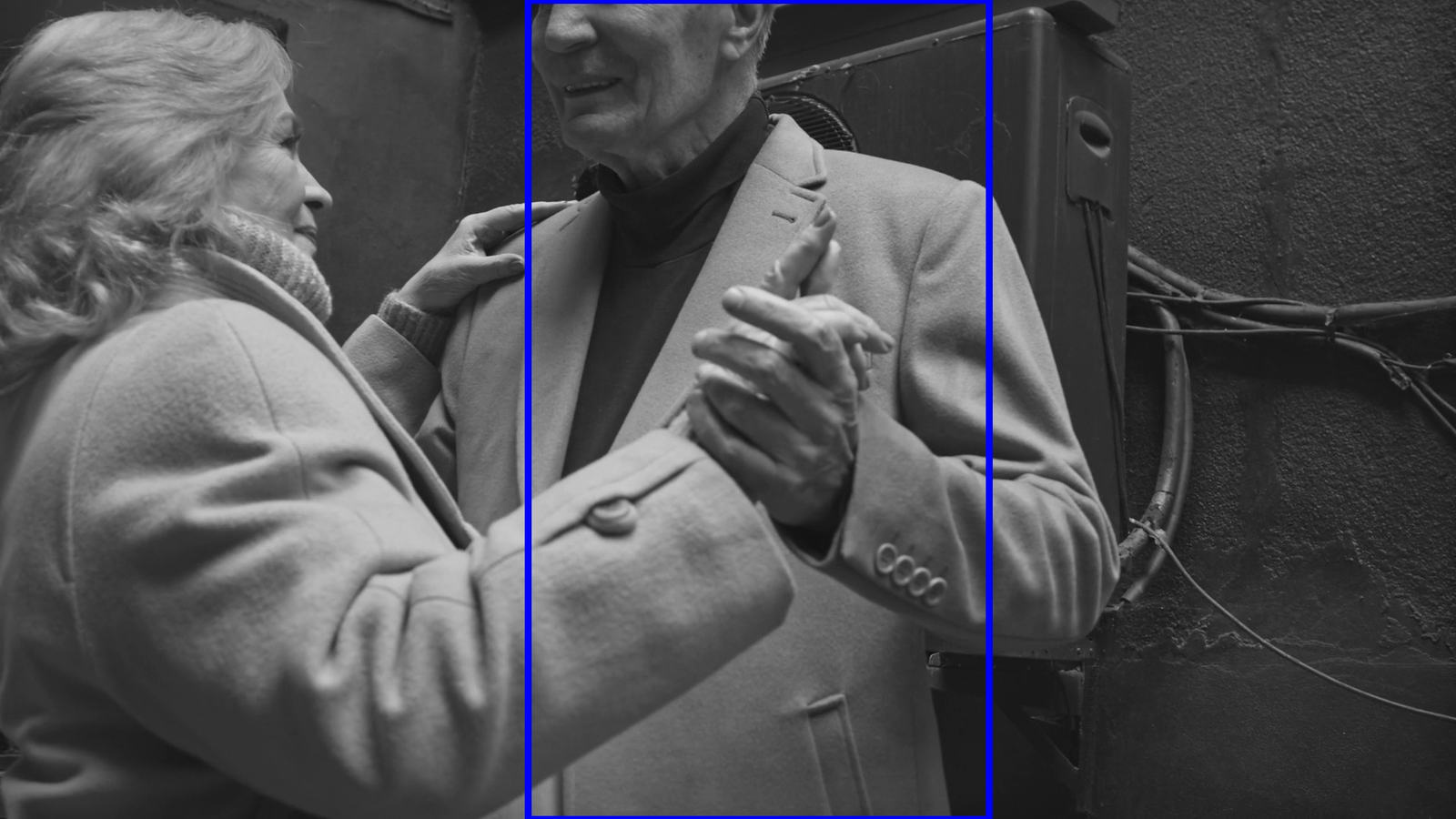} &
            \includegraphics[width=\fw]{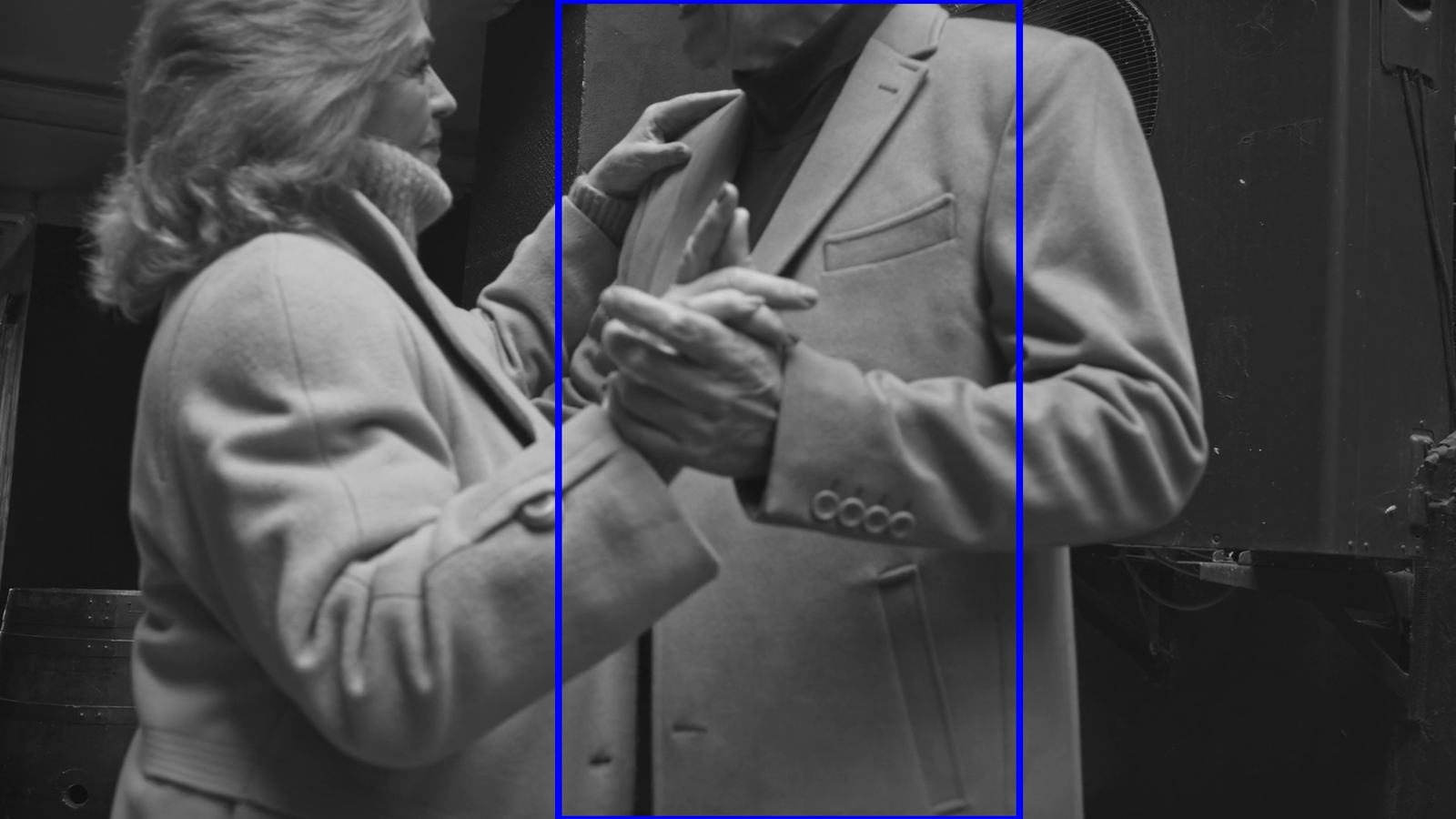} &
            \includegraphics[width=\fw]{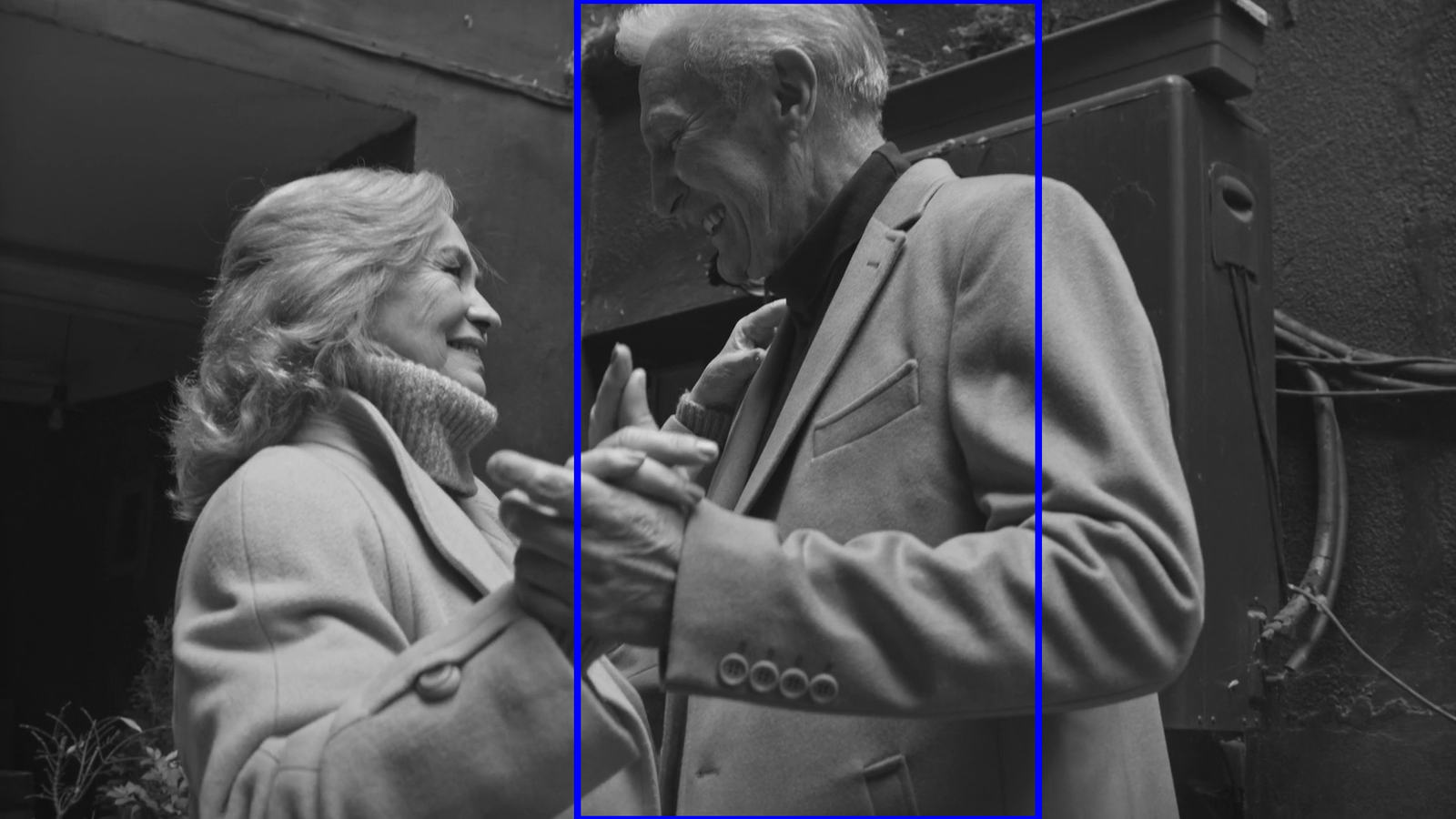} &
            \includegraphics[width=\fw]{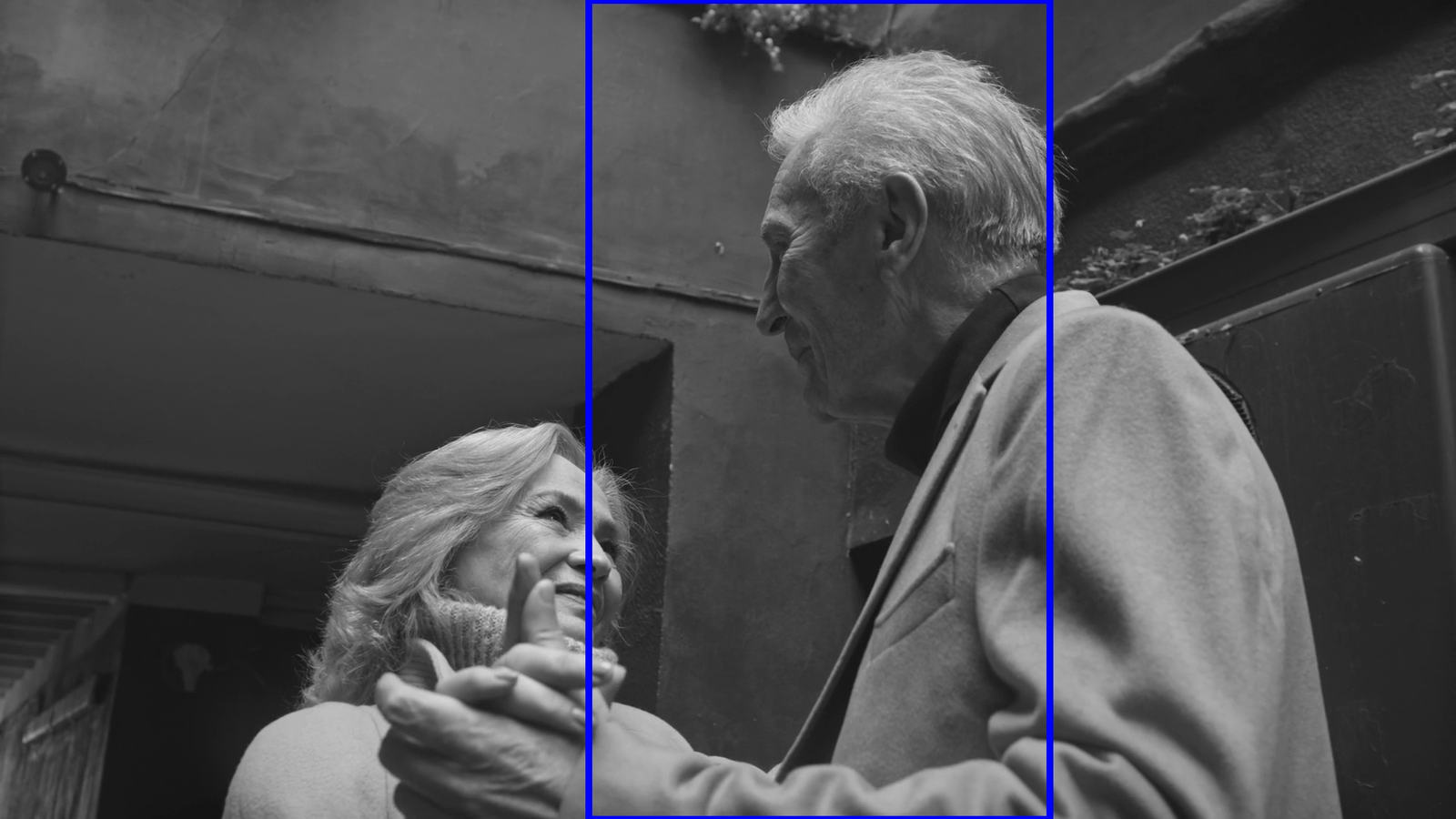} \\
            \rotatebox{90}{\footnotesize Ours} &
            \includegraphics[width=\fw]{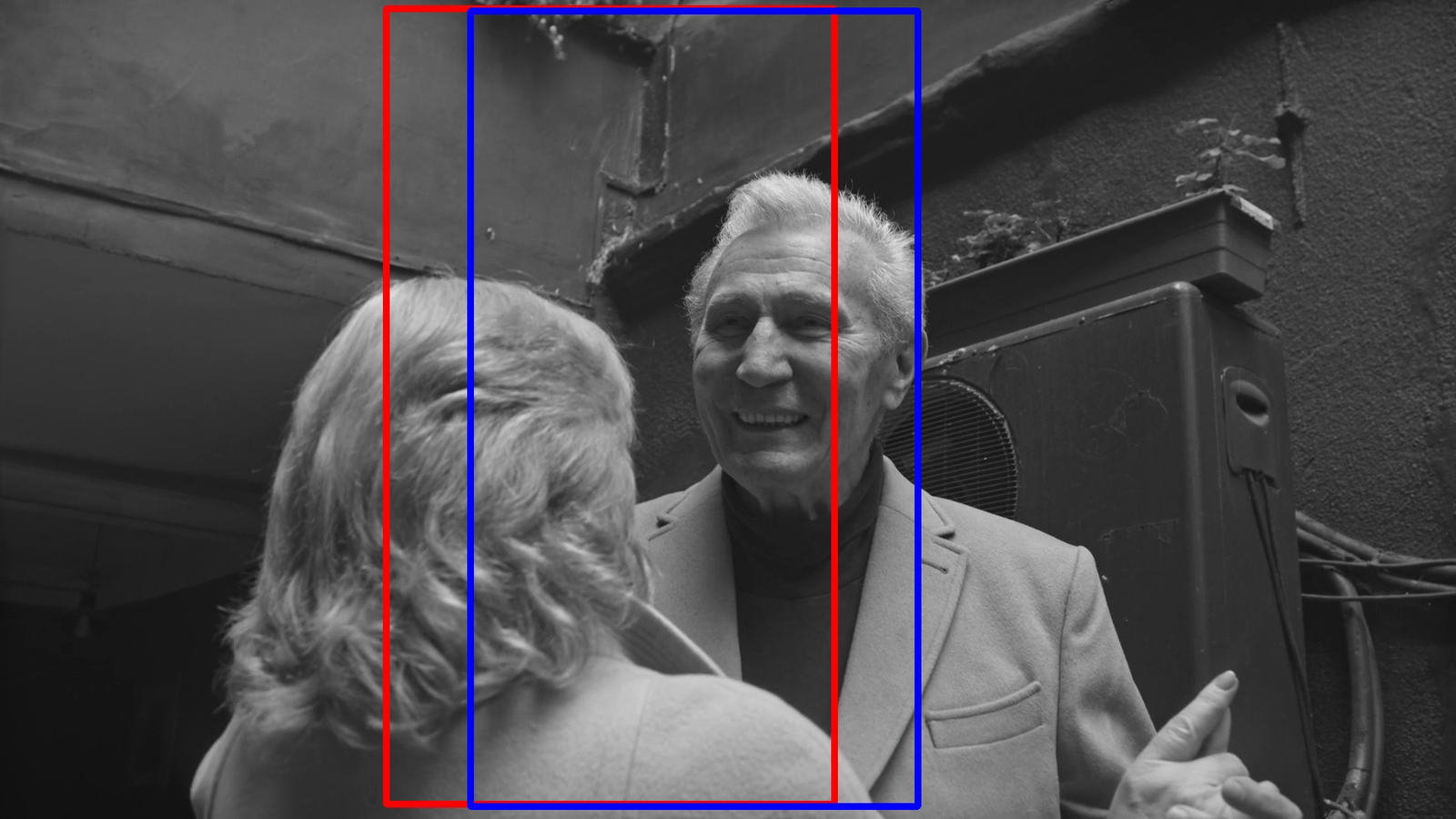} &
            \includegraphics[width=\fw]{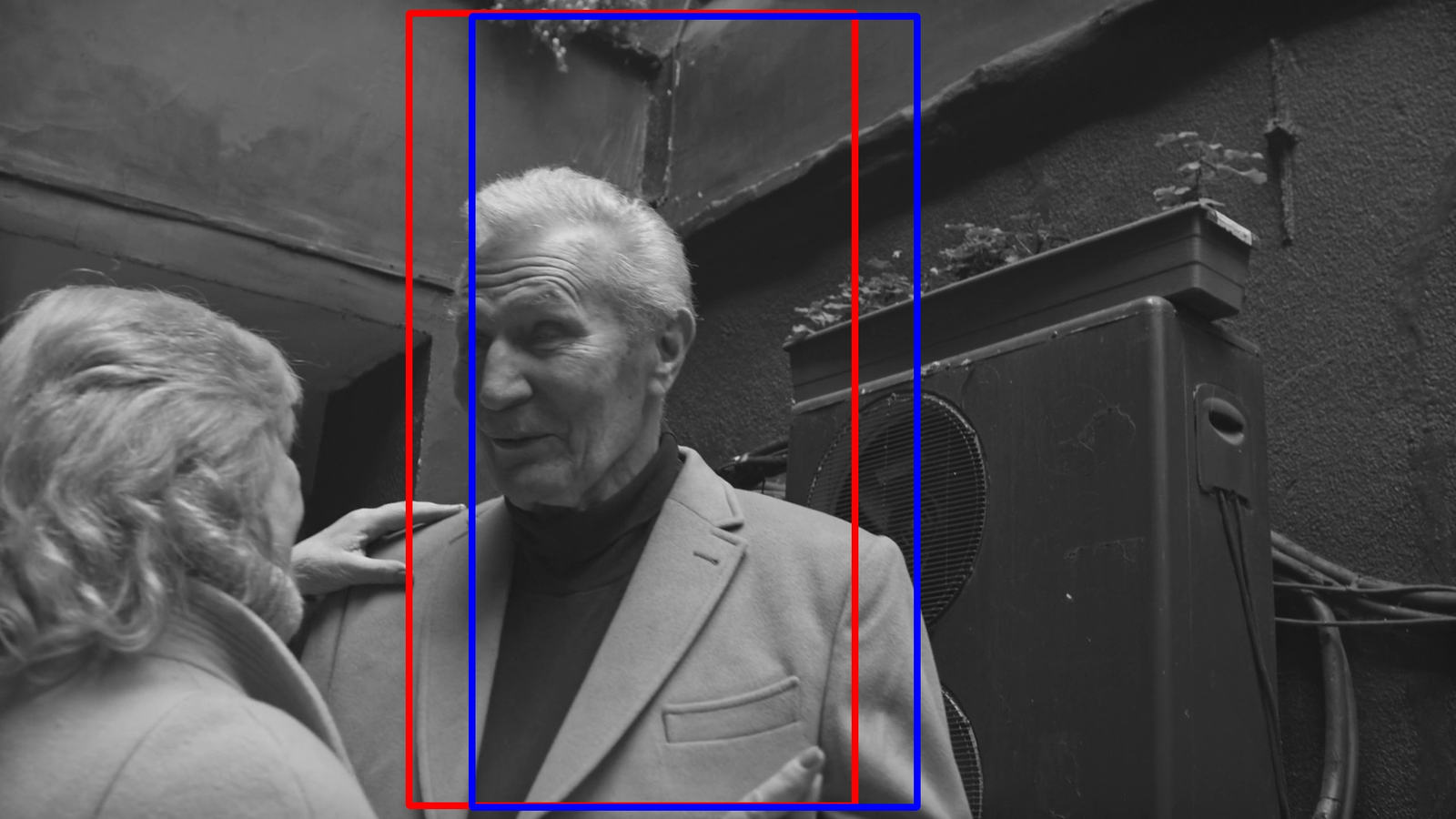} &
            \includegraphics[width=\fw]{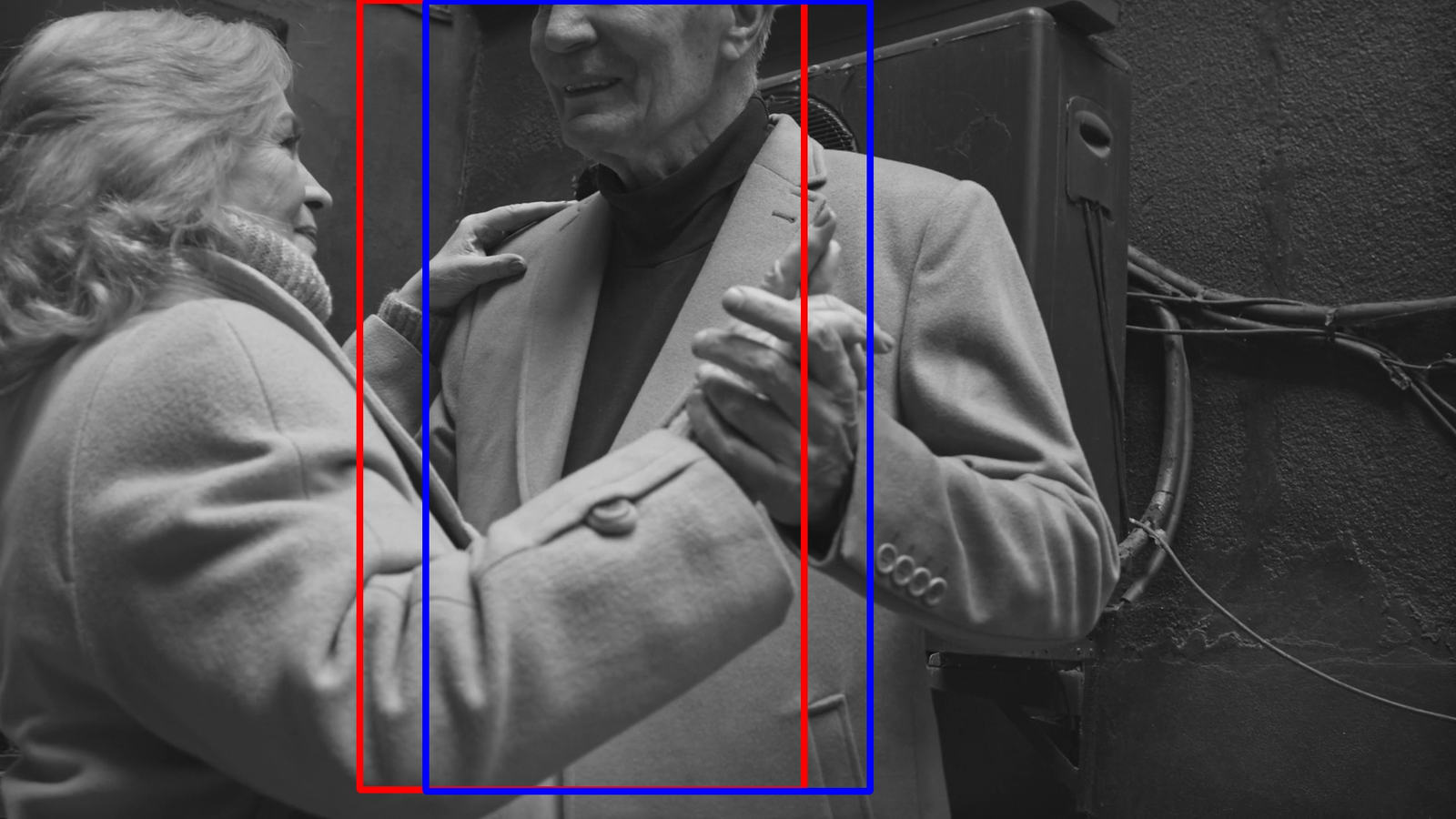} &
            \includegraphics[width=\fw]{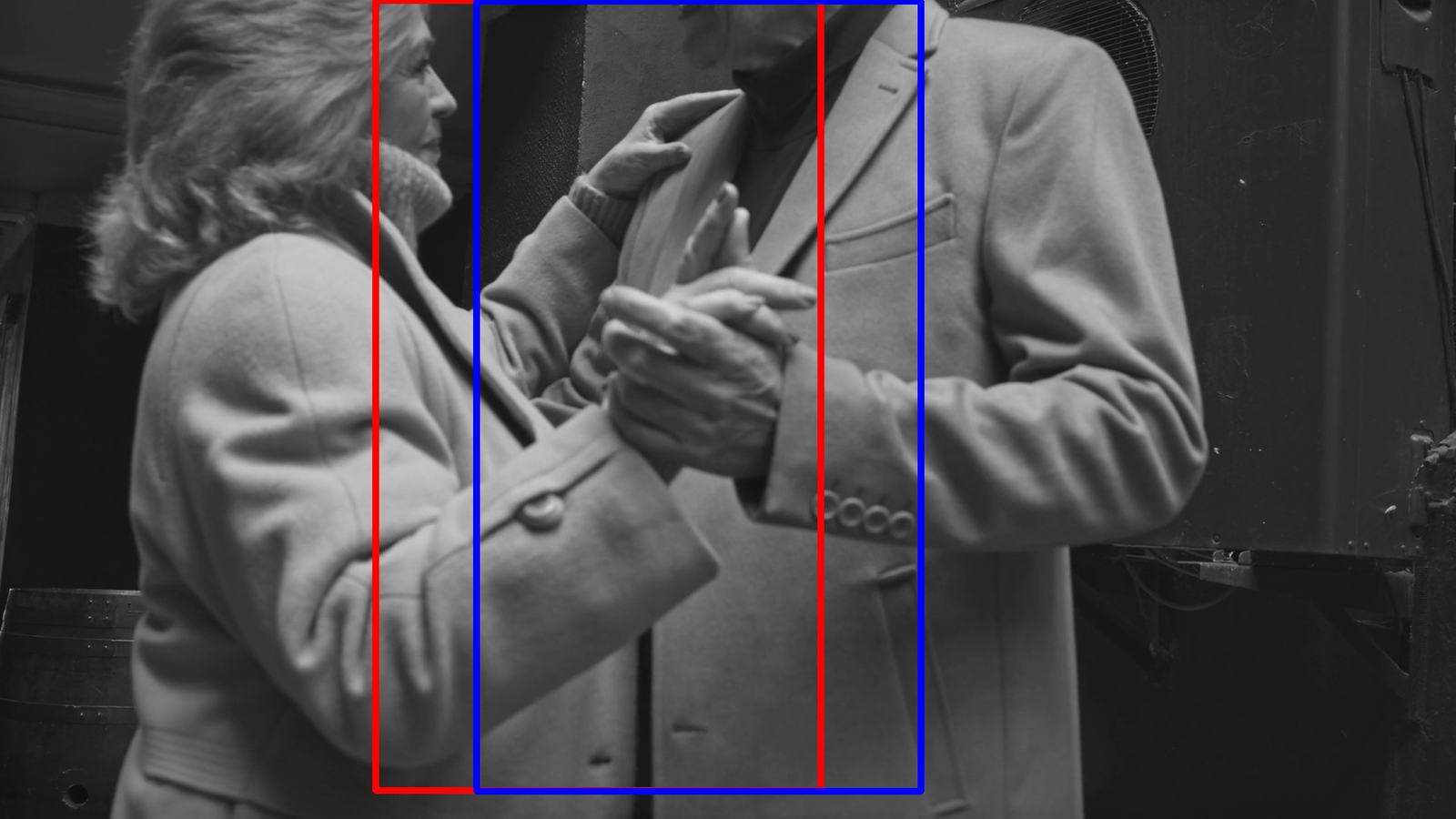} &
            \includegraphics[width=\fw]{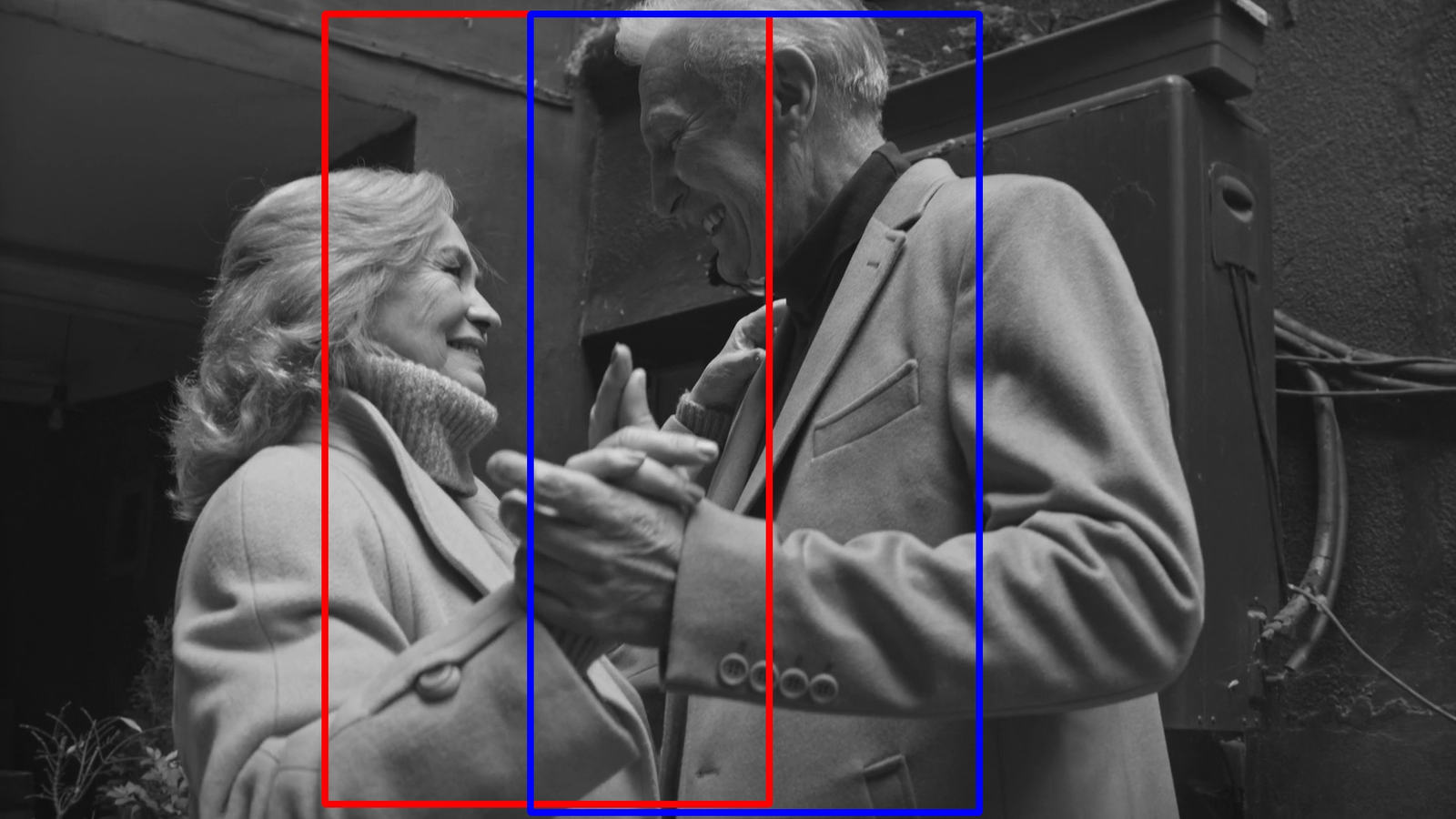} &
            \includegraphics[width=\fw]{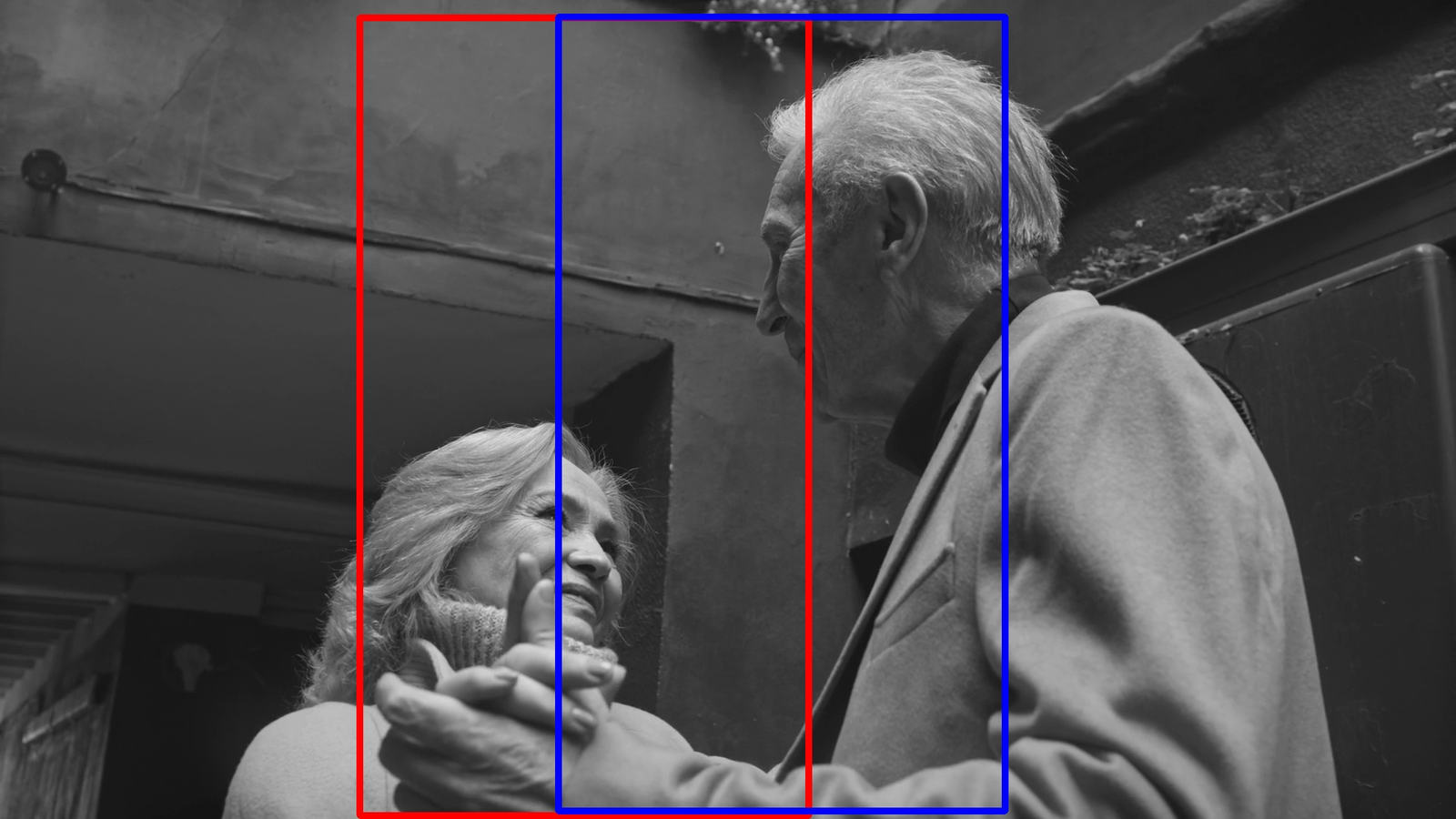} \\
            \rotatebox{90}{\footnotesize Ours + MLLM-based Agent} &
            \includegraphics[width=\fw]{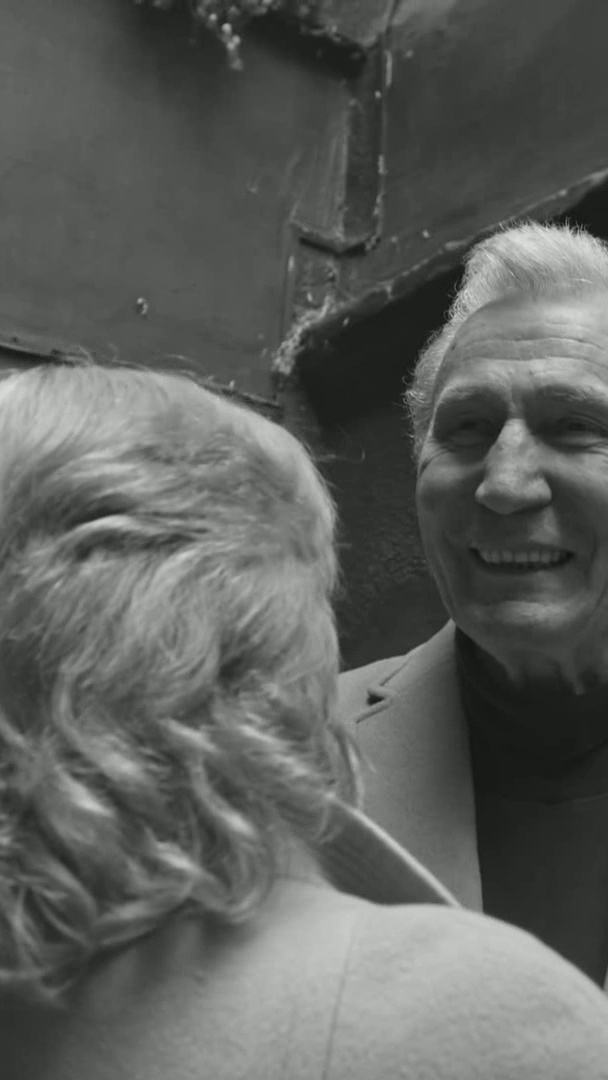} &
            \includegraphics[width=\fw]{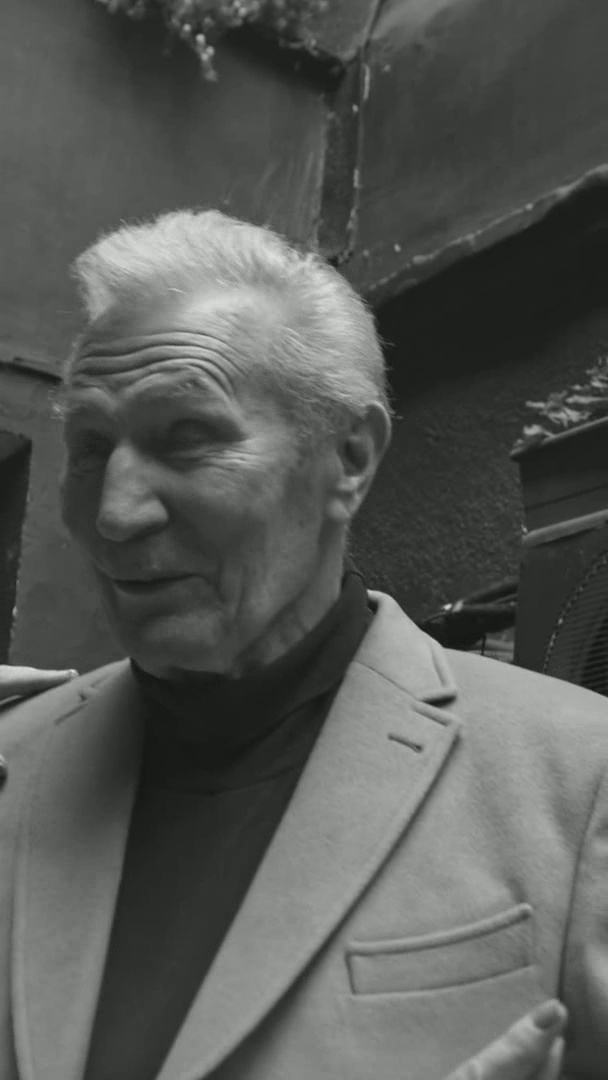} &
            \includegraphics[width=\fw]{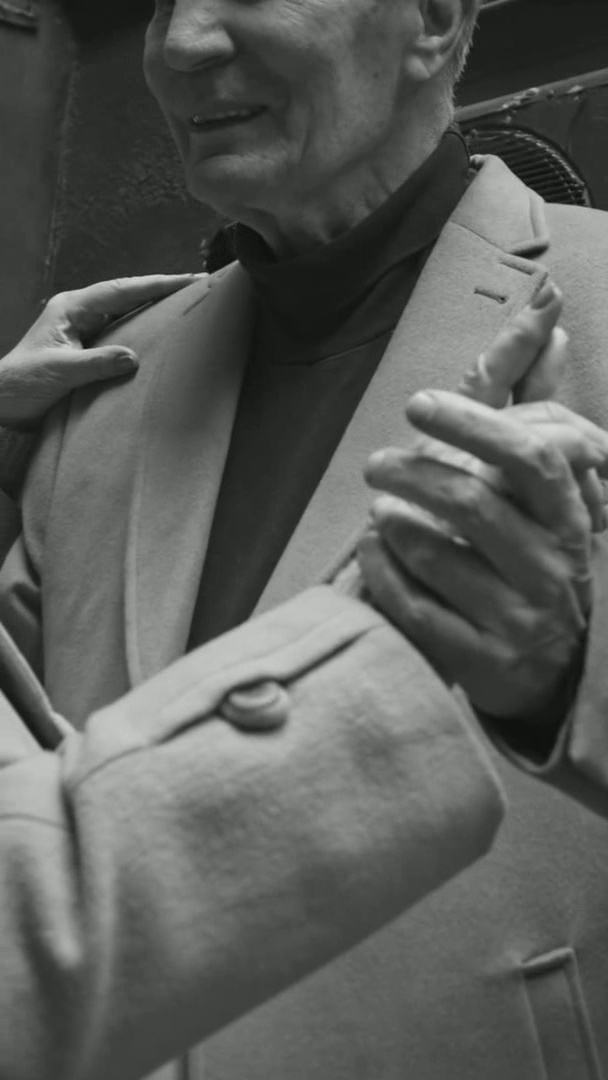} &
            \includegraphics[width=\fw]{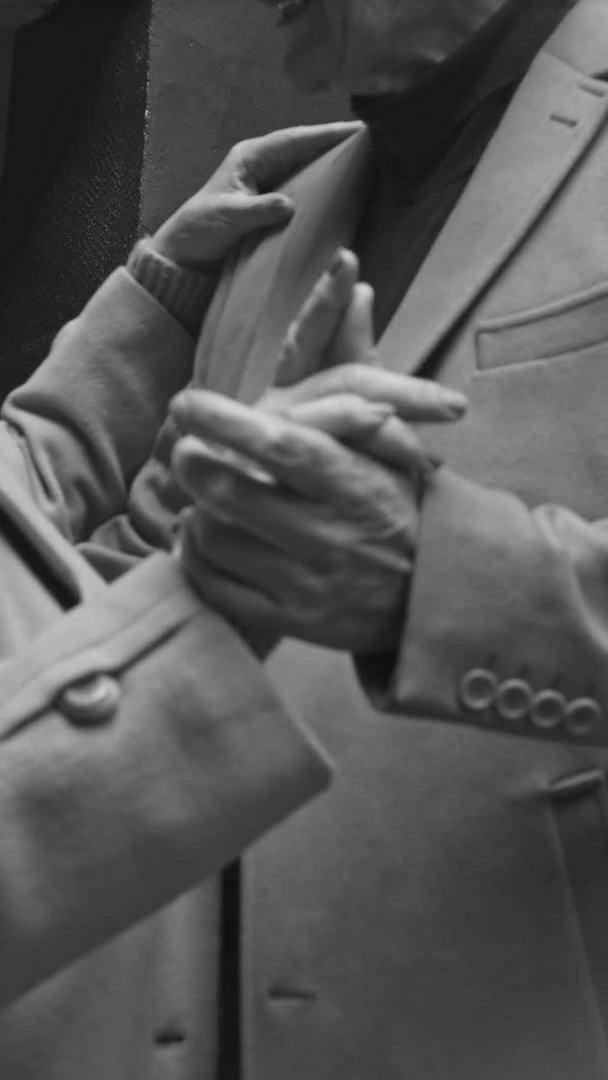} &
            \includegraphics[width=\fw]{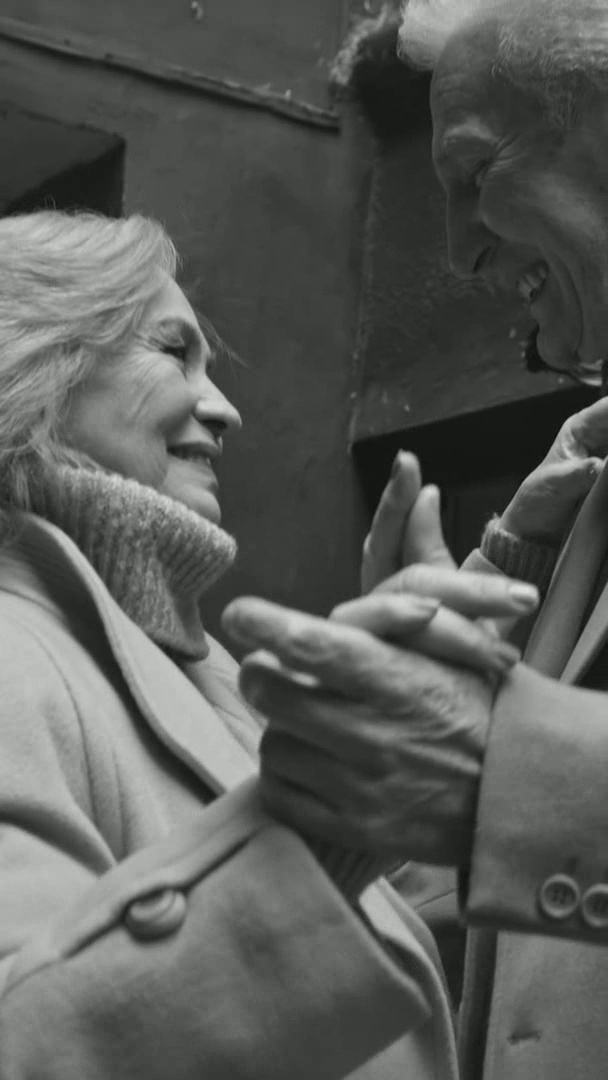} &
            \includegraphics[width=\fw]{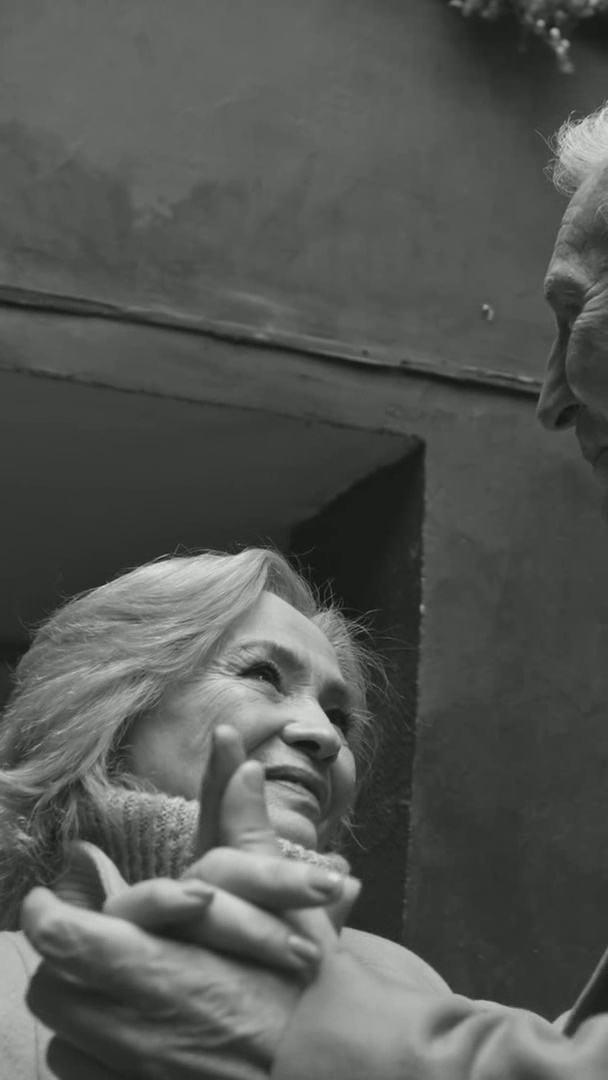} \\
        \end{tabular}
        \caption{Video ID: 4066364}
    \end{subfigure}

    \addtocounter{figure}{-1} 
    \caption{\textbf{Qualitative comparison of ROI extraction methods on the MiraData dataset \cite{ju2024miradata} (Part 3 of 3).} Subfigure (\textit{e}) demonstrates the final example of tracking and framing performance.}
    \label{fig:main_qualitative_comparison_part3}
\end{figure*}


\end{document}